\RequirePackage[l2tabu, orthodox]{nag}

\documentclass[12pt,phd,a4paper,twoside]{ucl_thesis}

\usepackage{blindtext}

\usepackage{emptypage}

\usepackage{graphicx}

\usepackage{float}
\usepackage{changepage}

\usepackage{amsmath}
\usepackage{amssymb}
\usepackage{amsthm}

\usepackage{gensymb}
\usepackage{textcomp}

\usepackage{setspace}

\usepackage{multirow}

\usepackage{bibentry} 

\usepackage[format=hang,font=small,labelfont=bf]{caption}

\usepackage{etoolbox}

\usepackage{lipsum}

\usepackage{xcolor}
\definecolor{nice-red}{HTML}{E41A1C}
\colorlet{dark-red}{nice-red!80!black}
\definecolor{nice-orange}{HTML}{FF7F00}
\colorlet{dark-orange}{orange!85!black}
\definecolor{nice-yellow}{HTML}{FFC020}
\definecolor{nice-green}{HTML}{4DAF4A}
\definecolor{nice-blue}{HTML}{377EB8}
\definecolor{nice-purple}{HTML}{984EA3}

\usepackage[xindy,acronym,sort=def]{glossaries}

\usepackage{booktabs}

\usepackage[ruled,noend,vlined]{algorithm2e}

\PassOptionsToPackage{hyphens}{url}
\usepackage{hyperref}
\hypersetup{
 unicode=false,
 pdftoolbar=true,
 pdfmenubar=true,
 pdffitwindow=false,
 pdfstartview={FitH},
 pdfauthor={Minqi Jiang},
 pdfnewwindow=true,
 colorlinks=true,
 linkcolor=magenta,
 citecolor=blue,
 filecolor=magenta,
 urlcolor=blue
}

\usepackage{subcaption}
\usepackage{microtype}
\usepackage{url}
\usepackage[numbers]{natbib}
\usepackage{soul}
\usepackage{changepage}
\usepackage{xargs}
\usepackage{bm}

\usepackage{tabularx}
\usepackage{wrapfig}
\usepackage{multicol}
\usepackage{rotating}
\usepackage[export]{adjustbox}

\usepackage[capitalize]{cleveref}

\usepackage{stmaryrd}

\usepackage[outline]{contour}

\usepackage{pdfpages}
\newtheorem{lemma}{Lemma}

\newtheorem{remark}[lemma]{Remark}

        {\medskip}

\newtheorem{innercustomgeneric}{\customgenericname}
\providecommand{\customgenericname}{}
\newcommand{\newcustomtheorem}[2]{%
  \newenvironment{#1}[1]
  {%
   \renewcommand\customgenericname{#2}%
   \renewcommand\theinnercustomgeneric{##1}%
   \innercustomgeneric
  }
  {\endinnercustomgeneric}
}

\newcustomtheorem{customthm}{Theorem}

        {\hspace*{\fill}$\Box$\par\vspace{4mm}}
        {\hspace*{\fill}$\Box$\par}
        
\DeclareMathOperator*{\argmax}{arg\!\max}
\DeclareMathOperator*{\argmin}{arg\!\min}

\newglossarystyle{acronyms}{%
    \setglossarystyle{listdotted}%
    \setlength{\glslistdottedwidth}{.23\linewidth}%
}

\newacronym{2p0s}{2p0s}{Two-player zero-sum}
\newacronym{a2c}{A2C}{Advantage Actor-Critic}
\newacronym{accel}{ACCEL}{Adversarially Compounding Complexity by Editing Levels}
\newacronym{acl}{ACL}{Automatic curriculum learning}
\newacronym{ai}{AI}{Artificial intelligence}
\newacronym{aoh}{AOH}{Action observation history}
\newacronym{cics}{CICS}{Curriculum-induced covariate shift}
\newacronym{cl}{CL}{Curriculum learning}
\newacronym{cnn}{CNN}{Convolutional neural network}
\newacronym{dcd}{DCD}{Dual Curriculum Design}
\newacronym{dr}{DR}{Domain randomization}
\newacronym{gae}{GAE}{Generalized Advantage Estimation}
\newacronym{iqm}{IQM}{Interquartile mean}
\newacronym{lstm}{LSTM}{Long Short-Term Memory}
\newacronym{marl}{MARL}{Multi-agent reinforcement learning}
\newacronym{mcc}{MCC}{Minimal-Criterion Coevolution}
\newacronym{mdp}{MDP}{Markov decision process}
\newacronym{ml}{ML}{Machine learning}
\newacronym{mlp}{MLP}{Multi-layer perceptron}
\newacronym{ne}{NE}{Nash equilibrium}
\newacronym{ood}{OOD}{Out of distribution}
\newacronym{pcg}{PCG}{Procedural content generation}
\newacronym{plr}{PLR}{Prioritized Level Replay}
\newacronym{robust_plr}{PLR$^{\perp}$}{Robust Prioritized Level Replay}
\newacronym{paired}{PAIRED}{Protagonist Antagonist Induced Regret Environment Design}
\newacronym{poet}{POET}{Paired Open-Ended Trailblazer}
\newacronym{pomdp}{POMDP}{Partially-observable Markov decision process}
\newacronym{posg}{POSG}{Partially-observable stochastic game}
\newacronym{ppo}{PPO}{Proximal Policy Optimization}
\newacronym{relu}{ReLU}{Rectified linear unit} 
\newacronym{repaired}{REPAIRED}{Replay-Enhanced PAIRED}
\newacronym{rl}{RL}{Reinforcement learning}
\newacronym{rnn}{RNN}{Recurrent neural network}
\newacronym{samplr}{SAMPLR}{Sample-Matched Prioritized Level Replay}
\newacronym{sl}{SL}{Supervised learning}
\newacronym{sps}{SPS}{Steps per second}
\newacronym{ssl}{SSL}{Self-supervised learning}
\newacronym{td}{TD}{Temporal difference}
\newacronym{ued}{UED}{Unsupervised Environment Design}
\newacronym{upomdp}{UPOMDP}{Underspecified partially-observable MDP}
\newglossarystyle{notationstyle}{%
    \setglossarystyle{listdotted}%
    \setlength{\glslistdottedwidth}{.33\linewidth}%
}
\newglossary[ng]{notation}{ns}{no}{Notation Used}

\newglossaryentry{Agent model parameters}
{
    type=notation,
    name=$\phi \in \Phi$,
    description={Agent model parameters.}
}

\newglossaryentry{Policy}
{
    type=notation,
    name=$\pi \in \Pi$,
    description={Agent policy and policy space.}
}

\newglossaryentry{Free parameters}
{
    type=notation,
    name=$\theta \in \Theta$,
    description={Free parameters of a UPOMDP.}
}

\newglossaryentry{MDP state}
{
    type=notation,
    name=$s_t \in \mathcal{S}$,
    description={State at timestep $t$ and state space.}
}

\newglossaryentry{MDP observation}
{
    type=notation,
    name=$o_t \in \Omega$,
    description={Observation at timestep $t$ and observation space.}
}

\newglossaryentry{MDP observation function}
{
    type=notation,
    name=\ensuremath{O(o_t|s_t)},
    description={Observation function.}
}

\newglossaryentry{MDP action}
{
    type=notation,
    name=$a_t \in \mathcal{A}$,
    description={Action at timestep $t$ and action space.}
}

\newglossaryentry{MDP transition function}
{
    type=notation,
    name=\ensuremath{P(s_{t+1}|s_t,a_t)},
    description={State transition function.}
}

\newglossaryentry{MDP reward}
{
    type=notation,
    name=$r_t \in \mathbb{R}$,
    description={Reward at timestep $t$.}
}

\newglossaryentry{Reward function}
{
    type=notation,
    name=\ensuremath{\mathcal{R}(s_t,a_t,s_{t+1})},
    description={Reward function.}
}

\newglossaryentry{Discount factor}
{
    type=notation,
    name=$\gamma$,
    description={Reward discount factor.}
}

\newglossaryentry{Timestep}
{
    type=notation,
    name=$t \in \mathbb{Z}^+$,
    description={Timestep of episode.}
}

\newglossaryentry{Length of episode}
{
    type=notation,
    name=$T$,
    description={Length of episode.}
}

\newglossaryentry{Max length of episode}
{
    type=notation,
    name=$T_{\text{max}}$,
    description={Maximum episode length.}
}

\newglossaryentry{AOH}
{
    type=notation,
    name=\ensuremath{\tau_t},
    description={Trajectory or AOH at timestep $t$.}
}

\newglossaryentry{Future return}
{
    type=notation,
    name=\ensuremath{R_t = \sum_{k=t}^{\infty}\gamma^{k-t}r_k},
    description={Future discounted return from timestep $t$.}
}

\newglossaryentry{Total return}
{
    type=notation,
    name=\ensuremath{J_{\theta}(\pi) = \mathbb{E}_{\pi}[\sum_{t=0}^{\infty} \gamma^{t}r_t]},
    description={Total return of policy $\pi$ in environment instance $\theta$.}
}

\newglossaryentry{State value function}
{
    type=notation,
    name=\ensuremath{V(s_t)},
    description={State value function.}
}

\newglossaryentry{Advantage function}
{
    type=notation,
    name=\ensuremath{A(s_t,a_t)},
    description={Advantage of action $a_t$ in state $s_t$.}
}

\newglossaryentry{1-step TD error}
{
    type=notation,
    name=\ensuremath{\delta_t},
    description={One-step TD error at timestep $t$.}
}

\newglossaryentry{GAE lambda}
{
    type=notation,
    name=$\lambda$,
    description={GAE discount factor.}
}

\newglossaryentry{Level replay buffer}
{
    type=notation,
    name=$\Lambda$,
    description={Level replay buffer (as used in PLR).}
}

\newglossaryentry{Conditional utility function}
{
    type=notation,
    name=\ensuremath{U(\pi|X)=\mathbb{E}_{\pi}[R_0 | X}],
    description={Utility of policy $\pi$ given $X$.}
}

\newglossaryentry{Belief model}
{
    type=notation,
    name=\ensuremath{\mathcal{B}(s_t|\tau)},
    description={Belief of state at timestep $t$ given history $\tau$.}
}

\newglossaryentry{Empirical estimate}
{
    type=notation,
    name=\ensuremath{\hat{x}},
    description={Empirically-derived estimate of $x$.}
}

\newglossaryentry{Stop gradient}
{
    type=notation,
    name=\ensuremath{x_{\perp}},
    description={Apply stop gradient to $x$.}
}

\makeglossaries
\glsaddall

\usepackage[utf8]{inputenc}

\title{Learning Curricula \\ in Open-Ended Worlds} 
\author{Minqi Jiang}
\department{Department of Computer Science}

\begin{document}

\includepdf[offset=1in -1in]{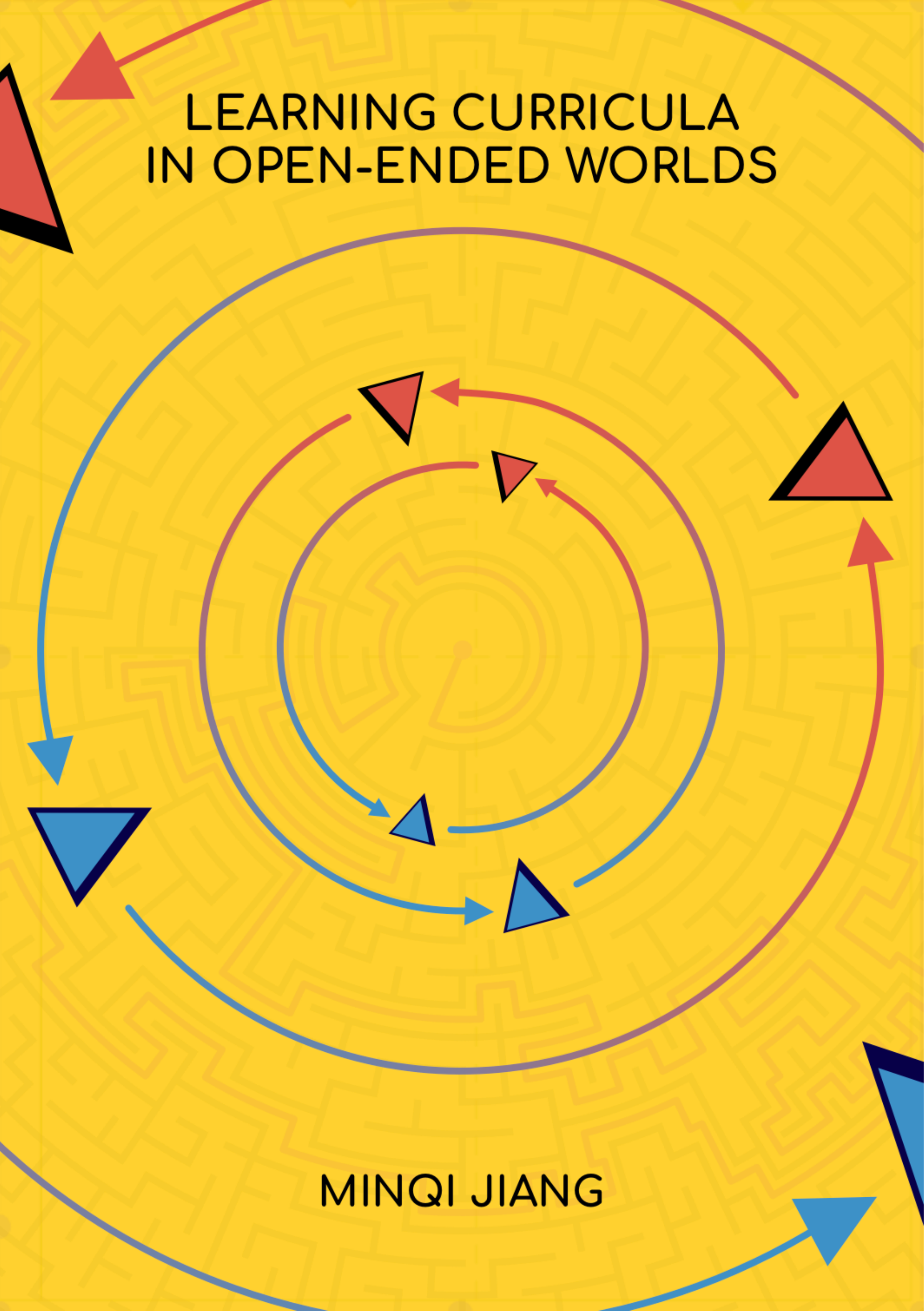}

\nobibliography*
\pagenumbering{arabic}

\maketitle
\frontmatter
\chapter*{}
\begin{center}
\emph{To my parents, who provided a good initialization, \\and friends, who reshaped my objective landscape.}
\end{center}

\chapter*{Acknowledgements}

While a PhD thesis usually comes across as a personal artifact, its creation is closer to that of a blockbuster movie: Years of collaborative effort by a whole cast of creative people. Stretching this analogy beyond its reasonable limits, I liken my role to that of a no-name wannabe director, and my advisor Tim Rocktäschel, as the seasoned producer that decided to take a chance on me with a Hollywood budget. Thank you, Tim, for believing in my abilities as a researcher, well before there was much to go off of, and for your endless patience in entertaining---and even encouraging---my often outlandish or impractical ideas for research. You helped cultivate in me a true fearlessness in chasing after big ideas and in confronting my own stupidity. Our discussions often led to somewhere truly fascinating. I had the further, outrageous fortune of being co-advised by Edward Grefenstette, from whom I learned a great deal about the life of the mind and the life outside it. Your conviction in my ideas was instrumental to their seeing the light of day, and our conversations shaped them into something much better. Additionally, I thank Laura Toni and Matt Kusner for their guidance at important junctions of my studies. I thank Jeff Clune and Dimitrios Kanoulas for examining this thesis. %
Jeff's research has been a constant source of inspiration, and Dimitrios's work shines a blazing beacon on the incredible potential of our field and the road ahead. I am deeply grateful for their involvement.

An important, close collaborator throughout the years has been Michael Dennis, who introduced me to the power of decision theory and game theory, and the fundamental role they have to play in developing general AI systems. The ideas in this thesis owe much to these discussions, and I continue to learn much from our conversations. My continued collaboration and conversations with Jack Parker-Holder and Mikayel Samvelyan have also been important in furthering the ideas laid out in this thesis, pushing their reach into new places. 
\newpage

The questions I pursued in this PhD were largely enabled by my joint affiliation at Meta AI. There, I was lucky to be part of many collaborations that introduced me to new ideas and wonderful people, including Roberta Raileanu, Heinrich Küttler, Eric Hambro, Mikael Henaff, Andrei Lupu, Chris Bamford, Sam Earle, Yiding Jiang, Jesse Mu, Ishita Mediratta, Victor Zhong, Eugene Vinitsky, Iryna Korshunova, Christoforos Nalmpantis, Sharath Chandra, and Wojciech Galuba. Meanwhile at UCL, I found myself in a welcoming community, where there was always space for discussions and musings with Zhengyao Jiang, Robert Kirk, Laura Ruis, Akbir Khan, Yingchen Xu, Pasquale Minervini, Pontus Stenetorp, Yihong Chen, Max Bartolo, and Yuxiang Wu. I owe much to Sebastian Riedel and Pierre-Louis Xech for making this dream setup possible, and doubly so, to Sebastian who also welcomed me into the UCL NLP group my first year when I was the only student in UCL DARK. I also thank Jelena Luketina and Nantas Nardelli, who shepherded me through my first research project in the middle of a pandemic, and with whom I had many fun and insightful conversations. While visiting FLAIR, I was lucky to work alongside many brilliant people, including Chris Lu, Ben Ellis, Matthew Jackson, Marc Rigter, Jonny Cook, Ola Kalisz, Silvia Sapora, Sebastian Towers, Tim Franzmeyer, Irene Zhang, Timon Willi, and Christian Schroeder de Witt.

This path through the multiverse was instigated by my friend and now collaborator, Jakob Foerster, who invited me to visit him in Oxford in August 2018, encouraged me to ``go into AI," and introduced me to Tim, who was just about to start his new lab at UCL. Thank you, Jakob, not only for the introduction, but also for your continued presence as a force of reason throughout my PhD. Your instinct for getting to the core of a problem is both instructive and terrifying. I hope some of that has rubbed off on me. 

Importantly, I thank my parents, Yanqing and Dan, for instilling a sense of curiosity and determination in me from an early age, and friends who helped me weather the process of starting a PhD in a new country during a global pandemic: Ben, Peter, Joe, Derrick, Haibo, and Nish. Lastly, I thank my partner Vanda, whose constant love and support has made the, at times, turbulent waves of the PhD journey smooth sailing.

\chapter*{Declaration}

I, Minqi Jiang confirm that the work presented in this thesis is my own. Where information has been derived from other sources, I confirm that this has been indicated in the thesis.

\vspace{2cm}
\hfill\textsc{Minqi Jiang}

\chapter*{Abstract}
Deep reinforcement learning (RL) provides powerful methods for training optimal sequential decision-making agents. As collecting real-world interactions can entail additional costs and safety risks, the common paradigm of sim2real conducts training in a simulator, followed by real-world deployment. Unfortunately, RL agents easily overfit to the choice of simulated training environments, and worse still, learning ends when the agent masters the specific set of simulated environments. In contrast, the real-world is highly open-ended---featuring endlessly evolving environments and challenges, making such RL approaches unsuitable. Simply randomizing across a large space of simulated environments is insufficient, as it requires making arbitrary distributional assumptions, and as the design space grows, it can become combinatorially less likely to sample specific environment instances that are useful for learning. An ideal learning process should automatically adapt the training environment to maximize the learning potential of the agent over an open-ended task space that matches or surpasses the complexity of the real world. This thesis develops a class of methods called Unsupervised Environment Design (UED), which seeks to enable such an open-ended process via a principled approach for gradually improving the robustness and generality of the learning agent. Given a potentially open-ended environment design space, UED automatically generates an infinite sequence or curriculum of training environments at the frontier of the learning agent’s capabilities. Through both extensive empirical studies and theoretical arguments founded on minimax-regret decision theory and game theory, the findings in this thesis show that UED autocurricula can produce RL agents exhibiting significantly improved robustness and generalization to previously unseen environment instances. Such autocurricula are promising paths toward open-ended learning systems that approach general intelligence---a long sought-after ambition of artificial intelligence research---by continually generating and mastering additional challenges of their own design.

\chapter*{Impact Statement}

The successes of deep reinforcement learning (RL)---including exceeding human-level performance on strategic games, designing next-generation chipsets, and controlling nuclear fusion plasma---remain largely confined to domains amenable to the application of handcrafted reward functions or supervised pre-training to improve task learnability. Still, deep RL often overfits to the training domain, preventing successful deployment in real-world open-ended environments with many degrees of freedom that may not be fully-anticipated within simulation. This thesis develops novel methods for automatically generating curricula rooted in minimax-regret decision theory and game theory. These methods thus provide principled assurances around the robustness of the resulting agents over potentially large environment design spaces. The resulting autocurricula not only ensure learnability by adapting training tasks to lie at the frontier of the agent's capabilities, but also enable the progressive expansion of the tasks considered during training. Given a sufficiently expressive design space, these methods thereby provide a path to training deep RL agents in simulation for successful deployment in the face of the open-endedness of the real world.

Both the core theoretical and algorithmic ideas presented in this thesis are largely agnostic to the decision-making problem, allowing the possibility of extension to many different problem domains beyond those explored in this thesis. Already, across both academia and industry, these methods have been applied to several additional RL settings outside the scope of the works in this thesis, including multi-agent, model-based, and meta-learning settings. We foresee that these methods and their intellectual progeny may extend to problem settings even farther afield from RL, for example, to self-supervised learning. Further extensions of these concepts to more universal task spaces may enable the realization of increasingly general systems that continually self-improve via such autocurricula, allowing system capabilities to scale directly with the amount of available training compute.

\raggedcolumns
\newpage

\setcounter{tocdepth}{2} 
\hypersetup{linkcolor=black}
\tableofcontents
{\small{
    \listoffigures
    \listoftables
}}
\clearpage

\hypersetup{linkcolor=magenta}

\glstoctrue

\setglossarystyle{acronyms}\printglossary[title=List of Abbreviations, type=\acronymtype]

\setglossarystyle{notationstyle}\printglossary[title=Notation Used, type=notation]
\clearpage

\mainmatter
\chapter{Introduction}

``\emph{An unproblematic state is a state without creative thought.\\Its other name is death.}''
\begin{flushright}
--- David Deutsch
\end{flushright}

\section{A New Kind of Software}
A unique aspect of the human species is our ability to write software. This relatively recent technological development is a culmination of many other impressive abilities possessed by humans: abstract reasoning, language, and the opposable thumb.\footnote{Necessary for quickly writing new software.} From the financial clearinghouse systems that hold up the modern economy to the external brains in the form of smartphones in billions of pockets worldwide, modern software systems serve at once as critical infrastructure for the functioning of society and external organs that amplify our natural capabilities and instincts. In this sense, software creation is a deeply human pursuit. A well-crafted piece of software codifies the concerns and needs of real human beings. Yet, conversely, that every program must be coded by a human expert limits software to only embody solutions that are known in advance. The recent rise of deep learning introduces a pivotal dynamic: Software that, in a sense, programs itself by tuning the weights of an artificial neural network, given only a high-level specification of success~\citep{krizhevsky2017imagenet, goodfellow2016deep} and examples fulfilling this criterion. The consequences of this paradigm shift have yet to play out in full, but already, deep learning has unlocked breakthrough advances across nearly every domain of artificial intelligence (AI) research and application by allowing the machine to discover solutions beyond the ken of human engineers. Still, deep learning methods commonly rely on humans to fully specify the problem of interest, which risks overfitting to the specific problem provided~\citep{stanley2015greatness}. In this thesis, we develop methods that relax this requirement, allowing deep learning systems to generate their own problems from which to learn. The resulting algorithms step us closer to a new kind of software, one that self-improves by generating its own task data, toward more robust and general behaviors---in short, software that better reflects the distinct human capacity to universally explore, explain, and create our world~\citep{deutsch2011beginning}.

Before discussing how an AI system can come to generate its own tasks for self-improvement, we must first discuss the simpler case of AIs that learn specific, predefined tasks---that is, automatically and iteratively improve their performance on these tasks. Indeed, a valid question to ask is why we should build learning systems in the first place. After all, early work in the field did not concern itself with learning. Rather, earlier methods focused on \emph{symbolic AI}, a class of methods that seeks to produce intelligent behavior by executing human-specified rules, typically embedded within a logical system~\citep{newell2007computer, russell2010artificial, kolata1982can}. Despite early successes in automated theorem proving~\citep{newell1956logic} and playing simple board games~\citep{samuel1959some}, symbolic AI ultimately ran into a computational brick wall on many real-world problems. To see why, consider the simple problem of email spam detection. Devising a spam filter based on an enumerated ruleset would require an astronomical, potentially infinite number of rules. Complicating matters further, the cat-and-mouse game between email users and spammers makes these rules impossible to fully specify in advance. 

In contrast to symbolic AI, \emph{machine learning} approaches forgo predefined behavioral rules in favor of learning rules that update (or optimize) the model parameters (or weights), using any training data that becomes available, both immediately and over time . Under this paradigm, the logic for solving the task is replaced by an automatically-learned program that is implemented by the weights of the trained model---one that can further adapt over time by continuing the optimization procedure on any new data that becomes available. Of particular importance is the ML problem setting of \emph{supervised learning}, where the goal is to optimize the model so that for each datapoint, the model can accurately predict a corresponding target value, e.g. a binary label indicating whether the input email text is spam or not-spam. Common supervised learning methods include logistic regression~\citep{berkson1944application}, support vector machines~\citep[SVM,][]{boser1992training}, and Gaussian process models~\citep{rasmussen2006gaussian}.

However, previous ML systems struggled to model complex problem domains, as they were largely hindered by suboptimal input representations, which were typically hand-engineered. For example, to train an SVM to detect spam emails, the email text must be preprocessed into a numeric format. A typical choice is to define a vector of length equal to some predefined vocabulary size and assign a value of 1 to each component corresponding to a word in the email. The SVM then seeks a linear separating plane between spam and non-spam emails within this rather arbitrary representation space---in which approximating such linear separability is likely challenging. \emph{Deep learning} addresses this representational challenge by directly learning the input representations end-to-end as part of the optimization, typically via stochastic gradient descent~\citep{rumelhart1985learning,lecun1989backpropagation,goodfellow2016deep}. In a \emph{deep neural network} (DNN), input representations correspond to the intermediate activations in a sequence of neural net layers---each usually a linear operator over the outputs of the previous layer followed by an optional nonlinearity, thereby parameterizing a rich space of functions at each layer. This approach removes the need for the manual feature engineering that limited previous ML approaches. Rather, in optimizing the loss function, deep learning directly seeks representations that are optimal for the task at hand. Moreover, as DNNs map inputs to outputs through a series of matrix multiplications, they easily scale up in both model size and parallelization across data batches on modern \emph{graphics processing units} (GPUs)~\citep{krizhevsky2017imagenet}. Thus, assisted by an equally meteoric and mutualistic rise in GPU technology~\citep{perry2018move}, deep learning has rapidly taken over AI, becoming the basis of the state-of-the-art method in nearly every application domain and enabling new, previously unimaginable, use cases, such as general-purpose chatbots~\cite{thoppilan2022lamda, openai2021chatgpt,openai2023gpttechreport} and text-to-image generators~\citep{ramesh2021zero,ramesh2022hierarchical} trained on web-scale datasets.

Despite this roaring success, supervised deep learning is fundamentally limited, due to specific assumptions it makes about its training data: 
\begin{enumerate}
    \item \textbf{The Problem Assumption:} The task of interest is fully-specified by the system designer upfront.
    \item \textbf{The Data Assumption:} The task-specific data is provided a priori.
\end{enumerate}

\noindent Assumption 1 simply points out that in providing the dataset upfront, SL commits to a specific set of tasks for learning. Consequently, after training, the model cannot be expected to learn any tasks outside the scope of the dataset. The model will thus be limited in generality. Assumption 2 highlights how SL, in itself, offers no means to generate novel training data. An important consequence of this fact is that SL can only be applied to problem settings in which at least some solutions are already known (and therefore can be included in the training data). In contrast, the problem setting of \emph{reinforcement learning}~\citep{sutton2018reinforcement} typically assumes zero initial training data. Instead, the model---in this context, called the \emph{agent}---must learn to accomplish the task of interest through repeated trial-and-error. In the process, the agent generates its own training data from which it learns to improve its performance. In fact, when the agent's performance is approximately optimal for the task, its decisions can themselves be recorded and made into a dataset for supervised learning~\citep{ernst2005tree}. The high-quality data used to train large language models owe their conception to just such a process: The human authors of these utterances have learned to write publishable text over many years of trial-and-error, and the data generated reflects the fruits of this labor. Similarly, the collection of spam and not-spam emails can be seen as a similar result of many email users learning, over time via trial-and-error, to mark certain messages as spam.

Deep RL methods have also made great strides over the past decade, achieving such feats as matching or exceeding top human players in strategic games~\citep{silver2016mastering,berner2019dota,silver2017mastering,perolat2022mastering,meta2022human}, performing large-scale chipset design~\citep{mirhoseini2021graph}, and controlling nuclear fusion plasma~\citep{degrave2022magnetic}.
However, like SL, typical RL approaches also make the Problem Assumption, ultimately limiting the degree of intelligence that can emerge within such learning systems. 
    
This thesis seeks to develop deep learning algorithms that relax both Assumptions 1 and 2 around the problem and data of interest, resulting in more general-purpose learning systems that produce their own training tasks and data. In order to relax Assumption 1, the works in this thesis consider tasks produced via procedural-content generation---that is, according to some underlying algorithmic process exhibiting a wide range of possibilities in output. In particular, the methods developed in this thesis focus entirely on how to generate a sequence of such tasks most appropriate for facilitating the learning of the agent, such that at the end of training, the agent's behavior will exhibit a maximum degree of robustness and generalizability across different task variations. In contrast to algorithms that focus on model and optimizer changes, these methods result in improved performance purely by changing the nature of the training data. As such, they are, as will be demonstrated, easily combined with other improvements to further enhance the agent's performance.

In order to relax Assumption 2, we make use of deep reinforcement learning~\citep{mnih2015human,schulman2015trust}, in which the agent is modeled using a DNN and interacts with each generated task to collect its own training data. Consequently, this thesis conceives of \emph{tasks} within the framework of RL and thus views a specific task as equivalent to a particular instantiation of an environment with a corresponding reward function. Following terminology from the game AI community, we will also often refer to such instantiations as a \emph{level} of the environment~\citep{risi2020increasing}.

Taken together, the approach outlined above leads to algorithms that generate automatic curricula or \emph{autocurricula}~\citep{graves2017automated,leibo2019autocurriculum_manifesto} over some space of tasks. Each such curriculum can be viewed as a ``path" through this task space, generated on-the-fly during the course of training to adaptively guide the learning dynamics of the agent according to some criterion. In this thesis, we investigate both heuristic and mathematically principled criteria for generating informative autocurricula that lead to agents exhibiting improved robustness and generality across task variations. Such autocurricula are often necessary to produce performant agents, given that even toy environments can exhibit combinatorial complexity in terms of the number of possible instantiations. Thus the most informative instances for a particular learning outcome may be rarely sampled, if at all. In particular, this thesis focuses on a class of autocurriculum methods called \emph{unsupervised environment design}~\citep[UED,][]{dennis2020emergent}, which adapts the task distribution in order to produce generally-capable agents that can robustly succeed over the full task space, rather than for any specific task distribution. At a high level, such autocurricula unfold by presenting the agent with tasks at the frontier of its current capabilities, until no such tasks can be further proposed or---in the case that the task space contains unlimited complexity---continue forever, ever-robustifying the agent to new challenges.

Crucially, the systems developed here learn not only how to solve tasks, but also which tasks to solve by autonomously directing its own learning toward the most informative tasks. \citet{leibo2019autocurriculum_manifesto} describes this higher-order task of finding the most useful next task for training as \emph{The Problem Problem}, and \citet{clune2019ai} notes this bootstrapping behavior as an important component of \emph{AI-generating algorithms} (AI-GAs), which automatically produce a form of generally-capable intelligence. Relatedly, \citet{schmidhuber1999artificial,schmidhuber2006developmental} describes systems that autonomously direct their own learning as exhibiting \emph{artificial curiosity}. Indeed, autocurricula can be viewed as a form of exploration over the task space, with the aim of collecting the most informative training data.

Autocurricula are deeply related to the problem of \emph{open-endedness}~\citep{stanley2017open,stanley2015greatness}, which seeks to devise a system capable of generating endlessly novel outputs over time. Remarkably, while no artificial system has successfully sustained an open-ended process, many real-world systems seem to exhibit open-endness, including the tree of life (i.e. the phylogenetic tree)~\citep{hochberg2017innovation}, the set of invented technologies~\citep{bedau2019open}, and human culture~\citep{mesoudi2018cumulative}. When the task space contains unbounded complexity, autocurricula serve as promising paths to open-endedness, by co-evolving an infinite set of tasks for the agent. Importantly, such a process may circumvent a longstanding challenge of open-endedness, which is that definitions of open-endedness are necessarily subjective~\citep{stanley2016role}, e.g. the set of all real numbers is technically open-ended, but this kind of open-endedness is likely uninteresting to most people. By tying the open-ended generation of new tasks to focus on those at the frontier of the agent's capabilities, the resulting tasks are anchored in a novelty criterion that is both non-trivial and practical. As an autocurriculum expands across the task space, the agent may then develop increasingly general capabilities---a prospect that highlights the close connection between open-ended learning and previous notions of ``artificial general intelligence"~\citep{legg2007universal}, which loosely refer to AIs capable of achieving any task of practical importance. While the methods developed in this thesis do not address the open problem of how to programmatically represent a \emph{universal task space} in which such autocurricula can develop increasingly-general capabilities, the methods show promising results on limited task spaces that nevertheless parameterize an astronomical number of unique task instances. These methods may thus also be useful for generating autocurricula over more universal task spaces\footnote{One promising universal task space is the space of all Markov Decision Processes represented as programs, which can be approximated by the support of a large language model trained on web-scale datasets of code.}, thereby serving as useful steps toward achieving both open-endedness and more general AI systems that perpetually self-improve.

\section{Overall Structure and Contributions}

This thesis contributes several new methods for generating autocurricula via UED, with the purpose of producing agents capable of robust behaviors across an entire task space. These methods stem from findings obtained in pursuing the following research problems around autocurricula:
\begin{itemize}
\setlength{\itemsep}{-0.0pt} 
\item How does the order in which tasks are presented to an RL agent during training affect its sample efficiency and generalization to held-out tasks?

\item Can we devise simple and scalable autocurricula-generating algorithms that improve agent performance over a potentially infinite task space?

\item Can we provide formal guarantees on the robustifying effects of such autocurriculum methods?

\item How can we improve how efficiently autocurricula search for the most informative training tasks?

\item How can autocurricula fail, and how can such failings be addressed?

\end{itemize}

\noindent Before presenting the results of these investigations, Chapter~\ref{chapter:background} introduces the core concepts underlying this work, beginning with a formal description of RL across several important problem settings and a detailed discussion of policy gradient methods. While the experiments in this thesis focus on single-agent learning problems, it is important to highlight that curriculum learning is inherently a multi-agent problem setting: There is always the concept of a student and a teacher underlying such algorithms. Thus, Section~\ref{sec:ne} reviews key ideas in game theory used throughout this thesis to explore this intuition at length. In particular, we will view autocurricula as arising from the competitive dynamics of a teacher and student in a two-player zero sum game and present several deep connections between single-agent curriculum learning and multi-agent settings. Section~\ref{sec:decision_theory} then introduces key ideas from decision theory that are also foundational to the theoretical analysis of these methods. Section~\ref{sec:autocurricula} discusses the motivations of various classes of automatic curriculum algorithms, highlighting the distinct value of the UED approach taken in this thesis. The proceeding chapters then focus on specific contributions:
\medskip
\paragraph{Prioritized Level Replay} (PLR), introduced in Chapter~\ref{chapter:plr}, is a conceptually simple, yet highly-scalable, method for generating auto-curricula over environment instances in a potentially infinite task space. PLR selectively samples environment instances (or \emph{levels}) during training to prioritize those in which the agent incurs the highest average value prediction errors, based on recent episodes in those levels. Empirically, PLR induces autocurricula over levels that improve both the sample efficiency on the training distribution and generalization performance on held-out test levels. I led this project in terms of idea conception, algorithmic and experimental design and implementation, and paper writing. The contents of this chapter appeared in

\medskip
\begin{adjustwidth}{2em}{0em}
\noindent Prioritized Level Replay. Minqi Jiang, Edward Grefenstette, Tim Rocktäschel. 2021. In \emph{The International Conference on Machine Learning (ICML 2021)}.
\end{adjustwidth}
\medskip

\paragraph{Dual Curriculum Design} (DCD), presented in Chapter~\ref{chapter:dcd}, provides a more principled framework for understanding autocurriculum methods like PLR. The original value prediction error of PLR is motivated by a heuristic argument. Even so, agents trained with PLR outperform those trained with previous UED methods with strong theoretical justifications in terms of zero-shot transfer to held-out environment instances. DCD generalizes the decision and game theoretic foundations of previous UED methods by modeling the sequence of training levels as arising from two concurrent curricula, each produced by a distinct teacher. Viewing PLR as a special case of DCD reveals a version of PLR called \emph{Robust PLR} (PLR$^\perp$), which has minimax regret guarantees at the Nash equilibria (NE) of its DCD game. Moreover, DCD analysis shows that the replay mechanism of PLR$^\perp$ can be combined with previous UED algorithms to produce more effective versions that retain their minimax regret guarantee at NE. I co-led this work, driving the algorithmic design, empirical studies, and paper writing, as well as proposing the initial project concept of synthesizing PLR with UED. Content from this chapter appeared in 

\medskip
\begin{adjustwidth}{2em}{0em}
\noindent Replay-Guided Adversarial Environment Design. Minqi Jiang$^*$, Michael Dennis$^*$, Jack Parker-Holder, Jakob Foerster, Edward Grefenstette, Tim Rocktäschel. 2022. In \emph{Neural Information Processing Systems (NeurIPS 2022)}.
\end{adjustwidth}
\medskip

\paragraph{Evolving Curricula}, the subject of Chapter~\ref{chapter:accel}, presents a powerful extension of PLR. The standard PLR implementation performs random search to find high-regret levels for training. This approach can be ineffective in more complex environment design spaces, especially as high-regret levels become lower regret upon successive revisitations via level replay. By viewing the PLR level replay buffer as a ``population" of levels and regret estimates as their fitness scores, we can replace random search with evolutionary search, which can more effectively search for high-regret levels by continuing to mutate previously discovered structures in the current population. This variation of PLR, called \emph{Adversarially Compounding Complexity via Editing Levels} (ACCEL), empirically produces curricula with greater environment complexity and policies with improved zero-shot transfer performance in complex design spaces. I co-led this work, conceiving the initial idea to extend PLR with evolution, contributing to the  algorithmic implementation, experimental design, and paper writing, and developing the web demo. The contents of this chapter appeared in 

\medskip
\begin{adjustwidth}{2em}{0em}
\noindent Evolving Curricula with Regret-Based Environment Design. Jack Parker-Holder$^*$, Minqi Jiang$^*$, Michael Dennis, Mikayel Samvelyan, Jakob Foerster, Edward Grefenstette, Tim Rocktäschel. 2022. In \emph{The International Conference on Machine Learning (ICML 2022)}.
\end{adjustwidth}
\medskip

\paragraph{Aligning Curricula}, the subject of Chapter~\ref{chapter:samplr}, focuses on an important problem inherent to autocurricula: Autocurricula typically introduce covariate shifts with respect to the ground-truth distribution of environment configurations at deployment. The benefits of curriculum learning thus come at the cost of biased data. We formally characterize this phenomenon as \emph{curriculum-induced covariate shift} (CICS) and prove  it can result in suboptimal policies in stochastic environments when the covariate shift occurs over the aleatoric parameters of the environment at each time step---that is, those environment properties whose value cannot be fully determined at each point of the trajectory. We show, in both discrete and continuous control environments, how autocurricula over such parameters can result in policies with severely degraded performance. To fix this issue, we propose \emph{Sample-Matched PLR} (SAMPLR), which produces robustifying curricula that nevertheless preserves optimality on a ground-truth distribution. I led this project, contributing to the problem formulation and driving the algorithmic design, empirical studies, and paper writing. This chapter is based on the following publication:

\medskip
\begin{adjustwidth}{2em}{0em}
\noindent Grounding Aleatoric Uncertainty for Unsupervised Environment Design. Minqi Jiang, Michael Dennis, Jack Parker-Holder, Andrei Lupu, Heinrich Küttler, Edward Grefenstette, Tim Rocktäschel, Jakob Foerster. 2022. In \emph{Neural Information Processing Systems (NeurIPS 2022)}.
\end{adjustwidth}
\medskip

In concluding this thesis, Chapter~\ref{chapter:conclusions} discusses the limitations of the ideas developed here and open research directions. This chapter further presents a general perspective on how the autocurriculum methods developed in this thesis relate to more general ideas of exploration that are essential to all learning systems. Much of this discussion is based on the following position article:

\medskip
\begin{adjustwidth}{2em}{0em}
\noindent General Intelligence Requires Rethinking Exploration. Minqi Jiang, Tim Rocktäschel, Edward Grefenstette. 2023. \emph{Royal Society Open Science}.
\end{adjustwidth}
\medskip

Lastly, it is important to note that while this thesis specifically focuses on developing foundational concepts around autocurricula in single-agent problem settings, it does so with a broader view that such techniques, once established in this basic setting, can then be extended to more complex settings such as multi-agent RL, model-based RL, and meta-learning. Indeed, I have since been involved with successfully extending these ideas to all of these additional problem settings. While not included in the core set of results presented in this thesis, Chapter~\ref{chapter:conclusions} provides a brief description of these follow-up works as examples of how the ideas developed in this thesis can be broadly applied.

\chapter{Background}
\label{chapter:background}

This chapter introduces the common background concepts necessary for the rest of this thesis. First, Section~\ref{sec:rl} introduces reinforcement learning, including the relevant formalisms for various settings, including the partially-observable, multi-agent, and multi-task settings (Sections~\ref{subsec:mdp}--\ref{subsec:upomdp}), and standard approaches for policy evaluation (Section~\ref{subsec:value_functions}). Then, Section~\ref{sec:policy_gradients} introduces policy gradient methods, the class of RL algorithms that serves as the base policy optimization approach in the experiments throughout this thesis. Sections~\ref{sec:ne} and \ref{sec:decision_theory} introduce key ideas from game theory and decision theory that inform the design of algorithms developed in this work. Finally, Section~\ref{sec:autocurricula} provides an overview of autocurricula methods. When appropriate, subsequent chapters may revise the presentation of certain concepts introduced here.

\section{Reinforcement Learning}
\label{sec:rl}
\emph{Reinforcement learning}~\citep[RL,][]{sutton2018reinforcement,kaelbling1996reinforcement} considers the setting in which an agent interacts across multiple time steps with its environment in order to learn to maximize a reward signal that appears in response to the agent's actions. This reward signal may be \emph{sparse}, appearing in only few interactions, or \emph{dense}, appearing in many interactions. Here, \emph{agent} simply refers to a system that takes actions in response to the current state of its environment in order to accomplish some task, e.g. one whose success is communicated via the reward signal. Typically, the agent is assumed to begin with zero (or limited) knowledge of the environment, and hence sequential interaction with the environment is necessary for the agent to learn to accomplish the task of interest. In practice, the agent can perform RL directly within a physical environment or a simulated world. RL in simulation usually entails the goal of transferring any learned behaviors to a target, real-world task domain---a process called \emph{sim2real transfer}~\citep{peng2017dr,zhao2020sim}. Often, this target domain is itself another simulated, virtual environment, such as a video game~\citep{bellemare2013arcade} or other software application~\citep{shi2017world}. 

A fundamental challenge of RL is balancing \emph{exploration} with \emph{exploitation}: At each time step, the agent can explore by trying actions that may improve its performance at the risk of reducing its current performance. Alternatively, it can exploit, by taking the best action it has learned so far, forgoing the chance to discover even better choices. Exploration is often required to avoid local optima, as well as to thoroughly explore the state space of the environment, so to develop more robust decision-making capabilities across different situations.

\subsection{Markov Decision Processes}
\label{subsec:mdp}
The environment is typically modeled as a \emph{Markov decision process}~\citep[MDP,][]{bellman1957dynamic,puterman1994markov}: At each time step $t$, an MDP exists in a state $s_t$ in the state space $\mathcal{S}$, where the starting state $s_0$ is sampled from a distribution $\rho(s_0)$. Conditioning on state $s_t$, the agent takes an action $a_t$ in an action space $\mathcal{A}$ according to its \emph{policy} $\pi: \mathcal{S} \times \mathcal{A} \mapsto [0,1]$, which defines a state-conditional distribution over actions. In response to the agent's action $a_t \sim \pi(\cdot|s_t)$, the MDP transitions to the next state, $s_{t+1}$ according to the transition function $P: \mathcal{S} \times \mathcal{A} \times \mathcal{S} \mapsto [0,1]$, which defines a distribution over next states conditioned on the current state and action, so that $s_{t+1} \sim P(\cdot|s_t, a_t)$. In \emph{episodic control}, $s_{t+1}$ can be a \emph{terminal state}, which upon arrival, ends the sequence of interactions, called an \emph{episode}. Terminal states are formally modeled as an absorbing state in the MDP, for which all transitions simply map back to the absorbing state. Importantly, the transition function is assumed to be \emph{Markovian}, whereby $(s_t, a_t)$ is a sufficient statistic for $s_{t+1}$. Taking action $a_t$ in state $s_t$ is accompanied by an associated reward $r_t$, based on the reward function $\mathcal{R}: \mathcal{S} \times \mathcal{A} \times \mathcal{S} \times \mathbb{R} \mapsto [0,1]$, which, in this general form, defines a distribution over real-valued rewards given $(s_t, a_t, s_{t+1})$. A \emph{transition} of the MDP at time $t$ typically refers to the tuple $(s_t, a_t, s_{t+1}, r_t)$, and the sequence of transitions up to time $t$, $\tau_{t} = (s_0, a_0, r_0, s_1, ..., s_{t-1}, a_{t-1}, r_{t-1}, s_t)$, is called a \emph{trajectory} at time $t$, or simply trajectory when clear from context. The Markov assumption then implies that trajectories are distributed according to $P(\tau_t) = \rho(s_0) \prod_{k=0}^{t-1} \pi(a_k|s_k)P(s_{k+1}|s_k,a_k)\mathcal{R}(r_k|s_k,a_k,s_{k+1})$.

All RL algorithms seek to learn an optimal policy $\pi^*$ that maximizes the expected \emph{total return}, defined in Equation~\ref{eq:rl_total_return}, by updating the agent's policy online, using information collected from repeated interactions with the environment over some countable number of time steps or time horizon $T$: 

\begin{equation}
    \label{eq:rl_total_return}
    R_0 = \underset{\substack{s_0 \sim \rho \\ \tau \sim \pi}}{\mathbb{E}} \left[ \sum_{t=0}^{T-1} \gamma^t r_t \right],
\end{equation}
\noindent where $\gamma < 1$ is the \emph{discount factor} and the expectation over $\tau \sim \pi$ means the rewards are based on transitions in trajectory $\tau$, sampled by taking actions according to $\pi$. This thesis focuses on the setting of \emph{episodic control}, where $T < \infty$, due to episodes ending at terminal states or upon exceeding $T$ time steps. The case of \emph{continuing control}~\citep{sutton2018reinforcement}, where $T = \infty$, typically formulates solutions in terms of maximizing Equation~\ref{eq:rl_total_return} with $\gamma < 1$ or, alternatively, in terms of average reward when a stationary distribution of the MDP exists, i.e. the limiting distribution over $\mathcal{S}$ as $t \rightarrow \infty$. Importantly, the discount factor makes the future-looking return well-defined when $T = \infty$ and, more generally, introduces a locality bias, such that near-term rewards are weighted more highly than more distant rewards. Reward discounting can also serve to reduce the variance of empirical return estimates used in RL algorithms, as it effectively shrinks the time horizon over which rewards are summed by reducing the contribution of more distant rewards. In practice, the specific setting of $\gamma$ can make a significant difference in the how successfully the agent learns from rewards it receives in the environment~\citep{jiang2015dependence, amit2020discount}. 

Taking these various components of an MDP into account, it is common to specify an MDP $\mathcal{M}$ by the tuple $\mathcal{M} = \left(\mathcal{S}, \mathcal{A}, P, \mathcal{R}, \rho, \gamma \right)$.

\medskip
\subsection{Partial Observability}
\label{subsec:pomdp}
In many real-world settings, the agent does not observe the full state $s_t$, but only some subset of the information in $s_t$. This setting of \emph{partial observability}~\citep{aastrom1965optimal,kaelbling1998planning} is modeled by extending the standard MDP tuple with an additional \emph{observation function} $O: \mathcal{S} \times \Omega \mapsto [0,1]$, which in general, defines a state-conditional distribution over the observation space $\Omega$. Then, rather than conditioning on $s_t$, the policy conditions on $o_t \sim O(\cdot|s_t)$, so that $\pi: \Omega \times \mathcal{A} \mapsto [0,1]$ and actions are sampled as $a_t \sim \pi(\cdot|o_t)$. Importantly, partial observability is a constraint specific to the agent, and therefore the transition and reward functions still condition on the full state as in a standard MDP. This extension of an MDP is called a \emph{partially-observable MDP} (POMDP) and can be succinctly represented by the tuple $\mathcal{M} = (\mathcal{S}, \mathcal{A}, \Omega, P, \mathcal{R}, O, \gamma)$. In this setting, it is often necessary to model a sufficient statistic for the optimal policy by aggregating information across the \emph{action-observation history} (AOH), $\tau_t = (o_0, a_0, o_1, ..., o_{t-1}, a_{t-1}, o_t)$, typically using a recurrent neural network~\citep[RNN,][]{hochreiter1997long,chung2014empirical}. 

\smallskip
\subsection{Multi-Agent Settings}
\label{subsec:posg}
While this thesis focuses on autocurricula for single-agent RL, the methods introduced in subsequent chapters model autocurricula themselves multi-agent systems consisting of student and teacher agents. Therefore, we provide the necessary formalism here for the multi-agent setting. The \emph{Partially-Observable Stochastic Game}~\citep[POSG,][]{bernstein2002complexity,hansen2004dynamic} extends the POMDP to the multi-agent setting by incorporating an index over $n > 0$ participating agents, each denoted as $A_i$ for $i$ in $\{1,...,n\}$. In general, each agent $A_i$ may have its own distinct action space $\mathcal{A}_i$ and observation space $\Omega_i$, resulting in a joint action space $\mathcal{A} = \times_{i} \mathcal{A}_i$ and joint observation space $\Omega = \times_i \Omega_i$. At each time step, all agents simultaneously sample an action from their policy $\pi_i$, producing a joint action $\boldsymbol{a}_t$, whose $i$-th component $\boldsymbol{a}_t^i$ corresponds to the action of agent $A_i$. Similar to a POMDP, the POSG transitions in response to $\boldsymbol{a}_t$ according to the transition function $P: \mathcal{S} \times \mathcal{A} \times \mathcal{S} \mapsto [0,1]$, emits the next observation according to the observation function $O: \mathcal{S} \times \Omega \mapsto [0,1]$, and emits a reward according to the reward function $\mathcal{R}: \mathcal{S} \times \mathcal{A} \times \mathcal{S} \times \mathbb{R}^{n} \mapsto [0,1]$. Here, the reward output at time $t$ is a vector $\boldsymbol{r_t}$ in $\mathbb{R}^n$, where the $i$-th component is the reward of agent $A_i$. The set of all agent policies is called the \emph{strategy profile}, denoted simply by $\pi$, and $\pi_i$ refers to the policy of agent $A_i$. As shorthand, the index $-i$ refers to all agents aside from agent $A_i$, e.g. $\pi_{-i}$ refers to the policies of all agents besides $A_i$ in the profile, and $\boldsymbol{a}_t^{-i}$ refers to the action of all other agents besides $A_i$ in the joint action. A POSG can thus be represented by the tuple $\mathcal{P} = (\mathcal{S}, \mathcal{A}, \Omega, P, \mathcal{R}, O, \gamma, n)$.

\medskip
\subsection{Underspecified Environments}
\label{subsec:upomdp}
Thus far, the RL settings discussed all assume a single environment instantiation, in the sense that the underlying POMDP or MDP is fixed across all interactions. In contrast, most real-world settings feature a large degree of variability across many aspects of the environment. Even within simulation, many virtual environments of interest are generated via \emph{procedural content generation}~\citep[PCG,][]{shaker2016procedural,risi_togelius_pcg}, which is the algorithmic generation of data. Moreover, curriculum learning methods seek to produce sequences of environment instances to facilitate learning, and thus necessarily assume the possibility of such variation in the environment across episodes. Otherwise, the environment is a \emph{singleton}, only existing in a single, fixed instantiation, and any curriculum learning method within the environment reduces to simply RL within this single environment. Crucially, curriculum learning methods require the ability to directly specify a particular instantiation of the environment. 

To model such control over the environment, the \emph{Underspecified POMDP}~\citep[UPOMDP,][]{dennis2020emergent} extends the standard POMDP with an additional space of \emph{free parameters} $\Theta$, such that specific instantiations of free parameters $\theta$ in $\Theta$ correspond to specific settings of aspects of the environment that can vary, e.g. positions of obstacles in a 2D maze environment. In its most general formulation, the UPOMDP assumes the specific values of the free parameters, $\theta$ may vary not only across episodes, but across time steps. The specific environment setting $\theta$ is then incorporated into the transition function, $P: \mathcal{S} \times \mathcal{A} \times{S} \times \Theta \mapsto [0,1]$. A UPOMDP thus corresponds to the tuple $(\mathcal{S}, \mathcal{A}, \Omega, P, \mathcal{R}, O, \gamma, \Theta)$. Likewise, the same modification of the POSG results in an Underspecified POSG (UPOSG). Such underspecified decision processes are simple yet powerful models, capable of representing nearly any virtual environment: Any virtual environment is definitionally generated by a program, in which case $\Theta$ corresponds to the set of environment variables modifiable by any underlying PCG algorithm used by the program to produce variation. Following common terminology, this thesis uses the term $level$ and $task$ as synonyms for a specific setting of the free parameters $\theta$. Importantly, the standard UPOMDP does not expose the value of $\theta$ in the observation, as it was first devised to study zero-shot to unknown $\theta$ at test time, i.e. without taking any gradient updates in the environment instance $\theta$. Of course, the UPOMDP can be easily modified to include $\theta$ in the observation, in which case, the resulting decision process becomes equivalent to what \citet{hallak2015contextual} call a \emph{contextual MDP}.

\subsection{Estimating Future Return}
\label{subsec:value_functions}
For a given environment (e.g. a specific MDP or POMDP), a policy $\pi$ induces a \emph{state value function}, which maps each state $s$ to the expected future discounted return obtained by sampling the rest of the trajectory using $\pi$, starting from $s$:
\begin{equation}
    \label{eq:v_function}
    V_{\pi}(s) = \mathbb{E}_{\tau \sim \pi}\left[\sum_{k=t}^{\infty} \gamma^{k-t}r_{k} \middle\vert s_t = s \right].
\end{equation}
\noindent The expected future discounted return from state $s$ under policy $\pi$ is often simply called the \emph{value} of state $s$ for policy $\pi$. 

A closely related entity is the \emph{state-action value function} or simply \emph{Q-function}, which maps every state-action pair $(s, a)$ to the discounted future return obtained by following $\pi$ after taking action $a$ in state $s$.
\begin{equation}
    \label{eq:q_function}
    Q_{\pi}(s,a) = \mathbb{E}_{\tau \sim \pi}\left[\sum_{k=t}^{\infty} \gamma^{k-t}r_{k} \middle\vert s_t = s, a_t = a \right].
\end{equation}
\noindent In other words, the Q-function measures the value of taking action $a$ in state $s$ for policy $\pi$. Subtracting the Q-function from the state-value function then yields the \emph{advantage function}, which maps each state-action pair $(s,a)$ to the expected improvement from taking action $a$ in state $s$ compared to the average performance of policy $\pi$ in state $s$:
\vspace{-2mm}
\begin{equation}
    \label{eq:advantage_function}
    A(s,a) =  Q_{\pi}(s,a) - V_{\pi}(s).
\end{equation}

\vspace{-4mm}
One approach to search for the optimal policy $\pi^*$ is to iteratively update the policy to take actions that maximize the advantage in each state. If the advantage cannot be improved in any state, then the policy must be optimal over all states~\citep{sutton2018reinforcement}. In practice, the future discounted return from each state must be approximated through Monte Carlo sampling, by \emph{rolling out}, i.e. stepping the policy through the environment, for some fixed number of time steps and computing the forward-looking returns from each visited state. Many environments feature high-dimensional state spaces that cannot be enumerated within typical memory budgets; thus, it is common practice to approximate these value functions with neural %

An important bias-variance trade-off appears when estimating the value function of a policy $\pi$, a procedure called \emph{policy evaluation}: Averaging over Monte Carlo rollouts of $\pi$ across full episodes in the MDP yields unbiased estimates of the return, and thus advantage---assuming a suitable function approximation of $V_{\pi}$. However, summing rewards over many time steps, each the result of a potentially stochastic transition, can result in high variance~\citep{kearns2000bias}. A common approach to trade variance for bias is to truncate the Monte Carlo rollout after $T$ steps and estimate the remaining future-looking return starting at $s_T$ with the current value function approximation $\hat{V}(s_T)$. Under this ``bootstrapping" approach, the forward-looking return at time $t$ is estimated as 
\begin{equation}
    \hat{R}_t = \left(\sum_{k=t}^{T-1} \gamma^{k-t}r_k  \right) + \gamma^{T-t}\hat{V}(s_T),
\end{equation}
\raggedbottom
\noindent resulting in a recursive loss function for training the value approximator:
\begin{equation}
    \label{eq:tstep_td_error}
    \mathcal{L}_V =  \left(\sum_{k=t}^{T-1} \gamma^{k-t}r_k  \right) + \gamma^{T-t}\hat{V}(s_T) - \hat{V}(s_t).
\end{equation}
\noindent In practice, one rollout can contribute multiple error terms to the value loss by computing a one-step bootstrap error for each time step, so that 
\begin{equation}
    \label{eq:one_step_td_error}
    \mathcal{L}_V = \sum_{k=0}^{T-1} r_k + \gamma \hat{V}(s_{k+1}) - \hat{V}(s_k),
\end{equation}
\noindent where each summand $\delta_k = r_k + V(s_{k+1}) - V(s_k)$ is called a \emph{temporal difference error} or \emph{TD error}. 
Equation~\ref{eq:tstep_td_error} then defines a $T$-step TD error. Importantly, the value function $V$ for any policy $\pi$ must satisfy the Bellman Equation~\citep{bellman1954theory}:
\begin{align}
    \label{eq:bellman_equation}
    V(s_{t}) &= \mathbb{E}_{\pi}\left[ r_t + \gamma V(s_{t+1})\right ]  \\
    &= \sum_{a_t,s_{t+1},r_{t}} \pi(a_t|s_t)P(s_{t+1}|s_t,a_t)\mathcal{R}(r_t|s_{t},a_t,s_{t+1}) + \gamma V(s_{t+1})
\end{align}
\noindent In fact, the value function for $\pi$ is provably the unique fixed point for the Bellman Equation, thereby guaranteeing the existence of a well-defined global optimum for the loss defined in Equation~\ref{eq:one_step_td_error}.

While reducing variance, the advantage estimates described in Equation~\ref{eq:tstep_td_error} based on the $T$-step TD error introduces bias via the final bootstrap term. \emph{Generalized Advantage Estimation}~\citep[GAE,][]{schulman2018high} provides a simple estimator that balances bias and variance in advantage estimation, based on an exponential average of all $T$-step TD errors for $T=1,...,\infty$. Importantly, the simple form of the GAE derives from the observation that this average can be written in terms of purely one-step TD errors across all time steps:
\begin{align}
    \hat{A}_t^{\text{GAE}(\lambda)} &= (1-\lambda)\left(\hat{A}_t^{(1)} + \lambda\hat{A}_t^{(1)} + \lambda^2\hat{A}_t^{(1)} + ...\right) \\
    &= \sum_{k=0}^{\infty} (\gamma\lambda)^{k} \delta_{t+k}, \label{eq:gae_expansion_one_step_td_errors}
\end{align}
\noindent where $\delta_t$ is the one-step TD error at time $t$, $\gamma$ is the discount factor, and $0 \leq \lambda \leq 1$ is the key GAE hyperparameter. When $\lambda = 0$, GAE reduces to $\delta_{t}$, the one-step TD error. When $\lambda = 1$, Equation~\ref{eq:gae_expansion_one_step_td_errors} reduces via a telescoping sum to become
\raggedbottom

\begin{equation}
    \sum_{k=0}^{\infty}\gamma^k \delta_{t+k} = \sum_{k=0}^{\infty} \gamma^k r_{t+k} - \hat{V}(s_t), \label{eq:gae_lambda_1}
\end{equation}

\noindent which is equivalent to the Monte Carlo advantage estimator. Thus, $\lambda$, the single hyperparameter of GAE, provides a knob for trading off between higher bias (e.g. $\lambda = 0$) and higher variance ($\lambda = 1$).

The next section provides a detailed description of RL methods that seek to maximize the total return by making use of learned approximations of the state value function and advantage function.

\section{Policy Gradient Methods}
\label{sec:policy_gradients}
The methods developed in this thesis are tested primarily in combination with a class of RL algorithms known as \emph{policy gradient methods}~\citep{sutton1999policy,silver2014deterministic,schulman2018high}.~\footnote{In principle, the curriculum methods developed in this thesis are agnostic to the underlying RL algorithm, which may also be value-based.} In contrast to \emph{value-based} RL methods~\citep{watkins1992q,mnih2015human}, which learn the optimal policy by way of learning the optimal action-value function, policy gradient methods perform stochastic gradient descent directly over the weights of the policy network to optimize a noisy estimate of the discounted future return, assuming some distribution over starting states. 

\subsection{From REINFORCE to Actor-Critic}
\label{subsec:ac}
The first, and perhaps simplest, policy gradient method is REINFORCE~\citep{williams1992simple}, which estimates the gradient of the expected discounted return with respect to the policy weights as

\begin{equation}
    \nabla_{\theta}J(\theta) \propto
    \label{eq:reinforce}
    \underset{\substack{s_{0:\infty},\\a_{0:\infty \sim {\pi_{\theta}}}}}{\mathbb{E}} \big{[}\sum_{t=0}^{\infty} R_t\nabla_{\theta} \log \pi_{\theta}(a_t | s_t)\big{]}.
\end{equation}

\noindent The REINFORCE estimator effectively restates the Policy Gradient Theorem~\citep{sutton1999policy}: The gradient of the expected discounted return of the discounted return $J(\theta)$ with respect to the policy parameters $\theta$ is equal to Equation \ref{eq:reinforce}. Remarkably, this expectation is independent of the ergodic state distribution of $\pi$ within the MDP, making it tractable to estimate its value through Monte Carlo rollouts of the policy, as done by REINFORCE.

An important property of the REINFORCE estimator is that it remains unbiased in the presence of any baseline term $b_t$ that is a function of values occurring along the trajectory up to time $t$. Actor-critic methods~\citep{sutton2018reinforcement,degris2012model,peters2008natural,mnih2016asynchronous,espeholt18a} exploit this fact to reduce the variance of REINFORCE, by subtracting such a baseline from the forward-facing return at each time $t$ as follows
\begin{equation}
    \label{eq:actor_critic}
    \nabla_{\theta}J(\theta) \propto \underset{\substack{s_{0:\infty},\\a_{0:\infty \sim {\pi_{\theta}}}}}{\mathbb{E}} \big{[}\sum_{t=0}^{\infty} (R_t - b_t)\nabla_{\theta} \log \pi_{\theta}(a_t | s_t)\big{]}.
\end{equation}
\noindent The baseline is typically implemented as a neural network, $\hat{V}: \mathcal{S} \mapsto \mathbb{R}$, that is trained to predict $V(s_t)$. This network is typically called the \emph{value network} or the ``critic" (whereas the policy is dubbed the ``actor"). When the baseline is a value network, the difference between the future discounted return $R_t$ and $b_t = \hat{V}(s_t)$ is an unbiased estimator of the advantage $A(s_t, a_t)$. Intuitively, updating the policy weights along the direction of the gradient defined in Equation \ref{eq:actor_critic} increases the probability of taking actions that are better than average in terms of expected future discounted return. In practice, both the expected discounted return and baseline terms within the advantage must be estimated from empirical returns. Rollout data is typically collected over a fixed horizon during training, independent of whether an episode terminates. Therefore, the discounted return $R_t$ is approximated via a \emph{bootstrapped value estimate}, such that $R_t \approx G_t = \sum_{t=0}^{T-1} r_t + \hat{V}(s_T)$. The value network is trained alongside the policy, by minimizing the L2 loss: 
\begin{equation}
    \mathcal{L}_V = \frac{1}{T}\sum_{t=0}^{T-1} \left(G_t - \hat{V}(s_t)\right)^2.
\end{equation}
A downside of the L2 loss is that its gradient magnitude increases linearly with that of the loss, which can lead to more unstable training~\citep{mnih2015human}. To address this issue, the Huber loss~\citep{huber1992robust}, shown in Equation \ref{eq:huber_loss}), defines a smooth, piecewise function that replaces the quadratic loss with an absolute value for inputs beyond a threshold magnitude $\sigma$:
\begin{equation}
    \label{eq:huber_loss}
    \mathcal{L}_{\text{Huber}}(\Delta) = 
    \begin{cases}
        \frac{1}{2}|\Delta|^2 & \text{if } |\Delta| \leq \sigma, \\
        \sigma \cdot (|\Delta| - \frac{1}{2}\sigma) & \text{otherwise}.
    \end{cases}
\end{equation}
\vspace{-4mm}

The corresponding loss function giving rise to Equation \ref{eq:actor_critic}, up to a scaling factor (that is absorbed into the learning rate), can then be written as

\begin{equation}
    \label{eq:ac_loss}
    \mathcal{L_{\text{AC}}} = -\frac{1}{T}\underset{\substack{s_{0:T}, \\a_{0:T \sim {\pi_{\theta}}}}}{\mathbb{E}} \big{[}\sum_{t=0}^{T-1} (G_t - b_t) \log \pi_{\theta}(a_t | s_t)\big{]}.
\end{equation}

Additionally, it is common to include an \emph{entropy regularization} term in the total loss when training deep RL networks~\citep{williams1991function,mnih2016asynchronous}, as shown in Equation~\ref{eq:entropy_loss}, where $\mathcal{H}$ denotes the Shannon entropy. 
\begin{equation}
    \label{eq:entropy_loss}
    \mathcal{L_{\mathcal{H}}} = -\frac{1}{T} \sum_{t=0}^{T-1} \mathcal{H}(\pi(a_t|s_t)).
\end{equation}
\noindent By promoting higher entropy in the policy distribution over the action space, this term can encourage the agent to explore a greater portion of the state space~\citep{schulman2017equivalence, ahmed2019understanding}, and, in some environments, lead to improved sample efficiency and robustness when transferring to perturbations of the MDP used for training~\citep{eysenbach2021maximum}. Adding this final term to the total loss function, we obtain the standard policy gradient loss used in actor-critic algorithms:
\begin{equation}
\label{eq:total_ac_loss}
    \mathcal{L} = -\mathcal{\mathcal{L}_{\text{AC}}} + \lambda_V \mathcal{L}_V - \lambda_{\mathcal{H}} \mathcal{L}_{\mathcal{H}},
\end{equation}
\noindent where $\lambda_V$ and $\lambda_{\mathcal{H}}$ are weighted coefficients, typically set via hyperparameter tuning. This formulation, in which the advantage is estimated via bootstrapped return estimates $G_t$, is commonly called \emph{Advantage Actor-Critic} (A2C).

\subsection{Proximal Policy Optimization}
\label{subsec:ppo}
Training models with A2C can be unstable and sample inefficient, requiring many transitions to reach a useful policy. One source of instability in A2C is its sensitivity to the step size taken along the gradient. Too a large step size can cause the policy to stray into suboptimal behaviors that then self-reinforce, as in RL, the model trains on its own transitions~\citep{kakade2002approximately}. \emph{Trust region} methods enforce a constraint on the policy update step, such that the updated policy cannot deviate too far from the current policy, and when appropriately constrained in this way, provably results in monotonic policy improvement~\citep{schulman2015trust}. This optimization can be written as
\vspace{-2mm}
\begin{align}
    \label{eq:trust_region_objective}
    \underset{\theta}{\text{maximize}}\; & \mathbb{E}_t \left[ \frac{\pi(a_t|s_t)}{\pi_{\text{old}}(a_t | s_t)} A(s_t, a_t) \right], \\
    \text{subject to}\; & D_{\text{KL}}(\pi_{\text{old}} || \pi) \leq \delta. \notag
\end{align}
\noindent Here, the expectation is an importance sampling estimator and $\pi_{\text{old}}$ denotes the current iterate of the policy, which collects transitions for the next update.

\emph{Proximal Policy Optimization}~\citep[PPO,][]{schulman2017ppo} approximates the trust-region constraint via a simple first-order update based on maximizing the following ``clipped" objective:
\begin{align}
    \label{eq:ppo_objective}
    J_{\text{clip}}(\theta) &= \mathbb{E}_{t}\big{[}\min(\rho_t(\theta)\hat{A}_t, \text{clip}(\rho_t(\theta), 1-\epsilon, 1+\epsilon)\hat{A}_t)\big{]} \\
    &= \mathbb{E}_{t}\big{[}\min(\rho_t(\theta)\hat{A}_t, g(\epsilon, \hat{A}_t)\big{]},
\end{align}

\noindent where $\rho_t$ denotes the importance sampling ratio, $\pi_{\theta}(a_t|s_t)/\pi_{\text{old}}(a_t|s_t)$, and $\epsilon > 0$ is the \emph{clipping constant}. This objective can be understood by observing how $g$ behaves for positive and negative advantages, $\hat{A}_t$~\citep{achiam2018spinning}: 

\begin{equation}
    g(\epsilon, \hat{A}_t) = \\
    \begin{cases}
        \hat{A}_t \cdot \min \left(\frac{\pi(a_t |s_t)}{\pi_{\text{old}}(a|s_t)}, 1 + \epsilon \right) & \text{if } \hat{A}_t > 0, \\
        \hat{A}_t \cdot \max \left(\frac{\pi(a_t |s_t)}{\pi_{\text{old}}(a_t |s_t)}, 1 - \epsilon \right)  & \text{if } \hat{A}_t < 0, \\
        0 & \text{otherwise}.
    \end{cases}
\end{equation}

\noindent When $\hat{A}_t > 0$, the action $a_t$ is better than average when taken in $s_t$. As we expect, the objective increases as $\pi(a_t | s_t)$ becomes more likely, but only up to a maximum amount of $(1+\epsilon)\pi_{\text{old}}\cdot(a_t | s_t)$. Likewise, when $\hat{A}_t < 0$, the action $a_t$ is worse than average when taken in $s_t$, and the objective decreases as $\pi(a_t | s_t)$ becomes more likely---but only up to a limit, $(1-\epsilon)\cdot\pi_{\text{old}}(a_t|s_t)$. The clipping of $\rho_t$ thus heuristically approximates the trust region constraint by limiting how much large changes in the policy can contribute to increasing the objective~\citep{achiam2018spinning}. PPO is most commonly implemented by replacing the A2C loss term $-\mathcal{L}_{\text{AC}}$ in Equation~\ref{eq:total_ac_loss} with $-J_{\text{clip}}(\theta)$.

Empirically, this clipped objective allows for PPO to stably take multiple gradient updates over a given batch of transitions collected by $\pi_{\text{old}}$, enabling improved sample-efficiency through greater data reuse per batch. In practice, PPO performs multiple gradient updates per batch by subsampling minibatches of data without replacement, typically over multiple iterations through the dataset. In contrast, A2C and other prior policy gradient methods take only a single gradient update per batch. For its relative simplicity and strong performance across many domains, the autocurricula experiments in this thesis make use of PPO as the base RL optimization algorithm. 

\medskip
\subsection{Independent PPO}
\label{subsec:ippo}
A particularly simple formulation of a multi-agent POSG parameterizes each participating agent with its own, independent set of parameters. 
In this setting, PPO can simply be applied to each agent's parameters without any additional algorithmic modifications. Here, each agent's PPO update makes use of the transitions collected in the agent's last rollout in the POSG, and all agents are updated simultaneously after each rollout. Importantly, each agent only updates using its own experiences (e.g. observations). This instantiation of PPO is referred to as \emph{Independent PPO}~\citep[IPPO,][]{de2020independent}. In later chapters, such application of PPO is implied whenever PPO is stated as the RL algorithm used to optimize multiple RL agents in POSGs that model autocurricula between student and teacher agents.

\section{Nash Equilibria}
\label{sec:ne}
In multi-agent settings, like an autocurriculum unfolding between a student and teacher, each agent must strategically adapt in response to the other agents' actions. Each agent's optimal policy then depends on the policy implemented by all other agents, making an exact definition of an ``optimal policy" nontrivial to specify in advance. A common solution concept that defines a practical notion of optimal behavior for the multi-agent setting is the \emph{Nash equilibrium}~\citep[NE,][]{nash1950equilibrium}, which refers to any policy profile $\pi^*$ such that each agent $i$ cannot obtain higher total return by unilaterally deviating from $\pi^*$:
\begin{equation}
    \label{eq:nash_equilibrium}
    J^i(\pi^*_{i}, \pi^*_{-i}) \geq J^i(\pi_{i}, \pi^*_{-i})\; \forall \pi_i \in \Pi,
\end{equation}
where $J^i(\pi_i, \pi_{-i})$ is the total return obtained by agent $A_i$ when following policy $\pi_i$ and all other agents follow $\pi_{-i}$. Here, $\Pi$ is the space of policies. An important result from game theory is the Minimax Theorem~\citep{vonneumman1928theorie}, which states that in two-player zero sum (2p0s) games---which define a strictly competitive setting where the episodic returns of both agents always sums to 0---there always exists at least one NE. Moreover, all such NE are \emph{interchangeable}, so that for any two Nash profiles $\pi^{(1)}$ and $\pi^{(2)}$,
\begin{equation}
    J^i(\pi^{(1)}_i, \pi^{(1)}_{-i}) = J^i(\pi_i^{(1)}, \pi^{(2)}_{-i}).
\end{equation}
\noindent The autocurricula methods developed in this thesis directly exploit the existence and interchangeability of such equilibria in 2p0s games (between student and teacher agents), alongside the definition of NE in Equation~\ref{eq:nash_equilibrium}, in order to devise training algorithms that provably induce certain useful properties in the participating policies at NE.

\raggedcolumns
\newpage
\section{Decision Making Under Uncertainty}
\label{sec:decision_theory}
This thesis presents methods aiming to produce more \emph{robust} agents, but how is robustness defined?
At a high level, \emph{robustness} refers to the degree to which a policy, trained on some distribution $P_{\text{train}}(\Theta)$, can maintain its performance, on a test distribution $P_{\text{test}}(\Theta)$, according to some measure of success. The act of deploying a model on a distribution of data differing from its training distribution is called \emph{transfer}, and the set of methods seeking to train a model to succeed in transfer is called \emph{transfer learning}. When $P_{\text{test}}$ is known a priori, the training routine can incorporate this information to ensure some degree of performance on the test distribution. However, often, there is little to no knowledge of $P_{\text{test}}$ available at training. The methods developed in this thesis thus make the more general assumption in which $P_{\text{test}}$ is not known in advance.

\begin{table}[h!]
\centering
\begin{minipage}{.48\linewidth}
\centering
\begin{tabular}{ l | c | c | c  }
 & Cold & Warm & Hot \\
\hline
Small & 250 & 200 & 150  \\
\hline
Medium & 200 & 500 & 500  \\
\hline
Large & 100 & 300 & 750
\end{tabular}
\end{minipage}
\begin{minipage}{.48\linewidth}
\centering
\begin{tabular}{ l | c | c  }
 & EU & MR \\
\hline
Small & 200 & 600  \\
\hline
Medium & \textbf{400} & 250 \\
\hline
Large & 383 & \textbf{200}
\end{tabular}
\end{minipage}
\caption{\small{Left: A simple decision matrix showing the dollar profits for an ice cream vendor's choice of order purchase size depending on if the weather turns out to be cold, warm, or hot. Right: The expected utility (EU) and maximum regret (MR) of each action, with the optimal action value for each criterion in bold.}}
\label{table:decision_matrix_minimax_regret_example}
\end{table}

Decision theory provides a firm foundation on which to develop methods for robust transfer. In general, \emph{decision theory} studies how one can make choices to maximize some utility function (akin to the total return in RL) assuming some information about the world. The typical model of decision making employed by decision theory is the \emph{decision matrix} (see Table~\ref{table:decision_matrix_minimax_regret_example} for an example), whose rows correspond to the available actions that can be taken and whose columns, the possible outcomes corresponding to different states of the world. These possible outcomes can be known when deciding or only revealed after the fact. This model of decision making can be connected to RL by viewing the outcome columns as corresponding to different environment instances, i.e. specific values of $\theta \in \Theta$ in a UPOMDP. Standard RL training assumes some fixed distribution, $P_{\text{train}}(\Theta)$, and seeks the  policy $\pi^*$ maximizing the expected total return under $P_{\text{train}}(\Theta)$. The closer $P_{\text{train}}(\Theta)$ is to $P_{\text{test}}(\Theta)$, the stronger the expected transfer performance of $\pi^*$. The problem setting of \emph{decision-making under risk} corresponds to this problem setting in which decision-making is accompanied by a distribution over the outcomes, and the optimal \emph{decision rule} corresponds to maximizing the expected utility (or total return in the case of RL).

In contrast, the transfer problems considered in this thesis correspond to a problem setting called \emph{decision-making under ignorance}, which assumes no known distribution over outcomes, i.e. the RL agent has zero prior knowledge about which $\theta$ corresponds to the test environment instance in which its transfer performance is evaluated. Several decision rules have been considered in this setting (see \citet{peterson_2009} for a detailed discussion). An especially simple rule for decision making under ignorance is the \emph{Principle of Insufficient Reason}~\citep{marquis1825essai}, which simply transforms the decision problem into one of decision making under risk by assuming a uniform distribution over all outcomes. This rule is obviously nonideal in that it may assign probabilities to outcomes that rarely or never occur. Another simple rule is the \emph{maximin} rule, which chooses the action with the highest minimum utility across all outcomes~\citep{Wald1951statistical}. By optimizing for the worst case outcome, this rule tends to result in overly conservative behaviors, making it nonideal in many situations. Instead, the methods in this thesis build upon the \emph{minimax regret} decision rule~\citep{savage1951theory}, which seeks to make decisions that minimize the worst-case regret over all possible outcomes. For a specific outcome, \emph{regret} refers to the difference between the utility obtained in choosing the optimal action under that outcome and the action chosen by the agent. In terms of RL, given a specific environment instance $\theta$ (that is, the outcome), where the optimal policy is $\pi^*_{\theta}$, the regret of policy $\pi$ on $\theta$ is equal to 
\begin{equation}
\textsc{Regret}(\pi, \theta) = J_{\theta}(\pi^*_{\theta}) - J_{\theta}(\pi).   
\end{equation}

The minimax regret policy $\pi'$ over some space of environment instances $\Theta$ is then equal to:
\begin{equation}
    \label{eq:minimax_regret_policy}
   \pi' = \underset{\pi}{\min}\;\underset{\theta}{\max}\;J_{\theta}(\pi^*_{\theta}) - J_{\theta}(\pi).
\end{equation}

\noindent Section~\ref{subsec:ued} describes how the RL problem can be reframed as a competitive POSG, such that the agent implements the minimax regret policy defined in Equation~\ref{eq:minimax_regret_policy} at the NE of this game.

\section{Automatic Curricula}
\label{sec:autocurricula}
Many problem domains, such as those modeled by UPOMDPs, feature environment instances of varying difficulty, each determined by a specific setting of free parameters $\theta \in \Theta$. A naive way to train an agent over this space of tasks is \emph{domain randomization} (DR), which simply samples $\theta \sim P_{\text{train}}(\Theta)$, where $P_{\text{train}}(\Theta)$ is the corresponding distribution of $\Theta$ induced by sampling the simulator or an equivalent physical process for resetting the task to different configurations. DR can often be a strong baseline approach, but in practice, can result in suboptimal policies: The distribution $P_{\text{train}}(\Theta)$ can be arbitrary, subject to the quirks of the underlying environment generation algorithm, and tasks especially useful for learning may be sampled only rarely or not at all.

\emph{Curriculum learning} (CL) seeks to improve the learning dynamics of RL agents when training in such environments, by sequencing specific environment instances across the course of training, such that the agent always trains on environment instances for which it is likely to make the most learning progress, e.g. in terms of improvement in total return. The most rudimentary form of CL defines some segmentation over environment instances according to an externally-provided difficulty metric, e.g. the distance to the goal position in a goal-navigation environment, and such curricula can both expedite the agent's learning of useful behaviors and improve the agent's robustness in environment instances held-out during training \citep{justesen2018illuminating}. However, such notions of difficulty rely on domain knowledge that is generally not available in all cases. Moreover, manually specifying such a metric does not easily scale to more complex environments with potentially many interacting axes of difficulty. What would be the correct way to manually sequence a curriculum over possible environments in a simulation of a robot walking over varying terrain?

\subsection{Automatic Curriculum Learning}

\emph{Automatic Curriculum Learning} (ACL) methods selectively sample environment instances during training in order to maximize the agent's performance on some target distribution of environments \citep{portelas2020automatic}. In ACL algorithms, a teacher module proposes each training task to a student---the primary RL agent that is the focus of training. Typically, ACL methods prioritize sampling of environment instances where the agent achieves higher \emph{learning progress}, as measured by some proxy metric. For example, Teacher-Student Curriculum Learning~\citep[TSCL,][]{tscl}, upweighs the probabilities of sampling tasks based on the magnitude of the linear regression slope over total returns obtained across a recent window of episodes of that task. Similarly, ALP-GMM~\citep{portelas2020alpgmm} fits a Gaussian Mixture Model over the free parameters of the environment and uses the Exp4 bandit algorithm to sample Gaussian components that maximize the \emph{absolute learning progress} (ALP) metric, defined as $|r_{\text{new}} - r_{\text{old}}|$, where $r_{\text{new}}$ is the total return obtained on a newly sampled instance $\theta_{\text{new}}$ and $r_{\text{old}}$ is the most recent total return obtained on the nearest instance previously sampled (within a window of the last $N > 0$ parameter-ALP pairs). The GMM over $\Theta$ is periodically refit over the most recent parameter-ALP pairs. ACL methods can improve the sample-efficiency and final target task performance compared to naive random sampling of the task parameters. While ACL methods relax the assumption of an external notion of task difficulty, they assume prior knowledge of a target task distribution of interest. For example, TSCL and ALP-GMM both directly operate over a predefined target task distribution $P_{\text{train}}$, with the goal of training policies that perform well specifically on $P_{\text{train}}$.

\subsection{Unsupervised Environment Design}
\label{subsec:ued}

Rather than assume a set of target tasks known at training, \emph{Unsupervised Environment Design} (UED) \citep{dennis2020emergent} requires only the specification of a task space, i.e. $\Theta$ in the UPOMDP formalism. As there is no specific target task distribution, UED methods are then evaluated based on the performance of the trained policy on a wide range of task distributions over some free parameter space $\Theta' \supseteq \Theta$, which can include environment instances that are out-of-distribution (OOD) with respect to any that might be sampled in the training set in terms of certain properties of the environment that can vary across different values of $\theta$. For example, in a maze domain, $\Theta'$ might include mazes that are larger or that feature denser configurations of obstacles than maze instances in $\Theta$. Generalization to such OOD environments is still possible when they share a common observation space and environment dynamics (in terms of transitions and rewards) with those environment instances in $\Theta$. 

Like in ACL methods, UED methods typically include a teacher and a student. During training, the teacher agent proposes environment instances that the student must master. However, unlike ACL methods, UED assumes the absence of any specific target task distribution, making it unreasonable to directly maximize task performance or learning progress on $P_{\text{train}}$. Rather, UED seeks to directly maximize the student's robustness over any possible distribution of environments in $\Theta$---an objective independent of any specific $P_{\text{train}}$. This thesis focuses on UED methods that seek to produce policies that are robust in the sense of being minimax-regret optimal, i.e. that satisfy Equation~\ref{eq:minimax_regret_policy}. Such UED methods reduce the problem of searching for this minimax regret optimal policy to one of searching for the NE of a 2p0s game between the teacher and student. In this game, the payoff to the teacher for each proposed task instance $\theta$ is the regret incurred by the student on $\theta$. Assuming there is a clear definition of task success, the student must provably follow a minimax regret policy that solves all solvable environment instances at NE \citep{dennis2020emergent}.\footnote{One extra benefit of this arrangement is that regret-maximizing teachers are incentivized to avoid proposing impossible tasks, whose regret is always 0---thereby avoiding a degeneracy of maximin UED in which the teacher can optimize its minimax objective by proposing only impossible levels.} The teacher in minimax-regret UED methods then produces an autocurricula of adversarial tasks for the student as this 2p0s game unfolds.

A general method for computing the student's true regret for a task $\theta$ requires knowledge of the optimal policy for $\theta$. In practice, UED makes use of a \emph{regret estimator} to approximate the true regret. \citet{dennis2020emergent} introduces Protagonist Antagonist Induced Environment Design (PAIRED), which expands the 2p0s between teacher and student into a 3-player game, between the student, called the protagonist, and a teacher-antagonist team, where the antagonist is a second student. The PAIRED teacher $\pi^T$ seeks to propose tasks maximizing the \emph{relative regret}, which is the difference in expected total return obtained between the protagonist and antagonist policies, $\pi^A$ and $\pi^P$ respectively:
\begin{equation}
    \textsc{Regret}(\pi, \theta) \approx J_{\theta}(\pi^A) - J_{\theta}(\pi^P).
\end{equation}
As $J(\pi^*_{\theta}, \theta) \geq J(\pi, \theta)$ for any policy $\pi$, the relative regret defines a lower bound on the true regret. As the teacher maximizes the relative regret and the two students reduce their individual regrets in each task by performing RL, the 3-player PAIRED game approximates the original 2p0s game, in which the teacher's payoff is the student's true regret.\footnote{Technically, there exist NE of this 3-player game that differ from the 2p0s game with a regret-maximizing teacher, e.g. if both students perfectly solve some task $\theta$ and $\pi^T$ collapses to only proposing $\theta$. In practice, randomness in student agent initializations and injecting noise into the environment design process appear to alleviate this issue.}

The methods developed in this thesis offer new approaches to minimax-regret UED that significantly improve over the PAIRED algorithm, including contributing several, more computationally-efficient regret estimators that require only a single student to estimate.

One special case of UED is \emph{domain randomization}~\citep[DR,][]{evolutionary_dr,peng2018sim,domain_randomization}, which simply samples environment instances at random, e.g. according to a uniform distribution over the set of possible instances or some other arbitrary distribution. If the distribution is uniform, DR can be viewed as UED with a constant objective function (and similarly, in the case of an arbitrary distribution, as a suitably weighted objective corresponding to this distribution). DR has proven useful in improving the robustness of policies for sim2real transfer in robotics domains~\citep{james2017transferring,akkaya2019rubiks,haarnoja2023learning,margolis2023walk}. However, since in general, its underlying distribution is arbitrary, the resulting robustness of policies trained with DR may be hard to anticipate, and DR may sample useful instances for learning only rarely or not at all.  

\subsection{Connection to Intrinsic Motivation}

A common class of exploration methods in deep RL is \emph{intrinsic motivation}~\citep[IM,][]{chentanez2004intrinsically}. These methods introduce an \emph{intrinsic reward} function that is separate from the task-specific or \emph{extrinsic reward} function. The intrinsic reward for a transition is typically based on some measure of the transition's novelty, e.g. giving a higher reward for arriving in less frequently visited states~\citep{strehl2008analysis, bellemare2016unifying,ecoffet2019goexplore}, states where a concurrently-trained predictive model of state properties sees high error~\citep{burda2019exploration, pathak2017curiosity,raileanu2020ride}, or states where an ensemble thereof shows high disagreement~\citep{pathak2019self}. During training, the agent then maximizes a total return based on a weighted sum of extrinsic and intrinsic rewards. Typically, as the same state is visited multiple times, its associated intrinsic reward tends to zero; thus, in the limit of exploring all states, the optimization converges to maximizing the total return for the task-specific reward. These methods can be seen as inducing autocurricula over informative trajectories within an environment instance.

Intrinsic rewards encourage the agent to take actions that lead the way to novel parts of the environment, which can hold higher learning potential for the agent. Autocurricula make use of similar objectives to assess the value of training on each environment instance, and thus can be viewed as a form of IM for guiding exploration over the space of environment or task instances. Both classes of methods ultimately seek to find states that lead to the greatest learning potential for the agent. In IM methods, this search is conducted by a learning agent directly situated within the current environment, while in autocurricula, an external process (e.g. a UED teacher) conducts this search over a space of environments. However, the reset-based paradigm for exploration introduced by Go-Explore~\citep{ecoffet2019goexplore} blurs this distinction by directly resetting the simulator state to the most promising states for further exploration, rather than have the policy return to them by maximizing an intrinsic return. If we view each possible reset state as defining a different environment instance, then Go-Explore effectively induces an autocurriculum over these states (of a single environment) while continuing episodic exploration across this set of state-defined environment instances. Moreso, for any set of MDPs (each an environment instance), we can construct a new MDP that encompasses them all, by introducing additional controllable states that determine which included MDP the new MDP should behave as. In this new MDP, either the learning agent or an external autocurriculum can drive exploration within a single environment instance or across the set of environment instances encompassed. These perspectives suggest which aspects of the exploration process are driven by a situated learning agent (IM) and which, by an external process (autocurriculum) is a rich design space with much room for negotiation. Methods that harness the interplay between IM and autocurricula thus form a promising frontier for further research.

\chapter{Prioritized Level Replay}
\label{chapter:plr}

\section{Introduction}

We begin our journey toward increasingly powerful autocurriculum methods by studying the impact of a family of conceptually-simple prioritized sampling algorithms in procedurally-generated environments. These empirical investigations inform the development of a conceptually simple method called \emph{Prioritized Level Replay} (PLR), which effectively and scalably addresses the fundamental challenges of learning generalizable behaviors offered in such environments---a challenge that traditional deep RL methods struggle to overcome.

Deep reinforcement learning (RL) easily overfits to training experiences, making generalization a key open challenge to widespread deployment of these methods. Environments making use of \emph{procedural content generation} (PCG) have emerged as a promising class of problems with which to probe and address this core weakness \citep{Risi_2020, gym_minigrid, cobbe2019procgen, Juliani_2019, DBLP:conf/iclr/ZhongRG20, kttler2020nethack}. Unlike singleton environments, like the Arcade Learning Environment games~\citep{Bellemare_2013}, PCG environments take on algorithmically created configurations at the start of each training episode, potentially varying the layout, asset appearances, or even game dynamics. Throughout this thesis, we will refer to each environment instance generated this way as a \emph{level} or, synonymously, a \emph{task}. By mapping an identifier, such as a random seed, to each unique level, PCG environments allow us to measure a policy's generalization to held-out test levels. In this work, we assume only a blackbox generation process that returns a level given only such an identifier. We avoid the strong assumption of control over the generation procedure itself, explored in subsequent chapters as well as prior works (see Section \ref{section:plr_related_work}). PLR's minimal requirements in terms of environment generation allow for a nearly universal scope of application in PCG settings. More direct control of environment generation, of course, can enable more targeted autocurricula, which can lead to improved agent performance. We will explore such methods in detail in Chapters~\ref{chapter:dcd}–\ref{chapter:samplr}, where we will find that they too can benefit from the simple replay mechanism of PLR. Importantly, for the environments we consider in this chapter,  we assume a common set of states and dynamics underly each level, so that in aggregate, experiences collected in individual levels reveal general rules governing the entire set of levels.

Despite its humble origin in games, the PCG abstraction proves general and far-reaching: Most if not all control problems, such as teaching a robotic arm to stack blocks in a specific formation, directly conform to PCG. Here, each level may consist of a unique combination of initial block positions, arm state, and target formation. In a vastly different domain, Hanabi \citep{bard2020hanabi} also conforms to PCG, where levels map to initial deck orderings. These examples illustrate the general applicability of the PCG abstraction: Many if not most useful RL problems entail generalizing across instances (or levels) differing along some underlying factors of variation and thereby can be aptly framed as PCG. The ubiquity of PCG makes developing effective methods for PCG environments a critical undertaking for the real-world viability of deep RL.

\begin{figure*}[t!]
    \centering
    \includegraphics[width=1.0\linewidth]{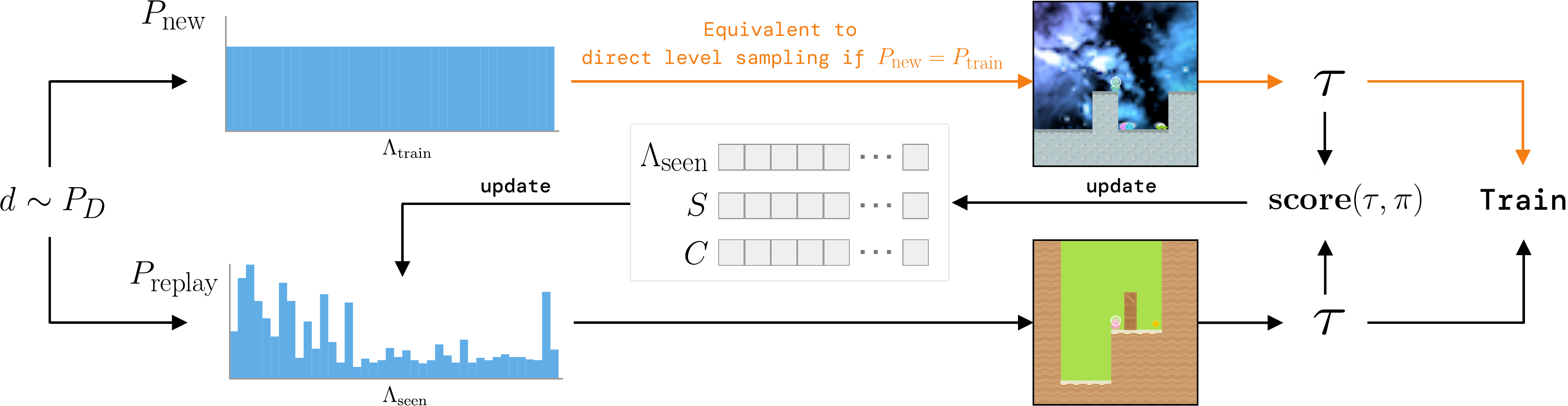}
    \caption{\small{Overview of Prioritized Level Replay. The next level is either sampled from a distribution with support over unseen levels (top), which could be the environment's (perhaps implicit) full training-level distribution, or alternatively, sampled from the replay distribution, which prioritizes levels based on future learning potential (bottom). In either case, a trajectory $\tau$ is sampled from the next level and used to update the replay distribution. This update depends on the lists of previously seen levels  $\Lambda_{\text{seen}}$, their latest estimated learning potentials $S$, and last sampled timestamps $C$.}}
    \label{fig:algo_fig}
\end{figure*}

Many techniques have been proposed to improve generalization in the PCG setting (see Section \ref{section:plr_related_work}), requiring changes to the model architecture, learning algorithm, observation space, or environment structure. Notably, these approaches default to uniform sampling of training levels. We instead hypothesize that the variation across levels implies that at each point of training, each level likely holds different potential for an agent to learn about the structure shared across levels to improve generalization. Inspired by this insight and selective sampling methods from active learning, we investigate whether sampling the next training level weighed by its learning potential can improve generalization.

We then introduce Prioritized Level Replay (PLR), illustrated in Figure~\ref{fig:algo_fig}, a method for sampling training levels that exploits the differences in learning potential among levels to improve both sample efficiency and generalization. PLR selectively samples the next training level based on an estimated learning potential of replaying each level anew. 
During training, our method updates scores estimating each level's learning potential as a function of the agent's policy and temporal-difference (TD) errors collected along the last trajectory sampled on that level. 
Our method then samples the next training level from a distribution derived from a normalization procedure over these level scores. PLR makes no assumptions about how the policy is updated, so it can be used in tandem with any RL algorithm and combined with many other methods such as data augmentation. 
Our method also does not assume any external, predefined ordering of levels by difficulty or other criteria, but instead derives level scores dynamically during training based on how the policy interacts with the environment. This allows PLR to be easily used with nearly any PCG simulator. The only requirements are as follows---satisfied by almost any problem that can be framed as PCG, including RL environments implemented as seeded simulators: (i) Some notion of ``level'' exists, such that levels follow common latent dynamics; (ii) such levels can be sampled from the environment in an identifiable way; and (iii) given a level identifier, the environment can be set to that level to collect new experiences from it.

While previous works in off-policy RL devised effective methods to directly reuse \emph{past} experiences for training \citep{schaul2015prioritized, andrychowicz2017hindsight}, PLR uses past experiences to inform the collection of \emph{future} experiences by estimating how much replaying each level anew will benefit learning. Our method can thus be seen as a forward-view variation of prioritized experience replay, and an online counterpart to this off-policy method.

In summary, this chapter presents the following contributions\footnote{PLR is open sourced at \url{https://github.com/facebookresearch/level-replay}.}: 
(i) we present a computationally-efficient algorithm for sampling levels during training based on an estimate of the future learning potential of collecting new experiences from each level;
(ii) we show our method significantly improves generalization on 10 of 16 environments in Procgen Benchmark and two challenging MiniGrid environments; (iii) we demonstrate our method combines with the previous leading method to set a new state-of-the-art on Procgen Benchmark; and (iv) we show our method induces an implicit curriculum over training levels in sparse reward settings.

\section{Background}
We refer to a \emph{PCG environment} as any computational process that, given a level identifier (e.g.~an index or a random seed), generates a \emph{level}, defined as an environment instance exhibiting a unique configuration of its underlying factors of variation, such as layout, asset appearances, or specific environment dynamics~\cite{Risi_2020}. For example, MiniGrid's MultiRoom environment instantiates mazes with varying numbers of rooms based on the seed \citep{gym_minigrid}. We refer to sampling a new trajectory generated from the agent's latest policy acting on a given level $l$ as \emph{replaying} that level $l$.

The level diversity of PCG environments makes them useful testbeds for studying the robustness and generalization ability of RL agents, measured by agent performance on unseen test levels. The standard test evaluation protocol for PCG environments consists of training the agent on a finite number of training levels $\Lambda_{\text{train}}$, and evaluating performance on unseen test levels $\Lambda_{\text{test}}$, drawn from the set of all levels. Training levels are sampled from an arbitrary distribution $P_{\text{train}}(l|\Lambda_{\text{train}})$. We call this training process \emph{direct level sampling}. A common variation of this protocol sets $\Lambda_{\text{train}}$ to the set of all levels, though in practice, the agent will still only effectively see a finite set of levels after training for a finite number of steps. In the case of a finite training set, typically $P_{\text{train}}(l|\Lambda_{\text{train}}) = \mathbf{Uniform}(l; \Lambda_{\text{train}})$.

PCG environments naturally lend themselves to curriculum learning. Prior works have shown that directly altering levels to match their difficulty to the agent's abilities can improve generalization \citep{justesen2018illuminating, dennis2020emergent,DBLP:journals/corr/abs-1810-08272,DBLP:conf/iclr/ZhongRG20}. These findings further suggest the levels most useful for improving an agent's policy vary throughout the course of training. In this work, we consider how to automatically discover a curriculum that improves generalization for a general blackbox PCG environment---crucially, without assuming any knowledge or control of how levels are generated (beyond providing the random seed or other indicial 
level identifier).
\newpage

\section{Prioritized Level Replay}
\label{section:methods}
In this section, we present~\emph{Prioritized Level Replay}~(PLR), an algorithm for selectively sampling the next training level given the current policy, by prioritizing levels with higher estimated learning potential when replayed (that is, revisited). 
PLR is a drop-in replacement for the experience-collection process used in a wide range of RL algorithms. Algorithm~\ref{alg:level_replay_mc} shows how it is straightforward to incorporate PLR into a generic policy-gradient training loop. The procedure for adding new levels into the level replay buffer is detailed in Algorithm~\ref{alg:plr_update_rule}. Though the pseudocode samples only a single level per training loop, level sampling and the subsequent rollouts and updates to the level buffer typically occur in parallel across a batch of levels.

Our method, illustrated in Figure~\ref{fig:algo_fig} and fully specified in Algorithm~\ref{alg:level_replay_mc}, induces a dynamic, nonparametric sampling distribution $P_{\text{replay}}(l | \Lambda_{\text{seen}})$ over previously visited training levels $\Lambda_{\text{seen}}$ that prioritizes visited levels with higher learning potential based on properties of the agent's past trajectories. We refer to $P_{\text{replay}}(l | \Lambda_{\text{seen}})$ as the \emph{replay distribution}. 
Throughout training, our method updates this replay distribution according to a heuristic score, assigning greater weight to visited levels with higher \emph{future} learning potential. 
Using dynamic arrays $S$ and $C$ of equal length to $\Lambda_{\text{seen}}$, PLR tracks level scores $S_i \in S$ for each visited training level $l_i$ based on the latest episode trajectory on $l_i$, as well as the episode count $C_i \in C$ at which each level $l_i \in \Lambda_{\text{seen}}$ was last sampled. Our method updates $P_{\text{replay}}$ after each terminated episode by computing a mixture of two distributions, $P_S$, based on the level scores, and $P_C$, based on how long ago each level was last sampled:
\begin{equation}
\label{eq:replay}
\begin{aligned}
P_{\text{replay}} = (1-\rho) \cdot P_S + \rho \cdot P_C,
\end{aligned}
\end{equation}
where the staleness coefficient $\rho \in [0,1]$ is a hyperparameter. 
We discuss how we compute level scores $S_i$, parameterizing the scoring distribution $P_S$, and the staleness distribution $P_C$, in Sections~\ref{sec:scoring} and~\ref{sec:staleness}, respectively.

PLR chooses the next level at the start of every training episode by first sampling a replay-decision from a Bernoulli (or similar) distribution $P_D(d)$ to determine whether to replay a level sampled from the replay distribution $P_{\text{replay}}(l|\Lambda_{\text{seen}})$ or to experience a new, unseen level from $\Lambda_{\text{train}}$, according to some distribution $P_{\text{new}}(l|\Lambda_{\text{seen}}; \Theta_{\text{train}})$. In practice, for the case of a finite number of training levels, we implement $P_{\text{new}}$ as a uniform distribution over the remaining unseen levels. For the case of a countably infinite number of training levels, we simulate $P_{\text{new}}$ by sampling levels from $P_{\text{train}}$ until encountering an unseen level. 
In our experiments based on a finite number of training levels, we opt to naturally anneal $P_D(d=1)$ as $|\Lambda_{\text{seen}}|/|\Lambda_{\text{train}}|$, so replay occurs more often as more training levels are visited.

The following sections describe how PLR updates the replay distribution $P_{\text{replay}}(l | \Lambda_{\text{seen}})$ via Equation~\ref{eq:replay} in detail.

\begin{figure}[h]
\begin{minipage}{\linewidth}
\begin{algorithm}[H]
\SetAlgoLined
\SetArgSty{textnormal}

\SetKwBlock{Input}{input:}{end}
\SetKwBlock{Initialize}{initialize:}{end}

\caption{Prioritized Level Replay (PLR)}
\label{alg:level_replay_mc}

\Input{
    Training levels $\Theta_{\text{train}}$ \\
    Policy $\pi_{\phi}$ \\
    Policy update function $\mathcal{U}(\mathcal{B}, \phi) \rightarrow \phi'$
}

\Initialize{
Level scores $S$ and level timestamps $C$ \\
Global episode counter $c \gets 0$ \\
Level replay buffer $\Lambda = \varnothing$ \\
Experience buffer $\mathcal{E} = \varnothing$ \\
}

\vspace{2mm}
\While{training}{
Sample replay decision $d \sim{} P_D(d)$ \\
    \eIf{$d=0$ \textbf{and} $\left| \Theta_{\text{train}} \setminus \Lambda\right| > 0$}{
        Define new index $i \gets |S| + 1$ \\
        Sample $\theta_i \sim{} P_{\text{new}}(\theta | \Lambda; \Theta_{\text{train}})$ \\
        Add $\theta_i$ to $\Lambda$ \\
        Add initial value $S_i = 0$ to $S$ and $C_i = 0$ to $C$ \\
    }
    {
        Sample $\theta_i \sim{} P_{\textnormal{replay}}(\theta)$ (via Equation~\ref{eq:replay}) \\
    }
    Sample $\tau \sim{} P_\pi(\tau | \theta_i)$ \\
    Update score $S_i \gets \mathbf{score}(\tau, \pi)$ and timestamp $C_i \gets c$ \\
    Update $\mathcal{E}$ with $ \mathcal{\tau}$ \\
    Update the policy $\phi \leftarrow \mathcal{U}(\mathcal{E}, \phi)$
}
\end{algorithm}
\end{minipage}
\end{figure}

\begin{figure}[h!] 
\vskip -0.1in
\begin{minipage}{\linewidth}
\begin{algorithm}[H]
{
\label{alg:plr_update_rule}
\SetAlgoLined
\caption{PLR level-buffer update rule}
    \KwIn{Level buffer $\bm{\Lambda}$ of size $K$ with scores $S$ and timestamps $C$; level $\theta$; level score $S_{\theta}$; and current episode count $c$}
    \eIf{$|\bm{\Lambda}| < K$}{
        Insert $\theta$ into $\bm{\Lambda}$, and set $S(\theta) = S_{\theta}$, $C(\theta) = c$ \\
    }
    {
        Find level with minimal support,    $\theta_{\textnormal{min}} = \underset{\theta}{\arg\min\;}  P_{\textnormal{replay}}(\theta)$ \\
        \If{$S(\theta_{\textnormal{min}}) < S_{\theta}$}{
            Remove $\theta_{\textnormal{min}}$ from $\bm{\Lambda}$ \\
            Insert $\theta$ into $\bm{\Lambda}$, and set $S(\theta) = S_{\theta}$, $C(\theta) = c$ \\
            Update $P_{\textnormal{replay}}$ with latest scores $S$ and timestamps $C$ \\
        }
    }
}
\end{algorithm}
\end{minipage}
\end{figure}

\subsection{Scoring Levels for Learning Potential}
\label{sec:scoring}
After collecting each complete episode trajectory $\tau$ on level $l_i$ using policy $\pi$, our method assigns $l_i$ a score \mbox{$S_i = \mathbf{score}(\tau, \pi)$} measuring the learning potential of replaying $l_i$ in the future. We employ a function of the TD-error at timestep $t$, $\delta_t = r_t + \gamma V(s_{t+1}) - V(s_t)$, as a proxy for this learning potential. The expectation of the TD-error over next states is equivalent to the advantage estimate, and therefore higher-magnitude TD-errors imply greater discrepancy between expected and actual returns, making $\delta_t$ a useful measure of the learning potential in revisiting a particular state transition. To prioritize the learning potential of future experiences resulting from replaying a level, we use a scoring function based on the \emph{average magnitude} of the Generalized Advantage Estimate~\citep[GAE;][]{schulman2018high} over each of $T$ time steps in the latest trajectory $\tau$ from that level:
\begin{equation}
S_i = \mathbf{score}(\tau, \pi) = \frac{1}{T}\sum_{t=0}^{T} \left|\sum_{k=t}^T(\gamma\lambda)^{k-t}\delta_k\right|. \label{eq:average_scoring_function}
\end{equation}

While the GAE at time $t$ is most commonly expressed as the discounted sum of all $1$-step TD-errors starting at $t$ as in Equation~\ref{eq:average_scoring_function}, it is equivalent to an exponentially-discounted sum of all $k$-step TD-errors from $t$, with discount factor $\lambda$. By considering all $k$-step TD-errors, the GAE mitigates the bias introduced by the bootstrap term in $1$-step TD-errors. The discount factor $\lambda$ then controls the trade-off between bias and variance. Our scoring function considers the absolute value of the GAE, as we assume the learning potential grows with the magnitude of the TD-error irrespective of its sign. This also avoids opposite signed errors canceling out.

Another useful interpretation of Equation~\ref{eq:average_scoring_function} comes from observing that the GAE magnitude at $t$ is equivalent to the L1 value loss $|\hat{V}_t - V_t|$ under a policy-gradient algorithm that uses GAE for its own advantage estimates (and therefore value targets $\hat{V}_t$), as done in state-of-the-art implementations of PPO \citep{DBLP:journals/corr/SchulmanWDRK17} used in our experiments. 
Unless otherwise indicated, PLR refers to the instantiation of our algorithm with L1 value loss as the scoring function.

We further formulate the \emph{Value Correction Hypothesis} to motivate our approach: In sparse reward settings, prioritizing the sampling of training levels with greatest average absolute value loss leads to a curriculum that improves both sample efficiency and generalization. We reason that on threshold levels (i.e.~those at the limit of the agent's current abilities) the agent will see non-stationary returns (or value targets)---and therefore incur relatively high value errors---until it learns to solve them consistently.
In contrast, levels beyond the agent's current abilities tend to result in stationary value targets signaling failure and therefore low value errors, until the agent learns useful, transferable behaviors from threshold levels.
Prioritizing levels by value loss then naturally guides the agent along the expanding threshold of its ability---without the need for any externally provided measure of difficulty. We believe that learning behaviors systematically aligned with the inherent complexities of the environment in this way may lead to better generalization, and will seek to verify this empirically in Section~\ref{subsection:minigrid_results}.

While we provide principled motivations for our specific choice of scoring function, we emphasize that in general, the scoring function can be any approximation of learning potential based on trajectory values. Note that candidate scoring functions should asymptotically decrease with frequency of level visitation to avoid mode collapse of $P_{\text{replay}}$ to a limited set of levels and possible overfitting. In Section~\ref{sec:experiments}, we compare our choice of the GAE magnitude, or equivalently, the L1 value loss, to alternative TD-error-based and uncertainty-based scoring approaches, listed in Table~\ref{table:scoring_metrics}.

Given level scores, we use normalized outputs of a prioritization function $h$ of these scores and a temperature parameter $\beta$ to define the score-prioritized distribution $P_S(\Lambda_{\text{train}})$ over the training levels, under which
\begin{equation}
\label{eq:score}
\begin{aligned}
P_S(l_i|\Lambda_{\text{seen}}, S) = \frac{h(S_i)^{1/\beta}}{\sum_j h(S_j)^{1/\beta}}. 
\end{aligned}
\end{equation}
The function $h$ defines how differences in level scores translate into differences in prioritization. The temperature parameter $\beta$ allows us to  tune how much $h(S)$ ultimately determines the resulting distribution. We make the design choice of using rank prioritization, for which $h(S_i) = 1/\text{rank}(S_i)$, where $\text{rank}(S_i)$ is the rank of level score $S_i$ among all scores sorted in descending order. We also experimented with proportional prioritization ($h(S_i)=S_i$) as well as greedy prioritization (the level with the highest score receives probability~$1$), both of which tend to perform worse.

\subsection{Staleness-Aware Prioritization}
\label{sec:staleness}
As the scores used to parameterize $P_S$ are a function of the state of the policy at the time the associated level was last played, they come to reflect a gradually more off-policy measure the longer they remain without an update through replay. We mitigate this drift towards ``off-policy-ness'' by explicitly mixing the sampling distribution with a staleness-prioritized distribution $P_C$:
\begin{equation}
\label{eq:staleness}
\begin{aligned}
P_C(l_i | \Lambda_{\text{seen}}, C, c) = \frac{c - C_i}{\sum_{C_j \in C} c - C_j}
\end{aligned}
\end{equation}
which assigns probability mass to each level $l_i$ in proportion to the level's \emph{staleness} $c - C_i$. 
Here, $c$ is the count of total episodes sampled so far in training and $C_i$ (referred to as the level's timestamp) is the episode count at which $l_i$ was last sampled. By pushing support to levels with staler scores, $P_C$ ensures no score drifts too far off-policy. 

Plugging Equations~\ref{eq:score} and~\ref{eq:staleness} into Equation~\ref{eq:replay} gives us a replay distribution that is calculated as
\[
P_{\text{replay}}(l_i) = (1-\rho) \cdot P_S(l_i | \Lambda_{\text{seen}}, S) + \rho \cdot P_C(l_i | \Lambda_{\text{seen}}, C, c).
\]
Thus, a level has a greater chance of being sampled when its score is high or it has not been sampled for a long time. 

\section{Experimental Setting}
\label{sec:experiments}
We evaluate PLR on several PCG environments with various combinations of scoring functions and prioritization schemes, and compare to the most common direct level sampling baseline of $P_{\text{train}}(l | \Lambda_{\text{train}}) = \mathbf{Uniform}(l; \Lambda_{\text{train}})$. We train and test on all 16 environments in the Procgen Benchmark on easy and hard difficulties, but focus discussion on the easy results, which allow direct comparison to several prior studies. We compare to UCB-DrAC~\citep{raileanu2021automatic}, the state-of-the-art image augmentation method on this benchmark, and mixreg~\citep{wang2020mixreg}, a recently introduced data augmentation method. We also compare to TSCL Window~\citep{Matiisen_2020}, which resembles PLR with an alternative scoring function using the slope of recent returns and no staleness sampling. For fair comparison, we also evaluate a custom TSCL Window variant that mixes in the staleness distribution $P_{C}$ weighted by $\rho > 0$. Further, to demonstrate the ease of combining PLR with other methods, we evaluate UCB-DrAC using PLR for sampling training levels. Finally, we test the Value Correction Hypothesis on two challenging MiniGrid environments.

We measure episodic \emph{test returns} per game throughout training, as well as the performance of the final policy over 100 unseen test levels of each game relative to PPO with uniform sampling. We also evaluate the mean normalized episodic test return and mean generalization gap, averaged over all games (10 runs each). We normalize returns according to \citet{cobbe2019quantifying} and compute the generalization gap as train returns minus test returns. Thus, a larger gap indicates more overfitting, making it an apt measure of generalization. We assess statistical significance at $p=0.05$, using the Welch t-test.

In line with the standard baseline for these environments, all experiments use PPO with GAE for training. 
For \mbox{Procgen}, we use the same ResBlock architecture as \citet{cobbe2019procgen} and train for 25M total steps on 200 levels on the easy setting as in the original baselines. 
For MiniGrid, we use a 3-layer CNN architecture based on \citet{igl2019generalization}, and provide approximately 1000 levels of each difficulty per environment during training. Detailed descriptions of the environments can be found in Appendices~\ref{appendix:env_procgen}–\ref{appendix:env_minigrid}. Choice of architectures and hyperparameters used in our experiments can be found in Appendix~\ref{appendix:exp_plr}. See Table~\ref{table:scoring_metrics} for the full set of scoring functions investigated in our experiments. 

Additionally, in Section~\ref{section:training_on_full_dist}, we extend PLR to support training on an unbounded number of levels by tracking a rolling, finite buffer of the top levels so far encountered by learning potential, and demonstrate that it improves the sample efficiency and generalization performance of the resultant policy in the MiniGrid environments studied.

\begingroup
\setlength{\tabcolsep}{1pt}
\begin{table}[hbtp]
\caption{\small{Scoring functions investigated in this work.}}
\label{table:scoring_metrics}
\begin{center}
\begin{tabularx}{\linewidth}{m{0.35\linewidth}m{0.65\linewidth}} 
\toprule
\renewcommand{\arraystretch}{1.8}
\textbf{Scoring function} & \textbf{$\mathbf{score}(\tau, \pi)$} \\ 
\midrule
Policy entropy & $-\frac{1}{T}\sum_{t=0}^{T} \sum_a{\pi(a, s_t) \log \pi(a, s_t)}$ \\[8pt]
Policy min-margin & \mbox{$\frac{1}{T}\sum_{t=0}^{T} \left(\max_{a} \pi(a, s_t) -  \max_{a \neq \max_{a} \pi(a, s_t)} \pi(a, s_t)\right)$} \\ [8pt]
Policy least-confidence & $\frac{1}{T}\sum_{t=0}^{T} (1 - \max_{a}\pi(a, s_t))$ \\[8pt]
1-step TD error & $\frac{1}{T}\sum_{t=0}^{T} |\delta_t|$ \\[8pt]
GAE & $\frac{1}{T}\sum_{t=0}^{T} \sum_{k=t}^T(\gamma\lambda)^{k-t}\delta_k$ \\[8pt]
L1 value loss, $\lvert$GAE$\rvert$    & $\frac{1}{T}\sum_{t=0}^{T} \left|\sum_{k=t}^T(\gamma\lambda)^{k-t}\delta_k\right|$ \\[8pt]
\bottomrule
\end{tabularx}
\end{center}
\end{table}
\endgroup

\section{Results and Discussion}

\begin{figure*}[t!]
    \centering
    \includegraphics[width=1.0\textwidth]{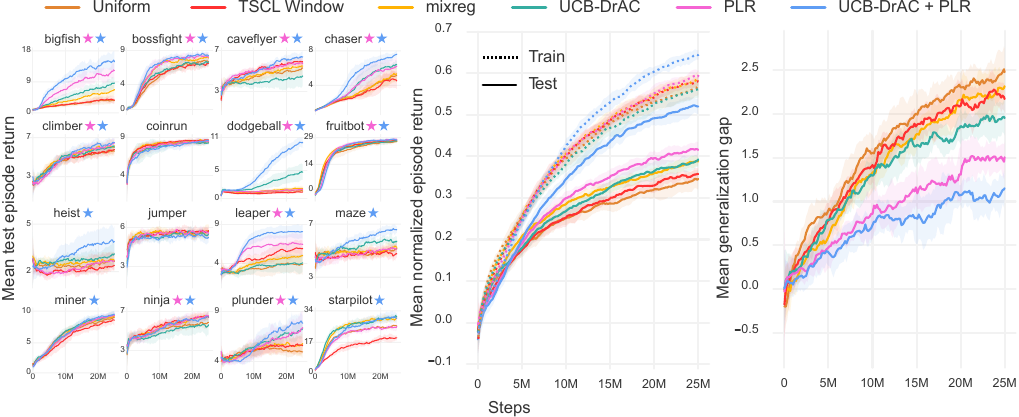}
    \caption{\small{Left: Mean episodic test returns (10 runs) of each method. Each colored $\bigstar$ indicates statistically significant ($p<0.05$) gains in final test performance or sample complexity along the curve, relative to uniform sampling, for the PLR-based method of the same color. Center: Mean normalized train and test returns averaged across all games. Right: Mean generalization gaps averaged across all games.}}
    \label{fig:procgen_summary}
\end{figure*}

Our main findings are that
(i) Both PLR with L1 value loss and 1-step TD errors significantly improves both sample efficiency and generalization, and the L1 value loss variant attains the highest normalized mean test and train returns and mean reduction in generalization gap on Procgen out of all individual methods evaluated, while matching UCB-DrAC in test improvement relative to PPO;
(ii) alternative scoring functions based on classifier uncertainty metrics lead to inconsistent improvements across environments; (iii) PLR combined with UCB-DrAC sets a new state-of-the-art on Procgen; and (iv) PLR induces an implicit curriculum over training levels, which substantially aids training in two challenging MiniGrid environments.

\subsection{Procgen Benchmark}
\label{subsection:procgen_benchmark}
\begin{figure}[t!]
    \centering
    \includegraphics[width=.75\linewidth]{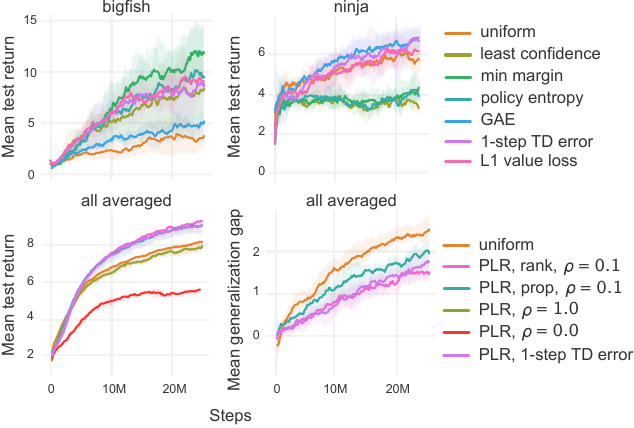}
    \caption{\small{Top: Two example Procgen environments, between which all scoring functions except L1 value loss and 1-step TD error show inconsistent improvements to test performance (rank prioritization, \mbox{$\beta=0.1$}, $\rho=0.3$). This inconsistency holds across settings in our grid search. Bottom: Mean unnormalized episodic test returns (left) and mean generalization gap (right) for various PLR settings.}}
    \label{fig:alternative-settings-results}
\end{figure}

Our results, summarized in Figure~\ref{fig:procgen_summary}, show PLR with rank prioritization ($\beta = 0.1$, $\rho = 0.1$) leads to the largest statistically significant gains in mean normalized test and train returns and reduction in generalization gap compared to uniform sampling, outperforming all other methods besides UCB-DrAC + PLR. PLR combined with UCB-DrAC sees the most drastic improvements in these metrics. As reported in Table \ref{tab:ucb-drac-plr-sota}, UCB-DrAC + PLR yields a 76\% improvement in mean test return relative to PPO with uniform sampling, and a 28\% improvement relative to the previous state-of-the-art set by UCB-DrAC. While PLR with rank prioritization leads to statistically significant gains in test return on 10 of 16 environments and proportional prioritization, on 11 of 16 games, we prefer rank prioritization: While we find the two comparable in mean normalized returns, Figure \ref{fig:alternative-settings-results} shows rank prioritization results in higher mean \emph{unnormalized} test returns and a significantly lower mean generalization gap, averaged over all environments.

\begin{figure*}[t!]
    \centering
    \includegraphics[width=\textwidth]{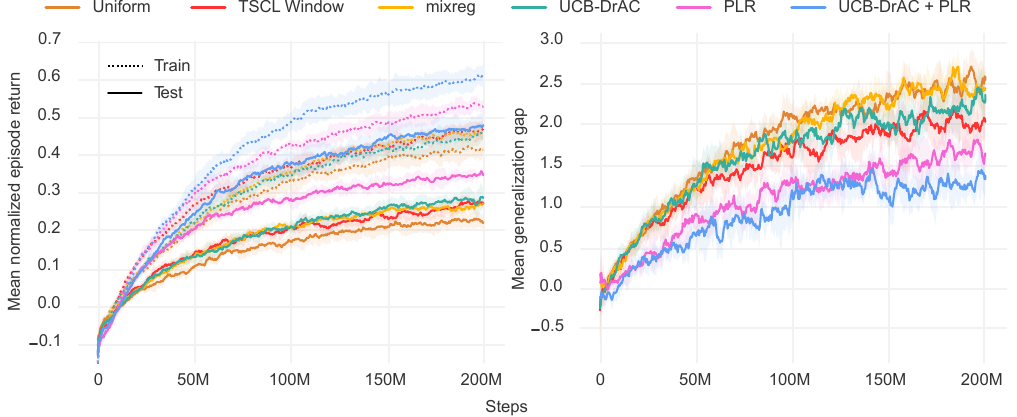}
    \caption{\small{Left: Mean normalized train and test episode returns on Procgen Benchmark (hard). Right: Corresponding generalization gaps during training. All curves are averaged across all environments over 5 runs. The shaded area indicates one standard deviation around the mean. PLR-based methods statistically significantly outperform all others in both train and test returns. Only the PLR-based methods statistically significantly reduce the generalization gap ($p < 0.05$).}}
    \label{fig:procgen_hard_summary}
\end{figure*}

Further, Figure \ref{fig:alternative-settings-results} shows that gains only occur when $P_{\text{replay}}$ considers \emph{both} level scores and staleness ($0 < \rho < 1$), highlighting the importance of staleness-based sampling in keeping scores from drifting off-policy. Lastly, we also benchmarked PLR on the hard setting against the same set of methods, where it again leads with 35\% greater test returns relative to uniform sampling and 83\% greater test returns when combined with UCB-DrAC. Figure~\ref{fig:procgen_hard_summary}
and Table~\ref{tab:ucb-drac-plr-hard} report additional details on these results.

\begin{figure*}[h]
    \centering
    \includegraphics[width=\linewidth]{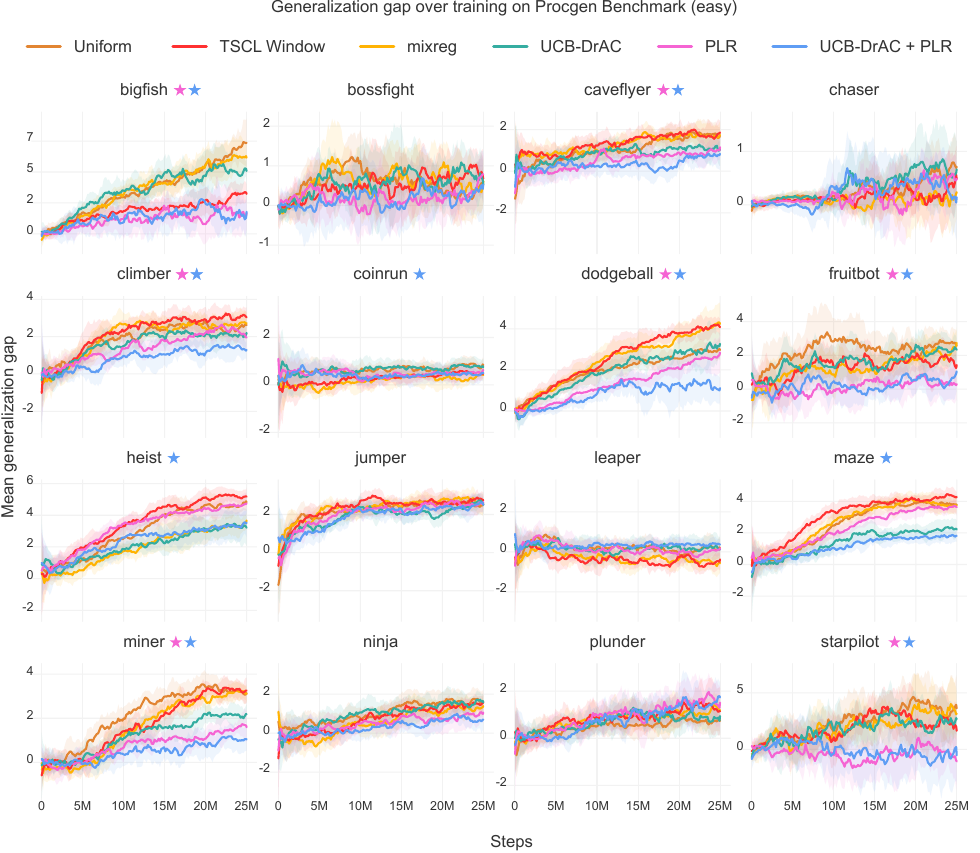}
    \caption{\small{Mean generalization gaps throughout training (10 runs) on each Procgen Benchmark game (easy). The shaded area indicates one standard deviation around the mean. A $\bigstar$ indicates the method of matching color results in a statistically significant ($p < 0.05$) reduction in generalization gap compared to the uniform-sampling baseline. By itself, PLR significantly reduces the generalization gap on 7 games, and UCB-DrAC, on 5 games. This number jumps to 10 of 16 games when these two methods are combined. TSCL only significantly reduces generalization gap on 2 of 16 games relative to uniform sampling, while increasing it on others, most notably on Dodgeball.}}
    \label{fig:procgen_gen_gap_large}
\end{figure*}

We find both TD-error-based scoring functions, based on L1 value loss (equivalent to GAE magnitude) and 1-step TD errors respectively, lead to significant improvements in sample efficiency and generalization across the Procgen benchmark. However, as seen in Figure~\ref{fig:alternative-settings-results}, prioritizing based on 1-step TD errors leads to slightly lower mean test return and higher generalization gap across the benchmark, and thus, we make use of the L1 value loss as the default scoring function for PLR throughout the other experiments in this study. The alternative scoring metrics based on classifier uncertainty perform inconsistently across games. While certain games, such as BigFish, see improved sample-efficiency and generalization under various scoring functions, others, such as Ninja, see no improvement or worse, degraded performance. See Figure~\ref{fig:alternative-settings-results} for an example of this inconsistent effect across games. We find the best-performing variant of TSCL Window does not incorporate staleness information ($\rho = 0$) and similarly leads to inconsistent outcomes across games at test time, notably significantly worsening performance on StarPilot, as seen in Figure \ref{fig:procgen_summary}, and increasing the generalization gap on some environments as revealed in Figure \ref{fig:procgen_gen_gap_large}.

\begin{table*}[t!]
\vskip -0.1in
\newcommand\mc[1]{\multicolumn{1}{l}{#1}}

\setlength{\tabcolsep}{2.5pt}
\caption{\small{Test returns of policies trained using each method with its best hyperparameters. Following \citet{raileanu2021automatic}, the reported mean and standard deviations per environment are computed by evaluating the final policy's average return on 100 test episodes, aggregated across multiple training runs (10 runs for Procgen Benchmark and 3 for MiniGrid, each initialized with a different training seed). Normalized test returns per run are computed by dividing the average test return per run for each environment by the corresponding average test return of the uniform-sampling baseline over all runs. We then report the means and standard deviations of normalized test returns aggregated across runs. We report the normalized return statistics for Procgen and MiniGrid environments separately. Bolded methods are not significantly different from the method with highest mean, unless all are, in which case none are bolded. PLR+ denotes the combined PLR and UCB-DrAC method.}}
\label{tab:ucb-drac-plr-sota}

\begin{center}
\begin{subtable}{\linewidth}
\footnotesize
\begin{tabular}{l|r r r r| r r}
\toprule
Environment &Uniform &TSCL &mixreg &UCB-DrAC &PLR &PLR+\\
\midrule
BigFish&$3.7\pm1.2$&$4.3\pm1.3$&$6.9\pm1.6$&$8.7\pm1.1$&$10.9\pm2.8$&$\mathbf{14.3\pm2.1}$\\
BossFight&$7.7\pm0.4$&$7.4\pm0.8$&$8.1\pm0.7$&$7.7\pm0.7$&$\mathbf{8.9\pm0.4}$&$\mathbf{8.8\pm0.8}$\\
CaveFlyer&$5.4\pm0.8$&$\mathbf{6.3\pm0.6}$&$6.0\pm0.6$&$4.6\pm0.9$&$\mathbf{6.3\pm0.5}$&$\mathbf{6.8\pm0.7}$\\
Chaser&$5.2\pm0.7$&$4.9\pm1.0$&$5.7\pm1.1$&$6.8\pm0.9$&$6.9\pm1.2$&$\mathbf{8.0\pm0.6}$\\
Climber&$5.9\pm0.6$&$6.0\pm0.8$&$\mathbf{6.6\pm0.7}$&$\mathbf{6.4\pm0.9}$&$\mathbf{6.3\pm0.8}$&$\mathbf{6.8\pm0.7}$\\
CoinRun&$8.6\pm0.4$&$\mathbf{9.2\pm0.2}$&$8.6\pm0.3$&$8.6\pm0.4$&$8.8\pm0.5$&$\mathbf{9.0\pm0.4}$\\
Dodgeball&$1.7\pm0.2$&$1.2\pm0.4$&$1.8\pm0.4$&$5.1\pm1.6$&$1.8\pm0.5$&$\mathbf{10.3\pm1.4}$\\
FruitBot&$27.3\pm0.9$&$27.1\pm1.6$&$27.7\pm0.8$&$27.0\pm1.3$&$28.0\pm1.3$&$27.6\pm1.5$\\
Heist&$2.8\pm0.9$&$2.5\pm0.6$&$2.7\pm0.4$&$3.2\pm0.7$&$2.9\pm0.5$&$\mathbf{4.9\pm1.3}$\\
Jumper&$5.7\pm0.4$&$6.1\pm0.6$&$6.1\pm0.3$&$5.6\pm0.5$&$5.8\pm0.5$&$5.9\pm0.3$\\
Leaper&$4.2\pm1.3$&$6.4\pm1.2$&$5.2\pm1.1$&$4.4\pm1.4$&$6.8\pm1.2$&$\mathbf{8.7\pm1.0}$\\
Maze&$5.5\pm0.4$&$5.0\pm0.3$&$5.4\pm0.5$&$6.2\pm0.5$&$5.5\pm0.8$&$\mathbf{7.2\pm0.8}$\\
Miner&$8.7\pm0.7$&$8.9\pm0.6$&$9.5\pm0.4$&$\mathbf{10.1\pm0.6}$&$\mathbf{9.6\pm0.6}$&$\mathbf{10.0\pm0.5}$\\
Ninja&$6.0\pm0.4$&$\mathbf{6.8\pm0.5}$&$\mathbf{6.9\pm0.5}$&$5.8\pm0.8$&$\mathbf{7.2\pm0.4}$&$\mathbf{7.0\pm0.5}$\\
Plunder&$5.1\pm0.6$&$5.9\pm1.1$&$5.7\pm0.5$&$\mathbf{7.8\pm0.9}$&$\mathbf{8.7\pm2.2}$&$\mathbf{7.7\pm0.9}$\\
StarPilot&$26.8\pm1.5$&$19.8\pm3.4$&$\mathbf{32.7\pm1.5}$&$\mathbf{31.7\pm2.4}$&$27.9\pm4.4$&$29.6\pm2.2$\\
\midrule
Norm. mean (\%)&$100.0\pm4.5$&$103.0\pm3.6$&$113.8\pm2.8$&$129.8\pm8.2$&$128.3\pm5.8$&$\mathbf{176.4\pm6.1}$\\
\midrule
\addlinespace[0.25cm]
MultiRoom-N4 & $0.80\pm0.04$ & -- & -- & -- & $\mathbf{0.81\pm0.01}$ & -- \\
OMG-Easy&$0.53\pm0.04$& -- & -- & -- & $\mathbf{0.85\pm0.04}$& -- \\
OMG-Med&$0.65\pm0.01$& -- & -- & -- & $\mathbf{0.73\pm0.07}$& -- \\
\midrule
Norm. mean (\%)&$100.0\pm2.5$& -- & -- & -- &$\mathbf{124.3\pm4.7}$ & --\\
\bottomrule
\end{tabular}
\end{subtable}
\end{center}
\end{table*}

We present an overview of the improvements in test performance of each method across all 16 Procgen Benchmark games over 10 runs in Figure~\ref{fig:procgen_summary}. For each game, Figure \ref{fig:procgen_gen_gap_large} further shows how the  generalization gap changes over the course of training under each method tested. 
The results in Figure~\ref{fig:alternative-settings-results} show the mean test episodic returns averaged over all games of the Procgen Benchmark (easy) for various ablations of PLR, including no prioritization and varying degrees of staleness sampling.  Using only staleness ($\rho=1$) or only L1 value loss scores ($\rho=0$) is considerably worse than direct level sampling. Thus, we only observe gains compared to the baseline when both level scores and staleness are used for the sampling distribution. Moreover, we see that PLR with rank prioritization leads to a slightly larger mean improvements on several games.

Finally, we also benchmarked PLR and UCB-DrAC + PLR (denoted PLR+) against uniform sampling, TSCL Window, mixreg, and UCB-DrAC on Procgen hard across 5 runs per environment. Due to the high computational cost of the evaluation protocol for Procgen hard, which entails 200M training steps, we directly use the best hyperparameters found in the easy setting for each method. The results in Figure \ref{fig:procgen_hard_summary} show the two PLR-based methods significantly outperform all other methods in terms of normalized mean train and test episodic return, as well as reduction in mean generalization gap, attaining even greater margins of improvement than in the easy setting. As summarized by Table~\ref{tab:ucb-drac-plr-hard}, the gains of PLR and UCB + PLR in mean normalized test return relative to uniform sampling in the hard setting are comparable to those in the easy setting.

\begin{table*}[t!]
\setlength{\tabcolsep}{2.5pt}
\caption{\small{Comparison of test scores of PPO with PLR against PPO with uniform-sampling on the hard setting of Procgen Benchmark. Following \citep{raileanu2021automatic}, reported figures represent the mean and standard deviation of average test scores over 100 episodes aggregated across 5 runs, each initialized with a unique training seed. For each run, a normalized average return is computed by dividing the average test return for each game by the corresponding average test return of the uniform-sampling baseline over all 500 test episodes of that game, followed by averaging these normalized returns over all 16 games. The final row reports the mean and standard deviation of the normalized returns aggregated across runs. Bolded methods are not significantly different from the method with highest mean, unless all are, in which case none are bolded.}}
\label{tab:ucb-drac-plr-hard}
\begin{center}
\footnotesize
\begin{tabular}{l|r r r r | r r}
\toprule
Environment &Uniform &TSCL &mixreg &UCB-DrAC &PLR &PLR+\\
\midrule
BigFish&$9.7\pm1.8$&$\mathbf{11.9\pm2.5}$&$\mathbf{12.0\pm2.5}$&$10.9\pm1.6$&$\mathbf{15.3\pm3.6}$&$\mathbf{15.5\pm2.8}$\\
BossFight&$\mathbf{9.6\pm0.2}$&$8.4\pm0.7$&$\mathbf{9.3\pm0.9}$&$9.0\pm0.2$&$\mathbf{9.7\pm0.4}$&$\mathbf{9.5\pm1.1}$\\
CaveFlyer&$3.5\pm0.8$&$6.3\pm0.6$&$4.0\pm1.0$&$2.6\pm0.8$&$6.4\pm0.6$&$\mathbf{8.0\pm0.9}$\\
Chaser&$5.9\pm0.5$&$6.2\pm1.0$&$\mathbf{6.5\pm0.8}$&$\mathbf{7.0\pm0.6}$&$\mathbf{6.8\pm2.2}$&$\mathbf{7.6\pm0.2}$\\
Climber&$ 5.3\pm1.1$&$ 5.2\pm0.7$&$ 5.7\pm0.7$&$ 6.1\pm1.0$&$ 7.4\pm0.6$&$ 7.6\pm1.8$\\
CoinRun&$4.5\pm0.4$&$5.8\pm0.8$&$\mathbf{6.2\pm1.0}$&$5.2\pm1.0$&$\mathbf{6.8\pm0.6}$&$\mathbf{7.1\pm0.5}$\\
Dodgeball&$3.9\pm0.6$&$1.9\pm0.9$&$4.7\pm1.0$&$9.9\pm1.2$&$7.4\pm1.3$&$\mathbf{12.4\pm0.7}$\\
FruitBot&$\mathbf{11.9\pm4.2}$&$13.1\pm2.3$&$\mathbf{14.7\pm2.2}$&$\mathbf{15.6\pm3.7}$&$\mathbf{16.7\pm1.0}$&$\mathbf{12.9\pm5.1}$\\
Heist&$1.5\pm0.4$&$0.9\pm0.3$&$1.2\pm0.4$&$1.1\pm0.3$&$1.3\pm0.4$&$2.6\pm2.2$\\
Jumper&$3.2\pm0.3$&$3.2\pm0.3$&$3.3\pm0.4$&$2.9\pm0.9$&$3.5\pm0.5$&$3.3\pm0.8$\\
Leaper&$7.1\pm0.3$&$\mathbf{7.5\pm0.5}$&$\mathbf{7.5\pm0.5}$&$3.8\pm1.6$&$\mathbf{7.4\pm0.2}$&$\mathbf{8.2\pm0.7}$\\
Maze&$3.6\pm0.7$&$3.8\pm0.6$&$3.9\pm0.5$&$4.4\pm0.2$&$4.0\pm0.4$&$\mathbf{6.2\pm0.4}$\\
Miner&$12.8\pm1.4$&$11.7\pm0.9$&$13.3\pm1.6$&$\mathbf{16.1\pm0.6}$&$11.3\pm0.7$&$\mathbf{15.3\pm0.8}$\\
Ninja&$5.2\pm0.1$&$5.9\pm0.8$&$5.0\pm1.0$&$5.2\pm1.0$&$\mathbf{6.1\pm0.6}$&$\mathbf{6.9\pm0.3}$\\
Plunder&$3.2\pm0.1$&$5.4\pm1.1$&$3.7\pm0.4$&$7.8\pm1.1$&$8.6\pm2.7$&$\mathbf{17.5\pm1.3}$\\
StarPilot&$5.5\pm0.6$&$2.1\pm0.4$&$6.9\pm0.6$&$\mathbf{11.2\pm1.7}$&$5.4\pm0.8$&$\mathbf{12.3\pm1.5}$\\
\midrule
Norm. mean (\%)&$100.0\pm2.0$&$103.9\pm3.5$&$110.6\pm3.9$&$126.6\pm3.0$&$135.0\pm6.1$&$\mathbf{182.9\pm8.2}$\\
\bottomrule
\end{tabular}
\end{center}
\end{table*}

\subsection{MiniGrid}
\label{subsection:minigrid_results}
\begin{figure*}
    \centering
    \begin{subfigure}[b]{1.0\textwidth}
        \includegraphics[width=\textwidth]{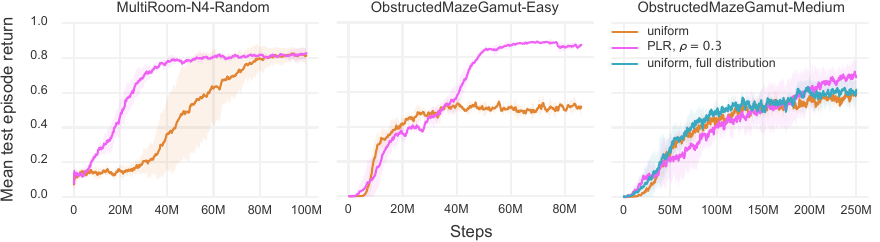}
        \label{fig:multiroom-value-testreturns}
    \end{subfigure}\vspace{-1em}
    \begin{subfigure}[b]{1.0\textwidth}
        \includegraphics[width=\textwidth]{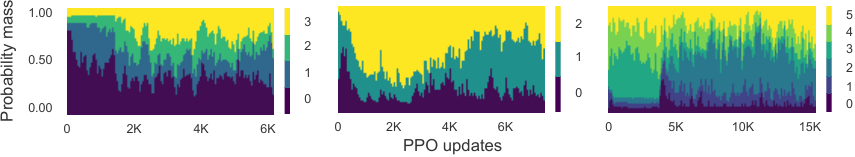}
        \label{fig:minigrid-autocurricula-final}
    \end{subfigure}
    \vspace{-.5cm}
    \caption{\small{Top: Mean episodic test returns of PLR and the uniform-sampling baseline on MultiRoom-N4-Random (4 runs), ObstructedMazeGamut-Easy (3 runs), and ObstructedMazeGamut-Medium (3 runs). Bottom: The probability mass assigned to levels of varying difficulty over the course of training in a single, randomly selected run for the respective environment.}}
    \label{fig:minigrid-results}
\end{figure*}

We provide empirical support for the Value Correction Hypothesis (defined in Section \ref{section:methods}) on two challenging MiniGrid environments, whose levels fall into discrete difficulties (e.g. by number of rooms to be traversed). In both, PLR with rank prioritization significantly improves sample efficiency and generalization over uniform sampling, demonstrating our method also works well in discrete state spaces. We find a staleness coefficient of $\rho = 0.3$ leads to the best test performance on MiniGrid. The top row of Figure~\ref{fig:minigrid-results} summarizes these results.

To test our hypothesis, we bin each level into its corresponding difficulty, expressed as ascending, discrete values (note that PLR does not have access to this privileged information). In the bottom row of Figure~\ref{fig:minigrid-results}, we see how the expected difficulty of levels sampled using PLR changes during training for each environment. We observe that as $P_{\text{replay}}$ is updated, levels become sampled according to an implicit curriculum over the training levels that prioritizes progressively harder levels. 
Of particular note, PLR seems to struggle to discover a useful curriculum for around the first $4{,}000$ updates on ObstructedMazeGamut-Medium, at which point it discovers a curriculum that gradually assigns more weight to harder levels. This curriculum enables PLR with access to only $6{,}000$ training levels to attain even higher mean test returns than the uniform-sampling baseline with access to the full set of training levels, of which there are roughly $4$ billion (so our training levels constitute $0.00015\%$ of the total number).

\subsection{Training on the Full Level Distribution}
\label{section:training_on_full_dist}

\begin{figure}[t!]
\centering
\begin{subfigure}[b]{0.8\linewidth}
    \includegraphics[width=\linewidth]{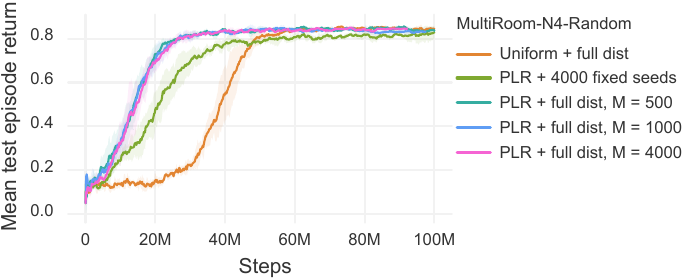}
\end{subfigure}
\begin{subfigure}[b]{0.8\linewidth}
    \includegraphics[width=\linewidth]{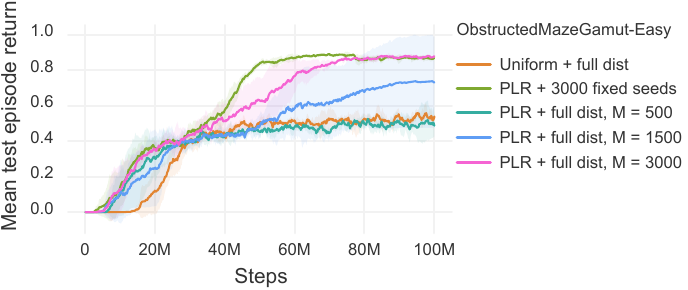}
\end{subfigure}

\caption{\small{Mean test episodic returns on MultiRoom-N4-Random (top) and ObstructedMazeGamut-Easy (bottom) with access to the full level distribution at training. Plots are averaged over 3 runs. We set $P_{D}$ to a Bernoulli parameterized as $p = 0.5$ for MultiRoom-N4-Random and $p=0.95$ for ObstructedMazeGamut-Easy (found via grid search). As with all MiniGrid experiments using PLR, we use rank prioritization, $\beta = 0.1$, and $\rho = 0.3$.}}

\label{fig:minigrid-full-dist}
\end{figure}

While assessing generalization performance calls for using a fixed set of training levels, ideally our method can also make use of the full level distribution if given access to it. We take advantage of an unbounded number of training levels by modifying the list structures for storing scores and timestamps to track the top $M$ levels by learning potential in our finite level buffer (see Algorithm~\ref{alg:plr_update_rule}). When the lists are full, we set the next level for replacement as
\[
l_{\text{min}} = \underset{l}{\arg \min}\;P_{\text{replay}}(l).
\]
When the outcome of the Bernoulli $P_{D}$ entails sampling a new level $l$, the score and timestamps of $l$ replace those of $l_{\text{min}}$ only if the score of $l_{\text{min}}$ is lower than that of $l$. In this way, PLR keeps a running buffer during training of the top $M$ levels appraised to have the highest learning potential for replaying anew.

Figure \ref{fig:minigrid-full-dist} shows that with access to the full level distribution at training, PLR improves sample efficiency and generalization performance in both environments compared to uniform sampling on the full distribution. In MultiRoom-N4-Random, the value $M$ makes little difference to test performance, and training with PLR on the full level distribution leads to a policy outperforming one trained with PLR on a fixed set of training levels. However, on ObstructedMazeGamut-Easy, a smaller $M$ leads to worse test performance. Nevertheless, for all but $M=500$, including the case of a fixed set of 3,000 training levels, PLR leads to better mean test performance than uniform sampling on the full level distribution.

\section{Related Work}
\label{section:plr_related_work}
Several methods for improving generalization in deep RL adapt techniques from supervised learning, including stochastic regularization \citep{igl2019generalization, cobbe2019procgen}, data augmentation \citep{kostrikov2020image, raileanu2021automatic, wang2020mixreg}, and feature distillation \citep{igl2020impact, cobbe2020phasic}. In contrast, PLR modifies only how the next training level is sampled, thereby easily combining with any model or RL algorithm. 

The selective-sampling performed by PLR makes it a form of active learning \citep{cohn1994improving, settles2009active}. Our work also echoes ideas from \citet{DBLP:conf/icml/GravesBMMK17}, who train a multi-armed bandit to choose the next task in multi-task supervised learning, so to maximize gradient-based progress signals. \citet{DBLP:conf/iclr/SharmaJHR18} extend these ideas to multi-task RL, but add the additional requirement of knowing a maximum target return for each task a priori, restricting its applicability to more open-ended 
environment spaces. 
\citet{zhang2020automatic} use an ensemble of value functions for selective goal sampling in the off-policy continuous control setting, which requires prior knowledge of the environment structure to generate candidate goals. Unlike PLR, these methods thus assume the ability to sample tasks or levels based on their structural properties, an assumption that does not hold generally for all PCG simulators. Instead, our method automatically uncovers tractable yet difficult levels, giving rise to a curriculum without prior knowledge of the environment.

A recent theme in the PCG setting explores adaptively generating levels to facilitate learning \citep{powerplay2013, sukhbaatar2017intrinsic, openai2021asymmetric, DBLP:conf/gecco/WangLCS19, enhanced_poet, khalifa2020pcgrl, akkaya2019rubiks, campero2020learning, dennis2020emergent}. Unlike these approaches, our method does not assume control over level generation, requiring only the ability to replay previously visited levels. Further, these methods require parameterizing level generation with additional learning modules. In contrast, our approach does not require such extensions of the environment, for example including teacher-specific action spaces~\citep{campero2020learning}. 
Similar adaptive approaches have focused on the goal-based setting, where the agent policy conditions on a task-specific goal that is adaptively set across training levels in order to facilitate favorable learning dynamics. Otheres have made progress here using generative modeling~\citep{goalgan,Racaniere2020Automated}, latent skill learning~\citep{carml}, and exploiting model disagreement~\citep{NEURIPS2020_566f0ea4}. These methods are less generally applicable than PLR due to their reliance on goal information that is provided before each episode. Moreover many of these methods require a well-behaved, learned generative model.

Most similar to our method, \citet{Matiisen_2020} proposes a teacher-student curriculum learning (TSCL) algorithm that samples training levels by considering the change in episodic returns per level, though they neither design nor test the method for generalization. As shown in Section \ref{subsection:procgen_benchmark}, TSCL provides inconsistent benefits at test time. Recent related work has studied curricula similar to TSCL, but based on changes in task success rate~\citep{kanitscheider2021multi} rather than task returns. A general limitation of such learning progress metrics is the need to track individual values per task variant or level, which may introduce scaling challenges in more open-ended environment spaces. Such TSCL-like curricula typically assume a priori knowledge of a target task set, for which learning progress can be tracked. Unlike these prior TSCL-like approaches, PLR does not assume access to all levels at the start of training, and as we show in Section~\ref{section:training_on_full_dist}, PLR can be extended to improve sample efficiency and generalization by training on an unbounded number of training levels. 

Like our method, \citet{schaul2015prioritized} and \citet{kapturowski2019r2d2} use TD-errors to estimate learning potential. While these methods make use of TD-errors to prioritize learning from  \emph{past} experiences, our method uses such estimates to prioritize revisiting levels for generating entirely new \emph{future} experiences for learning.

Generalization requires sufficient exploration of environment states and dynamics. Thus, recent exploration strategies \citep[e.g.][]{raileanu2020ride, campero2020learning, zhang2020bebold, zha2021rank} shown to benefit simple PCG settings are complementary to the aims of this work. However, as these studies focus on PCG environments with low-dimensional state spaces, whether such methods can be successfully applied to more complex PCG environments like Procgen Benchmark remains to be seen. If so, they may potentially combine with PLR to yield additive improvements. We believe the interplay between such exploration methods and PLR to be a promising direction for future research.

\section{Conclusion and Future Work}
We introduced Prioritized Level Replay (PLR), an algorithm for selectively sampling the next training level in PCG environments based on the estimated learning potential of revisiting each level for the current policy. We showed that our method remarkably improves both the sample efficiency and generalization of deep RL agents in PCG environments, including the majority of environments in Procgen Benchmark and two challenging MiniGrid environments. We further combined PLR with the prior leading method to set a new state-of-the-art on Procgen Benchmark. Further, on MiniGrid environments, we showed PLR induces an emergent curriculum of increasingly more difficult levels.

The flexibility of the PCG abstraction makes PLR applicable to many problems of practical importance, for example, robotic object manipulation tasks, where domain randomized environment instances map to the notion of levels. We believe PLR may even be applicable to singleton environments, given a procedure for generating variations of the underlying MDP as a function of a level identifier, for example, by varying the starting positions of entities. Another natural extension of PLR is to adapt the method to operate in the goal-conditioned setting, by incorporating goals into the level parameterization. 

Despite the wide applicability of PCG and consequently PLR, not all problem domains can be effectively represented in seed-based simulation. The open-ended nature of many real world problem domains, like car driving, cannot be adequately captured by a PCG simulation. Moreover, in such multi-agent settings, realizing a completely faithful simulation would entail solving the very same control problem of interest, as it would require modeling the presence of other agents in the environment of an already suitable skill level, creating a chicken-and-egg dilemma. Combining PLR with self-play autocurricula over co-players may be a promising path for training robust agents in such domains. Further, environment resets are not universally available, such as in the continual learning setting, where the agent interacts with the environment without explicit episode boundaries---arguably, a more realistic interaction model for a learning agent deployed in the wild. Still, pre-training in simulation with resets can nevertheless benefit such settings, where the target domain is rife with open-ended complexity and where resets are unavailable, especially as training through real-world interactions can be slow, expensive, and precarious. For these reasons, in practice, deep RL agents are typically trained in simulation. In more complex domains that are hard to hand-specify, the simulator can conceivably be learned as a world model~\citep{worldmodels,hafner2020dreamer} for the domain of interest. As PLR provides a simple method to more fully exploit the simulator for improved test-time performance, we believe PLR can be adapted to improve learning in these settings.

We further note that while we empirically demonstrated that L1 value loss acts as a highly effective scoring function, there likely exist even more potent choices. Directly learning such functions may reveal even better alternatives. Lastly, combining PLR with various exploration strategies may further improve test performance in hard exploration environments. We look forward to future investigations along these promising directions, prioritized accordingly, by learning potential.

\newcommand{\plrabbrev}{$\textnormal{PLR}^{\perp}$}

\chapter{Dual Curriculum Design}
\label{chapter:dcd}

\section{Introduction}
The training distribution of levels is crucial in learning robust and well-generalizing policies. However, it is not always feasible to specify an appropriate training distribution or a generator thereof. The experiments in Chapter~\ref{chapter:plr} show that PLR provides a way to automatically adapt the distribution over environment variations during training. However, PLR is largely motivated via a heuristic argument centered on viewing TD errors as a proxy for the learning potential of the agent. This chapter seeks to study PLR under a more principled lens, by using ideas from game theory and decision theory. We begin by considering the high-level structure of the PLR algorithm in relation to a concurrently-developed algorithm that produces single-agent autocurricula through the interplay between a student and teacher agent.

While PLR finds useful levels through random search, an  alternative option is to produce levels the levels directly via a generative model. Such an approach would confer greater control over the exact level design. One incarnation of this idea is Protagonist Antagonist Induced Regret Environment Design~\citep[PAIRED,][]{paired}, which trains a teacher agent to generate levels that challenge the student agent throughout training. PAIRED is couched in a self-supervised RL paradigm called Unsupervised Environment Design (UED).
Here, an environment generator (a \emph{teacher}) is co-evolved with a \emph{student} policy that trains on levels actively proposed by the teacher, leading to a form of adaptive curriculum learning. The aim of this coevolution is for the teacher to gradually learn to generate environments that exemplify properties of those that might be encountered at deployment time, and for the student to simultaneously learn a good policy that enables zero-shot transfer to such environments. PAIRED's specific adversarial approach to environment design ensures a useful robustness characterization of the final student policy in the form of a minimax regret guarantee \citep{savage1951theory}---assuming that its underlying teacher-student multi-agent system arrives at a Nash equilibrium~\cite[NE,][]{nash1950equilibrium}. 

In contast, PLR embodies an alternative form of dynamic curriculum learning that does not assume control of level generation. Instead, PLR assumes only the ability to selectively replay existing levels. PLR tracks levels previously proposed by a black-box environment generator, and for each, estimates the agent's learning potential in that level, in terms of how useful it would be to gather new experience from that level again in the future. The PLR algorithm exploits these scores to adapt a schedule for revisiting or \emph{replaying} levels to maximize learning potential. PLR has been shown to produce scalable and robust results, improving both sample complexity of agent training and the generalization of the learned policy in diverse environments. However, unlike PAIRED, PLR is motivated with heuristic arguments and lacks a useful theoretical characterization of its learning behavior.

In this chapter, we demonstrate that PLR is, in and of itself, an effective form of UED: By curating even randomly generated levels, PLR can generate novel, complex levels for learning robust policies. This insight leads to a natural class of UED methods, which we call \emph{Dual Curriculum Design} (DCD). In DCD, a student is challenged by a team of two co-evolving teachers. One teacher actively generates new, challenging levels, while the other passively curates existing levels for replaying, by prioritizing those estimated to be most suitably challenging for the student. We show that PAIRED and PLR are distinct members of the DCD class of algorithms and prove in Section~\ref{sec:dual} that all DCD algorithms enjoy similar minimax regret guarantees to that of PAIRED.
 
We make use of this result to provide the first theoretical characterization of PLR, which immediately suggests a simple yet highly counterintuitive adjustment to PLR: By only training on trajectories in replay levels, PLR becomes provably robust at NE. We call this resulting variant \plrabbrev{} (Section~\ref{sec:robust_plr}). From this perspective, PLR effectively performs level design in a diametrically opposite manner to PAIRED---through prioritized selection rather than active generation. A second corollary to the provable robustness of DCD algorithms shows that \plrabbrev{} can be extended to make use of the PAIRED teacher as a level generator while preserving the robustness guarantee of PAIRED, resulting in a method we call \emph{Replay-Enhanced PAIRED} (REPAIRED) (Section~\ref{sec:repaired}). We hypothesize that in this arrangement, \plrabbrev{} plays a complementary role to PAIRED in robustifying student policies.

Our experiments in Section~\ref{sec:experiments}  investigate the learning dynamics of \plrabbrev{}, REPAIRED, and their replay-free counterparts on a challenging maze domain and a novel continuous control UED setting based on the popular CarRacing environment \citep{gym}. In both of these highly distinct settings, our methods provide significant improvements over PLR and PAIRED,  producing agents that can perform out-of-distribution (OOD) generalization to a variety of human designed mazes and Formula 1 tracks.

\begin{figure*}
    \centering
    \begin{subfigure}{.18\textwidth}
        \includegraphics[width=\textwidth]{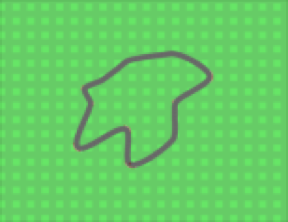}  
        \caption{DR}
    \end{subfigure}
    \begin{subfigure}[b]{.18\textwidth}
        \includegraphics[width=\textwidth]{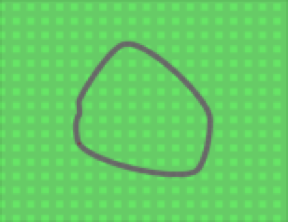}
        \caption{PAIRED}
    \end{subfigure}
    \begin{subfigure}[b]{.18\textwidth}
        \includegraphics[width=\textwidth]{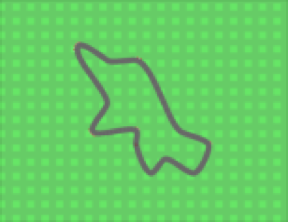}
        \caption{REPAIRED}
    \end{subfigure}
    \begin{subfigure}[b]{.18\textwidth}
        \includegraphics[width=\textwidth]{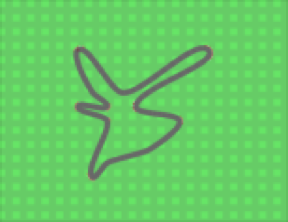}
        \caption{\plrabbrev{}}
    \end{subfigure}
    \begin{subfigure}[b]{.18\textwidth}
        \includegraphics[width=\textwidth]{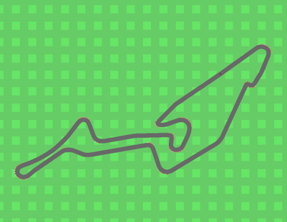}
        \caption{Human}
    \end{subfigure}
    \caption{\small{Randomly drawn samples of CarRacing tracks produced by different methods. (a)~Domain Randomization (DR) produces tracks of average complexity, with few sharp turns. (b)~PAIRED often overexploits the difference in the students, leading to simple tracks that incidentally favor the antagonist. (c)~REPAIRED mitigates this degeneracy, recovering track complexity. (d)~\plrabbrev{} selects the most challenging randomly generated tracks, resulting in tracks that more closely resemble human-designed tracks, such as  (e)~the N\"{u}rburgring Grand Prix.
    }}
    \label{figure:carracing_pics}
\end{figure*}

In summary, we present the following contributions in this chapter:
(i) We establish a common framework, Dual Curriculum Design, that encompasses PLR and PAIRED. This allows us to develop new theory, which provides the first robustness guarantees for PLR at NE as well as for REPAIRED, which augments PAIRED with a PLR-based replay mechanism.
(ii) Crucially, our theory suggests a highly counterintuitive improvement to PLR: the convergence to NE should be assisted by training on less data when using PLR---namely by only taking gradient updates from data that originates from the PLR buffer, using the samples from the environment distribution only for computing the prioritization of levels in the buffer.
(iii) Our experiments in a maze domain and a novel car racing domain show that our methods significantly outperform their replay-free counterparts in zero-shot generalization. We open source our methods at 
\url{https://github.com/facebookresearch/dcd}.

\raggedcolumns
\newpage

\section{Robustness in Dual Curriculum Design}
\label{sec:dual}
\begin{figure}[h]
    \centering{\includegraphics[width=1\linewidth]{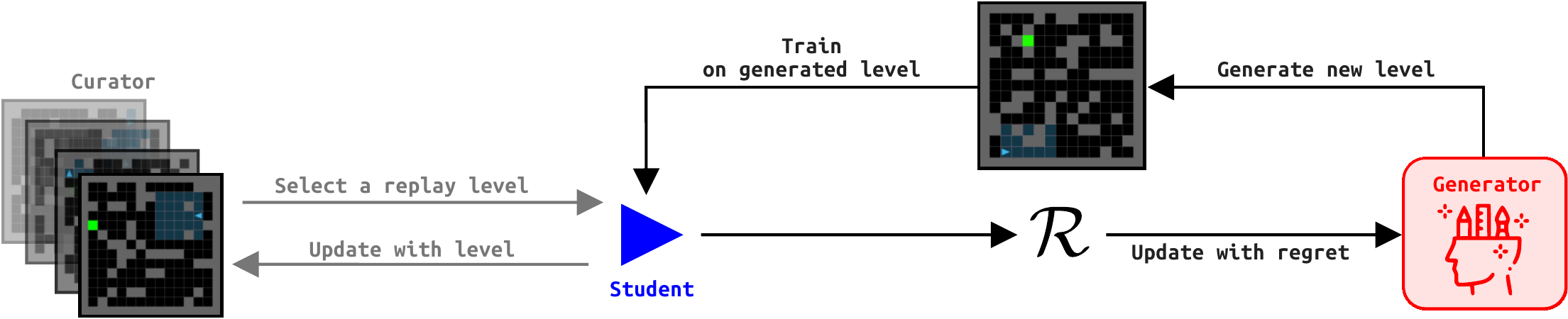}}
    \caption{\small{Overview of Dual Curriculum Design (DCD). The student learns in the presence of two co-adapting teachers that aim to maximize the student's regret: The generator teacher designs new levels to challenge the agent, and the curator teacher prioritizes a set of levels already created, selectively sampling them for replay.}} 
    \label{figure:repaired_overview}
\end{figure}

The previous approaches of PAIRED and PLR reveal a natural duality: Approaches that gradually learn to generate  levels like PAIRED, and methods which cannot generate levels, but instead, quickly curate existing ones, like PLR. This duality suggests combining slow level generators with fast level curators. We call this novel class of UED algorithms Dual Curriculum Design (DCD). For instance, PLR can be seen as curator with a prioritized sampling mechanism with a random generator, while PAIRED, as a regret-maximizing generator without a curator.
DCD can further consider Domain Randomization (DR) as a degenerate case of a random level generator without a curator.

To theoretically analyze this space of methods, we model DCD as a three player game among a student agent and two teachers called the \emph{dual curriculum game}. However, to formalize this game, we must first formalize the single-teacher setting: Suppose the UPOMDP is clear from context. Then, given a utility function for a single teacher, $U_t(\pi,\theta)$, we can naturally define the \emph{base game} between the student $s$ and teacher $t$ as $G = \langle S = S_s \times S_t, U = U_s \times U_t \rangle$, where $S_s= \Pi$ is the strategy set of the student, $S_t = \Theta$ is the strategy set of the teacher, and $U_s(\pi, \theta) = V^\theta(\pi)$ is the utility function of the student.  In Sections \ref{sec:robust_plr} and \ref{sec:repaired}, we will study settings corresponding to different choices of utility functions for the teacher agents, namely the maximum-regret objective $U_t^R(\pi, \theta)$ and the uniform objective $U_t^U(\pi, \theta)$. These two objectives are defined as follows (for any constant $C$):
\begin{align}
U_t^R(\pi, \theta) &= \argmax_{\pi^* \in \Pi}\{V^\theta(\pi^*)-V^\theta(\pi)\} \\
U_t^U(\pi, \theta) &= C
\end{align} 

In the dual curriculum game $\overline{G}$, the first teacher plays the game with probability $p$, and the second, with probability $(1-p)$---or more formally, $\overline{G} = \langle \overline{S} = S_s \times S_t \times S_t, U = \overline{U}_s \times \overline{U}^1_t \times \overline{U}^2_t\rangle$, where the utility functions for the student and two teachers respectively, $\overline{U}_s, \overline{U}^1_t, \overline{U}^2_t$, are defined as follows:
\begin{align}
\overline{U}^1_t(\pi, \theta^1, \theta^2) &= p U_t^1(\pi, \theta^1) \\
\overline{U}^2_t(\pi, \theta^1, \theta^2)  &= (1-p)U_t^2(\pi, \theta^2) \\
\overline{U}_s(\pi, \theta^1, \theta^2)  &= p U_s(\pi, \theta^1) + (1-p)U_s(\pi, \theta^2)
\end{align}

Our main theoretical result, summarized by Theorem~\ref{theorem:main} in Section~\ref{sec:dcd_theory}, is that NE in the dual curriculum game are approximate NE of both the base game for either of the original teachers and the base game with a teacher maximizing the joint-reward of $pU^1_t + (1-p)U^2_t$, where the quality of the approximations depends on the mixing probability $p$.

The intuition behind this theorem is that, since the two teachers do not affect each other's behavior, their best response to a fixed $\pi_s$ is to choose a strategy $\theta$ that maximizes $U_t^1$ and $U_t^2$ respectively. Moreover, the two teachers' strategies can be viewed as a single combined strategy for the base game with the joint-objective, or with each teacher's own objective. In fact, the teachers provide an approximate best-response to each case of the base game simply by playing their individual best responses. Thus, when we reach a NE of the dual curriculum game, the teachers arrive at approximate best responses for both the base game with the joint objective and with their own objectives, meaning they are also in an approximate NE of the base game with either teacher. The full proof of this result is presented in Section~\ref{sec:dcd_theory}.

\section{Robustifying PLR}
In this section, we provide theoretical justification for the empirically observed effectiveness of PLR, and in the process, motivate a counterintuitive adjustment to the algorithm.

\label{sec:robust_plr}

\begin{figure}[h!] 
\vskip -0.1in
\begin{minipage}{\linewidth}
\begin{algorithm}[H]
{\small
\label{alg:plr+}
\SetAlgoLined
\caption{Robust PLR (PLR$^\bot$)}
Randomly initialize policy $\pi(\phi)$ and an empty level buffer, $\bm{\Lambda}$ of size $K$. \\
    \While{not converged}{
        Sample replay-decision Bernoulli, $d \sim P_{D}(d)$ \\
        \eIf{$d=0$}{
            Sample level $\theta$ from level generator\\
            Collect $\pi$'s trajectory $\tau$ on $\theta$, {\color{blue} with a stop-gradient $\phi_{\bot}$} \hfill {\color{gray} \emph{i.e. Suppress policy update}} 
        }
        {
        Use PLR to sample a replay level from the level store, $\theta \sim \bm{\Lambda}$ \\
        Collect policy trajectory $\tau$ on $\theta$ and update $\pi$ with rewards $\bm{R}(\tau)$
        }
        
        Compute PLR score, $S = \textbf{score}(\tau, \pi)$ \\
        Update $\bm{\Lambda}$ with $\theta$ using score $S$
    }
}
\end{algorithm}
\end{minipage}
\end{figure}

\subsection{Achieving Robustness Guarantees with PLR}
PLR provides strong empirical gains in generalization, but lacks any theoretical guarantees of robustness. One step towards achieving such a guarantee is to replace its L1 value-loss prioritizaton with a regret prioritization, using the methods we discuss in Section \ref{subsec:estimating_regret}: While L1 value loss may be good for quickly training the value function, it can bias the long-term training behavior toward high-variance policies. However, even with this change, PLR holds weaker theoretical guarantees because the random generating teacher can bias the student away from minimax regret policies and instead, toward policies that sacrifice robustness in order to excel in unstructured levels. We formalize this intuitive argument in Section~\ref{sec:dcd_theory} as Corollary~\ref{corollary:plr}. This result follows from a direct application of Theorem \ref{theorem:main} to show that a NE of $\overline{G}$ is an approximate NE for the base game of the first teacher, and through constructing a simple example where the student's best response in $\overline{G}$ fails to attain the minimax regret in $G$. These arguments are described in full in Section~\ref{sec:dcd_theory}. This corollary provides some justification for why PLR improves robustness of the equilibrium policy, as it biases the resulting policy toward a minimax regret policy.  However, it also points a way towards further improving PLR: If the probability $p$ of using a teacher-generated level directly was set to $0$, then in equilibrium, the resulting policy converges to a minimax regret policy.
Consequently, we arrive at the counterintuitive idea of avoiding gradient updates from trajectories collected from randomly sampled levels, to ensure that at NE, we find a minimax regret policy. From a robustness standpoint, it is therefore optimal to train on less data. The modified PLR algorithm \plrabbrev{} with this counterintuitive adjustment is summarized in Algorithm \ref{alg:plr+}, in which this small change relative to the original algorithm is highlighted in {\color{blue} blue}.

\subsection{Estimating Regret}
\label{subsec:estimating_regret}
In general, levels may differ in maximum achievable returns, making it impossible to know the true regret of a level without access to an oracle. As the L1 value loss typically employed by PLR does not generally correspond to regret, we turn to alternative scoring functions that better approximate regret. Two approaches, both effective in practice, are discussed below.

\medskip
\noindent \textbf{Positive Value Loss (PVL):\medspace}Averaging over all transitions with positive value loss amounts to estimating regret as the difference between maximum achieved return and predicted return on an episodic basis. However, this estimate is highly biased, as the value targets are tied to the agent's current, potentially suboptimal policy. As it only considers positive value losses, this scoring function leads to optimistic sampling of levels with respect to the current policy. When using GAE \citep{gae} to estimate bootstrapped value targets, this loss takes the following form, where $\lambda$ and $\gamma$ are the GAE and MDP discount factors respectively, and $\delta_t$, the TD-error at timestep $t$:
\begin{equation}
\label{eq:pvl}
\frac{1}{T}\sum_{t=0}^{T} \max \left(\sum_{k=t}^T(\gamma\lambda)^{k-t}\delta_k, 0\right).
\end{equation}

\medskip
\noindent \textbf{Maximum Monte Carlo (MaxMC):\medspace} We can mitigate some of the bias of the positive value loss by replacing the value target with the highest return achieved on the given level so far during training. By using this maximal return, the regret estimates no longer depend on the agent's current policy. This estimator takes the simple form of $(1/T)\sum_{t=0}^{T} R_{\rm{max}} - V(s_t)$. In our dense-reward experiments, we compute this score as the difference between the maximum achieved return and $V(s_0)$.

\begin{figure}[h!] 
\vskip -0.1in
\begin{minipage}{\linewidth}
\begin{algorithm}[H]
{
\label{alg:repaired}
\SetAlgoLined
\caption{REPAIRED}

Randomly initialize Protagonist, Antagonist, and Generator policies $\pi^A(\phi^A)$, $\pi^B(\phi^B)$, and $\tilde{\theta}$ \\
Initialize Protagonist and Antagonist PLR level buffers $\bm{\Lambda}^A$ and $\bm{\Lambda}^B$ \\
    \While{not converged}{
        Sample replay-decision Bernoulli, $d \sim P_{D}(d)$ \\
        \eIf{$d = 0$}{
            Teacher policy $\tilde{\theta}$ generates the next level, $\theta$ \\
            Set $\theta^A = \theta^B = \theta$ \\
            Collect trajectory $\tau^A$ on $\theta^A$ and $\tau^B$ on $\theta^B$ with stop-gradients $\phi^A_{\bot}$, $\phi^B_{\bot}$ \\
            Update $\tilde{\theta}$ with $\textsc{Regret}^{\theta}(\pi^A, \pi^B)$ \\
        } 
        {
            PLR samples replay levels, $\theta^A \sim \bm{\Lambda}^A$ and $\theta^B \sim \bm{\Lambda}^B$ \\
            Collect trajectory $\tau^A$ on $\theta^A$ and $\tau^B$ on $\theta^B$ \\
            Update $\pi^A$ with rewards $\bm{R}(\tau^A)$, and $\pi^B$, with rewards $\bm{R}(\tau^B)$ \\
        }

        Compute PLR score $S^A = \textbf{score}(\tau^A, \tau^B, \pi^A)$ \\ Compute PLR score $S^B = \textbf{score}(\tau^B, \tau^A, \pi^B)$ \\
        Update $\bm{\Lambda}^A$ with $\theta^A$ using score $S^A$ \\
        Update $\bm{\Lambda}^B$ with $\theta^B$ using score $S^B$ \\
    }
}
\end{algorithm}
\end{minipage}
\end{figure}

\section{Replay-Enhanced PAIRED (REPAIRED)}
\label{sec:repaired}

We can replace the random generator teacher used by \plrabbrev{} with the PAIRED teacher. This extension entails a second student agent, the antagonist, also equipped with its own PLR level buffer.
In each episode, with probability $p$, the students evaluate their performances (but do not train) on a newly generated level and, with probability $1-p$, train on a level sampled from each student's own regret-prioritizing PLR buffer. Training only on the highest regret levels should mitigate inefficiencies in the PAIRED teacher's optimization procedure. We refer to this extension as \emph{Replay-Enhanced PAIRED} (REPAIRED), depicted by the black arrows in Figure~\ref{figure:repaired_overview}, with the students being the protagonist and antagonist, while the full pseudocode is outlined in Algorithm~\ref{alg:repaired}.

Since \plrabbrev{} and PAIRED both promote regret in equilibrium, it is reasonable to believe that the combination of the two does the same. A straightforward corollary of Theorem \ref{theorem:main} (stated and proven as Corollary~\ref{corollary:repaired} in Section~\ref{sec:dcd_theory}), shows that in a theoretically ideal setting, combining these two algorithms as is done in REPAIRED indeed finds minimax regret strategies in equilibrium.

This result gives us some amount of assurance that, if our method arrives at NE, then the protagonist has converged to a minimax regret strategy, which has the benefits outlined in~\citep{paired}: Since a minimax regret policy solves all solvable environments, whenever this is possible and sufficiently well-defined, we should expect policies resulting from the equilibrium behavior of REPAIRED to be robust and versatile across all environments in the domain.

\section{Theoretical Results}
\label{sec:dcd_theory}
In this section we prove the theoretical results around the dual curriculum game and use these results to show approximation bounds for our methods, given that they have reached a Nash equilibrium (NE).

The first theorem is the main result that allows us to analyze dual curriculum games. The high-level result says that the NE of a dual curriculum game are approximate NE of the base game from the perspective of any of the individual players, or from the perspective of the joint strategy.

\newpage 

\newtheorem{main_theorem}{Theorem}
\setcounter{main_theorem}{0}
\begin{main_theorem}
\label{theorem:main}
Let $B$ be the maximum difference between $U^1_t$ and $U^2_t$, and let $(\pi, \theta^1, \theta^2)$ be a NE for $\overline{G}$.  Then $(\pi, p\theta^1+(1-p)\theta^2)$ is an approximate NE for the base game with either teacher or for a teacher optimizing their joint objective. More precisely, it is a $2Bp(1-p)$-approximate NE when $U_t = pU^1_t+(1-p)U^2_t$, a $2B(1-p)$-approximate NE when $U_t = U^1_t$, and a $2Bp$-approximate NE when $U_t = U^2_t$.
\end{main_theorem}

At a high level, this is true because, for low values of $p$, the best-response strategies for the individual players can be thought of as approximate-best response strategies for the joint-player, and vis-versa.  Since the Nash Equilibrium consists of each of the players playing their own best response, they must be playing an approximate best response for the joint-player. We provide a formal proof below:

\begin{proof}
Let $B$ be the maximum difference between $U^1_t$ and $U^2_t$, and let $(\pi, \theta^1, \theta^2)$ be a Nash Equilibrium for $\overline{G}$.  Then consider $p\theta^1+(1-p)\theta^2$ as a strategy in the base game for the joint player $pU^1_t+(1-p)U^2_t$.  Let $\theta^{1+2}$ be the best response for the joint player to $\pi$. Since $\pi$ is a best response by assumption, it is sufficient to show that $p\theta^1+(1-p)\theta^2$ is an approximate best response. We then have
\begin{align}
    U_t(\pi,p\theta^1+(1-p)\theta^2) &= p^2U^1_t(\pi,\theta^1)+p(1-p)U^2_t(\pi,\theta^1) \\
    &\quad + p(1-p)U^1_t(\pi,\theta^2)+(1-p)^2U^2_t(\pi,\theta^2) \nonumber \\
    &\geq p^2U^1_t(\pi,\theta^1) \\
    &\quad +p(1-p)(U^1_t(\pi,\theta^1)-B) \nonumber \\
    &\quad +p(1-p)(U^2_t(\pi,\theta^2)-B) \nonumber \\
    &\quad +(1-p)^2U^2_t(\pi,\theta^2) \nonumber \\
    &= pU^1_t(\pi,\theta^1)+(1-p)U^2_t(\pi,\theta^2)-2Bp(1-p)\\
    &\geq U_t(\pi,\theta^{1+2})-2Bp(1-p)
\end{align}

Thus, we have shown that $(\pi,p\theta^1+(1-p)\theta^2)$ represents an $2Bp(1-p)$-Nash equilibrium for the joint player. For the first teacher we have the opposite condition trivially, the teacher is doing a best response to the student. We must now show that the student is doing an approximate best response to the teacher.

Let $\pi^{1}$ be the best response to the first teacher (with utility $U^1_t$) and let $\pi^{1+2}$ be the best response policy to the joint teacher.  In this argument we will start with the observation that $U_s(\pi^1, \theta^{1+2}) \leq U_s(\pi^{1+2}, \theta^{1+2})$ by definition, and then argue that we can construct an upper bound on the performance of $\pi^1$ on $\theta^1$, $U_s(\pi^1,\theta^1)$, and a lower bound on the performance of $\pi^{1+2}$ on $\theta^1$, $U_s(\pi^{1+2},\theta^1)$.  We get the desired result by combining these two arguments.

First we use $U_s(\pi^1, \theta^{1+2})$ to upper bound $U_s(\pi^1,\theta^1)$: 
\begin{align}
    U_s(\pi^1, \theta^{1+2}) &= p U_s(\pi^1, \theta^1) + (1-p) U_s(\pi^1, \theta^2) \\
    &\geq p U_s(\pi^1, \theta^1) + (1-p) (U_s(\pi^1, \theta^1) - B) \\
    &= U_s(\pi^1, \theta^1) - (1-p) B
\end{align}

Second we can use $U_s(\pi^{1+2}, \theta^{1+2})$ to lower bound $U_s(\pi^{1+2},\theta^1)$:
\begin{align}
    U_s(\pi^{1+2}, \theta^{1+2}) &= p U_s(\pi^{1+2}, \theta^1) + (1-p) U_s(\pi^{1+2}, \theta^2) \\
    &\leq p U_s(\pi^{1+2}, \theta^1) + (1-p) (U_s(\pi^{1+2}, \theta^1) + B) \\
    &= U_s(\pi^{1+2}, \theta^1) + (1-p) B
\end{align}

Putting this all together, we have
\[
 U_s(\pi^{1+2}, \theta^1) + (1-p) B \geq U_s(\pi^1, \theta^1) - (1-p) B. 
\]

Which, after rearranging terms, gives 
\[
 U_s(\pi^{1+2}, \theta^1) \geq U_s(\pi^1, \theta^1) - 2 (1-p) B
\]
as desired.  Repeating the symmetric argument shows the desired property for the second teacher.
\end{proof}

We can apply Theorem~\ref{theorem:main} to both standard PLR and REPAIRED. Standard PLR trains on a mixture of a uniformly random teacher (DR) with utility function $U_t^C$ and the PLR teacher with utility function $U_t^R$. Intuitively, applying Theorem~\ref{theorem:main} to PLR then shows that as we reduce the number of random teacher episodes, the approximation to a minimax regret strategy improves. Consequently, this approximation becomes exact when the number of random teacher episodes goes to zero, thereby motivating \plrabbrev{}. In the discussion that follows, this argument is formalized in the proof of Corollary~\ref{corollary:plr}. In the case of REPAIRED, in which both teachers are regret-maximizing, Theorem~\ref{theorem:main} shows that the student must follow a minimax regret strategy at NE. This result is formally stated and proven as Corollary~\ref{corollary:repaired}.

\newtheorem{plr-corollary}{Corollary}
\setcounter{plr-corollary}{0}
\begin{plr-corollary}
\label{corollary:plr}
Let $\overline{G}$ be the dual curriculum game in which the first teacher maximizes regret, so $U^1_t = U^R_t$, and the second teacher plays randomly, so  $U^2_t = U^U_t$.  Let $V^\theta(\pi)$ be bounded in $[B^-,B^+]$ for all $\theta, \pi$.  Further, suppose that $(\pi, \theta^1,\theta^2)$ is a Nash equilibrium of $\overline{G}$.  Let $R^* = \min_{\pi_A \in \Pi}\{\max_{\theta,\pi_B \in \Theta , \Pi}\{\textsc{Regret}^{\theta}(\pi_A,\pi_B)\}\}$ be the optimal worst-case regret.  Then $\pi$ is $2(B^+-B^-)(1-p)$ close to having optimal worst-case regret, or formally, $\max_{\theta,\pi_B \in \Theta , \Pi}\{\textsc{Regret}^{\theta}(\pi_A,\pi)\} \geq R^* - 2(B^+-B^-)(1-p)$.  Moreover, there exists environments for all values of $p$ within a constant factor of achieving this bound.
\end{plr-corollary}

\begin{proof}
Since $V^\theta(\pi)$ is bounded in $[B^-,B^+]$ for all $\theta, \pi$, we know that  $U^1_t$ and $U^2_t$ are within $(B^+ - B^-)$ of each other.  Thus by Theorem \ref{theorem:main} we have that $(\pi, \theta^1,\theta^2)$ is a $2(B^+ - B^-)(1-p)$-Nash equilibrium of the base game when $U_t = U_t^1$. Thus $\pi$ is a $2(B^+ - B^-)(1-p)$ approximate best-response to $\theta^1$.  However, since $\theta^1$ is a best response it chooses a regret maximizing parameter distribution.  Thus the $2(B^+ - B^-)(1-p)$ does not just measure the sub-optimally of $\pi$ with respect to $\theta^1$, but the worst-case regret of $\pi$ across all $\theta$, as desired.

The intuition for the existence of examples in which this approximation of regret decays linearly in $p$ is that a random level and the maximal regret level can be very different, and so the two measures may diverge drastically.  For an example environment where $\pi$ deviates strongly from the minimax regret strategy, consider the one-step UMDP described in Table \ref{table:edge_case_game}.

\begin{table}[h!]
\begin{center}
\begin{tabular}{ c | c | c  | c  }
 & $\theta_0$ & $\theta_1$ & $\theta_2 \dotsc \theta_n$ \\
\hline
$\pi_0$ & $B$ & $0$ & $0$ \\
\hline
$\pi_1$ & $0$ & $B$ & $0$ \\
\hline
$\pi_2$ & $Bp+2\epsilon$ & $0$ & $\frac{Bp}{2}+\epsilon$ \\
\hline
$\pi_3$ & $0$ & $Bp+2\epsilon$ & $\frac{Bp}{2}+\epsilon$ \\
\end{tabular}
\caption{\small{In this environment all payoffs are between $0$ and $B$(for $p \in (0,1)$ and $\epsilon<\frac{B(1-p)}{2}$), where $B$ is assumed to be positive.  Randomizing between $\pi_0$ and $\pi_1$ minimizes regret, but choosing $\pi_2$ or $\pi_3$ is better in expectation under the uniform distribution.  For large $n$ it is especially clear that $\pi_2$ and $\pi_3$ have better expected value under the uniform distribution, though we show that even for $n=2$, the optimal joint policy can mix between $\pi_2$ and $\pi_3$ incurring high regret.}}
\label{table:edge_case_game}
\end{center}
\end{table}

Note that in Table \ref{table:edge_case_game}, no policy has less than $\frac{B}{2}$ regret, since every policy will have to incur $B$ regret on either $\{\theta_0,\theta_1\}$ at least half the time. The minimax regret policy mixes uniformly between $\pi_0$ and $\pi_1$ to achieve regret of exactly $\frac{B}{2}$.  We can ignore $\theta_2 \dotsc \theta_n$ for the regret calculations by assuming that $\epsilon < \frac{B(1-p)}{2}$, since every policy achieves less than $\frac{B}{2}$ regret on these levels.

Our claim is that in equilibrium of $\overline{G}$ in this environment, the student policy can incur $\frac{B}{2} + \frac{B(1-p)}{2}-\epsilon$ regret, which is $\frac{B(1-p)}{2}- \epsilon$ more than the minimax regret policy.  An example of such an equilibrium point would be when the student policy uniformly randomizes between $\pi_2$ and $\pi_3$, which we will call $\pi_{2+3}$,  when the minimax teacher uniformly randomizes between $\theta_{0}$ and $\theta_{1}$ which we will call $\theta_{0+1}$, and when the uniform teacher randomizes exactly which we call $\tilde{\theta}$.  To check this we must show that $(\pi_{2+3}, \theta_{0+1}, \tilde{\theta})$ is in fact a NE of $\overline{G}$.  Then we must show that $\pi_{2+3}$ incurs $\frac{B}{2} + \frac{B(1-p)}{2}-\epsilon$ regret.  

To show that $(\pi_{2+3}, \theta_{0+1}, \tilde{\theta})$ is a NE of $\overline{G}$ first note that $\tilde{\theta}$ is trivially a best response for the uniform utility function.  Also note that $\theta_{0+1}$ maximizes the regret of $\pi_{2+3}$ since $\theta_0$ and $\theta_1$ are the only two parameters on which $\pi_{2+3}$ incur regret, and they incur the same regret; thus, any mixture over them will be optimal for the regret-based teacher.  Finally, we need to show that $\pi_{2+3}$ is optimal for the student.  To do this we will calculate the expected value of each policy and notice that the expected values for $\pi_2$ and $\pi_3$ are higher than for $\pi_0$ and $\pi_1$.  Thus any optimal policy will place no weight on $\pi_0$ and $\pi_1$, but any distribution over $\pi_2$ and $\pi_3$ will be equivalently optimal. 
By symmetry, we can show only the calculations for $\pi_0$ and $\pi_2$:
\begin{align}
    \pi_0 &= p(\frac{1}{2}B + \frac{1}{2}0)+ (1-p) 0 = \frac{Bp}{2} \\
    \pi_2 &= p(\frac{1}{2} (Bp+2\epsilon) + \frac{1}{2}0) + (1-p)(\frac{Bp}{2}+\epsilon) = \frac{Bp}{2}+\epsilon
\end{align}

Thus $\pi_2$ and $\pi_3$ achieve $\epsilon$ higher expected value by the joint distribution.  Thus, we know that $\pi_{2+3}$ is a best response and $(\pi_{2+3}, \theta_{0+1}, \tilde{\theta})$ is in fact a NE of $\overline{G}$.   

Finally, we simply need to show that $\pi_{2+3}$ incurs $\frac{B}{2} + \frac{B(1-p)}{2}-\epsilon$ regret. WLOG, we can evaluate its regret on $\theta_0$.  On $\theta_0$, $\pi_{2+3}$ achieves $\frac{Bp}{2}+\epsilon$ reward while $\pi_0$ achieves $B$.  Thus $\pi_{2+3}$ incurs regret of $B - (\frac{Bp}{2}+\epsilon) = \frac{B}{2} + \frac{B - Bp}{2}-\epsilon = \frac{B}{2} + \frac{B(1-p)}{2}-\epsilon$ as desired.  As discussed before, since the minimax regret policy achieves $\frac{B}{2}$, this is $\frac{B(1-p)}{2}-\epsilon$ more regret than optimal. 
\end{proof}

Lastly, we can also apply Theorem~\ref{theorem:main} to prove that REPAIRED achieves a minimax regret strategy at NE. The intuition behind this corollary is that, since the utility functions of both teachers are the same, the approximate NE ensured by Theorem \ref{theorem:main} is actually a true NE; there the minimax theorem applies.  

\newtheorem{repaired-corollary}[]{Corollary}
\setcounter{repaired-corollary}{1}
\begin{repaired-corollary}
\label{corollary:repaired}
Let $\overline{G}$ be the dual curriculum game in which both teachers maximize regret, so $U^1_t = U^2_t = U^R_t$. Further, suppose that $(\pi, \theta^1,\theta^2)$ is a Nash equilibrium of $\overline{G}$. Then, $\pi \in \argmin_{\pi_A \in \Pi}\{\max_{\theta,\pi_B \in \Theta , \Pi}\{\textsc{Regret}^{\theta}(\pi_A,\pi_B)\}\}$.
\end{repaired-corollary}
\begin{proof}
Since $U^1_t = U^2_t = U^R_t$ the joint objective is $p U^1_t + (1-p)U^2_t = U^R_t$.  Note that since $U^1_t = U^2_t$, $B=0$.  Thus by Theorem \ref{theorem:main} $(\pi, p\theta^1+(1-p)\theta^2)$ is a $0$-Nash Equilibrium of the base game with teacher objective $U^R_t$, thus by the minimax theorem, $\pi \in \argmin_{\pi_A \in \Pi}\{\max_{\theta,\pi_B \in \Theta , \Pi}\{\textsc{Regret}^{\theta}(\pi_A,\pi_B)\}\}$ as desired.
\end{proof}

\section{Experiments}
\label{sec:experiments}

Our experiments firstly aim to (1) assess the empirical performance of the theoretically motivated \plrabbrev{}, and secondly, seek to better understand the effect of replay on unsupervised environment design, specifically (2) its impact on the zero-shot generalization performance of the induced student policies, and (3) the complexity of the levels designed by the teacher. To do so, we compare PLR and REPAIRED against their replay-free counterparts, DR and PAIRED in two challenging environments. As we seek comparison with key baselines, like PAIRED, which require direct control of environment generation, we cannot make use of the Procgen Benchmark, featured in Chapter~\ref{chapter:plr}. Instead, we use the extended version of the maze domain introduced in \citet{dennis2020emergent}. To further test our methods outside of discrete environments, we turn to a continuous-control car racing environment, with pixel-based observations and dense rewards. We provide full environment details in Appendices~\ref{appendix:env_maze}–\ref{appendix:env_carracing} and model and hyperparameter choices in Appendix~\ref{appendix:exp_dcd}. 

\subsection{Partially-Observable Navigation}
\label{sec:exp_minigrid}

Each navigation level is a partially-observable maze requiring student agents to take discrete actions to reach a goal and receive a sparse reward. Our agents use PPO~\citep{schulman2017proximal} with an LSTM-based recurrent policy to handle partial observability. Before each episode, the teacher designs the level in this order: beginning with an empty maze, it places one obstructing block per time step up to a predefined block budget, and finally places the agent followed by the goal.

\begin{figure}[h]
    \includegraphics[width=1\linewidth]{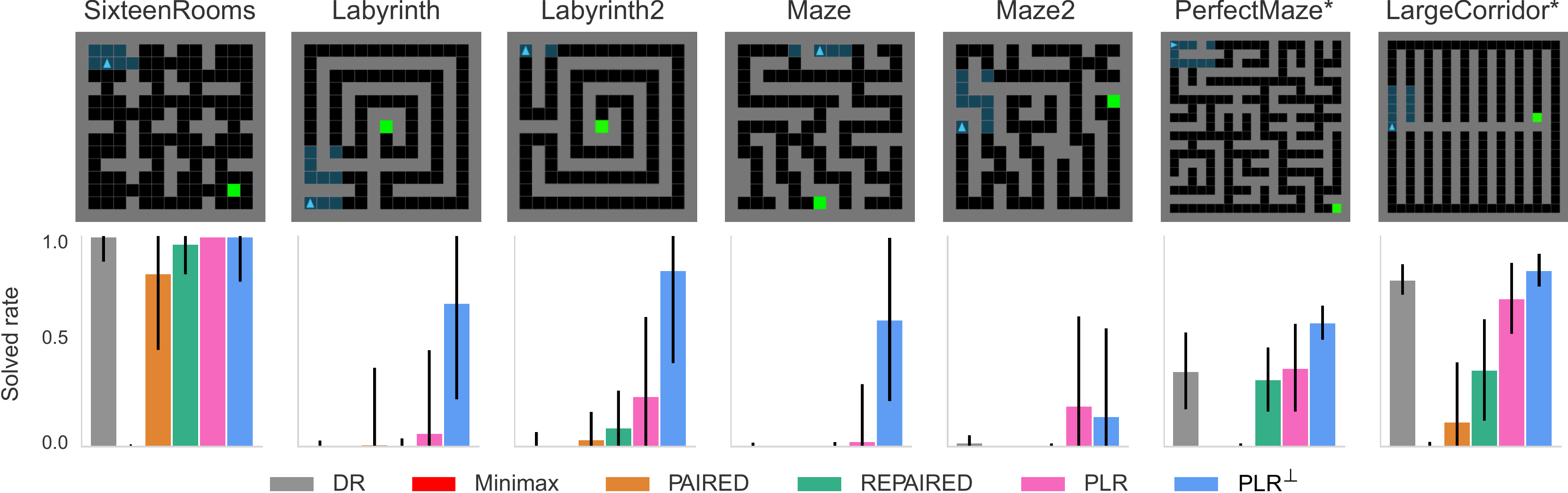}
    \caption{\small{Zero-shot transfer performance in challenging test environments after 250M training steps. The plots show median and interquartile range of solved rates over 10 runs. An asterisk (*) next to the maze name indicates the maze is procedurally-generated, and thus each attempt corresponds to a random configuration of the maze.}}
    \label{figure:main_minigrid}
\end{figure}

\medskip
\noindent \textbf{Zero-shot generalization:} We train policies with each method for 250M steps and evaluate zero-shot generalization on several challenging, human-designed OOD environments, in addition to levels from the full distribution of two procedurally-generated environments, PerfectMaze and LargeCorridor (See Appendix~\ref{appendix:env_maze} for a full description of these test environments). We also compare against DR and minimax baselines. 

Unlike the original maze experiments used to evaluate PAIRED~\citep{paired}, we conduct our main maze experiments with a block budget of 25 blocks (reported in Section \ref{sec:exp_minigrid}), rather than 50 blocks. Following the environment parameterization in \citet{paired}, for a block budget of $B$, the teacher attempts to place $B$ blocks that act as obstacles when designing each maze level. However, the teacher can place fewer than $B$ blocks, as placing a block in a location already occupied by a block results in a no-opt. We found that PAIRED underperforms DR when both methods are given a budget of 50 blocks, a setting in which randomly sampled mazes exhibit enough structural complexity to allow DR to learn highly robust policies. Note that \citet{paired} used a DR baseline with a 25-block budget. With a 50-block budget, DR and all replay-based methods are able to fully solve nearly all test mazes after around 500M steps of training, making UED of mazes with a 50-block budget too simple of a setting to provide an informative comparison among the methods studied. We thus focus on the more challenging 25-block setting.

In assessing our experimental results, we test for statistical significance in differences between methods via the Welch t-test \citep{welch1947generalization}. We report the results of evaluating policies produced by each method after 250M training steps on each of the zero-shot transfer environments in Figure~\ref{figure:main_minigrid} and Table~\ref{table:minigrid_eval}. Each test environment is visualized in Figure~\ref{fig:minigrid_ood_envs}. All replay-based UED methods lead to policies with statistically significantly ($p < 0.05$) higher test performance than PAIRED, and \plrabbrev{}, after 500M training steps, similarly improves over PLR when trained for an equivalent number of gradient updates (as replay rate is set to $0.5$). Note that for PAIRED and REPAIRED, we evaluate the protagonist policy, which we refer to as the student.

\begin{figure}[t!]
    \centering
    \includegraphics[width=1\linewidth]{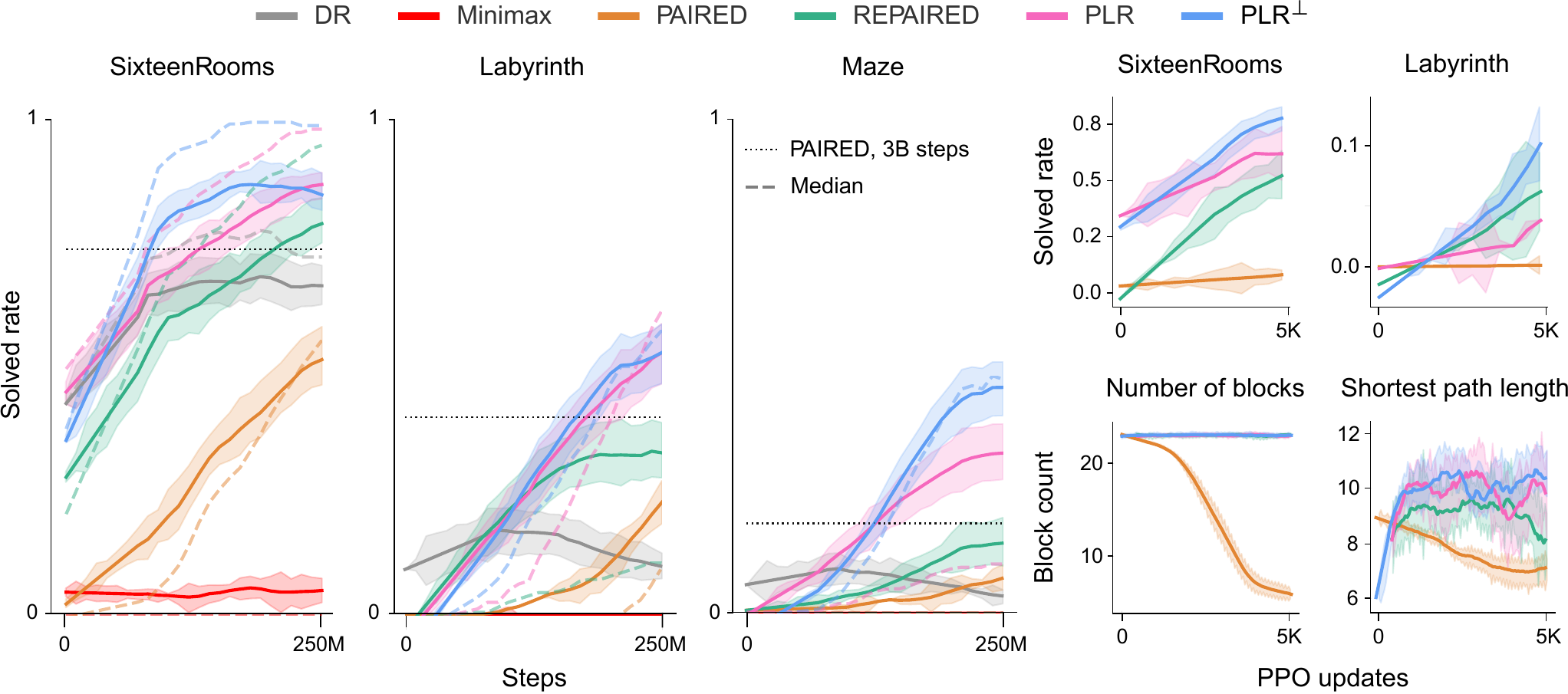}
    \caption{\small{Zero-shot transfer performance during training for PAIRED and REPAIRED variants. The plots show mean and standard error across 10 runs. The dotted lines mark the mean performance of PAIRED after 3B training steps, as reported in~\citet{paired}, while dashed lines indicate median returns.}}
    \label{figure:minigrid_curves}
\end{figure}

\begin{figure}[t]
    \centering
    \includegraphics[width=0.75\linewidth]{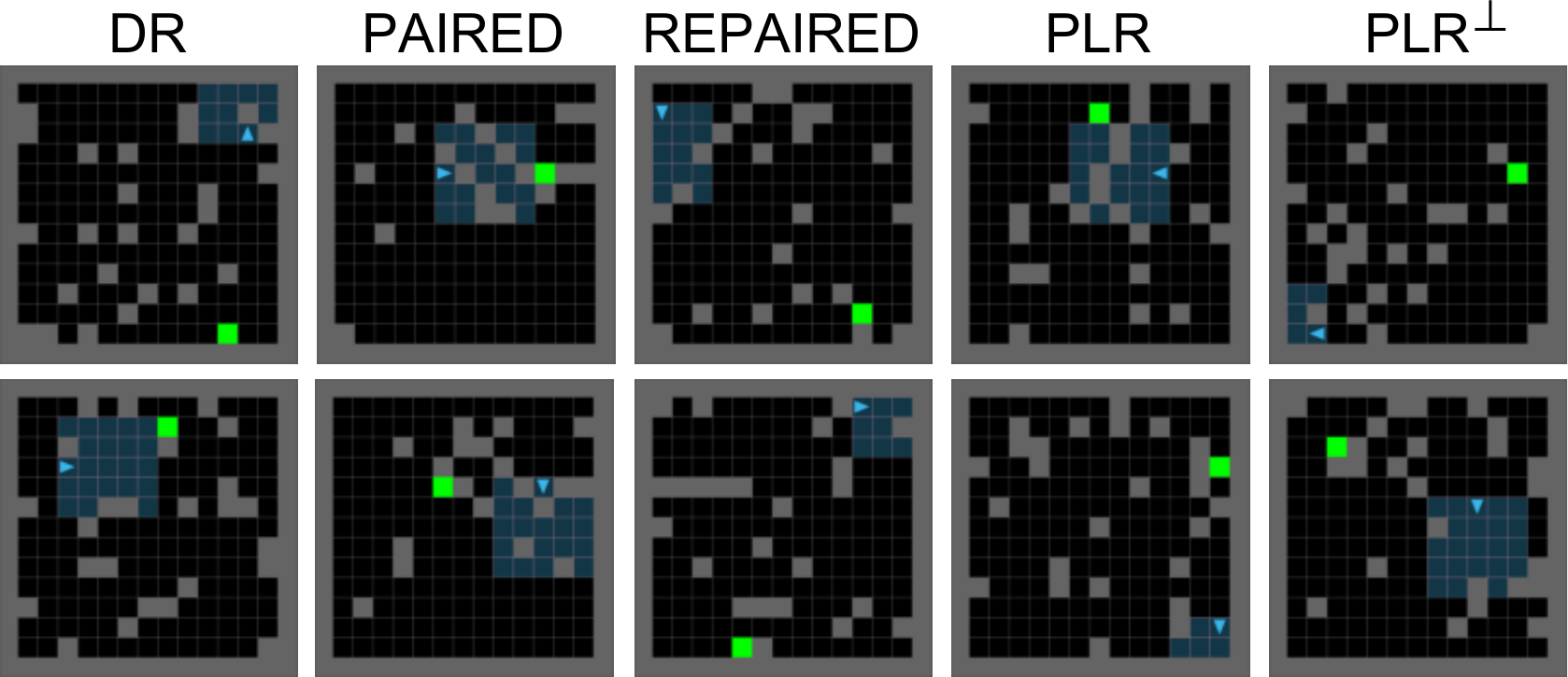}
   \caption{\small{Examples of emergent structures generated by each method.}}
   \label{figure:minigrid_complexity_examples}
\end{figure}

Our results in Figure~\ref{figure:main_minigrid} and~\ref{figure:minigrid_curves} show that \plrabbrev{} and REPAIRED both achieve greater sample-efficiency and zero-shot generalization than their replay-free counterparts. The improved test performance achieved by \plrabbrev{} over both DR and PLR when trained for an equivalent number of gradient updates, aggregated over all test mazes, is statistically significant ($p < 0.05$), as is the improved test performance of REPAIRED over PAIRED.  Well before 250 million steps, both PLR and \plrabbrev{} significantly outperform PAIRED after 3 billion training steps, as reported in~\citet{paired}. Further, both PLR variants lead to policies exhibiting greater zero-shot transfer than the PAIRED variants. Notably, the \plrabbrev{} agent learns to solve mazes by approximately conducting a wall-following strategy. Table~\ref{table:minigrid_eval} reports performance across all test mazes. The success of designing regret-maximizing levels via random search and successive level replay (curation) over learning a generator with RL suggests that for some UPOMDPs, the regret landscape, as a function of the free parameters $\theta$, has a low effective dimensionality \citep{bergstra2012random}. Foregoing gradient-based learning in favor of random search may then lead to faster adaptation to the changing regret landscape, as the policy evolves during training.

\begin{table}[h]
\small
      \caption{\small{Mean test solved rates and standard errors on zero-shot transfer mazes for each method using a 25-block budget after 250M training steps. Results are aggregated over 100 attempts for each maze across 10 runs per method. Bolded figures overlap in standard error with the method attaining the maximum mean solved rate in each row. The asterisk $*$ indicates training for 500M steps.}}
      \centering
      \scalebox{0.85}{
        \begin{tabular}{lllllllr}
\toprule
Environment &DR &Minimax &PAIRED &REP. &PLR & \plrabbrev{} & \plrabbrev{}* \\
\midrule
Labyrinth&$0.2\pm0.1$&$0.0\pm0.0$&$0.3\pm0.1$&$0.1\pm0.0$&$0.3\pm0.1$&$\mathbf{0.5\pm0.1}$&$\mathbf{0.7\pm0.1}$\\
Labyrinth2&$0.2\pm0.1$&$0.0\pm0.0$&$0.2\pm0.1$&$0.2\pm0.1$&$0.4\pm0.1$&$\mathbf{0.6\pm0.1}$&$\mathbf{0.8\pm0.1}$\\
LargeCorridor&$\mathbf{0.7\pm0.1}$&$0.1\pm0.1$&$0.3\pm0.1$&$0.5\pm0.1$&$\mathbf{0.7\pm0.1}$&$\mathbf{0.8\pm0.1}$&$\mathbf{0.8\pm0.1}$\\
Maze&$0.0\pm0.0$&$0.0\pm0.0$&$0.0\pm0.0$&$0.2\pm0.1$&$0.3\pm0.1$&$\mathbf{0.6\pm0.1}$&$\mathbf{0.5\pm0.1}$\\
Maze2&$0.0\pm0.0$&$0.0\pm0.0$&$0.1\pm0.1$&$0.1\pm0.1$&$\mathbf{0.4\pm0.1}$&$\mathbf{0.4\pm0.1}$&$\mathbf{0.5\pm0.1}$\\
PerfectMaze&$0.3\pm0.1$&$0.0\pm0.0$&$0.0\pm0.0$&$0.4\pm0.1$&$0.4\pm0.1$&$\mathbf{0.6\pm0.1}$&$\mathbf{0.5\pm0.1}$\\
SixteenRooms&$0.9\pm0.0$&$0.1\pm0.1$&$0.7\pm0.1$&$0.9\pm0.1$&$\mathbf{1.0\pm0.0}$&$0.8\pm0.1$&$\mathbf{1.0\pm0.0}$\\
SixteenRooms2&$\mathbf{0.7\pm0.1}$&$0.0\pm0.0$&$0.0\pm0.0$&$\mathbf{0.6\pm0.1}$&$\mathbf{0.5\pm0.1}$&$\mathbf{0.7\pm0.1}$&$\mathbf{0.7\pm0.1}$\\
\midrule
Mean &$0.4\pm0.0$&$0.0\pm0.0$&$0.2\pm0.0$&$0.4\pm0.0$&$0.5\pm0.1$&$\mathbf{0.6\pm0.1}$&$\mathbf{0.7\pm0.1}$\\
        \bottomrule
        \end{tabular}}
    \label{table:minigrid_eval}
\end{table}

\begin{figure}[h!]
    \includegraphics[width=\linewidth]{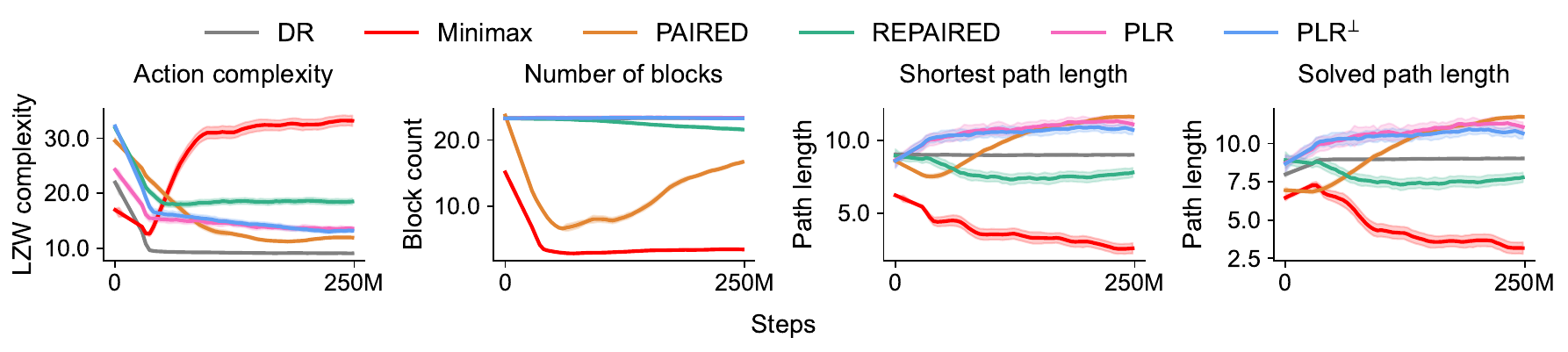}
    \caption{\small{Complexity metrics of environments generated by the teacher throughout training with a 25-block budget. Plots show the mean and standard error of 10 runs.}}
    \label{figure:minigrid_complexity_all_250M}
\end{figure}

\medskip
\noindent \textbf{Emergent complexity:} As the student agents improve, the teachers must generate more challenging levels to maintain regret. We measure the resultant emergent complexity by tracking the number of blocks in each level and the shortest path length to the goal (where unsolvable levels are assigned a length of 0). Figure~\ref{figure:minigrid_curves}~(right) shows that over the first 5000 PPO updates, PAIRED slowly adapts the complexity over training while REPAIRED initially quickly grows complexity, before being overtaken by PAIRED. This more rapid onset of complexity may be due to REPAIRED's fast replay mechanism, and the long-term slowdown relative to PAIRED may be explained by its less frequent gradient updates due to the use of a high level replay rate ($p=0.95$). Notably, both PLR and \plrabbrev{} begin to produce levels with longer solution paths significantly earlier in training. This result shows that random search is surprisingly efficient at continually discovering levels of increasing complexity, given an appropriate curation mechanism. Figure~\ref{figure:minigrid_complexity_examples} shows that, similar to methods with a regret-maximizing teacher, PLR and \plrabbrev{} can find levels exhibiting complex structure. 

In addition to these two metrics, we also track the mean solved path length, which averages the shortest path length to the goal over levels successfully solved by the student. Further, we track the student's action complexity, corresponding to the Lempel-Ziv-Welch (LZW) complexity of the action sequence taken. LZW complexity is a commonly used measure of string compressibility. The evolution all of these metrics over the course of 250M training steps is shown in Figure~\ref{figure:minigrid_complexity_all_250M}. We see the initial complexity trends in solution path lengths shown in Figure~\ref{figure:minigrid_curves} persist throughout training, and PAIRED eventually matches the solution path complexity of PLR and \plrabbrev{}. Despite the REPAIRED teacher performing far fewer gradient updates than that of PAIRED in the same number of environment steps, the REPAIRED teacher's shortest path lengths exceed that of PAIRED after adjusting proportionately by replay rate. Foreseeably, over a longer period, the shortest path lengths generated by REPAIRED may meet or exceed that of PAIRED. In all cases, except for the minimax baseline, the action complexity reduces as the agent becomes more decisive. We see that both PAIRED and REPAIRED lead to more decisive and robust policies---as indicated by the simultaneously lower action complexity and higher block counts (relative to DR) and, in the case of PAIRED, higher path length metrics. Notably, the minimax teacher begins to produce imposible levels (indicated by solution paths going to 0) using only a few blocks, which leads the student to take more random action sequences (indicated by increasing action complexity).

\begin{figure}[h!]
    \includegraphics[width=\linewidth]{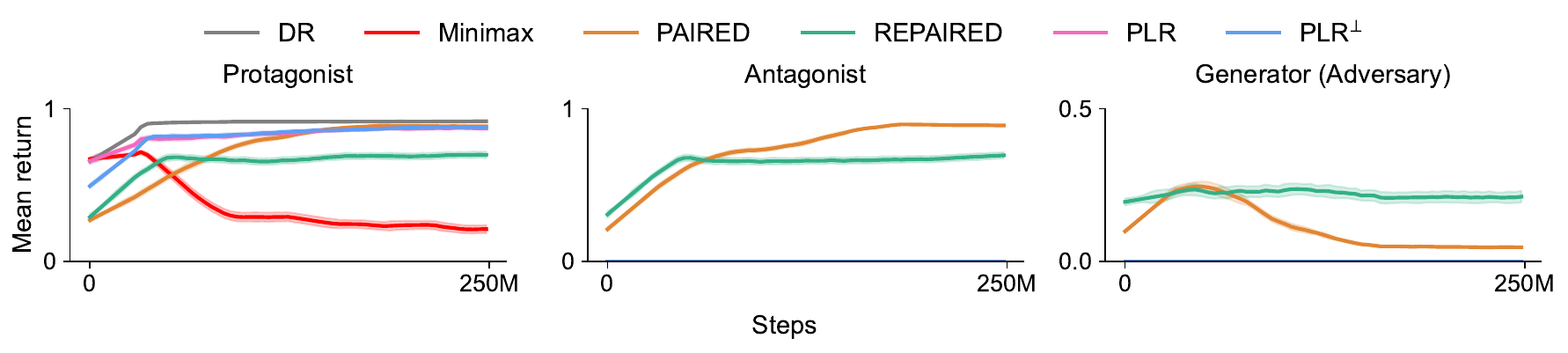}
    \caption{\small{Training returns for each participating agent in each method, when trained with a 25-block budget. Plots show the mean and standard error over 10 runs.}}
    \label{figure:minigrid_training_returns_250M}
\end{figure}

To provide a further sense of the training dynamics, we present the per-agent training returns for each method in Figure~\ref{figure:minigrid_training_returns_250M}. Notably PAIRED results in antagonists that attain higher returns than the protagonist as expected. This dynamic takes on a mild oscillation, visible in the training return curve of the generator (teacher). As the protagonist adapts to the adversarial levels, the generator's return reduces, until the generator discovers new configurations that better exploit the relative differences between the two student policies. Notably, the adversary under REPAIRED seems to propose more difficult levels for both the protagonist and antagonist, while the resulting protagonist policy exhibits improved test performance, as seen in Figure~\ref{figure:minigrid_curves}.

\subsection{Pixel-Based Car Racing with Continuous Control}
\label{sec:exp_carracing}

\begin{figure}[h]
    \centering\includegraphics[width=1\linewidth]{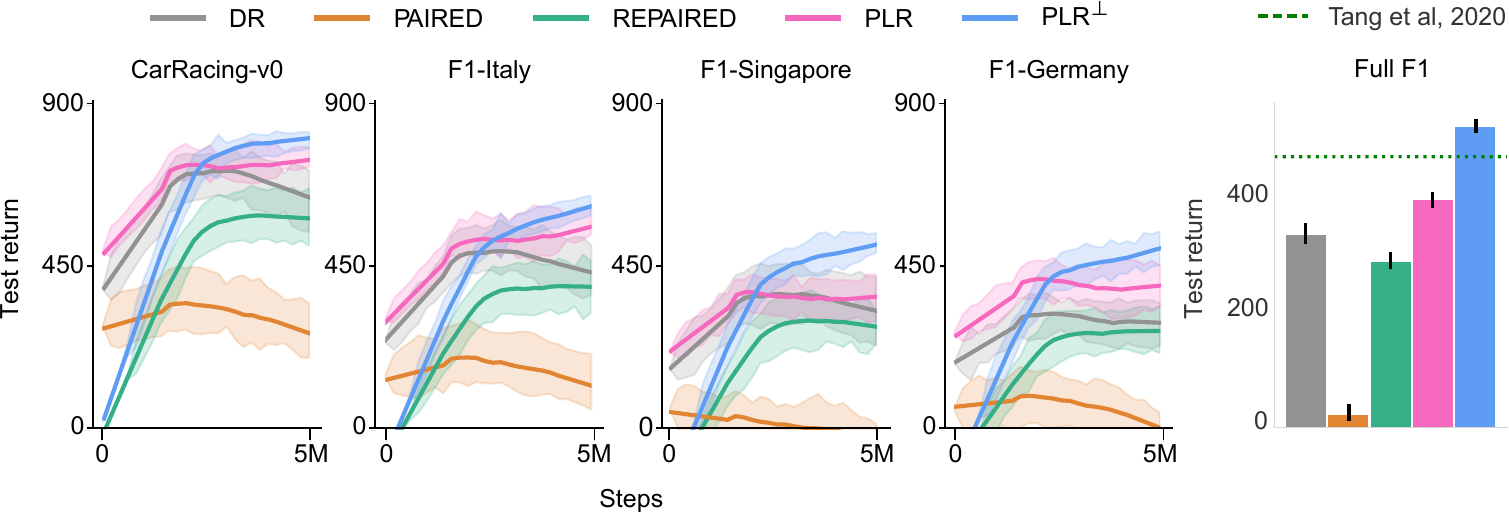}
    \caption{\small{Zero-shot transfer performance. Plots show mean and standard error over 10 runs.}}
    \label{figure:carracing_main_results}
\end{figure}

To test the versatility and scalability of our methods, we turn to an extended version of the CarRacing environment from OpenAI Gym~\citep{gym}. This environment entails continuous control with dense rewards, a 3-dimensional action space, and partial, pixel observations, with the goal of driving a full lap around a track. To enable UED of any closed-loop track, we reparameterize CarRacing to generate tracks as B{\'e}zier curves~\citep{bezier_ref} with arbitrary control points. The teacher generates levels by choosing a sequence of up to 12 control points, which uniquely defines a B{\'e}zier track within specific, predefined curvature constraints. After 5M steps of training, we test the zero-shot transfer performance of policies trained by each method on 20 levels replicating official human-designed Formula One (F1) tracks (see Figure~\ref{figure:f1_tracks} for a visualization of the tracks). Note that these tracks are significantly OOD, as they cannot be defined with just 12 control points. In Figure~\ref{figure:carracing_main_results} we show the progression of zero-shot transfer performance for the original CarRacing environment, as well as three F1 tracks of varying difficulty, while also including the final performance on the full F1 benchmark. For the final performance, we also evaluated the state-of-the-art CarRacing agent from \citet{attentionagent} on our new F1 benchmark. 

Unlike in the sparse, discrete navigation setting, we find DR leads to moderately successful policies for zero-shot transfer in CarRacing. Dense rewards simplify the learning problem and random Bezier tracks occasionally contain the challenges seen in F1 tracks, such as hairpin turns and observations showing parallel tracks due to high local curvature. Still, we see that policies trained by selectively sampling tracks to maximize regret significantly outperform those trained by uniformly sampling from randomly generated tracks, in terms of zero-shot transfer to the OOD F1 tracks. Remarkably, with a replay rate of 0.5, \plrabbrev{} sees statistically significant ($p < 0.001$) gains over PLR in zero-shot performance over the full F1 benchmark, despite directly training on only half the rollout data using half as many gradient updates. Once again, we see that random search with curation via PLR produces a rich  selection of levels and an effective curriculum.  

\begin{figure}[t!]
    \includegraphics[width=\linewidth]{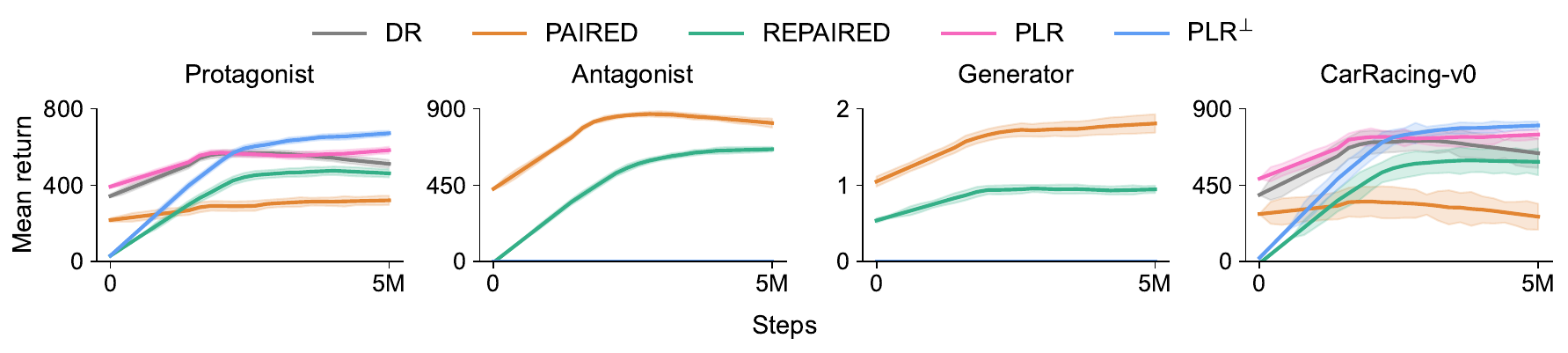}
    \caption{\small{From left to right: Returns attained by the protagonist, antagonist, and generator (adversary) throughout training; the protagonist's zero-shot transfer performance on the original CarRacing-v0 during training. The mean and standard error over 10 runs are shown.}}
    \label{figure:carracing_training_returns_5M}
\end{figure}

\begin{figure}[t!]
    \centering
    \includegraphics[width=\linewidth]{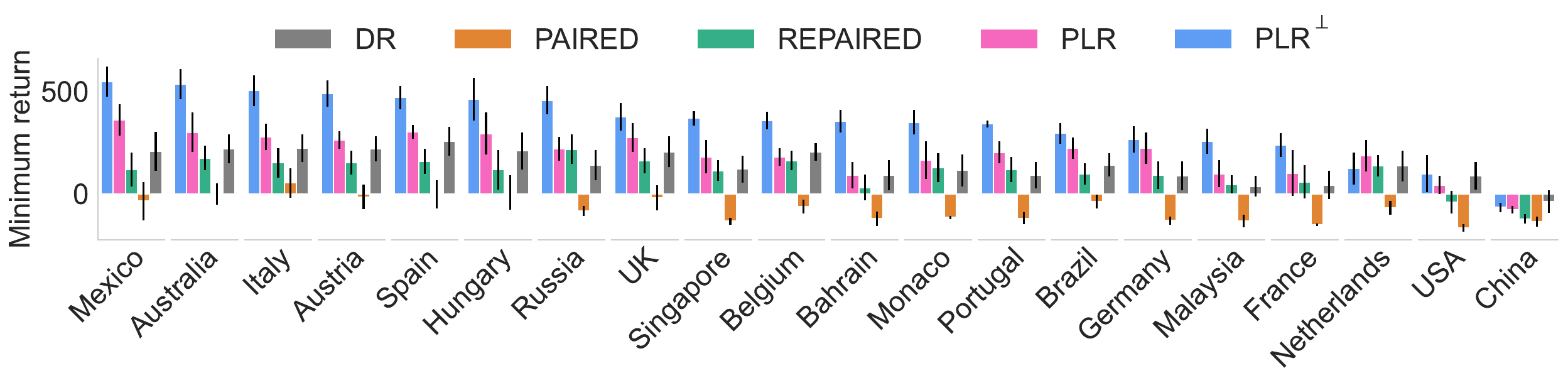}
    \caption{\small{Minimum returns attained across 10 test episodes per track per seed. Bars report mean and standard error over 10 training runs.}}
    \label{figure:carracing_min_returns}
\end{figure}

We also observe that PAIRED struggles to train a robust protagonist in CarRacing. Specifically, PAIRED overexploits the relative strengths of the antagonist over the protagonist, finding curricula that steer the protagonist towards policies that ultimately perform poorly even on simple tracks, leading to a gradual reduction in level complexity. This dynamic can be seen in the per-agent training curves in Figure~\ref{figure:carracing_training_returns_5M} and leads to degenerate, overly-simple tracks, as shown in Figure~\ref{figure:carracing_pics}, which visualizes sample tracks generated by each method. As shown in Figure~\ref{figure:carracing_main_results}, REPAIRED mitigates this degeneracy substantially, though not completely, inducing a policy that significantly outperforms PAIRED ($p < 0.001$) in mean performance on the full F1 benchmark, but underperforms DR. Notably, \plrabbrev{} exceeds the performance of the state-of-the-art AttentionAgent~\citep{attentionagent}, despite not using a self-attention policy and training on less than 0.25\% of the number of environment steps in comparison. These gains come purely from the induced curriculum. Figure~\ref{figure:carracing_min_returns} further reveals that \plrabbrev{} produces CarRacing policies that tend to achieve higher minimum returns on average compared to the baselines, providing further evidence of the benefits of the minimax regret property coupled with a fast level replay mechanism for efficiently finding high-regret levels.

\begin{table}[!htb]
      \caption{\small{Mean test returns and standard errors of each method on the full F1 benchmark. Results are aggregated over 10 attempts for each track across 10 runs per method. Bolded figures overlap in standard error with the method attaining the maximum mean test return in each row. We see that \plrabbrev{} consistently either outperforms the other methods or matches PLR, the next best performing method. Note that we separately report the results of a single run for AttentionAgent due to its high compute overhead.
      }}
      \centering
      \scalebox{0.87}{
        \begin{tabular}{*{6}l | r}
\toprule
Track &DR &PAIRED &REPAIRED &PLR &PLR$^{\bot}$ & AA \\
\midrule
Australia&$484\pm29$&$100\pm22$&$414\pm27$&$545\pm23$&$\mathbf{692\pm15}$ &826\\
Austria&$409\pm21$&$92\pm24$&$345\pm19$&$442\pm18$&$\mathbf{615\pm13}$ &511\\
Bahrain&$298\pm27$&$-35\pm19$&$295\pm23$&$411\pm22$&$\mathbf{590\pm15}$ &372\\
Belgium&$328\pm16$&$72\pm20$&$293\pm19$&$327\pm15$&$\mathbf{474\pm12}$ &668\\
Brazil&$309\pm23$&$76\pm18$&$256\pm19$&$387\pm17$&$\mathbf{455\pm13}$ &145\\
China&$115\pm24$&$-101\pm9$&$7\pm18$&$84\pm20$&$\mathbf{228\pm24}$& 344\\
France&$279\pm32$&$-81\pm13$&$240\pm29$&$290\pm35$&$\mathbf{478\pm22}$&153\\
Germany&$274\pm23$&$-33\pm16$&$272\pm22$&$388\pm20$&$\mathbf{499\pm18}$& 214\\
Hungary&$465\pm32$&$98\pm29$&$414\pm29$&$533\pm26$&$\mathbf{708\pm17}$& 769\\
Italy&$461\pm27$&$132\pm24$&$371\pm25$&$588\pm20$&$\mathbf{625\pm12}$&798\\
Malaysia&$236\pm25$&$-26\pm17$&$200\pm17$&$283\pm20$&$\mathbf{400\pm18}$ & 300\\
Mexico&$458\pm33$&$67\pm31$&$415\pm30$&$561\pm21$&$\mathbf{712\pm12}$&580\\
Monaco&$268\pm28$&$-28\pm18$&$256\pm26$&$360\pm32$&$\mathbf{486\pm19}$ &835\\
Netherlands&$328\pm26$&$70\pm20$&$307\pm21$&$\mathbf{418\pm21}$&$\mathbf{419\pm25}$& 131\\
Portugal&$324\pm27$&$-49\pm13$&$265\pm21$&$407\pm15$&$\mathbf{483\pm13}$& 606\\
Russia&$382\pm30$&$51\pm21$&$419\pm25$&$479\pm24$&$\mathbf{649\pm14}$ & 732\\
Singapore&$336\pm29$&$-35\pm14$&$274\pm21$&$386\pm22$&$\mathbf{566\pm15}$& 276\\
Spain&$433\pm24$&$134\pm24$&$358\pm24$&$482\pm17$&$\mathbf{622\pm14}$&759\\
UK&$393\pm28$&$138\pm25$&$380\pm22$&$456\pm16$&$\mathbf{538\pm17}$&729\\
USA&$263\pm31$&$-119\pm11$&$120\pm25$&$243\pm28$&$\mathbf{381\pm33}$& -192\\
\midrule
Mean&$341\pm22$&$19\pm15$&$293\pm18$&$408\pm12$&$\mathbf{534\pm7}$ & 477 \\
\bottomrule
        \end{tabular}}
    \label{table:carracing_f1_benchmark}
\end{table}

We report per-track zero-shot transfer returns for the entire CarRacing-F1 benchmark in Table~\ref{table:carracing_f1_benchmark}. While DR acts as a strong baseline in terms of zero-shot generalization in this setting, \plrabbrev{} either attains the highest mean return, or matches the method achieving the highest return within standard error on all tracks. The mean performance of \plrabbrev{} across the full benchmark is statistically significantly higher ($p < 0.001$) than that of all other methods. Notably, the PAIRED teacher's ability to overexploit the differences between antagonist and protagonist is highly detrimental to zero-shot transfer performance. We see that REPAIRED mitigates this effect to a degree, resulting in more competitive policies. Note that due to the high compute overhead of training the AttentionAgent (8.2 billion steps of training over a population 256 agents) \citep{attentionagent}, we resorted to evaluating its mean F1 performance using the pre-trained model weights provided by the authors with their public code release. As a result, we only have a single training run for AttentionAgent. This means we cannot reliably compute standard errors for this baseline, but we believe that showing the performance for a single training seed of AttentionAgent on the F1 benchmark alongside our methods, as done in Figure \ref{figure:carracing_main_results}, nonetheless provides a useful comparison for further contextualizing the efficacy of our methods. This comparison highlights how, by only modifying the training curriculum, our methods produce policies with test returns exceeding that of AttentionAgent---which in contrast, uses a powerful attention-based policy and a much larger number of training steps.

\begin{figure*}
    \centering
    \begin{subfigure}{.18\textwidth}
        \includegraphics[width=\textwidth]{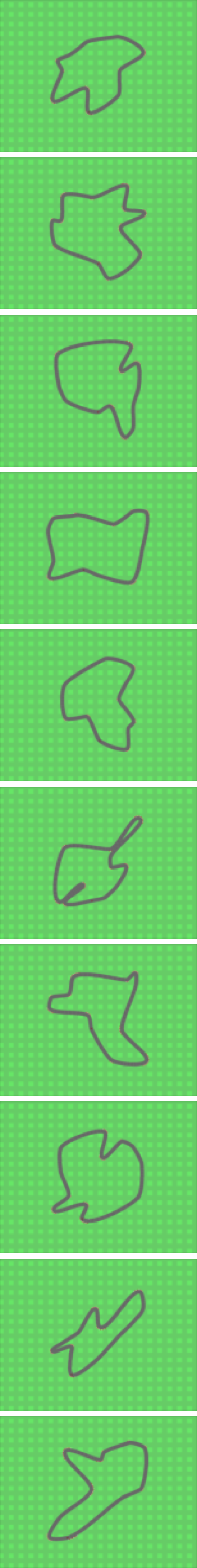}  
        \caption{DR}
    \end{subfigure}
    \begin{subfigure}{.18\textwidth}
        \includegraphics[width=\textwidth]{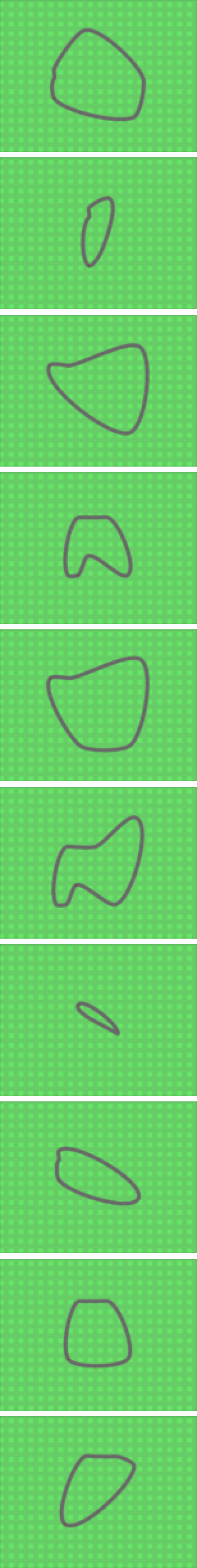}
    \caption{PAIRED}
    \end{subfigure}
    \begin{subfigure}{.18\textwidth}
        \includegraphics[width=\textwidth]{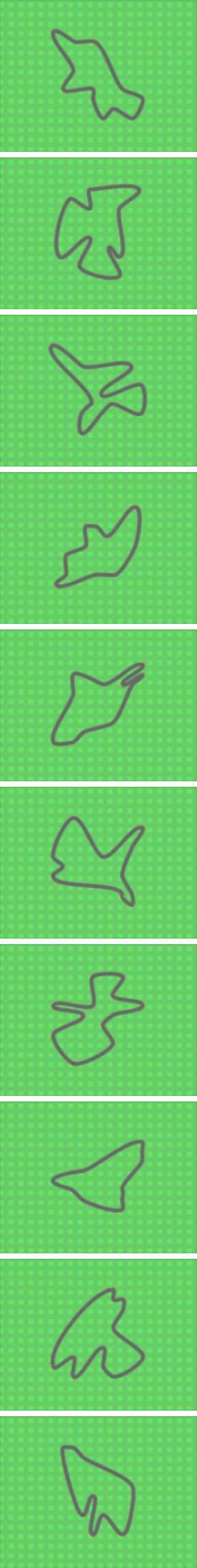}
    \caption{REPAIRED}
    \end{subfigure}
    \begin{subfigure}{.18\textwidth}
        \includegraphics[width=\textwidth]{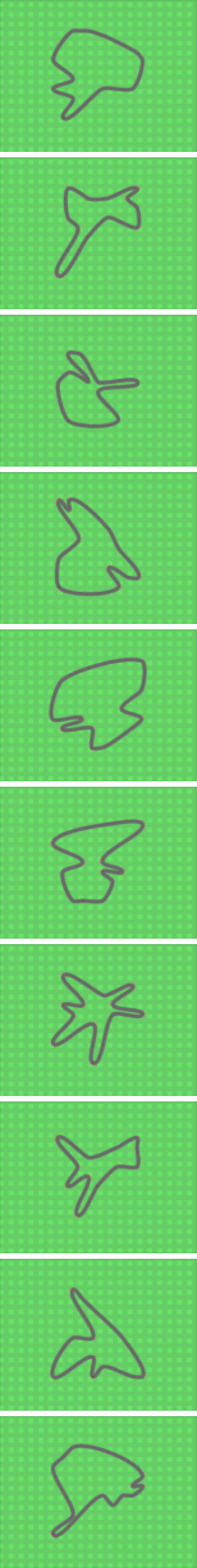}
        \caption{PLR}
    \end{subfigure}
    \begin{subfigure}{.18\textwidth}
        \includegraphics[width=\textwidth]{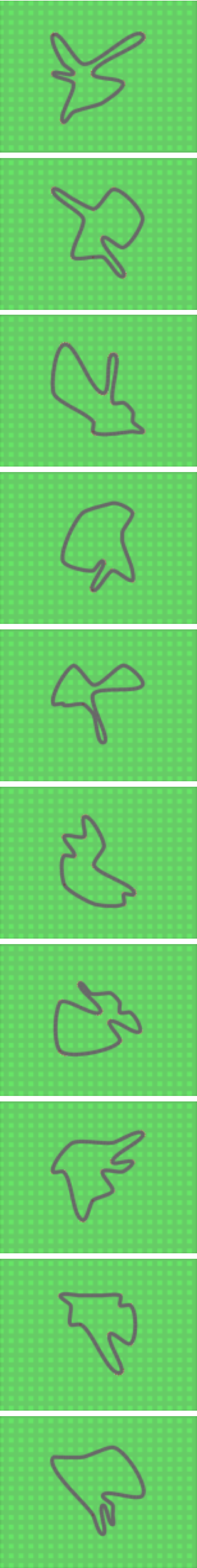}
        \caption{PLR$^{\perp}$}
    \end{subfigure}
    \caption{\small{A randomly-selected set of CarRacing tracks generated by each method. (a)~Domain Randomization (DR) produces tracks of average complexity, with few sharp turns. (b)~PAIRED often overexploits the difference in the students, leading to simple tracks that incidentally favor the antagonist. (c)~REPAIRED mitigates this degeneracy, recovering track complexity. (d)~PLR and (e)~PLR$^\bot$ similarly generate tracks of considerable complexity, by prioritizing the most challenging randomly generated tracks.}}
    \label{figure:carracing_pics}
\end{figure*}

As a further analysis of robustness, we inspect the minimum returns over 10 attempts per track, averaged over 10 runs per method. We present these results (mean and standard error) in Figure~\ref{figure:carracing_min_returns}. \plrabbrev{} achieves consistently higher minimum returns on average for many of the tracks compared to the other methods, including on the challenging Russia and USA tracks. The fact that simply curating random levels, as done by \plrabbrev{}, more reliably approaches a minimax regret policy than PAIRED and REPAIRED suggests that RL may not be an effective means for optimizing the PAIRED teacher.

\newpage

\section{Related Work}
In inducing parallel curricula, DCD follows a rich lineage of curriculum learning methods~\citep{bengio_curriculum, powerplay2013, curriculum_rl_survey1,portelas2020automatic}. Many previous automatic curriculum learning (ACL) algorithms resemble the curator in DCD, sharing similar underlying selective-sampling mechanisms as \plrabbrev{}. Most similar is TSCL~\citep{tscl}, which prioritizes levels based on return rather than value loss, and has been shown to overfit to training levels in some settings~\citep{plr}. In our setting, replayed levels can be viewed as past strategies from a level-generating teacher. This multi-agent perspective from DCD links our replay-based methods to fictitious self-play~\cite[FSP,][]{fictitious_sp}, and more closely, Prioritized FSP \citep{vinyals2019grandmaster}, which selectively samples opponents based on historic win ratios.

As discussed in Section~\ref{section:plr_related_work}, many previous ACL methods make use of a generating adversary include Asymmetric Self-Play~\citep{sukhbaatar2018intrinsic, openai2021asymmetric}, wherein one agent proposes tasks for another in the form of environment trajectories, and AMIGo~\citep{amigo}, wherein the teacher is rewarded for proposing reachable goals. However, unlike the DCD approaches developed in this chapter, these prior methods, including the original PLR algorithm, are largely heuristically-motivated and lack principled robustness guarantees.

Other recent algorithms can be understood as forms of UED and like DCD, framed in the lens of decision theory. POET~\citep{poet, enhanced_poet}, a coevolutionary approach~\citep{Popovici2012}, uses a population of \emph{minimax} (rather than minimax regret) adversaries to construct terrain for a BipedalWalker agent. In contrast to our methods, POET requires training a large population of both agents and environments and consequently, a sizable compute overhead. APT-Gen~\citep{fang2021adaptive} also procedurally generates tasks, but requires access to target tasks, whereas our methods seek to improve zero-shot transfer.

The DCD framework also encompasses adaptive domain randomization methods~\cite[DR,][]{mehta2019activedomain,evolutionary_dr}. The success of DR-based methods in sim2real transfer for robotics~\citep{domain_randomization, james2017transferring, dexterity, rubics_cube} suggests that DCD and, more broadly, UED approaches can help further the robustness in such real-world applications. DR itself is subsumed by procedural content generation~\citep[PCG,][]{risi_togelius_pcg}, for which UED and DCD may be seen as providing a formal, decision-theoretic framework, enabling the development of more principled algorithms.

\section{Discussion}
\label{sec:conclusion}
We established a novel connection between PLR and minimax regret UED approaches like PAIRED, by developing the theory of Dual Curriculum Design (DCD). In this setting, a student policy is challenged by a team of two co-adapting, regret-maximizing teachers: one, a generator that creates new levels, and the other, a curator that selectively samples previously generated levels for replay. This view unifies PLR and PAIRED, which are both instances of DCD. Our theoretical results on DCD then enabled us to prove that PLR attains a minimax regret policy at NE, thereby providing the first theoretical characterization of the robustness of PLR. Notably our theory leads to the counterintuitive result that PLR can be made provably robust by training on less data, specifically, by only using the trajectories on levels sampled for replay. In addition, we developed Replay-Enhanced PAIRED (REPAIRED), which extends the selective replay-based updates of \plrabbrev{} to PAIRED, and proved it shares the same robustness guarantee at NE.
Empirically, in two highly distinct environments, we found that \plrabbrev{} significantly improves zero-shot generalization over PLR, and REPAIRED, over PAIRED. As our methods solely modify the order of levels visited during training, they can, in principle, be combined with many other RL methods to yield potentially orthogonal improvements in sample-efficiency and generalization.

While these DCD-based improvements to PLR and PAIRED empirically lead to more robust policies, it is important to emphasize that our theoretical results only prove a minimax regret guarantee at NE for these methods; however, they provide no explicit guarantee of convergence to such NE. Further, it is worth highlighting that replay-based methods like \plrabbrev{} are completely dependent on the quality of levels proposed by the generator. Our results show that simply curating high regret levels discovered via random search is enough to outperform the RL-based PAIRED teacher in the domains studied. We expect that advancing methods for defining or adapting the generator's proposal distribution holds great potential to improve the efficacy of our methods, especially in more complex, higher-dimensional domains, where random search may prove ineffective for finding useful training levels. Crucial to this endeavor is the design of the regret estimator. While effective in practice, both the positive value loss (PVL) and maximum Monte Carlo (MaxMC) estimators may be strongly biased toward lower regret estimates, as they both approximate regret by the performance gap between the current student's return and the value prediction, a measure of historical performance, in each state. These estimates will be skewed toward lower values than the true regret, as the student can be expected to be suboptimal. Conversely, when the student is optimal, both estimators can still estimate a positive regret as long as the value loss is nonzero. Developing more accurate regret estimators can be expected to improve the performance of UED methods. Lastly, but importantly, our methods assume an appropriate choice of the UPOMDP's free parameters. These methods cannot be expected to produce robust policies for zero-shot transfer if the set of environments defined by the free parameters does not sufficiently align with the transfer domain of interest. Designing an environment parameterization for successful zero-shot transfer to a specific target domain, can be highly non-trivial, posing an important problem for future research. More ambitious is the challenge of designing an environment parameterization that can tractably encompass a universal task space, allowing for autocurricula that produce increasingly capable agents. Chapter~\ref{chapter:conclusions} provides a more detailed discussion of this exciting direction.

Looking beyond environment design, we notice that long-running UED processes in expansive UPDOMPs closely resemble continual learning in open-ended domains. The congruency of these settings suggests our contributions around DCD may extend to more general continual learning problems in which agents must learn to master a diverse sequence of tasks with predefined (or inferred) episode boundaries---if tasks are assumed to be designed by a regret-maximizing teacher. Thus, DCD-based methods like \plrabbrev{} may yield more general policies for continual learning. We anticipate many exciting crossovers between these areas of research in the years to come.

\newcommand{\accellong}{Adversarially Compounding Complexity by Editing Levels}

\chapter{Evolving Curricula}
\label{chapter:accel}

\section{Introduction}
\label{sec:intro}

Autocurricula hold great promise for producing an open-ended learning process~\citep{chromaria,stanley2017open}, given the curriculum can be continually steered toward novel, challenging tasks for the agent to solve. However, the UED methods studied so far all require the teacher to generate new environment instances from scratch. While effective in practice on some domains, such strategies are likely to run into computational limitations in more complex design spaces. A more efficient search procedure in richer design spaces should take advantage of useful structures previously discovered. Methods in the evolutionary computing community have long pursued this direction in many optimization problem settings. Recent methods like Minimal Criteria Coevolution~\citep[MCC,][]{mcc_og}) and POET~\citep{poet, enhanced_poet} show that evolving levels can effectively produce agents capable of solving a diverse range of challenging tasks. In contrast to the UED algorithms in the preceding chapters, these evolutionary methods directly take advantage of the most useful structures found so far in a constant process of mutation and selection. However, key drawbacks of these methods are their reliance on domain specific heuristics and need for vast computational resources, making it challenging for the community to make progress in this direction.
    
\begin{figure*}[t!]
    \centering
    \begin{subfigure}{\textwidth}
    \includegraphics[width=\textwidth]{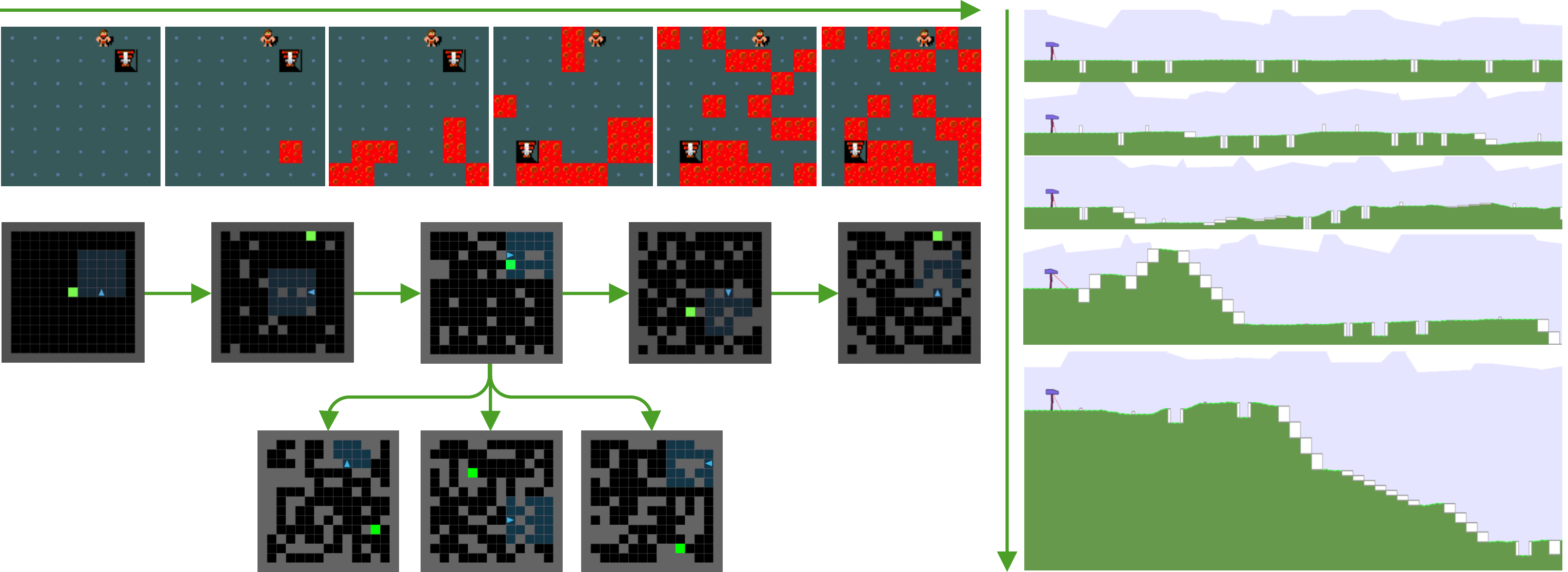}
    \end{subfigure}
    \caption{\small{The evolution of a level in three different environments: MiniHack lava grids, MiniGrid mazes and BipedalWalker terrains. In each case, the direction of the green arrows indicate the sequence of edits to an initial simple level. Each level along the evolutionary path has a high regret for the student agent at that point in time. Thus the level difficulty co-evolves with the agent's capabilities. In each environment, we see that despite starting with simple levels, the pursuit of high regret leads to increasingly complex challenges. This complexity emerges entirely without relying on any environment-specific exploration heuristics. Note that since the agent can move diagonally in the lava environment, the final level in the top row is solvable.}}
    \label{fig:level_evolution}
\end{figure*}

\begin{figure}[t!]
    \centering
    \includegraphics[width=0.85\textwidth]{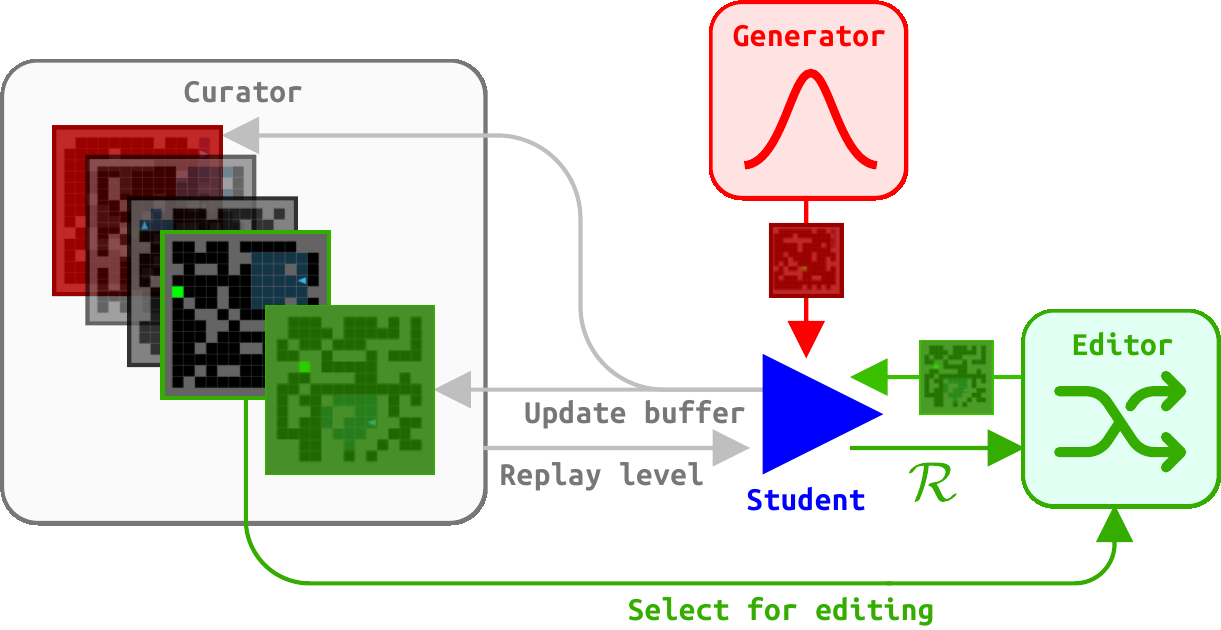}
    \caption{\small{An overview of ACCEL. Levels are randomly sampled from a generator and evaluated, with high-regret levels added to the level replay buffer. The curator selects levels to replay, and the student only trains on replay levels. After training, the regret of replayed levels are edited and evaluated again for level replay. }}
    \label{fig:accel}
\end{figure}

In this work, we seek to harness the power and potential open-endedness of evolution in a principled regret-based curriculum. We introduce a new algorithm, called \emph{\accellong{}}, or ACCEL. This method evolves a curriculum by making small \emph{edits} (e.g. mutations) to previously high-regret levels, thus constantly producing new levels at the frontier of the student agent's capabilities (see Figure~\ref{fig:accel}). Levels generated by ACCEL begin simple but quickly become more complex. This dynamic benefits the beginning of training where the student can then learn more quickly \citep{Berthouze04adaptive,powerplay2013}, and encourages the policy to rapidly co-evolve with the environment to solve increasingly complex levels (see Figure~\ref{fig:level_evolution}). An interactive web demo of ACCEL is available at \url{https://accelagent.github.io}.

We believe ACCEL provides the best of both worlds: an evolutionary approach that can generate increasingly complex environments, combined with a regret-based curator that reduces the need for domain-specific heuristics and provides theoretical robustness guarantees in equilibrium. ACCEL leads to strong empirical gains in both sparse-reward navigation tasks and a 2D bipedal locomotion task over challenging terrain. In both domains, ACCEL demonstrates the ability to rapidly increase level complexity while producing highly capable agents. ACCEL produces and solves highly challenging levels with a fraction of the compute of previous approaches, reaching comparable level complexity as POET while training on less than 0.05\% of the total number of environment interaction samples, on a single GPU. An open source implementation of ACCEL reproducing our experiments is available at \mbox{\url{https://github.com/facebookresearch/dcd}}.

\section{Adversarially Compounding Complexity}

In this section we introduce a new algorithm for UED, combining an evolutionary environment generator with a principled regret-based curator. Unlike PLR which relies on random sampling to produce new batches of training levels, we instead propose to make edits (e.g. mutations) to previously curated ones. Evolutionary methods have been effective in a variety of challenging optimization problems \citep{neuronature, qdnature}, yet typically rely on handcrafted, domain-specific rules. For example, POET manually filters BipedalWalker levels to have a return in the range $[50,300]$. The key insight in this work is that with regret as a domain-agnostic fitness function for evolution, evolution can be harnessed to continually generate levels at the frontier of agent capabilities. Indeed, by iteratively editing and curating the resulting levels, the content of the level replay buffer quickly increases in complexity. As such, we call our method \emph{\accellong{}} (ACCEL). 

ACCEL does not assume a specific editing mechanism, which can be any mutation process used in other open-ended evolutionary approaches \citep{chromaria}. In our experiments, editing involves making small changes (e.g. adding or removing obstacles in a maze), which can operate directly on environment elements within the level or on a more indirect encoding such as the latent-space representation of the level under a generative model of the environment. In general, editing may rely on more advanced mechanisms, such as search-based methods, but in this work we predominantly make use of simple, random mutations. ACCEL makes the key assumption that regret varies smoothly with the environment parameters $\Theta$, such that the regret of a level is close to the regret of others within a small edit distance. If this is the case, then small edits to a single high-regret level should lead to the discovery of entire batches of high-regret levels---an otherwise challenging task in high-dimensional design spaces. 

Building on PLR$^{\perp}$, we do not immediately train on edited levels. Instead, we first evaluate them and only add them to the level replay buffer if they have high regret, estimated by positive value loss (Equation~\ref{eq:pvl}). We consider two different criteria for selecting which replayed levels to edit: Under the \emph{hard} criterion, we edit a subsample of levels in which the agent both incurs high regret and has difficulty solving, approximated as the agent's regret minus its return. Under the \emph{batch} criterion, we simply edit the entire batch of levels most recently sampled for replay. The full procedure is shown in Algorithm~\ref{alg:accel}.

\begin{figure}[h]
\begin{minipage}{\linewidth}
\begin{algorithm}[H]
\SetAlgoLined
\caption{\mbox{\accellong{}}}
\label{alg:accel}
\textbf{Input:} Level buffer size $K$, initial fill ratio $\rho$, level generator \\
\textbf{Initialize:} Initialize policy $\pi(\phi)$, level buffer $\bm{\Lambda}$ \\
    Sample $K * \rho$ initial levels to populate $\bm{\Lambda}$ \\
    \While{not converged}{
        Sample replay decision $d \sim P_{D}(d)$ \\
        \eIf{$d=0$}{
            Sample level $\theta$ from level generator\\
            Collect $\pi$'s trajectory $\tau$ on $\theta$, with stop-gradient $\phi_{\bot}$ \\  
            Compute regret score $S$ for $\theta$ (Equation~\ref{eq:pvl}) \\
            Update $\bm{\Lambda}$ with $\theta$ if score $S$ meets threshold
        }
        {
        Sample a replay level, $\theta \sim \bm{\Lambda}$ \\
        Collect policy trajectory $\tau$ on $\theta$ \\
        Update $\pi$ with rewards $\bm{R}(\tau)$ \\
        Edit $\theta$ to produce $\theta'$ \\
        Collect $\pi$'s trajectory $\tau$ on $\theta'$, with stop-gradient $\phi_{\bot}$ \\  
        Compute regret score $S$ ($S'$) for $\theta$ ($\theta'$) \\       
        Update $\bm{\Lambda}$ with $\theta$ ($\theta'$) if score $S$ ($S'$) meets threshold \\
        (Optionally) Update level editor using score $S$
        }
    }
\end{algorithm}
\end{minipage}
\end{figure}

ACCEL can be viewed as an open-ended evolutionary search algorithm~\citep{stanley2017open}, whereby the fitness is estimated regret, as levels only stay in the population (that is, the level replay buffer) if they meet the high-regret criterion for curation. However, ACCEL avoids some important weaknesses of evolutionary algorithms such as POET: First, ACCEL maintains a population of levels, but not a population of agents. Thus, ACCEL requires only a single desktop GPU for training. In contrast, evolutionary approaches typically require a CPU cluster. Moreoever, forgoing an agent population allows ACCEL to avoid the agent selection problem. Instead, ACCEL directly trains a single \emph{generalist} agent. Finally, since ACCEL uses a minimax \emph{regret}  objective (rather than minimax as in POET), it naturally promotes levels at the frontier of agent's capabilities, without relying on domain-specific knowledge (such as reward ranges). Training on high regret levels also means that ACCEL inherits the robustness guarantees in equilbrium from PLR$^{\perp}$ (Corollary 1 in Chapter~\ref{chapter:plr}):

\newtheorem{accelremark}{Remark}
\begin{accelremark}
If ACCEL reaches a Nash equilibrium, then the student follows a minimax regret strategy.
\end{accelremark}

\noindent In contrast, other evolutionary approaches primarily justify their applicability solely via empirical results on specific domains. As our experiments show, a key strength of ACCEL is its generality. It can produce highly capable agents in a diverse range of environments, without domain knowledge. 

\section{Experiments}
\label{sec:experiments}
Our experiments compare agents trained with ACCEL to those trained with other UED baselines. In all cases, we train a student agent via Proximal Policy Optimization~\citep[PPO,][] {schulman2017proximal}. Our primary baseline is Robust PLR \citep[PLR$^{\perp}$,][]{jiang2021robustplr}, which combines the random search with a regret-based curation mechanism. For convenience, in the remainder of this chapter, we refer to Robust PLR simply as ``PLR." The other baselines are domain randomization (DR), PAIRED \citep{paired}, and a minimax adversarial teacher. The minimax baseline corresponds to the objective used in POET without any hand-coded constraints. We leave a full comparison to population-based methods to future work due to the additional computational expense required. We report results in a consistent manner across environments: In each case, we show the emergent complexity during training and report test performance in terms of the aggregate inter-quartile mean (IQM) and optimality gap using the recently introduced RLiable library~\citep{agarwal2021deep}. To evaluate the quality of the resulting curricula, we report all performance with respect to the number of gradient updates for the student policy, as opposed to total number of environment interactions, which is, in any case, often comparable for PLR and ACCEL (see Table~\ref{table:accel_stepcount}). Full details on choice of hyperparameters for each experiment is listed in Table~\ref{table:hyperparams}. 

\subsection{Learning with Lava}
\label{app:minihack}

We begin with LavaGrid, a simple proof-of-concept environment to assess the impact of supplementing PLR with evolutionary search: Here an agent must navigate to a goal while avoiding lava tiles in a fully-observable grid-based environment based on the NetHack runtime~\citep{kuttler2020nethack} and built using MiniHack~\citep{samvelyan2021minihack}. The reward is sparse, with the agent receiving $+1$ reward for reaching the goal and a per timestep penalty of $-0.01$. The grid is only $7\times7$, but remains challenging as the episode terminates with zero reward if the agent touches the lava. This dynamic makes exploration more difficult by penalizing random actions. Moreover, while toy, such challenges may be relevant in real-world, safety-critical settings, where agents may wish to avoid events causing early termination during training. For DR and PLR$^{\perp}$, the random generator samples the number of lava tiles to place from the range $[0,20]$. For ACCEL, we use a generator that outputs only the empty room. Subsequent edits then produce new levels by adding or removing lava tiles.

Figure \ref{fig:level_evolution} shows the results of running each method over 5 runs. Despite starting with empty rooms, ACCEL quickly produces levels with more lava than the other methods, while also achieving higher training returns, reaching near-perfect performance on its training distribution. PLR$^{\perp}$ is able to produce a similar training profile to ACCEL, but attains lower complexity metrics.

\begin{figure}[H]
    \centering
    \includegraphics[width=.75\linewidth]{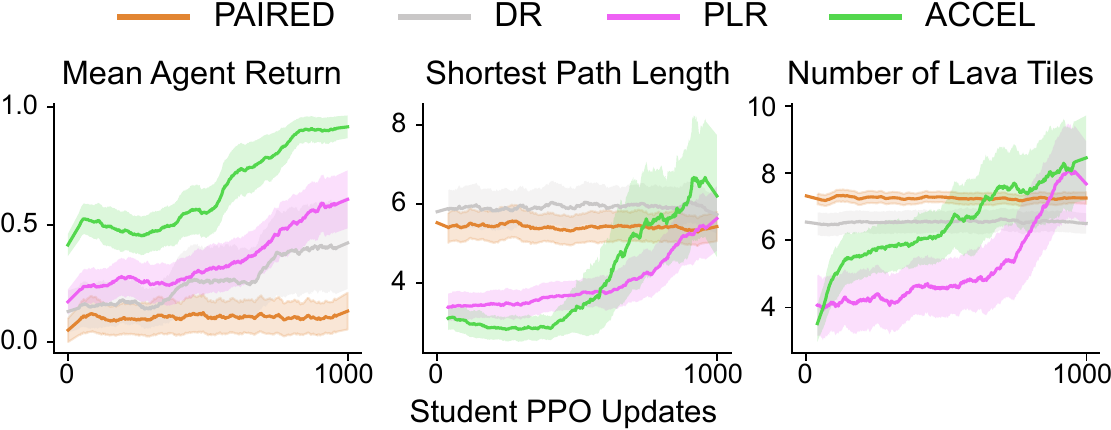}
    \caption{\small{Training return and emergent complexity in LavaGrid. The plots report the mean and standard error over 5 seeds.}}
    \label{fig:lava7x7_complexity}
\end{figure}

\begin{figure}[H]
    \centering
    \includegraphics[width=0.75\textwidth]{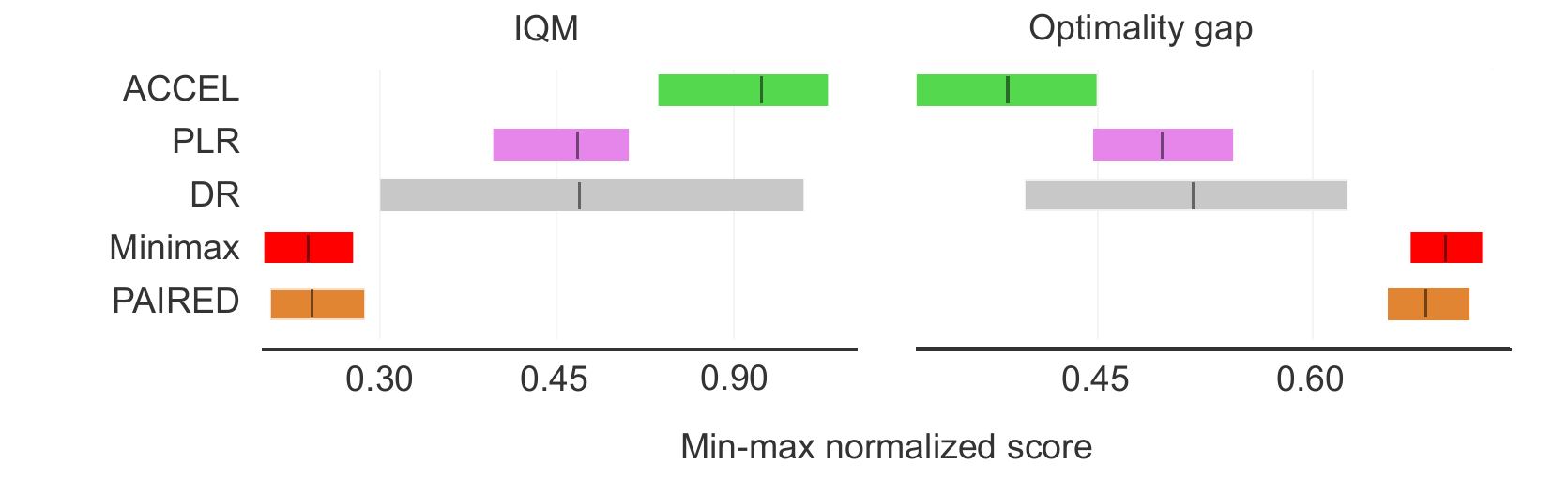}
    \caption{\small{Lava Grid aggregate test performance.}}
    \label{fig:lava_rliable}
\end{figure}

To test robustness, we evaluate each agent on held-out test levels after 1000 PPO updates ($\approx$20M timesteps), and report the aggregate results in Figure \ref{fig:lava_rliable}, where we see that ACCEL is the best performing method. Extended results are shown in Table \ref{table:lava7x7results}. The first three test environments (Empty, 10 Tiles and 20 Tiles) evaluate in-distribution robustness, as these levels can be sampled in the training distribution. In contrast, LavaCrossing-S9N1 (LavaX) tests generalization to an OOD environment.
Across both in-distribution and OOD evaluations, ACCEL agents obtain the best performance.

\begin{table}[h!]
\begin{center}
\caption{\small{Test performance in in-distribution and out-of-distribution environments. Each entry is the mean (and standard error) of 5 training runs, where each run is evaluated for 100 trials on each environment. Bold values are within one standard error of the best mean.}
}
\label{table:lava7x7results}
\scalebox{0.9}{
\begin{tabular}{ l|llllllr } 
\toprule
\textbf{Env.} & PAIRED  & Minimax & DR & PLR & ACCEL \\ 
\midrule 
Empty & $0.77 \pm 0.03$ & $0.76 \pm 0.02$ & $0.89 \pm 0.05$ & $0.96 \pm 0.04$ & $\mathbf{1.0 \pm 0.0}$ \\ 
10 Tiles & $0.12 \pm 0.03$ & $0.05 \pm 0.01$  & $0.33 \pm 0.15$ & $0.3 \pm 0.05$ & $\mathbf{0.49 \pm 0.07}$ \\ 
20 Tiles & $0.06 \pm 0.01$ & $0.11 \pm 0.04$ & $0.23 \pm 0.12$ & $0.25 \pm 0.06$ & $\mathbf{0.35 \pm 0.08}$ \\ 
LavaX & $0.0 \pm 0.0$ & $0.0 \pm 0.0$ & $\mathbf{0.05 \pm 0.05}$ & $0.01 \pm 0.0$ & $\mathbf{0.05 \pm 0.04}$ \\ 
\bottomrule
\end{tabular}}
\end{center}
\end{table}

\subsection{Partially Observable Navigation}

\begin{figure}[t!]
    \centering
    \includegraphics[width=0.75\textwidth]{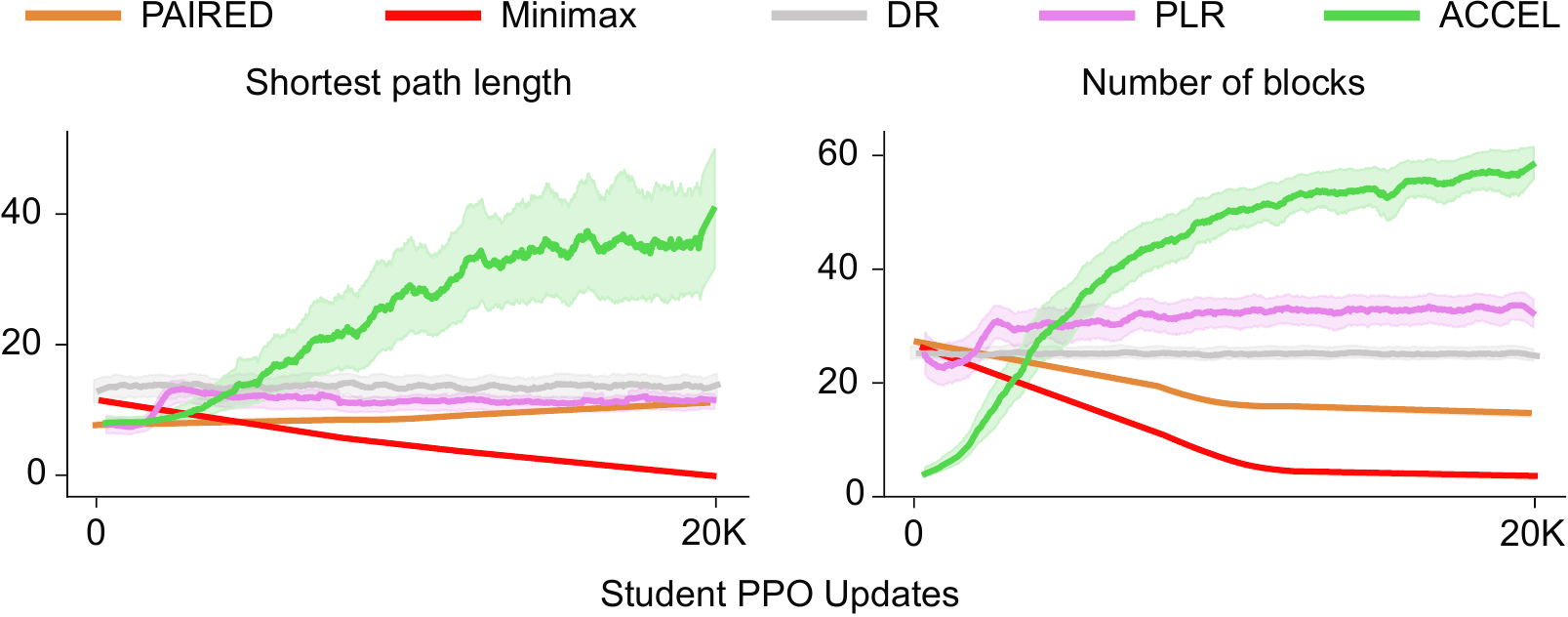}
    \caption{\small{Emergent complexity metrics for mazes generated during training. Mean and standard error across 5 training seeds are shown.}}
    \label{fig:minigrid_complexity}
\end{figure}

Next we move to the larger $15\times15$ maze environments from Chapter~\ref{chapter:dcd}, allowing us to directly compare against previous baselines. This domain is based on MiniGrid~\citep{gym_minigrid} and was originally introduced in \citet{paired}. In this environment, the agent has access to a $5\times5$ forward-facing, partial observation and must navigate through a maze, consisting of multiple wall blocks, to reach a goal. Upon reaching the goal, the episode terminates and the agent receives a sparse reward equal to $1 - 0.9(T/T_{\text{max}})$, where $T$ is the episode length and $T_{\text{max}} = 250$ is the maximum episode length allowed. Despite being conceptually simple, this maze domain entails large-scale compute: Our agents train for 20k updates ($\approx$350M steps, see Table~\ref{table:accel_stepcount}), learning an LSTM-based policy with a 75-dimensional partially-observable observation. The random generator used by both DR and PLR$^{\perp}$ samples between 0—60 walls to place. For ACCEL we begin with empty rooms and randomly edit wall locations (by adding or removing wall blocks), as well as the goal location. After replay, we edit levels selected via the hard criterion---effectively moving the most difficult levels closer to the learning frontier, where the student can make progress. In Figure~\ref{fig:minigrid_complexity}, we report training performance and complexity metrics. We see that ACCEL rapidly grows complexity, leading to training levels with significantly higher wall-block counts and longer solution paths than other methods.

\begin{figure}[t!]
    \centering
    \includegraphics[width=0.75\textwidth]{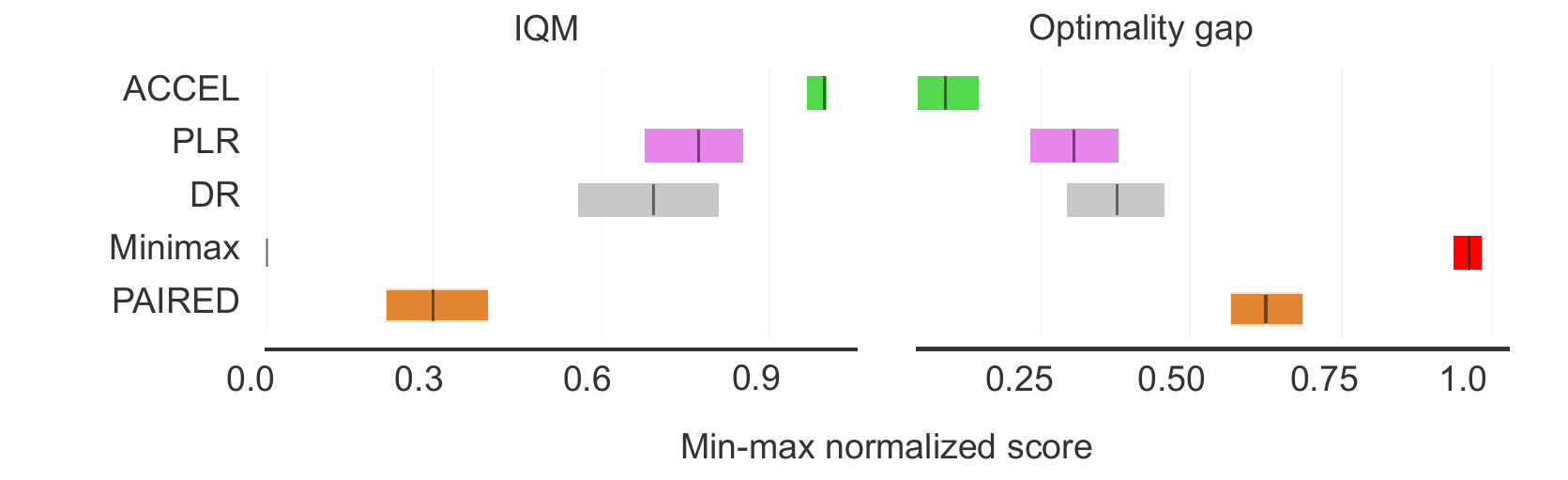}
    \caption{\small{Aggregate zero-shot test performance in the maze domain.}}
    \label{fig:minigrid_rliable}
\end{figure}

\begin{figure}[t!]
    \centering
    \includegraphics[width=0.65\textwidth]{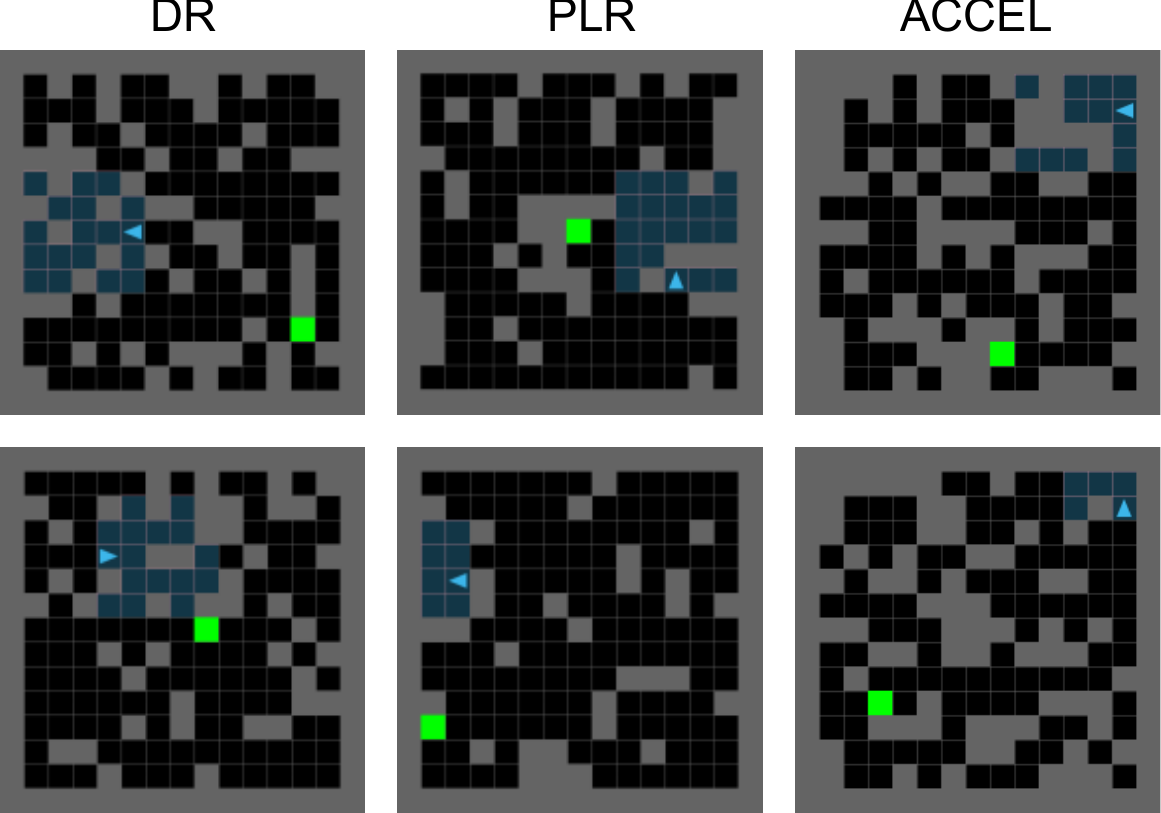}
    \caption{\small{Example levels generated by DR, PLR, and ACCEL.}}
    \label{fig:minigrid_levels_by_methods}
\end{figure}

We evaluate the zero-shot transfer performance of each method on a series of OOD test environments, as done in prior works. For DR, PLR, and ACCEL, evaluation occurs after 20k student PPO updates, thereby focusing the comparison on the effect of the curriculum. The minimax and PAIRED results are those reported in Chapter~\ref{chapter:dcd} at 250M training steps ($\approx$30k updates). As we see, ACCEL performs at least as well as the next best method in almost all test environments, with particularly strong performance in Labyrinth and Maze. As reported in Figure~\ref{fig:minigrid_rliable}, ACCEL achieves drastically stronger performance than all other methods in aggregate across all test environments: Its IQM approaches a perfect solved rate compared to below 80\% for the next best method, PLR, and demonstrates an 80.2\% probability of improvement over PLR. Per-environment test results are reported in Table~\ref{table:minigrid_ood_results}. The random samples of levels generated by each method in Figure~\ref{fig:minigrid_levels_by_methods} show that ACCEL produces mazes with greater average block count and longer solution paths.

\begin{table}[H]
\small
\begin{center}
\caption{\small{Zero-shot transfer to human-designed environments. Each entry corresponds to the mean and standard error of 5 training runs, where each run is evaluated for 100 trials on each environment. $\dagger$ indicates the generator first samples the number of blocks to place in $[0, 60]$, then places that many at random locations. $\ddagger$ indicates the generator produces only empty rooms. Bold values are within one standard error of the best mean. $\star$ indicates a statistically significant improvement against PLR ($p<0.05$ via Welch's t-test). All methods are evaluated after 20k student updates, aside from PAIRED and Minimax, which are evaluated at $\approx$30k updates.}}
\label{table:minigrid_ood_results}
\scalebox{0.8}{
\begin{tabular}{ l|lllllr } 
\toprule
\textbf{Environment} & PAIRED  & Minimax & DR$\dagger$ & PLR$\dagger$ & ACCEL$\dagger$  & ACCEL$\ddagger$ \\ 
\midrule
16Rooms & $0.63 \pm 0.14$ & $0.01 \pm 0.01$  & $0.87 \pm 0.06$&  $0.95 \pm 0.03$ & $\mathbf{1.0 \pm 0.0}$ & $\mathbf{1.0 \pm 0.0}$ \\
16Rooms2 & $0.53 \pm 0.15$ & $0.0 \pm 0.0$ & $0.53 \pm 0.18$ & $0.49 \pm 0.17$ & $0.62 \pm 0.22$ & $\mathbf{0.92 \pm 0.06}$ \\
SimpleCrossing & $0.55 \pm 0.11$ & $0.11 \pm 0.04$ & $0.57 \pm 0.15$ & $\mathbf{0.87 \pm 0.05}$ & $\mathbf{0.92 \pm 0.08}$ & $\mathbf{0.84 \pm 0.16}$ \\
FourRooms & $0.46 \pm 0.06$ & $0.14 \pm 0.03$ & $0.77 \pm 0.1$ & $0.64 \pm 0.04$ & $\mathbf{0.9 \pm 0.08}$ & $0.72 \pm 0.07$ \\
SmallCorridor & $0.37 \pm 0.09$ & $0.14 \pm 0.09$ & $\mathbf{1.0 \pm 0.0}$ & $0.89 \pm 0.05$ & $0.88 \pm 0.11$ & $\mathbf{1.0 \pm 0.0}$ \\
LargeCorridor & $0.27 \pm 0.08$ & $0.14 \pm 0.09$ & $0.64 \pm 0.05$ & $0.79 \pm 0.13$ & $0.94 \pm 0.05$ & $\mathbf{1.0 \pm 0.0}$ \\
Labyrinth & $0.45 \pm 0.14$ & $0.0 \pm 0.0$ & $0.45 \pm 0.23$ & $0.55 \pm 0.23$ & $\mathbf{0.97 \pm 0.03}$ & $0.86 \pm 0.14$ \\
Labyrinth2 & $0.38 \pm 0.12$ & $0.0 \pm 0.0$ & $0.54 \pm 0.18$ & $0.66 \pm 0.18$ & $\mathbf{1.0 \pm 0.01}$ & $\mathbf{1.0 \pm 0.0}$ \\
Maze & $0.02 \pm 0.01$ & $0.0 \pm 0.0$ & $0.43 \pm 0.23$ & $\mathbf{0.54 \pm 0.19}$ & $\mathbf{0.52 \pm 0.26}$ & $\mathbf{0.72 \pm 0.24}$ \\
Maze2 & $0.37 \pm 0.13$ & $0.0 \pm 0.0$ & $0.49 \pm 0.16$ & $0.74 \pm 0.13$ & $0.93 \pm 0.04$ & $\mathbf{1.0 \pm 0.0}$ \\
Maze3 & $0.3 \pm 0.12$ & $0.0 \pm 0.0$ & $0.69 \pm 0.19$ & $0.75 \pm 0.12$ & $\mathbf{0.94 \pm 0.06}$ & $0.8 \pm 0.1$ \\
PerfectMaze(M) & $0.32 \pm 0.06$ & $0.01 \pm 0.0$ & $0.45 \pm 0.1$ & $0.62 \pm 0.09$ & $\mathbf{0.88 \pm 0.12}$ & $\mathbf{0.93 \pm 0.07}$ \\
\midrule
\textbf{Mean} & $0.39 \pm 0.03$ & $0.05 \pm 0.01$ & $0.62 \pm 0.05$ & $0.71 \pm 0.04$ & $\mathbf{0.88 \pm 0.04}^\star$ & $\mathbf{0.9 \pm 0.03}^\star$ \\
\bottomrule
\end{tabular}}
\end{center}
\end{table}

\begin{figure}[t!]
    \centering
    \includegraphics[width=0.65\textwidth]{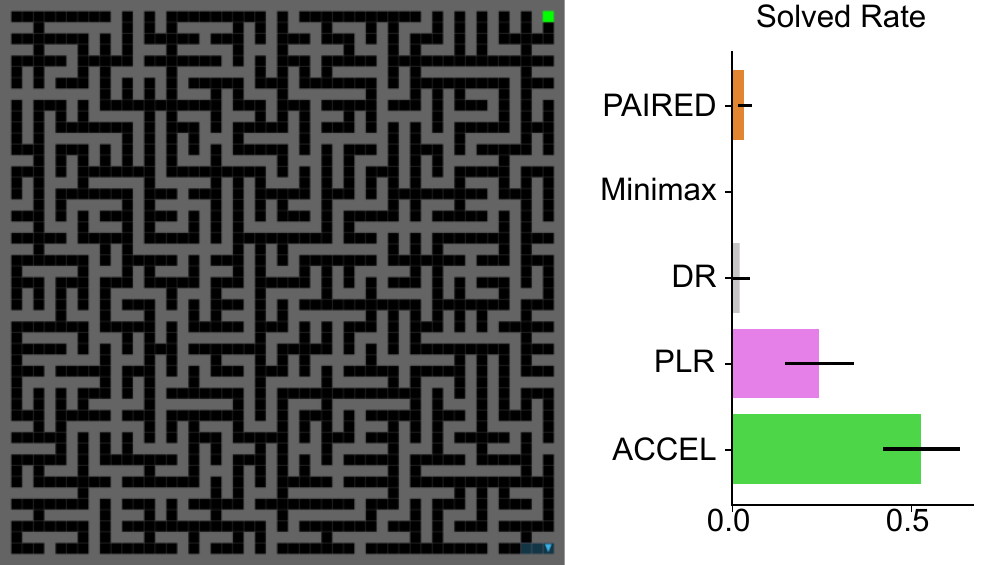}
    \caption{\small{Zero-shot performance on a large procedurally-generated maze environment. The bars show mean and standard error over 5 training seeds, each evaluated over 100 episodes. ACCEL achieves over twice the success rate of the next best method.}}
    \label{fig:perfect_maze_results}
\end{figure}

Next, we evaluate each method on much larger variants of the PerfectMaze PCG environment: PerfectMaze-L (shown in Figure~\ref{fig:perfect_maze_results}) features levels with $51\times51$ tiles and maximum episode lengths of 5K steps, while PerfectMaze-XL (shown in Figure~\ref{figure:perfectmaze_scale}) features levels with $101\times101$ tiles and maximum episode lengths of 20k steps---sizes that are orders of magnitude larger than seen in training. Such a large partially-observable maze would be challenging even for humans. We evaluate agents for 100 episodes (per training seed), using the same checkpoints after 20k PPO updates as compared in Figure~\ref{fig:minigrid_rliable}. In PerfectMaze-L (see Figure~\ref{fig:perfect_maze_results}), ACCEL significantly outperforms all baselines with a success rate of $53\%$ compared to the next best method, PLR, which has a success rate of 25\%, while all other methods fail. On the larger PerfectMaze-XL, the performance of all methods is significantly weaker, with DR and PLR achieving a mean success rate of 4\%. However, ACCEL still outperforms all baselines, achieving 8\% and 7\% mean success rates when using the empty and DR generators respectively. Notably, we observe that successful agents in both environments follow an approximate wall-following strategy for solving these single-component mazes. 

\begin{figure}[t!]
    \centering
    \includegraphics[width=0.65\textwidth]{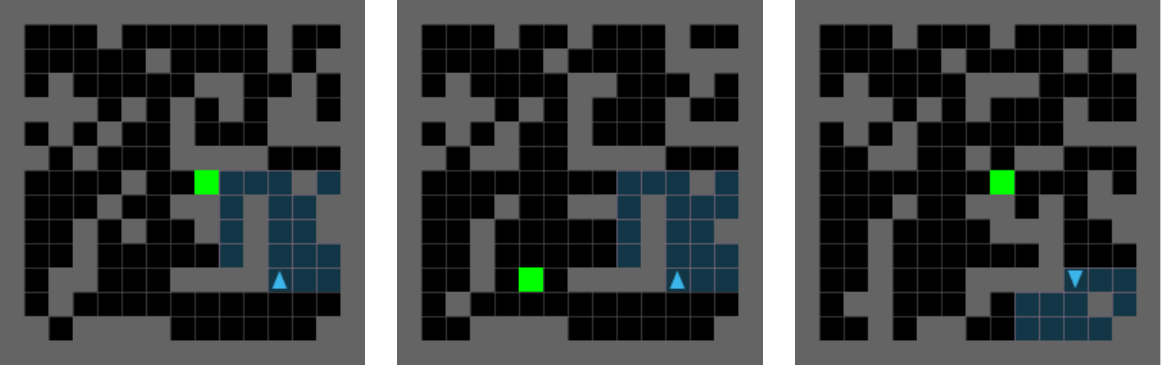}
    \caption{\small{Despite sharing a common ancestor, each of these levels requires different behaviors to solve. Left: The agent can approach the goal by moving upwards or leftwards. Middle: The goal is on the left. Right: The left path is blocked.}}
    \label{fig:minigrid_edit_variations}
\end{figure}

Figure~\ref{fig:minigrid_edit_variations} provides a peek into what may drive ACCEL's strong performance compared to other UED methods: Here, we see three edits of the same level produced by ACCEL. Each has a similar initial observation, yet requires the agent to explore in different directions to reach the goal, thereby pressuring the agent to actively explore the environment. 
These variations demonstrate how incremental changes to a level can lead to a diverse batch of new ones \citep{sturtevant2020unexpected}, which may move those that are currently too hard or too easy towards the frontier of the agent's capabilities. This diversity may prevent overfitting. Further, making edits often do not change the optimal solution path and thus can be seen as a form of data augmentation that changes the observation but not the optimal policy. Such data augmentations have been shown to improve sample efficiency and robustness in RL \citep{rad, drq, ucb_drac}.

\subsection{Walking in Challenging Terrain}
\begin{figure}[t!]
    \centering
    \includegraphics[width=\textwidth]{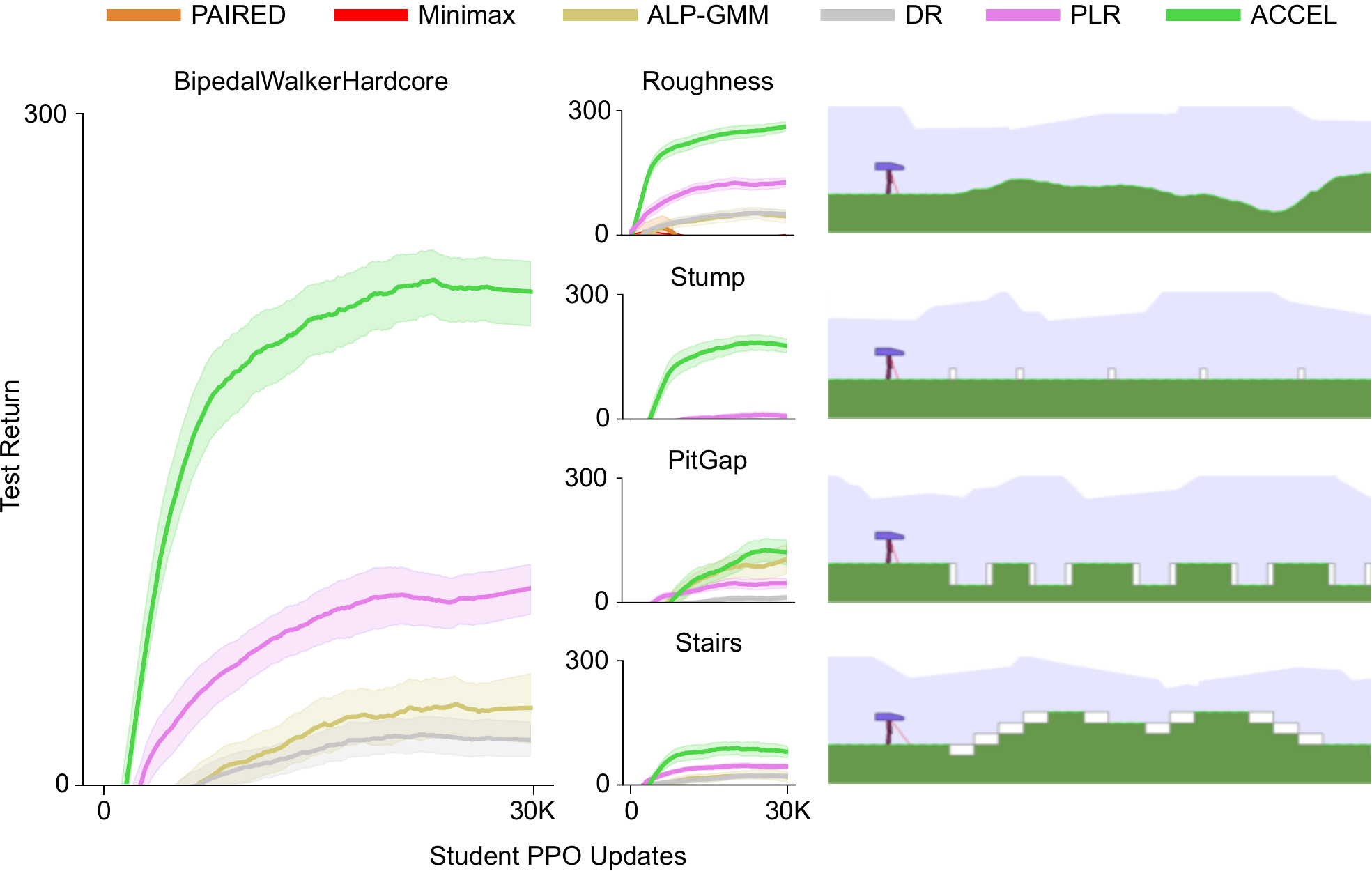}
    \caption{\small{Left: Performance on test environments during training (mean and standard error). Negative returns are omitted. Right: Example levels from the per-obstacle challenge environments.}}
    \label{fig:bipedal_ood_test_curves}
\end{figure}

Our final set of experiments evaluate ACCEL in the BipedalWalker environment from \citet{poet}, a challenging continuous-control environment with dense rewards. This environment serves as a more challenging continual control task to benchmark ACCEL against previous methods, including POET, a powerful autocurriculum method that shows state-of-the-art performance in this domain. As in \citet{poet}, we use a modified version of BipedalWalker-Hardcore~\citep{gym}; however, we include all eight parameters in the design space, rather than only the subset used in \citet{poet}. This environment is detailed at length in Appendix~\ref{appendix:env_bipedal}. We run all baselines from previous experiments, in addition to ALP-GMM~\citep{portelas2019teacher}, an ACL method originally tested in BipedalWalker. We train agents for 30k student updates, equivalent to between 1B–2B total environment steps, depending on the method (see Table~\ref{table:accel_stepcount}). During training we evaluate agents on both the simple BipedalWalker and more challenging BipedalWalker-Hardcore environments, in addition to four environments testing the agent's effectiveness against specific, isolated challenges that are otherwise presented together to varying degrees in training levels: ground roughness, pit gaps, stumps, and stairs (shown in Figure~\ref{fig:bipedal_ood_test_curves}).

\begin{figure}[h!]
    \centering
    \includegraphics[width=0.65\textwidth]{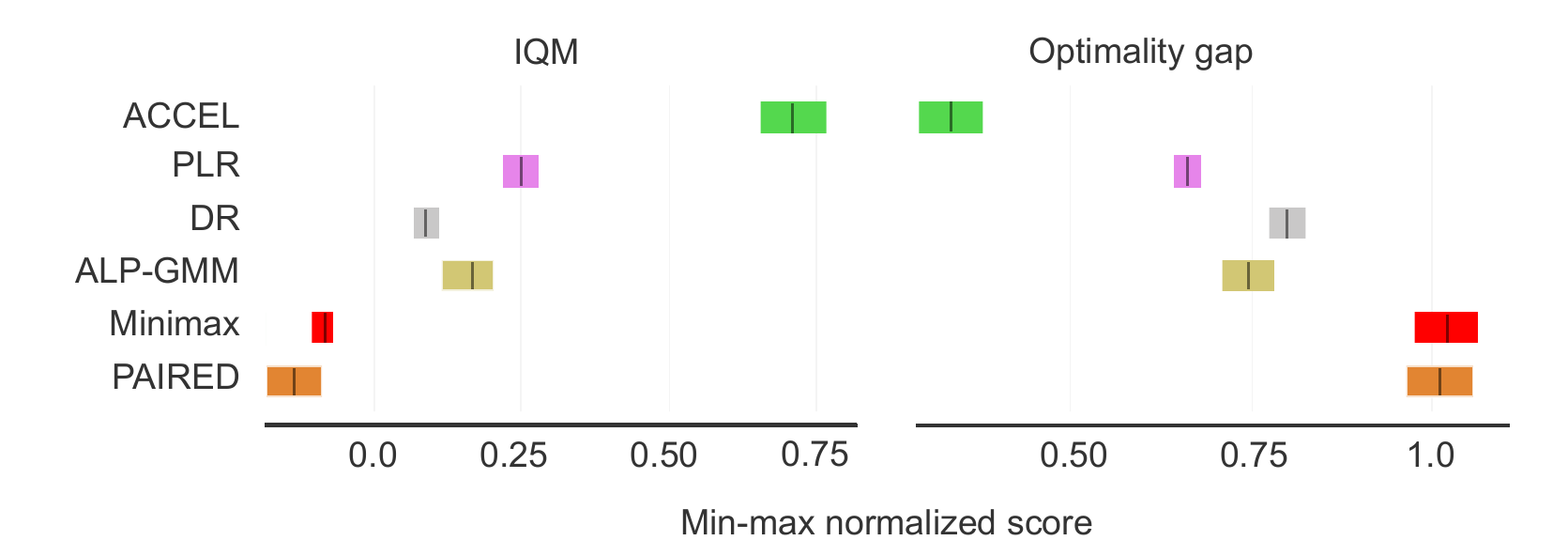}
    \caption{\small{Aggregate performance for ten seeds across all five BipedalWalker test environments.}}
    \label{fig:bipedal_rliable}
\end{figure}

After 30k PPO updates, we evaluate each agent based on 100 episodes in each test environment. Figure~\ref{fig:bipedal_rliable} reports the aggregate results, normalized according to the return range of $[0,300]$. ACCEL significantly outperforms all baselines, achieving close to 75\% of optimal performance, almost three times the performance of the best baseline, PLR$^{\perp}$. All other baselines struggle, likely due to the environment design space containing a high proportion of levels not useful for learning. Faced with such challenging levels, agents may learn to resort to the locally optimal behavior of preventing itself from falling (avoiding a -100 penalty), rather than attempt forward locomotion. Finally, we see that ALP-GMM performs poorly when the design space is increased from 2D (as in \citet{portelas2019teacher}) to 8D.

\begin{figure}[t!]
    \begin{subfigure}{\textwidth}
    \centering
    \includegraphics[width=0.85\textwidth]{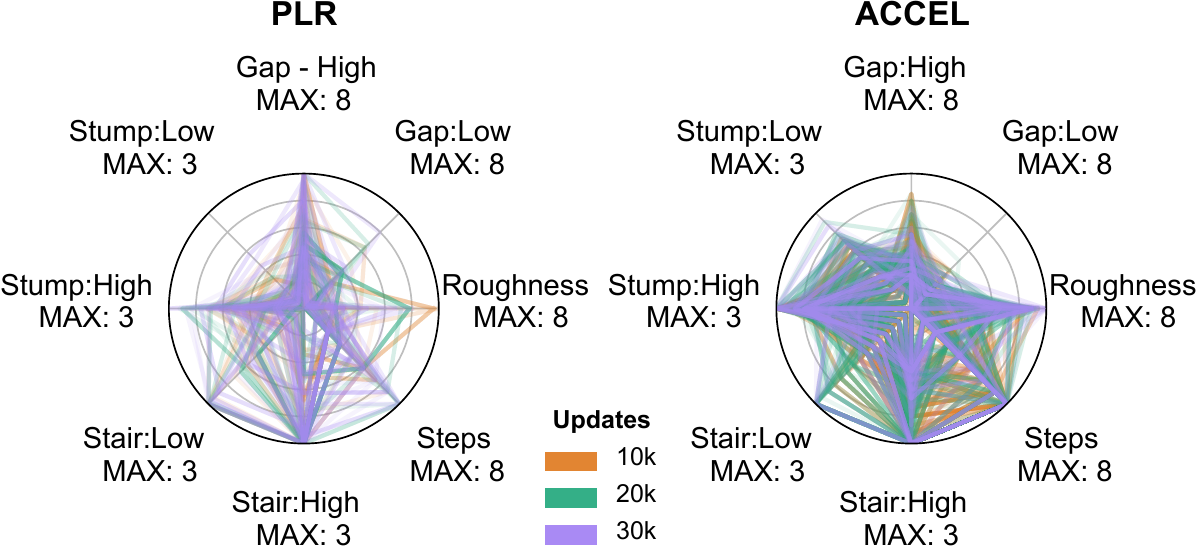}
    \end{subfigure}
    \caption{\small{Top: Rose plots of complexity metrics of BipedalWalker levels discovered by PLR and ACCEL. Each line represents a solved level from the associated checkpoint. All levels are among the top-100 highest regret levels for the given checkpoint. Bottom: Two levels created and solved by ACCEL.}}
    \label{fig:bipedal_complexity}
\end{figure}

Next we seek to understand the properties of the evolving distribution of high-regret levels. We analyze the set of all solved levels from the top-100 highest regret levels in the level replay buffer of ACCEL and PLR$^{\perp}$ training runs after 10k, 20k, and 30k student updates. For each level we show all eight parameters in Figure~\ref{fig:bipedal_complexity} (top). ACCEL agents solve many levels of comparable difficulty with other methods such as POET, but uses a fraction of the compute: ACCEL sees a total of 2.07B environment steps after 30k student updates, less than 0.5\% of that used by POET as reported in \citet{poet}. 

\begin{table}[H]
\begin{center}
\caption{\small{Test performance on challenging evaluation environments. Each entry corresponds to the mean and standard error of 10 independent runs, where each run is evaluated for 100 trials on each environment. $\dagger$ indicates the generator creates each level with obstacle parameters uniformly sampled between the corresponding minimum value of the ``Easy Init'' range and max value defined in Table~\ref{table:bipedal_params}. $\ddagger$ indicates the generator instead uniformly samples obstacle parameters within the ``Easy Init'' ranges. Bold indicates being within one standard error of the best mean. All methods are evaluated at 30k updates.}}
\label{table:bipedalresults}
\scalebox{0.67}{
\begin{tabular}{ l | llllllr } 
\toprule
\textbf{Env.} & PAIRED  & Minimax & ALP-GMM & DR$\dagger$ & PLR$\dagger$ & ACCEL$\dagger$  & ACCEL$\ddagger$ \\ 

\midrule
Basic & $206.5 \pm 30.3$ & $154.3 \pm 59.2$ & $301.5 \pm 11.6$ & $261.9 \pm 19.3$ & $304.1 \pm 1.8$ & $316.9 \pm 2.1$ & $\mathbf{318.1 \pm 1.0}$ \\
Hardcore & $-47.2 \pm 10.6$ & $-44.3 \pm 1.6$ & $29.7 \pm 9.9$ & $23.8 \pm 8.3$ & $82.6 \pm 8.5$ & $163.3 \pm 30.9$ & $\mathbf{236.0 \pm 8.9}$ \\
Stairs & $-27.4 \pm 12.1$ & $-2.6 \pm 2.6$ & $22.1 \pm 6.3$ & $23.3 \pm 4.4$ & $48.0 \pm 4.3$ & $59.4 \pm 10.5$ & $\mathbf{91.7 \pm 8.9}$ \\
PitGap & $-68.2 \pm 9.7$ & $-79.3 \pm 0.5$ & $\mathbf{98.8 \pm 24.9}$ & $11.0 \pm 7.6$ & $46.2 \pm 11.3$ & $49.6 \pm 12.6$ & $\mathbf{133.3 \pm 39.1}$ \\
Stump & $-76.0 \pm 10.3$ & $-65.0 \pm 18.4$ & $-22.4 \pm 17.2$ & $-5.4 \pm 5.5$ & $7.5 \pm 6.4$ & $44.6 \pm 49.8$ & $\mathbf{188.8 \pm 10.9}$ \\
Roughness & $-5.1 \pm 25.9$ & $-1.2 \pm 7.7$ & $44.7 \pm 11.6$ & $52.3 \pm 9.0$ & $126.7 \pm 7.3$ & $211.7 \pm 21.5$ & $\mathbf{248.9 \pm 12.3}$ \\
\midrule
\textbf{Mean} & $-2.9 \pm 14.5$ & $-6.3 \pm 24.6$ & $79.1 \pm 17.5$ & $61.1 \pm 12.6$ & $102.5 \pm 13.0$ & $140.9 \pm 23.0$ & $\mathbf{202.8 \pm 13.6}$ \\
\bottomrule
\end{tabular}}
\end{center}
\end{table}

\subsection{Ablations}
\begin{figure}[t!]
    \centering
    \includegraphics[width=0.8\textwidth]{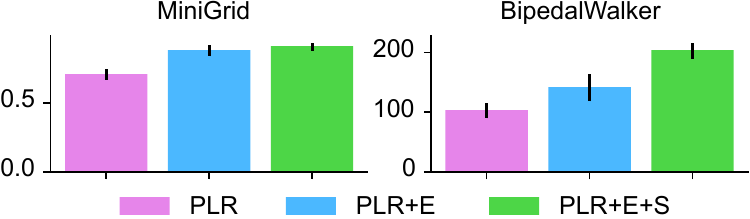}
    \caption{\small{Aggregate returns for Editing ablations in MiniGrid and BipedalWalker. E=editing, S=start simple.}}
    \label{fig:accel_ablation_editing}
\end{figure}

We conduct a simple ablation study to test the importance of ACCEL's editing mechanism and its inductive bias of starting simple. In Figure~\ref{fig:accel_ablation_editing} we show the performance of three approaches: PLR (sample and replay DR levels), PLR+E (sample, replay, and edit DR levels) and finally PLR+E+S (i.e. ACCEL). As we see, editing levels leads to improved performance, while starting simple is more important in BipedalWalker environments.

Next, we investigate additional design variations of ACCEL's editing mechanism: via random mutations applied to a subsample of size 1 of the last batch selected for level replay (standard ACCEL), via random mutations applied to the full batch of levels sampled for level replay (``Edit Batch"), via a learned editing policy trained with PPO to maximize the PVL incurred by the student (``Learned Editor"), and finally, by considering the ``No Editor" ablation, where the editing step is replaced by simply sampling an equivalent number of additional levels from the DR generator. For a fair comparison to this last configuration, all ablations use the DR generator, rather than the empty generator. The level replay rate is set to $10\%$ for all methods. We train each ablation for 10k PPO updates and evaluate each on the full set of $15\times15$ OOD maze environments. The results in Table \ref{table:minigrid_editing_ablations} show that editing the full batch of replay levels results in slightly worse zero-shot performance than editing only a single level from the replay batch. Likewise, using a learned editor that seeks to maximize the PVL of the resulting levels degrades zero-shot performance. Still, each of the ablations that actively edit levels still outperforms the next best baseline, PLR$^{\perp}$, which sees a mean solved rate of 0.69 over all OOD environments. Finally, the ``No Editor" ablation performs worse than PLR, showing that ACCEL's strong performance derives from level editing.

\begin{table}[h]
\begin{center}
\caption{\small{Zero-shot transfer to human-designed environments. Each entry is the mean and standard error of five independent runs, where each run is evaluated for 100 trials on each environment. All methods use a DR generator that places between 0 and 60 blocks.}}
\label{table:minigrid_editing_ablations}
\scalebox{0.9}{
\begin{tabular}{ l | lllr } 
\toprule
\textbf{Env.} & ACCEL & Edit Batch & Learned Editor & No Editor \\ 
\midrule 
16Rooms & $1.0 \pm 0.0$ & $0.76 \pm 0.19$ & $0.9 \pm 0.07$ & $0.84 \pm 0.06$ \\
16Rooms2 &  $0.51 \pm 0.28$ & $0.23 \pm 0.16$ & $0.41 \pm 0.19$ & $0.68 \pm 0.18$ \\
SimpleCrossing & $0.8 \pm 0.05$ & $1.0 \pm 0.0$ & $0.9 \pm 0.1$ & $0.75 \pm 0.05$ \\
FourRooms &  $0.85 \pm 0.05$ & $0.85 \pm 0.06$ & $0.88 \pm 0.04$ & $0.88 \pm 0.05$ \\
SmallCorridor &  $0.72 \pm 0.1$ & $0.74 \pm 0.1$ & $0.6 \pm 0.17$ & $0.7 \pm 0.18$ \\
LargeCorridor &  $0.91 \pm 0.05$ & $0.75 \pm 0.08$ & $0.56 \pm 0.18$ & $0.63 \pm 0.18$ \\
Labyrinth &  $0.98 \pm 0.02$ & $0.85 \pm 0.11$ & $0.99 \pm 0.01$ & $0.67 \pm 0.19$ \\
Labyrinth2 &  $0.97 \pm 0.03$ & $0.83 \pm 0.11$ & $0.7 \pm 0.15$ & $0.48 \pm 0.2$ \\
Maze &  $0.78 \pm 0.21$ & $0.87 \pm 0.05$ & $0.57 \pm 0.18$ & $0.15 \pm 0.08$ \\
Maze2 &  $0.5 \pm 0.24$ & $0.67 \pm 0.18$ & $0.65 \pm 0.15$ & $0.23 \pm 0.15$ \\
Maze3 & $0.79 \pm 0.14$ & $0.9 \pm 0.08$ & $0.95 \pm 0.05$ & $0.56 \pm 0.17$ \\
\midrule
\textbf{Mean} & $0.79 \pm 0.04$ & $0.76 \pm 0.04$ & $0.74 \pm 0.04$ & $0.58 \pm 0.05$ \\
\bottomrule
\end{tabular}}
\end{center}
\end{table}

\subsection{Comparison to POET}

For a more direct comparison to POET, we train each method using 10 training seeds for 50k student PPO updates with the smaller 5D BipedalWalker environment encoding used in \citet{poet}. We use the thresholds provided in \citet{poet}, summarized in Table~\ref{table:poet_thresholds}, to evaluate the difficulty of generated levels. A level meeting none of these thresholds is considered ``easy", while meeting one, two or three is considered ``challenging," ``very challenging," or ``extremely challenging" respectively. 

\begin{table}[h!]
\begin{center}
\caption{\small{Environment encoding thresholds for 5D BipedalWalker.}}
\label{table:poet_thresholds}
\scalebox{0.85}{
\begin{tabular}{ ccc } 
\toprule
\textbf{Stump height (high)} & \textbf{Pit gap (high)}  & \textbf{Ground roughness} \\ 
\midrule 
$\geq2.4$ & $\geq6$ & $\geq4.5$ \\
\bottomrule
\end{tabular}}
\end{center}
\end{table}

\begin{figure}[h!]
    \centering
    \includegraphics[width=.65\textwidth]{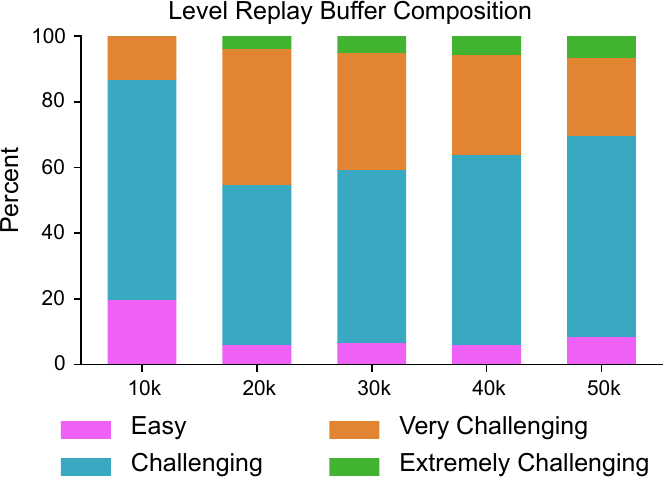}
    \caption{\small{Percent of ACCEL level replay buffer for each difficulty. This complexity emerges purely in pursuit of high-regret levels.}}
    \label{fig:bipedal_poet_progression}
\end{figure}

\noindent The difficulty composition of the ACCEL level replay buffer during training is shown in Figure~\ref{fig:bipedal_poet_progression}. ACCEL produces an increasing number of extremely challenging levels as training progresses. This is a significant achievement given that POET's evolutionary curriculum is unable to create levels in this category, without including a complex and computationally-costly stepping-stone procedure~\citep{poet}. We thus see minimax regret UED offers a simpler and cheaper alternative to producing such levels. Moreover, POET explicitly encourages the environment parameters to reach high values through a novelty bonus that relies on either a population~\citep{enhanced_poet} or domain-specific knowledge~\cite{poet} to compute, whereas the complexity discovered by ACCEL emerges purely through the pursuit of high-regret levels---as estimated via PVL, a domain-agnostic, single-agent regret estimator.

Despite the similar degrees of emergent complexity between POET and ACCEL, the underlying goals of each method, in some sense, take opposite approaches toward producing a potentially open-ended curriculum of challenges: While POET seeks to discover a diverse population of specialists, each capable of solving a specific extremely challenging level, ACCEL aims to train a single generalist. To evaluate the generality of the ACCEL agent, we test all agents trained in the 5D BipedalWalker environment on the settings outlined in Figure~\ref{fig:bipedal_ood_test_curves}, and report the results in Table~\ref{table:poet_exp_robustness}. Note that in this case, the Stairs environment is OOD, as the agent never sees stairs during training under the 5D environment parameterization. As we saw in the higher-dimensional setting, the resultant generalist ACCEL agent is able to perform well across test environments.

\begin{table}[h!]
\begin{center}
\caption{\small{Test solved rates at 50k updates (mean and standard error) for 10 runs of each method on 100 episodes. Extremely challenging evaluation uses 1000 episodes due to the high diversity of levels. Bold values are within one standard error of the best mean.}}
\label{table:poet_exp_robustness}
\scalebox{0.9}{
\begin{tabular}{ l ccc } 
\toprule
 & PLR & ALP-GMM & ACCEL \\ 
\midrule 
Stump & $0.04 \pm 0.02$ & $0.07 \pm 0.02$ & $\mathbf{0.44 \pm 0.08}$ \\
PitGap & $0.2 \pm 0.09$ & $\mathbf{0.58 \pm 0.08}$ & $\mathbf{0.61 \pm 0.08}$ \\
Roughness & $0.23 \pm 0.04$ & $0.13 \pm 0.03$ & $\mathbf{0.73 \pm 0.03}$ \\
Stairs & $0.02 \pm 0.0$ & $0.01 \pm 0.0$ & $\mathbf{0.12 \pm 0.02}$ \\
\midrule
Hardcore & $0.21 \pm 0.04$ & $0.2 \pm 0.04$ & $\mathbf{0.65 \pm 0.02}$ \\
Extreme &  $0.01 \pm 0.01$ & $0.02 \pm 0.01$ & $\mathbf{0.12 \pm 0.02}$ \\
\bottomrule
\end{tabular}}
\end{center}
\end{table}

We further test all methods on a held-out distribution of ``extremely challenging" levels. In this case, we resample the level parameters per episode so to ensure they meet all three criteria in Table~\ref{table:poet_thresholds}, leading to a highly diverse set of test levels. To mitigate the effect of policy stochasticity in influencing outcomes, we evaluate each method over 1000 episodes. The results are summarized in Table~\ref{table:poet_exp_robustness}, where we see ACCEL attains 12\% average solved rate, while PLR and ALP-GMM see 1\% and 2\% average solved rates respectively.

Finally, we seek to evaluate our agents on specific levels produced by POET. We used the rose plots from \citet{poet} to create six ``extremely challenging" environments, each solved by one of the three reported POET runs. It is important to emphasize that POET co-evolves its agent population with discovered levels to ensure each level is solved by at least one POET agent. As POET agents follow deterministic policies, once an agent and level pair are found, that agent will always successfully solve that level, which might grow arbitrarily complex. Unsurprisingly, the results in Table~\ref{table:poet_roseplot_test} show that ACCEL agents find these levels challenging, attaining low success rates. The difficulty posed to ACCEL agents by level settings co-evolved by POET highlights the relative benefits of specialists over generalists. Still, 9 out of 10 of our independent runs solved at least one of the 6 environments at least once out of 100 trials. 
Moreover, it is important to note that our experimental setup does not perform a perfect comparison: POET fixes the level generator's random seed, thereby producing a single fixed level per parameterization, while we repeatedly sample different levels using different random seeds per parameterization. 

\begin{table}[H]
\begin{center}
\caption{\small{Test performance on extremely challenging levels produced by POET. For each level, we run 100 trials with different random seeds. Mean shows the mean performance across all 10 ACCEL runs and trials. Max shows the best performance out of all runs and trials for each environment.}}
\label{table:poet_roseplot_test}
\scalebox{0.87}{
\begin{tabular}{ l | cccccc } 
\toprule
 & 1a  & 1b & 2a & 2b & 3a & 3b   \\ 
\midrule
\textbf{Mean} & $0.01$ & $0.01$ & $0.00$ & $0.03$ & $0.01$ & $0.12$ \\
\textbf{Max} & $0.03$ & $0.05$ & $0.00$ & $0.08$ & $0.03$ & $0.31$ \\
\bottomrule
\end{tabular}}
\end{center}
\end{table}

In light of these results, we believe ACCEL can produce levels of comparable complexity to POET at the fraction of the compute cost, without requiring a large population or domain-specific heuristics. Moreover ACCEL produces a single agent robust across environment challenges, while POET results in multiple agents, each tailored to individual challenges. Therefore, we believe our method produces agents that are more robust, and thus more generally capable. In practice, whether a generalist or a population of specialists should be favored largely depends on the application domain. These trade-offs are discussed at length in Section~\ref{sec:accel_discussion}.

\section{Related Work}

\begin{table*}[t!]
\footnotesize
\begin{center}
\caption{\small{The components of related approaches. Like POET, ACCEL evolves levels, but only trains a single agent while using a minimax-regret objective to ensure levels are solvable. PAIRED uses minimax regret to train the generator, and does not replay levels. Finally, PLR curates levels using minimax regret, but relies solely on domain randomization for generation.}}
\label{table:summary_of_diff}
\scalebox{0.95}{
\begin{tabular}{ ccccc } 
\toprule
\textbf{Algorithm}   & \textbf{Generator Strategy} & \textbf{Generator Obj} & \textbf{Curation Obj} &  \textbf{Output} \\ \midrule
POET \citep{poet}  & Evolution & Minimax  & MCC & Specialists \\ 
PAIRED \citep{paired}  & Reinforcement Learning & Minimax Regret  & None & Generalist \\ 
PLR \citep{plr, jiang2021robustplr}  & Random & None  & Minimax Regret & Generalist \\ 
\midrule
ACCEL  & Random + Evolution  & None & Minimax Regret & Generalist \\ 
\bottomrule
\end{tabular}}
\end{center}
\end{table*}

The evolutionary component of ACCEL is inspired by the open-ended creative potential of POET \citep{poet, enhanced_poet, mcc_og, pinsky}, which seeks to train a population of highly capable specialists. In contrast, ACCEL trains a single generally capable agent with a regret-based curriculum as in PAIRED~\citep{paired} and PLR$^{\perp}$. Table~\ref{table:summary_of_diff} summarizes the relationship between these diverse methods under the DCD framework introduced in Chapter~\ref{chapter:dcd}. 

In the field of procedural content generation (PCG), such evolutionary mechanisms have been applied to the design of videogame levels~\citep{khalifa2022learning}. We are particularly inspired by PCGRL \citep{pcgrl,controllablepcgrl2021earle} which frames level design as an RL problem, making incremental changes to a level to maximize some objective subject to game-specific constraints. Our work also closely relates to previous environment design literature in the symbolic AI commmunity \citep{zhang2008ed,zhang2009ed, keren2017equi, keren2019efficient}, which developed methods for making small changes to an environment in order to influence an agent's behavior. Unlike these previous works, ACCEL edits environments to produce an autocurriculum that facilitates the learning of robust behavior.

More broadly, the field of evolutionary computation presents a rich space of ideas that can likely be integrated into ACCEL-like autocurriculum methods to improve on some of ACCEL's shortcomings. We discuss how these ideas can improve ACCEL in Section~\ref{sec:accel_discussion}.

\section{Discussion and Limitations}
\label{sec:accel_discussion}
We proposed ACCEL, a new method for unsupervised environment design (UED), that evolves a curriculum by \emph{editing} previously curated, high-regret levels. Such edits induce an evolutionary process that leads to a wide variety of environments at the frontier of the agent's capabilities, resulting in autocurricula over training environments that start simple and quickly compound in complexity. Thus, ACCEL provides a principled regret-based curriculum that exploits an evolutionary process to produce training levels matched to the agent's current capabilities. Importantly, unlike many previous evolutionary methods, ACCEL avoids the need for domain-specific heuristics. Our experiments demonstrate that ACCEL is capable of training robust agents across several challenging domains, where ACCEL agents significantly outperform previous UED methods in OOD transfer. 

In comparison to POET~\citep{poet}, a population-based approach for generating an evolutionary autocurriculum across environment instances, ACCEL produces levels of comparable complexity. However, the end result of ACCEL differs from that of POET. The primary motivation of ACCEL is to produce a single robust agent that can solve a wide range of challenges. In contrast, POET co-evolves agent-environment pairs in order to find specialized policies that each act as the expert for a single, highly specialized task. In this way, POET's specialized agents can likely learn to solve challenging environments outside the capabilities of ACCEL's generalist agents, but at the cost of potentially overfitting to their paired levels. Thus, unlike ACCEL, the policies produced by POET should not be expected to be robust across the full distribution of levels. However, a specialist approach may be better for some application domains. For example, if the goal is to maximize performance on a particular set of tasks known in advance, then assigning a specialist to each task would yield the best performance. In contrast, a generalist may need to make performance trade-offs across these tasks, but may be expected to more robustly adapt in new scenarios. Moreover, specialist approaches like POET may benefit training, where a population of specialists may encode a broader set of behaviors that allow the autocurriculum to explore a wider set of environments where at least some part of the population can make learning progress. Given the trade-offs between these approaches, a particularly exciting direction for future research is to develop methods that effectively distill a population of specialist models into a single generalist model or that combines them dynamically as a mixture of experts. The relative merits of POET and ACCEL thus highlight the fundamental trade-offs between specialization and generalization, both of which play important roles in generalist systems that seek to solve a large variety of tasks.

While ACCEL's simplicity is appealing, larger design spaces may require additional mechanisms, like using more powerful evolutionary search algorithms~\citep{hansen2001completely,wierstra2014natural,salimans2017evolution} or actively promoting diversity in level design via novelty search~\citep{lehman2011novelty,lehman2011evolving,conti2018improving} and quality-diversity search~\citep{pugh2016quality,mouret2015illuminating,colas2020scaling,fontaine2020covariance,gaier2018data}. Such diversity-inducing methods may help mitigate the possibility of ACCEL's evolutionary search collapsing into specific environment subspaces. Moreover, ACCEL uses an inductive bias by starting with the simplest base case (e.g. an empty room), which may not always be a suitable idea in practice. In some settings, structurally-simple levels may be extremely difficult and thereby hinder the agent's learning, e.g. in a hide-and-seek game, where fewer objects in the environment makes the task more difficult. Thus, search methods like MAP-Elites~\citep{mouret2015illuminating}, which can segment the environment space into distinct classes can be used to ensure more comprehensive, structured exploration of the environment space. Ultimately, such methods may be necessary for ensuring enough diversity for robust sim2real transfer to the open-ended possibilities of the real world. It remains an open question whether producing sufficient diversity for such transfer would require a population, e.g. in order to use the domain-agnostic, population-based novelty objective in Enhanced POET \citep{enhanced_poet}. The core regret-based evolutionary curriculum of ACCEL can, in principle, be combined with a method like POET, to produce an MCC-based algorithm to produce a diverse population of minimax regret policies specialized to distinct subsets of the environment space.

There are also many possibilities for innovating on the variation operator that ACCEL uses to perform edits. For example, the editor itself might be parameterized as a population of controllable level generators~\citep{earle2021illuminating,controllablepcgrl2021earle} or even a large language model (LLM). The advent of powerful LLMs make it possible to perform evolutionary search with the LLM as an intelligent variation operator that encapsulates domain-relevant structure, e.g. generating code diffs to mutate a natural-language or programmatic encoding of a solution~\citep{lehman2022evolution,meyerson2023language,wang2023voyager,zhang2023omni}. LLMs thus present a promising means to extend ACCEL-style autocurricula to more complex environment design spaces, assuming the environment can be encoded in a natural or structured language representation. Such learned editors might be pre-trained and fine-tuned based on the actual regret estimates incurred by the student. Moreover, edits can occur in the compact latent space of a generative model~\citep{lsi_overcooked}, which may allow for more efficient search. Other potentially helpful ideas from evolutionary computation include directly searching for levels that are likely to produce more useful levels in the future \citep{evolvabilityes}, as well as introducing so-called \emph{extinction events}~\citep{raup1986biological, lehman2015extinction}, believed to play a crucial role in natural evolution, and which can lead to finding more robust solutions. The interplay between evolutionary computation and UED presents an fascinating frontier for future reseach.

Finally, we note that while ACCEL may be an effective approach for automatically generating an effective curriculum, it may still be necessary to likewise adapt the agent model and optimizer hyperparameters~\citep{pbt,autorl_survey} to most effectively train agents in open-ended autocurricula.

\newcommand{\PGround}{\overline{P}}
\newcommand{\PCur}{P}
\newcommand{\QGround}{\overline{Q}}
\newcommand{\QCur}{Q}

\newcommand{\ThirdSectionVGround}[1]{\overline{V}(\pi\vert \tau_t {#1})}
\newcommand{\ThirdSectionVOpt}{V^*(\pi \vert \tau_t)}
\newcommand{\ThirdSectionV}{V(\pi \vert \tau_t)}
\newcommand{\ThirdSectionChoiceSet}{\pi}

\newcommand{\VGround}{\overline{U}}
\newcommand{\VGroundOverD}[2]{\VGround_{#1}(#2)}
\newcommand{\VGroundPiAfterT}[2]{\VGround(#1 \vert #2)}
\newcommand{\GTU}{ground-truth utility function}

\newcommand{\GTUEqUnconditioned}{\mathbb{E}\left[\sum_{t=0}^{\infty}\gamma^t r_t\right]}

\newcommand{\GTUEqConditionedX}{\mathbb{E}_{\tau, \theta \sim \overline{P}(\theta \vert X)}\left[\sum_{t=0}^{\infty}\gamma^t r_t\right]}

\newcommand{\csabbrev}{CICS}

\chapter{Aligning Curricula}
\label{chapter:samplr}

\section{Introduction}
\label{section:samplr_intro}
On one hand, the test-time robustness induced by the autocurricula studied in previous chapters comes ``for free," deriving purely from changes to the sequence of tasks (and thus data) provided to the agent during training. No further changes to the agent model or optimizer are required. On the other hand, these changes come at the cost of training on biased data: By definition, curricula change the presentation of training data, which often alters the underlying training distribution with respect to some ground-truth reference distribution of tasks. 
Problematically, in partially-observable or stochastic settings, optimal policies may depend on the ground-truth distribution over certain random parameters of the environment in the intended deployment setting. As, curriculum learning necessarily shifts the training distribution, UED methods like PLR and ACCEL can thus result in suboptimal policies at deployment. 
Directly sampling these parameters from the ground-truth distribution avoids the issue, but prevents the application of curriculum learning. Ideally, we desire a method that can fully exploit the benefits of curriculum learning while avoiding any detrimental biases resulting from training on the resulting biased data. 

This chapter formalizes and presents a solution to this fundamental problem of curriculum learning in RL, which we call \emph{curriculum-induced covariate shift} (\csabbrev{}). Analogous to the covariate shift that occurs in supervised learning~\citep{huang2006correcting}, \csabbrev{} refers to a mismatch between the \textit{input distribution} at training and test time. In the case of RL, we will show this becomes problematic when the shift occurs over the \emph{aleatoric parameters} of the environment---those aspects of the environment holding irreducible uncertainty even in the limit of infinite experiential data~\citep{der2009aleatory}. Such parameters correspond to those factors of variation in the environment whose value cannot be fully determined at each point of the agent's trajectory. Devising autocurriculum methods that avoid \csabbrev{} thus presents an important alignment problem, whereby we wish to ensure the compatibility of the resultant policy with the inherent stochasticity of a particular test-time domain. This challenge embodies the fundamental tension between the creative, open-ended potential of autocurricula and the need for controlling such processes to ensure sensible and safe behaviors in specific test settings that can be known in advance.

As in previous chapters, we cast our discussion under the lens of Unsupervised Environment Design \citep[UED,][]{paired}, to establish precise language around adaptive curricula. UED allows us to view adaptive curricula as emerging via a multi-player game between a \emph{teacher} that proposes environments with parameters $\theta \sim P(\Theta)$ and a \emph{student} that learns to solve them. In addition to notational clarity, this formalism enables using game theoretic constructs, such as Nash equilibria \citep[NE,][]{nash1950equilibrium}, to analyze curricula.
This game-theoretic view has led to the development of curriculum methods with principled robustness guarantees, such as PAIRED \citep{paired}, PLR$^{\perp}$~\citep{jiang2021robustplr}, and ACCEL~\citep{accel}, which aim to maximize a student's regret and lead to minimax regret \citep{savage1951theory} policies at NE. Thus, at NE, the student can solve all solvable environments within the training domain. However, in their current form the UED robustness guarantees are misleading: if the UED curriculum deviates from a ground-truth distribution $\PGround(\Theta)$ of interest, i.e. the distribution at deployment, with respect to aleatoric parameters $\Theta' \subset \Theta$, the resulting policies may be suboptimal under the ground-truth distribution $\PGround$. 

\begin{figure}[t!]
    \includegraphics[width=\textwidth]{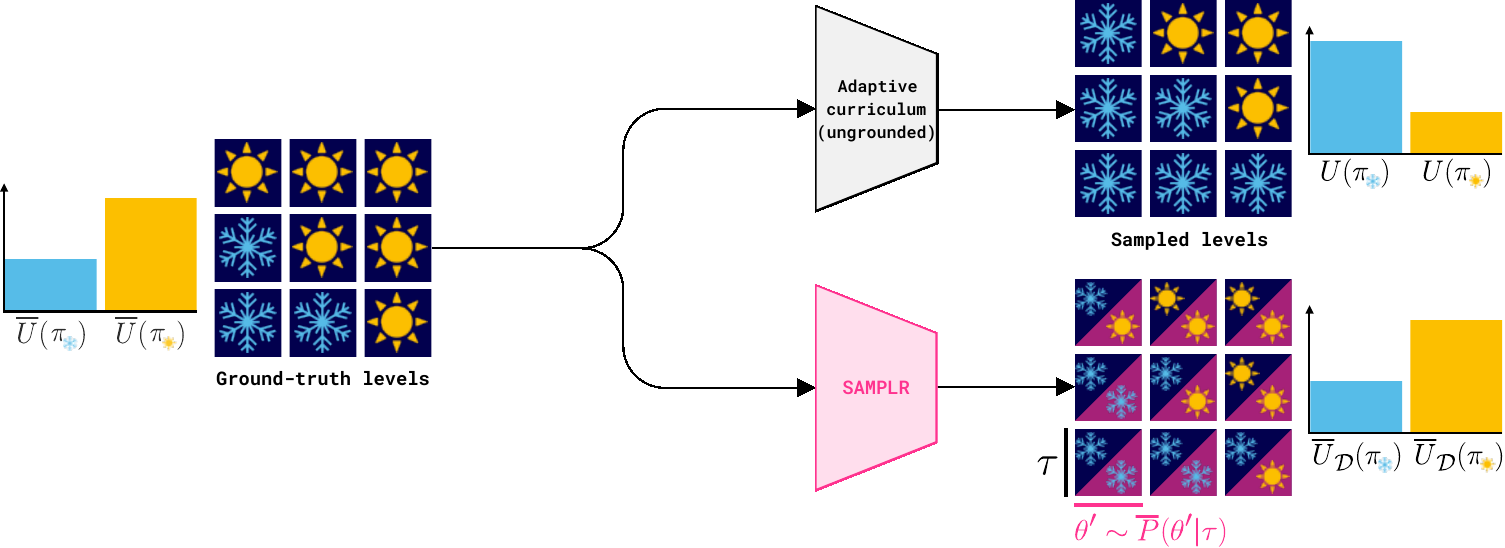}
    \caption{\small{Curricula can result in covariate shifts in environment parameters with respect to the ground-truth distribution $\overline{P}(\Theta)$ (top path), e.g. whether a road is icy or not, which can cause the policy to be optimized for a utility function $U$ differing from the ground-truth utility function $\overline{U}$ based on $\overline{P}$ (See Equation~\ref{equation:grounded_value_function}). Here, the policies $\pi$\textsubscript{\includegraphics[scale=2]{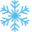}} and $\pi$\textsubscript{\includegraphics[scale=2]{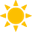}} drive assuming ice and no ice respectively. SAMPLR (bottom path) matches the distribution of training transitions to that under $\overline{P}(\Theta|\tau)$ (pink triangles), thereby ensuring the optimal policy trained under a biased curriculum retains optimality for the ground-truth distribution $\overline{P}$.
    }}
    \label{fig:covariate_shift_overview}
\end{figure}

For a concrete example of how \csabbrev{} can be problematic, consider the case of training a self-driving car to navigate potentially icy roads, when icy conditions rarely occur under $\PGround$. When present, the ice is typically hard to spot in advance; thus, the aleatoric parameters $\Theta'$ correspond to whether each section of the road is icy. 
A priori, a curriculum should selectively sample more challenging icy settings to facilitate the agent's mastery over such conditions. However, this approach risks producing an overly-pessimistic agent (i.e. one that assumes that ice is common), driving slowly even in fair weather. Such a policy leads to inadequate performance on $\PGround$, which features ice only rarely.

We can preserve optimality on $\PGround$ by \emph{grounding the policy}---that is, ensuring that the agent acts optimally with respect to the \emph{\GTU{}} for any action-observation history $\tau$ 
and the implied ground-truth posterior over $\Theta$:
\begin{equation}
\label{equation:grounded_value_function}
\VGroundPiAfterT{\pi}{\tau} = \mathbb{E}_{\theta \sim \PGround(\theta | \tau)} \left[\; \VGroundPiAfterT{\pi}{\tau, \theta}\right],
\vspace{-2mm}
\end{equation} 

where the ground-truth utility conditioned on $X$, $\VGround(\pi \vert X)$, is defined to be  $\GTUEqConditionedX$, for rewards $r_t$ and a discount $\gamma$. 

We can ground the policy by \emph{grounding the training distribution}, which means constraining the training distribution of aleatoric parameters $P(\Theta')$ to match $\PGround(\Theta')$. This is trivially accomplished by directly sampling $\theta' \sim \PGround(\Theta')$, which we call \emph{naive grounding}. 
Unfortunately, this approach makes many curricula infeasible by removing the ability to selectively sample environment settings over aleatoric parameters. Applying this strategy to the self-driving agent may result in a policy that is optimal in expectation under $\PGround$ where there is rarely ice, but nevertheless fails to drive safely on ice. 

We wish to maintain the ability to bias a training distribution, since it is required for curriculum learning, while ensuring the resulting decisions remain optimal in expectation under $\PGround$. This goal is captured by the following objective:
\begin{equation}
\label{equation:grounding_objective}
\VGroundOverD{\mathcal{D}}{\pi} = \mathbb{E}_{\tau \sim \mathcal{D}}\left[\; \VGroundPiAfterT{\pi}{\tau} \right],
\vspace{-2mm}
\end{equation}

\noindent where $\mathcal{D}$ is the training distribution of $\tau$. Under naive grounding, $\mathcal{D}$ is equal to $\PGround(\tau)$ and Equation \ref{equation:grounding_objective} reduces to $\VGround(\pi)$. To overcome the limitations of naive grounding, we develop an approach that allows $\mathcal{D}$ to deviate from $\PGround(\tau)$, e.g. by prioritizing levels most useful for learning, but still grounds the policy by evaluating decisions following potentially biased training trajectories $\tau$ according to $\VGround(\pi|\tau)$. Figure~\ref{fig:covariate_shift_overview} summarizes this approach, and contrasts it with an ungrounded adaptive curriculum. 

This chapter first develops the formalization of \csabbrev{} in Section~\ref{section:curriculum_bias}. Then, Section~\ref{section:method}, presents a general algorithmic solution for maintaining the grounding depicted in Figure~\ref{fig:covariate_shift_overview} and integrates this approach with PLR$^{\perp}$, resulting in a new algorithm called \emph{Sample-Matched PLR} that preserves optimality on $\PGround$. In Section~\ref{section:theory}, we prove that SAMPLR promotes Bayes-optimal policies that are robust over all environment settings $\theta \sim \PGround(\Theta)$. Finally, in \ref{section:experiments}, we demonstrate on several challenging environments, spanning stochastic partially-observable navigation and pixel-based continuous control, that SAMPLR learns highly robust policies, whereas \plrabbrev{} fails due to CICS.

\section{Curriculum-Induced Covariate Shift}
\label{section:curriculum_bias}

Since UED algorithms formulate curriculum learning as a multi-agent game between a teacher and a student agent, we can formalize when \csabbrev{} becomes problematic by considering the equilibrium point of this game: Let $\Theta$ be the environment parameters controlled by UED, $\PGround(\Theta)$, their ground-truth distribution, and $\PCur(\Theta)$, their curriculum distribution at equilibrium.
We use $\tau_t$ to refer to the joint action-observation history (AOH) of the student until time $t$ (and simply $\tau$ when clear from context).
Letting $\ThirdSectionV$ denote the value function under the curriculum distribution $\PCur(\Theta)$, we characterize an instance of CICS over $\Theta$ as \textit{problematic} if the optimal policy under $\PCur(\Theta)$ differs from that under the ground-truth $\PGround(\Theta)$ for some $\tau_t$, so that
\[
\underset{\ThirdSectionChoiceSet}{\arg \max}~ \ThirdSectionV \neq \underset{\ThirdSectionChoiceSet}{\arg \max}~ \ThirdSectionVGround{}.
\]
The value function $\ThirdSectionVGround{}$ with respect to $\PGround(\Theta)$ can be expressed as a marginalization over $\theta$:
\begin{align}
\label{eq:q_function_decomposition}
\ThirdSectionVGround{} 
&= \sum_{\theta} \PGround(\theta|\tau_t) \tilde{V}(\pi|\tau_t, \theta)
\propto \sum_{\theta}\PGround(\theta)\tilde{P}(\tau_t|\theta)\tilde{V}(\pi|\tau_t, \theta).
\vspace{-2mm}
\end{align}

Here, the notation $\overline{P}(\theta)$ means $\overline{P}(\Theta = \theta)$, and the tilde on the $\tilde{P}$ and $\tilde{V}$ terms indicates independence from any distribution over $\Theta$, as they both condition on $\theta$. Importantly, the value function under the curriculum distribution  $\ThirdSectionV$ corresponds to Equation \ref{eq:q_function_decomposition} with $\PGround$ replaced by $\PCur$. We see that $\ThirdSectionVGround{}$ is unchanged for a given $\tau_t$ when $\PGround(\theta)$ is replaced with $\PCur(\theta)$ if 1)  $\PGround(\theta^*|\tau_t) = 1$ for some $\theta^*$, and 2) $\PGround$ shares support with $\PCur$. Then $\tilde{P}(\tau_t|\theta) > 0$ iff $\theta = \theta^*$ and zero elsewhere. In this case, the sums each reduce to a constant multiple of $\ThirdSectionVGround{,\theta^*}$, regardless of changing the ground-truth distribution $\overline{P}$ to $P$. In other words, when $\Theta$ is fully determined given the current history $\tau$, covariate shifts over $\Theta$ with respect to $\PGround(\Theta)$ have no impact on policy evaluation and thus the value function for the optimal policy. If the first condition does not hold, the uncertainty over the value of some subset $\Theta' \subset \Theta$ is irreducible given $\tau$, making $\Theta'$ aleatoric parameters for the history $\tau$. Thus, assuming the curriculum shares support with the ground-truth distribution, covariate shifts only alter the optimal policy at $\tau$ when they occur over aleatoric parameters given $\tau$. Such parameters can arise when the environment is inherently stochastic or when the cost of reducing uncertainty is high. 

Crucially, our analysis assumes $\PCur$ and $\PGround$ share support over $\Theta$. When this assumption is broken, the policy trained under the curriculum can be suboptimal for environment settings $\theta$, for which $\PCur(\theta) = 0$ and $\PGround(\theta) > 0$. In this chapter, we specifically assume that $\PCur$ and $\PGround$ share support and focus on addressing suboptimality under the ground-truth $\PGround$ due to CICS over the aleatoric parameters $\Theta'$.

This discussion thus makes clear that problematic CICS can be resolved by \emph{grounding the training distribution}, i.e. enforcing the constraint \mbox{$\PCur(\Theta'|\tau) = \PGround(\Theta'|\tau)$} for the aleatoric parameters of the environment. This constraint results in \emph{grounding the policy}, i.e. ensuring it is optimal with respect to the ground-truth utility function based on $\overline{P}$ (Equation~\ref{equation:grounded_value_function}). As discussed, naive grounding satisfies this constraint by directly sampling $\theta' \sim \PGround (\Theta')$, at the cost of curricula over $\Theta'$. This work develops an alternative for satisfying this constraint while admitting curricula over $\Theta'$.

\section{Sample-Matched PLR (SAMPLR)}
\label{section:method}

\begin{figure}[t]
\begin{minipage}[t!]{\linewidth}
\begin{algorithm}[H] {
\small
\SetAlgoLined
\caption{Sample-Matched PLR (SAMPLR)}
\label{algo:samplr}
Randomly initialize policy $\pi(\phi)$, an empty level buffer $\bm{\Lambda}$ of size $K$, and belief model $\mathcal{B}(s_t|\tau)$. \\
    \While{not converged}{
        Sample replay-decision Bernoulli, $d \sim \PGround_{D}(d)$ \\
        \eIf{$d=0 \;\textup{or}\; |\Lambda| = 0$}{
            Sample level $\theta$ from level generator\\
            Collect $\pi$'s trajectory $\tau$ on $\theta$, with a stop-gradient $\phi_{\bot}$
        }
        {
        Use PLR to sample a replay level from the level store, $\theta \sim \bm{\Lambda}$ \\
        Collect fictitious trajectory $\tau'$ on $\theta$, based on $s'_t \sim \mathcal{B}$ \\
        Update $\pi$ with rewards $\bm{R}(\tau')$
        }
        
        Compute PLR score, $S = \textbf{score}(\tau', \pi)$ \\
        Update $\bm{\Lambda}$ with $\theta$ using score $S$
    }
}
\end{algorithm}
\end{minipage}
\end{figure}

We now describe a general strategy for addressing CICS, and apply it to \plrabbrev{}, resulting in Sample-Matched PLR (SAMPLR). This new UED method features the robustness properties of \plrabbrev{} while mitigating the potentially harmful effects of CICS over the aleatoric parameters $\Theta'$.

As discussed in Section \ref{section:curriculum_bias}, CICS become problematic when the covariate shift occurs over some aleatoric subset $\Theta'$ of the environment parameters $\Theta$, such that the expectation over $\Theta'$ influences the optimal policy. Adaptive curriculum methods like \plrabbrev{} prioritize sampling of environment settings where the agent experiences the most learning. While such a curriculum lets the agent focus on correcting its largest errors, the curriculum typically changes the distribution over aleatoric parameters $\Theta'$, inducing bias in the resulting decisions. Ideally, we can eliminate this bias, ensuring the resulting policy makes optimal decisions with respect to the \GTU{}, conditioned on the current trajectory:

\begin{equation}
\VGroundPiAfterT{\pi}{\tau} = \mathbb{E}_{\theta' \sim \PGround(\theta' | \tau)} \left[\; \VGroundPiAfterT{\pi}{\tau, \theta'}\right].
\end{equation} 

A naive solution for grounding is to simply exclude $\Theta'$ from the set of environment parameters under curriculum control. That is, for each environment setting proposed by the curriculum, we \mbox{resample $\theta' \sim \PGround$}. We refer to this approach as \emph{naive grounding}. Naive grounding forces the expected reward and next state under each transition at the current AOH $\tau$ to match that under $\PGround$. Thus, optimal policies under naive grounding must be optimal with respect to the ground-truth distribution over $\theta'$.

\begin{figure}{r}
    \centering\includegraphics[width=0.4\textwidth,valign=t]{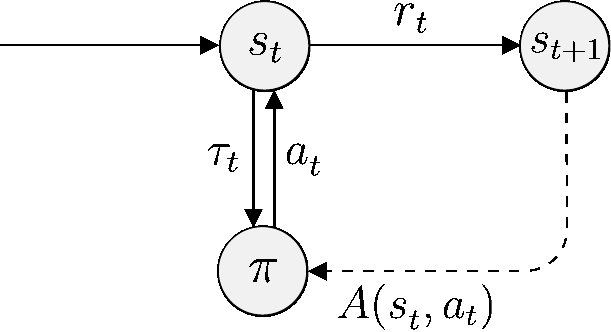}
    \quad\quad
    \centering\includegraphics[width=0.4\textwidth,valign=t]{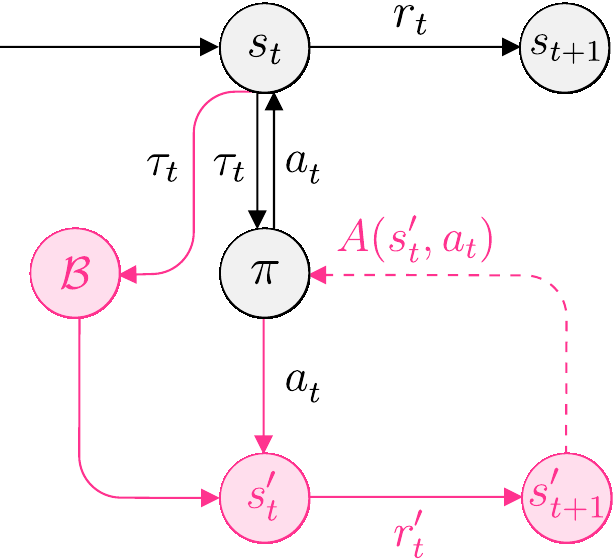}
    \caption{\small{A standard RL transition (top) and a fictitious transition used by SAMPLR (bottom). $A$ is the advantage function.
    }}
    \label{fig:fict_transition}
\end{figure}

While technically simple, naive grounding suffers from lack of control over $\Theta'$. This limitation is of no concern when the value of $\Theta'$ does not alter the distribution of $\tau$ until the terminal transition, e.g. when $\Theta'$ is the correct choice in a binary choice task, thereby only influencing the final, sparse reward when the right choice is made. In fact, our initial experiment in Section \ref{section:experiments} shows naive grounding performs well in such cases. However, when the value of $\Theta'$ changes the distribution of $\tau$ before the terminal transition, the agent may benefit from a curriculum that actively samples levels which promote learning robust behaviors under unlikely events. Enabling the full benefits of the curriculum in such cases requires the curriculum to selectively sample values of $\Theta'$.
Instead of naive grounding, we aim to ground only the policy updates, allowing the curriculum to bias the training distribution. This can be accomplished by optimizing the following objective: 
\begin{equation}
\label{equation:grounded_value_function_D}
\VGroundOverD{\mathcal{D}}{\pi} = \mathbb{E}_{\tau \sim \mathcal{D}}\left[\; \VGroundPiAfterT{\pi}{\tau} \right].
\end{equation}
To achieve this, we replace the reward $r_t$ and next state $s_{t+1}$ with counterfactual values that would be experienced if $\theta'$ were consistent with $\tau$ and $\PGround$, so that $\theta' \sim \PGround(\theta'|\tau)$. This substitution occurs by simulating a \emph{fictitious transition}, where the fictitious state is sampled as $s'_{t} \sim \mathcal{B}(s'_t|\tau)$, the action as $a_t \sim \pi(\cdot|\tau)$ (as per usual), the fictitious next state as $s'_{t+1} = \mathcal{T}(s'_t, a_t)$, and the fictitious reward as $r'_t = \mathcal{R}(s'_{t+1})$. The belief model $\mathcal{B}(s'_t|\tau)$ is the ground-truth posterior of the current state given $\tau$:
\begin{equation}
\label{eq:belief_model}
\vspace{-0.5mm}
\mathcal{B}(s_t|\tau) = \sum_{\theta'}\PGround(s_t|\tau, \theta')\PGround(\theta'|\tau).
\vspace{-0.5mm}
\end{equation}
Fictitious transitions, summarized in Figure~\ref{fig:fict_transition}, ground the observed rewards and state transitions to $\PGround$. Should training on these transitions lead to an optimal policy over $\Theta$, this policy will also be optimal with respect to $\PGround$. We prove this property in Section \ref{section:theory}. Fictitious transitions thus provide the benefit of naive grounding without giving up curriculum control over $\Theta'$.

In general, we implement $\mathcal{B}$ as follows: Given  $\PGround(\Theta')$ as a prior, we model the posterior $\PGround(\theta'|\tau)$ with Bayesian inference. The posterior could be learned via supervised learning with trajectories collected from the environment for a representative selection of $\theta'$. Further, we may only have limited access to $\PGround(\Theta)$ throughout training, for example, if sampling $\PGround(\Theta)$ is costly. In this case, we can learn an estimate $\hat{P}(\Theta')$ from samples we do collect from $\PGround(\Theta)$, which can occur online. We can then use $\hat{P}(\Theta')$ to inform the belief model.

SAMPLR, summarized in Algorithm \ref{algo:samplr}, incorporates this fictitious transition into \plrabbrev{} by replacing the transitions experienced in replay levels sampled by \plrabbrev{} with their fictitious counterparts, as \plrabbrev{} only trains on these trajectories. \plrabbrev{} uses PPO with the Generalized Advantage Estimator \citep[GAE,][]{gae} as the base RL algorithm, where both advantage estimates and value losses can be written in terms of one-step TD errors $\delta_t$ at time $t$. Training on fictitious transitions then amounts to computing these TD errors with fictitious states and rewards: $\delta_t = r'_t + V(s'_t) - V(s'_{t+1})$. Importantly, because \plrabbrev{} provably leads to policies that minimize worst-case regret over all $\theta$ at NE, SAMPLR enjoys the same property for $\theta \sim \PGround(\Theta)$, a fact proven in Section~\ref{section:theory}.

SAMPLR makes two key assumptions: First, the simulator can be reset to a specific state, which is often true, as RL largely occurs in resettable simulators or those that can be made to do so. When a resettable simulator is not available, a possible solution is to learn a model of the environment which we leave for future work. Second, we have knowledge of $\overline{P}(\Theta')$. Indeed, often we do know $\overline{P}$ \emph{a priori}, e.g. empirically or via the domain specification, as in games of chance. %

\section{The Grounded Optimality of SAMPLR}
\label{section:theory}
Training on fictitious transitions is a method for learning an optimal policy with respect to the \GTU{} $\VGroundOverD{\mathcal{D}}{\pi}$ over the distribution $\mathcal{D}$ of training trajectories $\tau$, defined in Equation \ref{equation:grounded_value_function_D}.

When $\mathcal{D}$ corresponds to the distribution of trajectories on levels $\theta \sim \PGround(\Theta)$, $\VGroundOverD{\mathcal{D}}{\pi}$ reduces to the \GTU{}, $\VGround(\pi)$. For any UED method, our approach ensures that, in equilibrium, the resulting policy is Bayes-optimal with respect to $\PGround(\Theta)$ for all trajectories in the support of $\mathcal{D}$.

\begin{remark}
  If $\pi^*$ is optimal with respect to the \GTU{} $\VGroundOverD{\mathcal{D}}{\pi}$ then it is optimal with respect to the ground-truth distribution $\PGround(\Theta)$ of environment parameters on the support of $\mathcal{D}$.
\end{remark}
\begin{proof}
    By definition we have $\pi^* \in \argmax\limits_{\pi \in \Pi}\{\VGroundOverD{\mathcal{D}}{\pi} \} = \argmax\limits_{\pi \in \Pi}\{ \mathbb{E}_{\tau \sim \mathcal{D}}\left[ \VGroundPiAfterT{\pi}{\tau} \right]\}$.  Since $\pi$ can condition on the initial trajectory $\tau$, the action selected after each trajectory can be independently optimized. Therefore, for all $\tau \in \mathcal{D}$, $\pi^* \in \argmax\limits_{\pi \in \Pi}\{\VGroundPiAfterT{\pi}{\tau}\}$ implying that $\pi^*$ is the optimal policy maximizing $\VGroundPiAfterT{\pi}{\tau}$.
\end{proof}

Thus, assuming the base RL algorithm finds Bayes-optimal policies, a UED method that optimizes the \GTU{}, as done by SAMPLR, results in Bayes-optimal performance over the ground-truth distribution. If the UED method maximizes worst-case regret, we can prove an even stronger property we call \emph{robust $\epsilon$-Bayes optimality}.

Let $\VGroundOverD{\theta}{\pi}$ be the \GTU{} for $\pi$ on the distribution $\mathcal{D}_{\theta}^{\pi}$ of initial trajectories sampled from level $\theta$, so that $\VGroundOverD{\theta}{\pi} = \VGroundOverD{\mathcal{D}_{\theta}^{\pi}}{\pi}$. Given a policy $\overline{\pi}$ maximizing $\VGroundOverD{\theta}{\pi}$, we say that $\overline{\pi}$ is robustly $\epsilon$-Bayes optimal iff for all $\theta$ in the domain of $\PGround(\Theta)$ and all $\pi'$, we have
\[
    \VGroundOverD{\theta}{\overline{\pi}} \geq \VGroundOverD{\theta}{\pi'} - \epsilon.
\]
Note how this property differs from being simply $\epsilon$-Bayes optimal, which would only imply that
\[
    \VGround(\overline{\pi}) \geq \VGround(\pi') - \epsilon.
\]
Robust $\epsilon$-Bayes optimality requires $\overline{\pi}$ to be $\epsilon$-optimal on all levels $\theta$ in the support of the ground-truth distribution, even those rarely sampled under $\PGround(\Theta)$. We will show that at $\epsilon$-Nash equilibrium, SAMPLR results in a robustly $\epsilon$-Bayes optimal policy for the \GTU{} $\VGround_{\theta}(\pi)$. In contrast, training directly on levels $\theta \sim \PGround(\Theta)$ results in a policy that is only $\epsilon$-Bayes optimal.

\newtheorem{main_theorem_samplr}{Theorem}
\begin{main_theorem_samplr}
  If $\pi^*$ is $\epsilon$-Bayes optimal with respect to $\VGroundOverD{\widehat{\mathcal{D}}}{\pi}$ for the distribution $\widehat{\mathcal{D}}$ of trajectories sampled under $\pi$ over levels maximizing the worst-case regret of $\pi$, as occurs under SAMPLR, then $\pi^*$ is robustly $\epsilon$-Bayes optimal with respect to the \GTU{}, $\VGround(\pi)$.
\end{main_theorem_samplr}
\begin{proof}

 Let $\pi^*$ be $\epsilon$-optimal with respect to $\VGroundOverD{\widehat{\mathcal{D}}}{\pi}$ where $\widehat{\mathcal{D}}$ is the trajectory distribution under $\pi$
 on levels maximizing the worst-case regret of $\pi$.
 Let $\overline{\pi}^*$ be an optimal grounded policy. Then for any $\theta$,
\begin{align}
     \VGroundOverD{\theta}{\overline{\pi}^*} - \VGroundOverD{\theta}{\pi^*} \leq   \VGroundOverD{\widehat{\mathcal{D}}}{\overline{\pi}^*}- \VGroundOverD{\widehat{\mathcal{D}}}{\pi^*} \leq \epsilon
\end{align}
The first inequality follows from $\widehat{\mathcal{D}}$ being trajectories from levels that maximize worst-case regret with respect to $\pi^*$, and the second follows from $\pi^*$ being $\epsilon$-optimal on $\VGroundOverD{\widehat{\mathcal{D}}}{\pi}$. Rearranging terms gives the desired condition.
\end{proof}

\section{Experiments}
\label{section:experiments}

Unlike in previous chapters, we turn to stochastic environments to evaluate SAMPLR. Our experiments first focus on a discrete, stochastic binary choice task, with which we validate our theoretical conclusions by demonstrating that CICS can indeed lead to suboptimal policies. Moreover, we show that naive grounding suffices for learning robustly optimal policies in this setting. However, as we have argued, naive grounding gives up control of the aleatoric parameters $\Theta'$ and thus lacks the ability to actively sample scenarios helpful for learning robust behaviors---especially important when such scenarios are infrequent under the ground-truth distribution $\PGround(\Theta)$. SAMPLR induces potentially biased curricula, but retains optimality under $\PGround(\Theta)$ by matching transitions under $P(\Theta')$ with those under $\PGround(\Theta')$. We evaluate this approach in a second domain, based on the introductory example of driving on icy roads. In this continuous-control driving domain, we seek to validate whether SAMPLR does in fact learn more robust policies that transfer to tail cases under $\PGround(\Theta')$, while retaining high expected performance on the whole distribution $\PGround(\Theta')$.

All agents are trained using PPO \citep{schulman2017proximal} with the best hyperparameters found via grid search using a set of validation levels. We provide full details on the environments in Appendices~\ref{appendix:env_sfc}–\ref{appendix:env_carracing} and choice of architecture and hyperparameters in Appendix~\ref{appendix:exp_samplr}.

\subsection{Stochastic Fruit Choice}

\begin{figure*}[t!]
    \centering
    \includegraphics[width=\textwidth]{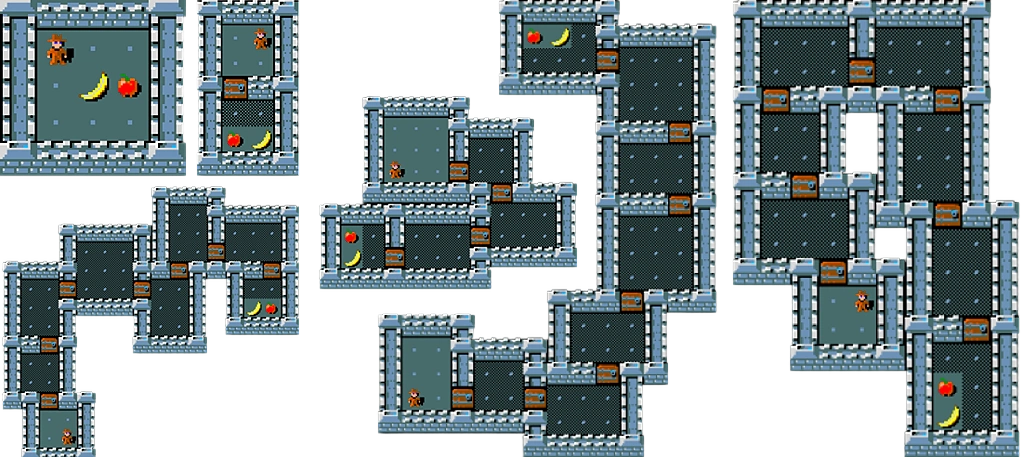}
    \caption{\small{Example Stochastic Fruit Choice levels.}}
    \label{fig:multiroom_n8_main_results}
\end{figure*}

\begin{figure}[t!]
    \centering
    \centering
    \includegraphics[width=\textwidth]{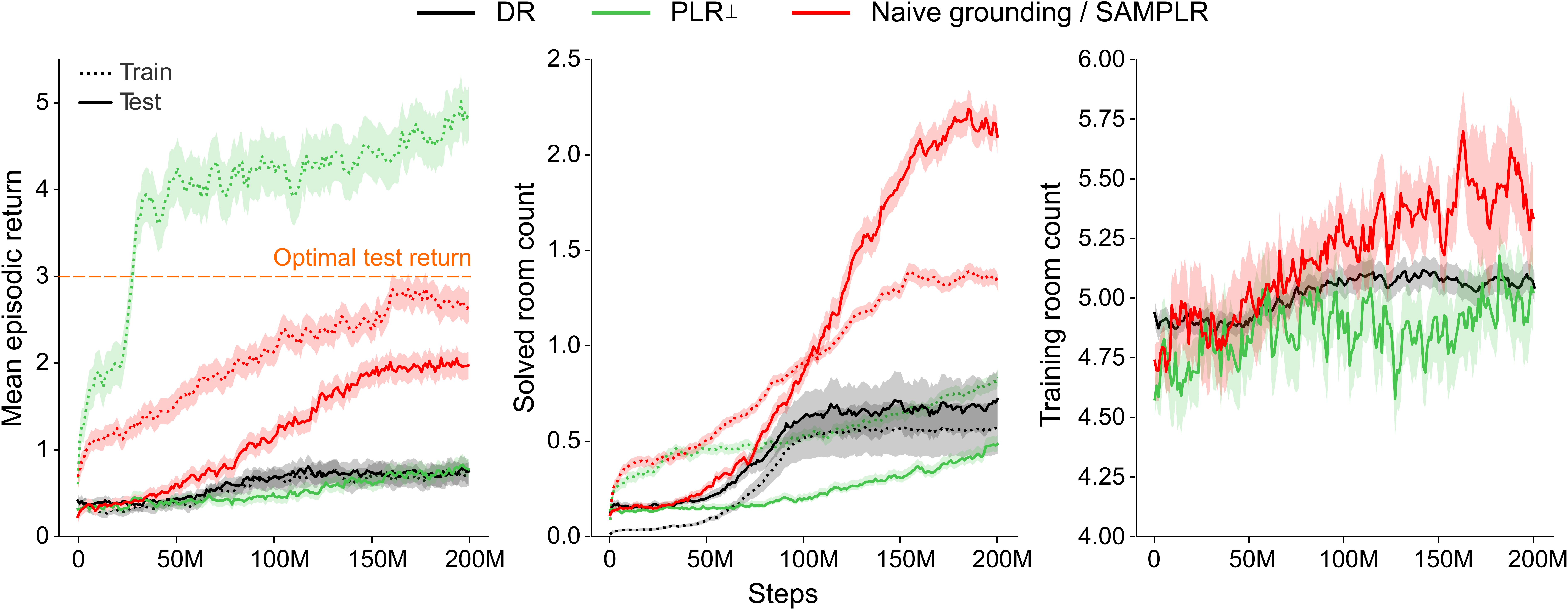}
    \caption{\small{Mean and standard error (over 10 runs) of episodic returns (left); room count of solved levels (middle), during training (dotted lines) and test on the ground-truth distribution (solid lines), for $q=0.7$; and the room count of levels presented at training (right).}}
    \label{fig:multiroom_n8_main_results}
\end{figure}

We aim to demonstrate the phenomenon of CICS in Stochastic Fruit Choice, a binary choice task, where the aleatoric parameter determines the correct choice. This task requires the agent to traverse up to eight rooms, and in the final room, decide to eat either the apple or banana. The correct choice $\theta'$ is fixed for each level, but hidden from the agent. Optimal decision-making depends on the ground-truth distribution over the correct fruit, $\PGround(\Theta')$. This task benefits from a curriculum over the number of rooms, but a curriculum that selectively samples over both room layout and correct fruit choice can lead to suboptimal policies. Figure \ref{fig:multiroom_n8_main_results} shows example levels from this environment.

This domain presents a hard exploration challenge for RL agents, requiring robust navigation across multiple rooms. Further, this environment is built on top of MiniHack \citep{samvelyan2021minihack}, enabling integration of select game dynamics from the NetHack Learning Environment \citep{nle}, which the agent must master to succeed: To go from one room to the next, the agent needs to learn to kick the locked door until it opens. Upon reaching the final room, the agent must then apply the eat action on the correct fruit. 

Let $\pi_A$ be the policy that always chooses the apple, and $\pi_B$, the banana. If the probability that the goal is the apple is $\PGround(A) = q$, then the expected return is $R_Aq$ under $\pi_A$ and $R_B(1-q)$ under $\pi_B$. The optimal policy is $\pi_A$ when $q > R_B/(R_A + R_B)$, and $\pi_B$ otherwise. Domain randomization (DR), which directly samples each level $\theta \sim \PGround(\theta)$, optimizes for the correct ground-truth $\PGround(\Theta')$, but will predictably struggle to solve the exploration challenge. \plrabbrev{} may induce curricula easing the exploration problem, but can be expected make the correct fruit choice oscillate throughout training to maximize regret, leading to problematic CICS. 

\begin{figure*}[t!]
    \centering{\includegraphics[width=0.8\textwidth]{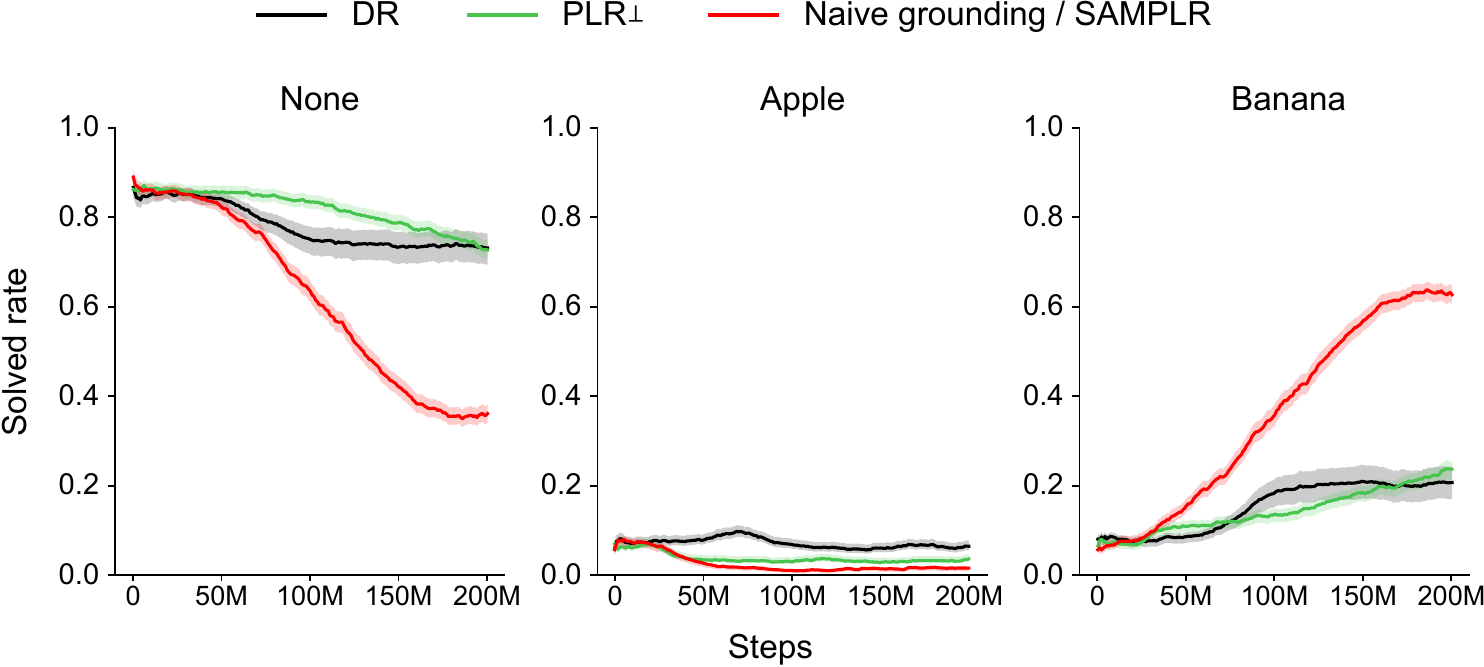}}
    \caption{\small{Left: Proportion of training episodes for $q=0.7$ in which the agent fails to eat any fruit; eats the apple; or eats the banana. Right: Number of rooms in levels during training. Plots show mean and standard error of 10 runs.}}
    \label{fig:fruit_choice_choices}
\end{figure*}

\begin{figure*}[h!]
    \centering{\includegraphics[width=0.9\textwidth]{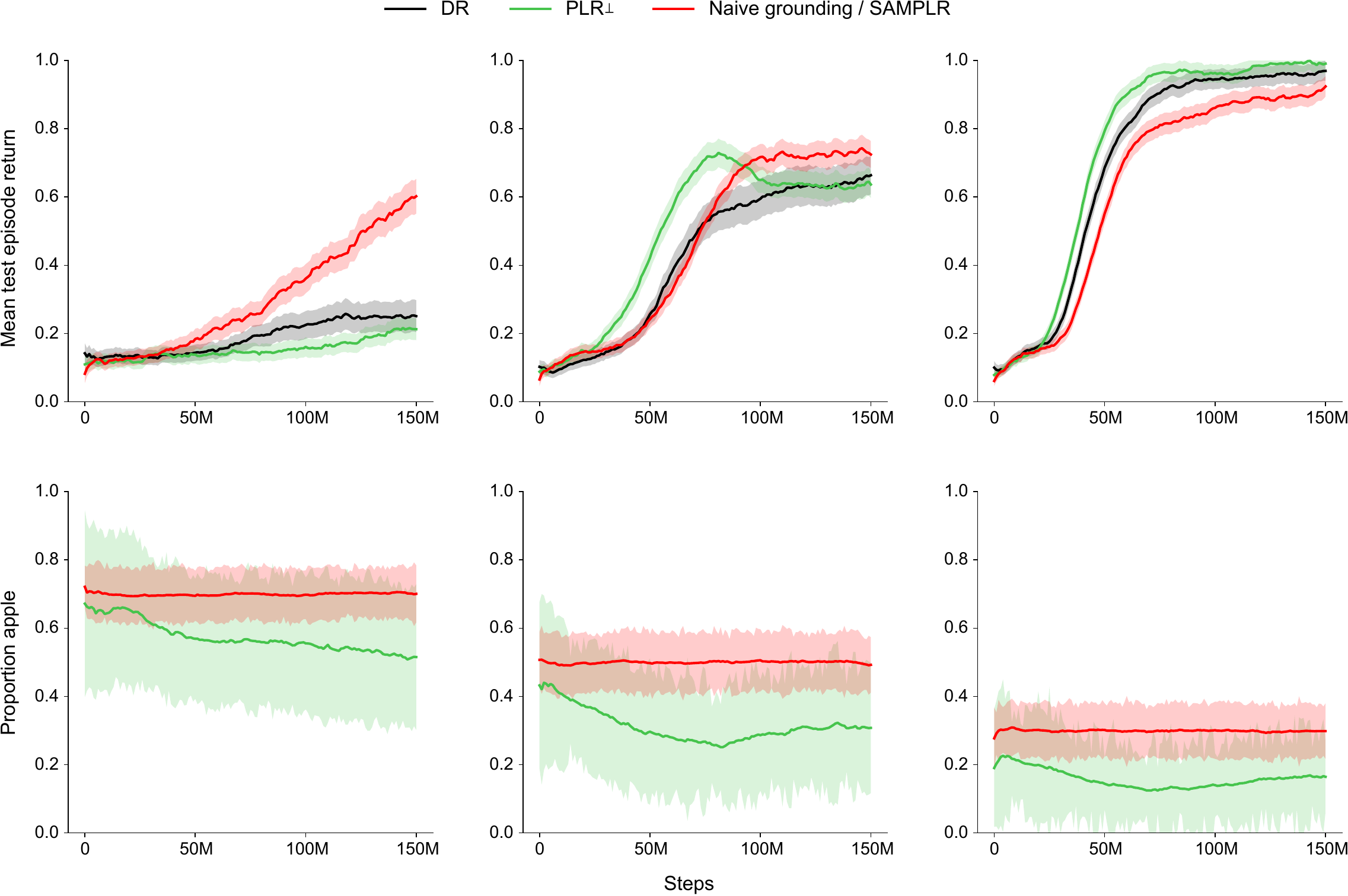}}
    \caption{\small{Top: Mean and standard error of episodic test returns as the probability $q$ of the apple being the correct choice takes on the values 0.7, 0.5, and 0.3. Bottom: The proportion of training levels chosen by each method where apple is the correct choice. The mean and standard deviation are shown.}}
    \label{fig:fruit_choice_vary_q}
\end{figure*}

We set $R_A = 3$, $R_B = 10$, and $q=0.7$, making the policy that always chooses banana, $\pi_B$, optimal with an expected return of $3.0$. We compare the train and test performance of agents trained with DR, \plrabbrev{}, and \plrabbrev{} with naive grounding over 200M training steps in Figure \ref{fig:multiroom_n8_main_results}. In this domain, SAMPLR reduces to naive grounding, as $\theta'$ only effects the reward of a terminal transition, making fictitious transitions equivalent to real transitions for all intermediate time steps. We see that DR struggles to learn an effective policy, plateauing at a mean return around $1.0$, while \plrabbrev{} performs the worst. Figure~\ref{fig:fruit_choice_vary_q} shows that the \plrabbrev{} curriculum exhibits much higher variance in $q$, rapidly switching the optimal choice of fruit to satisfy its regret-maximizing incentive, making learning more difficult. In contrast, \plrabbrev{} with naive grounding constrains $q=0.7$, while still exploiting a curriculum over an increasing number of rooms, as visible in Figure \ref{fig:fruit_choice_vary_q}. This grounded curriculum results in a policy that solves more complex room layouts at test time. Figure~\ref{fig:fruit_choice_choices} shows how the SAMPLR agent's choices converge to $\pi_B$, while both DR and \plrabbrev{} fail to learn to consistently eat the banana even after 200M steps of training.

Moreover, Figure~\ref{fig:fruit_choice_vary_q} shows how the size of SAMPLR's improvement varies under alternative choices of $q$ in $\{0.5, 0.3\}$. When $q = 10/13$, the expected returns for the policy always choosing apple ($\pi_B$) equals that for the policy always choosing banana ($\pi_B$). The top row of Figure \ref{fig:fruit_choice_vary_q} shows that the negative impact of CICS on \plrabbrev{} and thus the benefits of SAMPLR diminish the farther $q$ is from this equilibrium value. Intuitively, for $q$ closer to the equilibrium value, smaller covariate shifts suffice to flip the policy, making it easier for \plrabbrev{} to rapidly oscillate the optimal policy during training. We see in the bottom row of \ref{fig:fruit_choice_vary_q} that \plrabbrev{} indeed produces large adversarial oscillations in $q$. This makes it difficult for the agent to settle on the optimal policy with respect to any ground-truth distribution. In contrast, SAMPLR grounds PLR’s otherwise wild shifts in $q$ with respect to its ground-truth value, allowing the agent to learn a well-grounded policy.

\subsection{Zero-Shot Driving Icy Formula 1 Tracks}

\begin{figure}
    \centering
    \centering
    \begin{subfigure}{0.24\textwidth}               
    \includegraphics[width=\textwidth,valign=c]{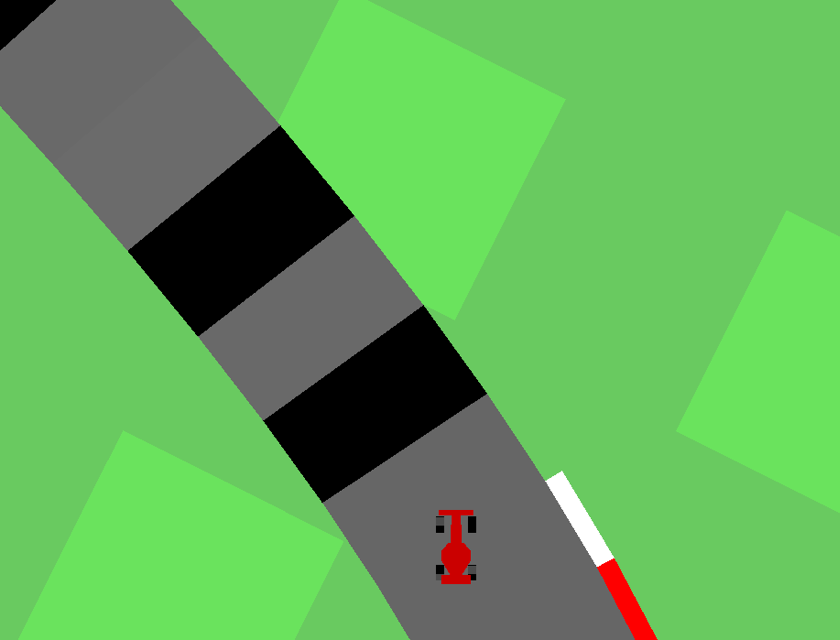}
    \end{subfigure}
    \centering
    \begin{subfigure}{0.24\textwidth}
    \includegraphics[width=\textwidth,valign=c]{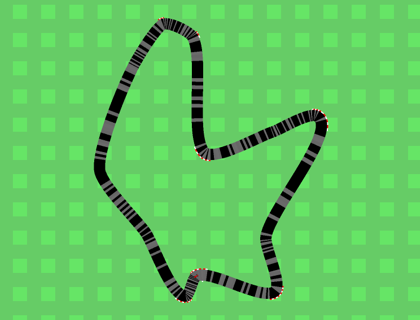}
    \end{subfigure}
    \centering
    \begin{subfigure}{0.24\textwidth}
        \includegraphics[width=\textwidth,valign=c]{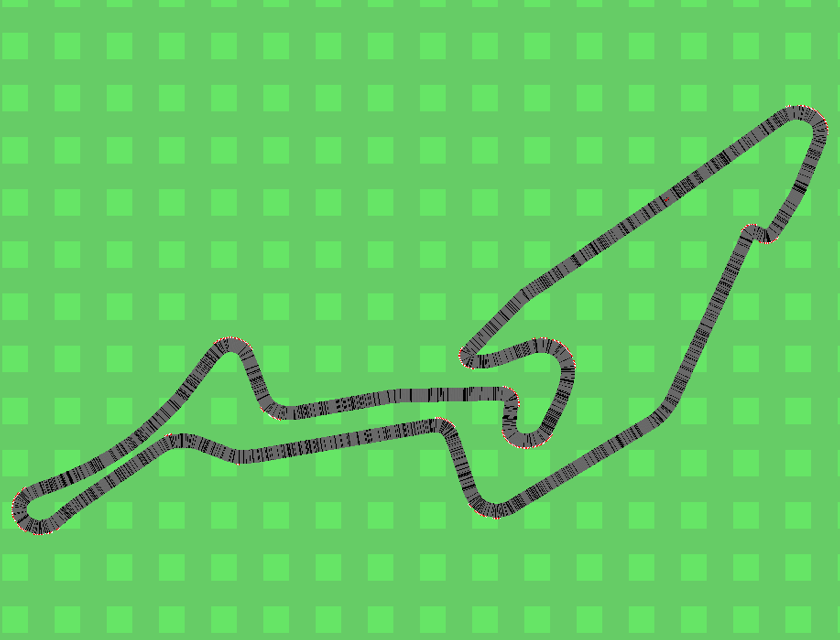}
    \end{subfigure}    
    \centering
    \begin{subfigure}{0.24\textwidth}
        \includegraphics[width=\textwidth,valign=c]{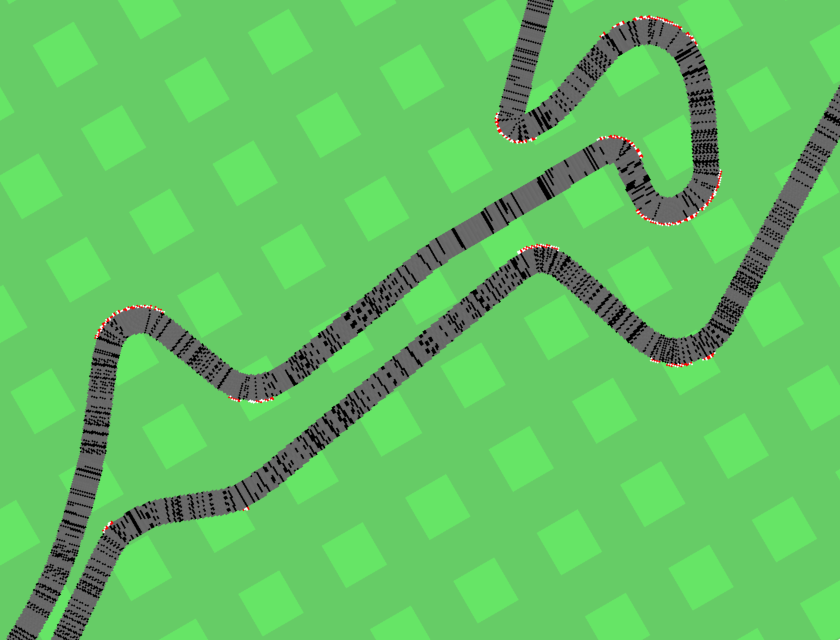}
    \end{subfigure}
    
    \caption{\small{Charts show mean and standard error (over 10 runs) of fraction of visited tiles with ice during training (left) and zero-shot performance on the full Formula 1 benchmark as a function of ice rate (right). Top row screenshots show the agent approaching black ice ($q=0.4$) and an example training track ($q=0.6$). Bottom row shows a Formula 1 track ($q=0.2$) at two zoom scales.
    }}
    \label{figure:blackice_examples}
\end{figure}

\begin{figure}[tbh]
\centering
    \centering
    \includegraphics[width=0.8\linewidth]{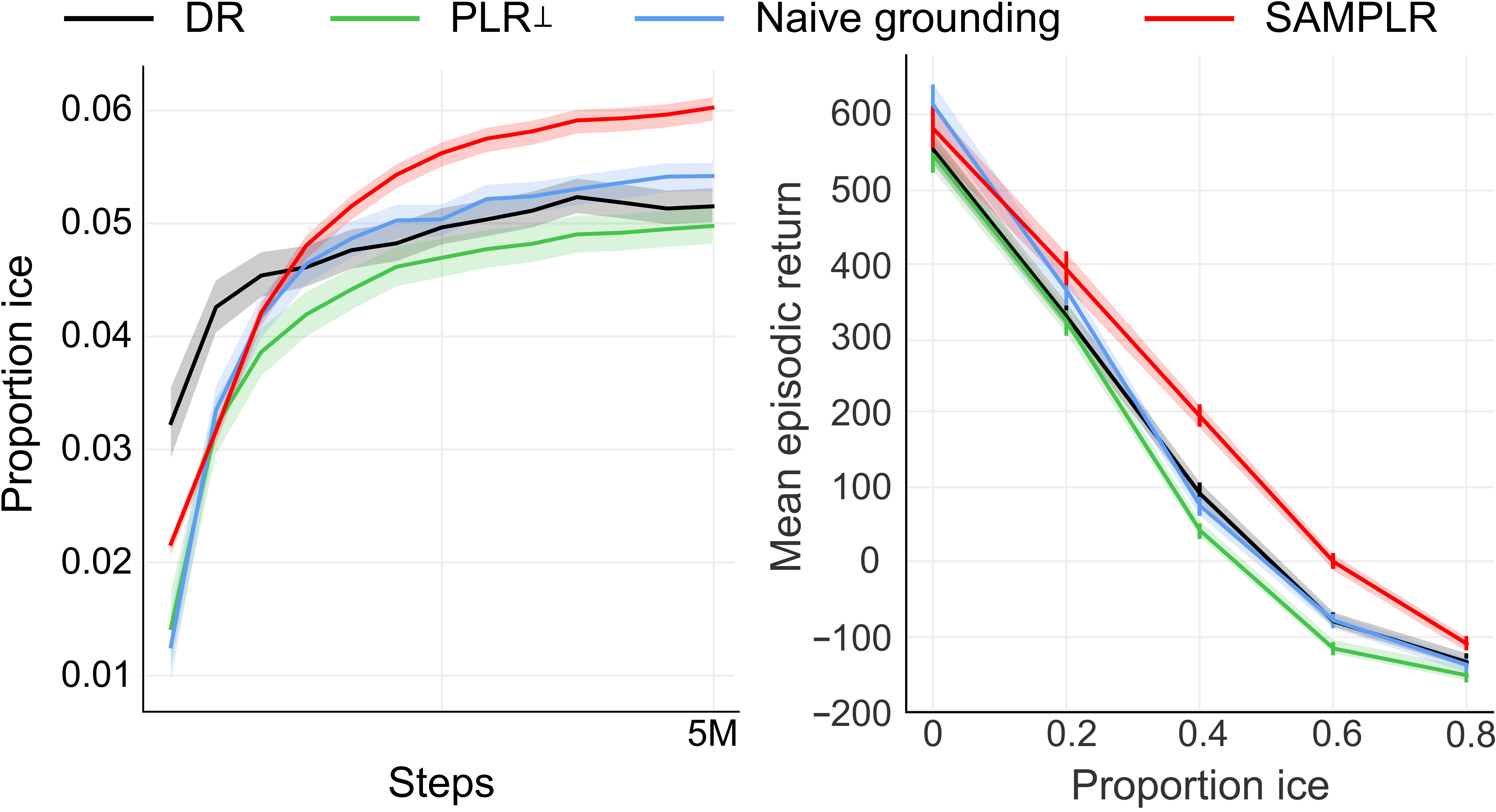}
    \caption{\small{Left: Fraction of visited tiles with ice during training. Right: Zero-shot performance on the full Formula 1 benchmark as a function of ice rate. The mean and standard error are shown.
    }}
    \label{figure:blackice_main_results}
\end{figure}

We now turn to a domain where the aleatoric parameters influence the distribution of $\tau_t$ at each $t$, thereby creating opportunities for a curriculum to actively sample specific $\theta'$ to promote learning on biased distributions of $\tau_t$. We base this domain on the introductory example of driving over black ice, by modifying the CarRacingBezier environment from Chapter~\ref{chapter:dcd}. In this version, each track tile has black ice with probability $q$, in which case its friction coefficient is 0, making acceleration and braking impossible. This task is especially difficult, since the agent cannot see black ice in its pixel observations. Figure \ref{figure:blackice_main_results} shows example tracks with ice rendered for illustration purposes. The episodic returns scale linearly with how much of the track is driven and how quickly this is accomplished. As success requires learning to navigate the challenging dynamics over ice patches, a curriculum targeting more difficult ice configurations should lead to policies more robust to black ice. Here, the ground-truth distribution $\PGround(\Theta')$ models the realistic assumption that most days see little to no ice. We therefore model the probability of ice per tile as $q \sim \text{Beta}(\alpha, \beta)$, where $\alpha=1$, $\beta=15$. 

\begin{table}
\caption{\small{Icy F1 returns, mean $\pm$ standard error over 10 runs.}}
\centering
\hspace*{-.75\columnsep}
\scalebox{0.75}{
\begin{tabular}{p{2.5cm}rrrr}
\toprule
Condition & DR & PLR & Naive  & SAMPLR\\
\midrule
& & & & \\[-0.3cm]
\emph{Ground truth} & & & & \\[0.05cm]
\mbox{$q \sim \text{Beta}(1,15)$}&$581\pm23$&$543\pm21$&$\mathbf{618\pm6}$&$\mathbf{616\pm6}$\\
\midrule
& & & & \\[-0.3cm]
\emph{Zero-shot} & & & & \\[0.05cm]
$q = 0.2$&$\mathbf{332\pm63}$&$\mathbf{323\pm60}$&$363\pm15$&$\mathbf{393\pm13}$\\
$q = 0.4$&$94.7\pm41$&$43\pm38$&$75\pm39$&$\mathbf{195\pm11}$\\
$q= 0.6$&$-76.3\pm24$&$-115\pm12$&$-79\pm25$&$\mathbf{-1\pm17}$\\
$q = 0.8$&$-131.1\pm11$&$-151\pm6.0$&$-139\pm9$&$\mathbf{-111\pm7}$\\
\bottomrule
        \end{tabular}}
    \label{table:carracing_results}
\end{table}

We test the hypothesis that SAMPLR's regret-maximizing curriculum results in policies that preserve optimal performance on the ground-truth distribution $\PGround(\Theta')$, while being more robust to tail cases compared to DR and \plrabbrev{} with naive grounding. We expect standard \plrabbrev{} to underperform all methods due to CICS, leading to policies that are either too pessimistic or too optimistic with respect to the amount of ice. These baselines provide the controls needed to distinguish performance changes due to the two grounding approaches and those due to the underlying curriculum learning method.

We train agents with each method for 5M and test zero-shot generalization performance on the Formula 1 (F1) tracks from the CarRacingF1 benchmark, extended to allow each track segment to have black ice, based on i.i.d. Bernoulli samples with mean $q$ in $\{0.0, 0.2, 0.4, 0.6, 0.8\}$. These test tracks are significantly longer and more complex than those seen at training, as well as having a higher rate of black ice.

To implement SAMPLR's belief model, we use a second simulator as a perfect model of the environment. At each time step, this second simulator, which we refer to as the \emph{fictitious simulator}, resets to the exact physics state of the primary simulator, and its icy tiles are resampled according to the exact posterior over the aleatoric parameter $q = \theta'$, such that $\theta' \sim \PGround(\theta' | \tau)$, ensuring the future uncertainty is consistent with the past. The agent decides on action $a_t$ based on the current real observation $o_t$, and observes the fictitious return $r'_t$ and next state $s'_{t+1}$ determined by the fictitious simulator after applying $a_t$ in state $s'_t \sim \PGround(s'_t | \tau, \theta')$. This dual simulator arrangement, fully detailed in Appendix~\ref{appendix:exp_samplr}, allows us to measure the impact of training on fictitious transitions independently of the efficacy of a model-based RL approach. Further, as the training environment in RL is most often simulation (e.g. in sim2real), this approach is widely applicable.

SAMPLR outperforms all baselines in zero-shot transfer to higher ice rates on the full F1 benchmark and attains a statistically significant improvement at $p < 0.001$ when transferring to $q=0.4$ and $q=0.6$, and $p < 0.05$ when $q=0.8$. Importantly, SAMPLR outperforms \plrabbrev{} with naive grounding, indicating that SAMPLR exploits specific settings of $\Theta'$ to better robustify the agent against rare icy conditions in the tail of $\PGround(\Theta')$. Indeed, Figure \ref{figure:blackice_main_results} shows that on average, SAMPLR exposes the agent to more ice per track tile driven, while \plrabbrev{} underexposes the agent to ice compared to DR and naive grounding, suggesting that under \plrabbrev{} agents attain higher regret on ice-free tracks---a likely outcome as ice-free tracks are easier to drive and lead to returns, with which regret scales. Unfortunately, this results in \plrabbrev{} being the worst out of all methods on the ground-truth distribution. SAMPLR and naive grounding avoid this issue by explicitly matching transitions to those under $\PGround$ at $\tau$. As reported in Table \ref{table:carracing_results}, SAMPLR matches the baselines in mean performance across all F1 tracks under $\PGround(\Theta')$, indicating that despite actively sampling challenging $\theta'$, it preserves performance under $\PGround(\Theta')$, i.e. the agent does not become overly cautious. 

\section{Connection to Off-Belief Learning}
In cooperative multi-agent reinforcement learning (MARL), self-play promotes the formation of cryptic conventions---arbitrary sequences of actions that allow agents to communicate information about the environment state. These conventions are learned jointly among all agents during training, but are arbitrary and hence, indecipherable to independently-trained agents or humans at test time. Crucially, this leads to policies that fail to perform zero-shot coordination \citep[ZSC,][]{hu2021otherplay}, where independently-trained agents must cooperate successfully without additional learning steps---a setting known as ad-hoc team play. Off-Belief Learning \citep[OBL;][]{hu2021obl} resolves this problem by forcing agents to assume their co-players act according to a fixed, known policy $\pi_0$ until the current time $t$, and optimally afterwards, conditioned on this assumption. 
If $\pi_0$ is playing uniformly random, this removes the possibility of forming arbitrary conventions. 

Formally, let $G$ be a decentralized, partially-observable MDP \citep[Dec-POMDP,][]{bernstein2002complexity}, with state $s$, joint action $a$, observation function $\mathcal{I}^i(s)$ for each player $i$, and transition function $\mathcal{T}(s,a)$. Let the historical trajectory $\tau = (s_1,a_1,...a_{t-1},s_t)$, and the action-observation history (AOH) for agent $i$ be $\tau^i = (\mathcal{I}^i(s_1), a_1, ..., a_{t-1}, \mathcal{I}^i(s_t))$. Further, let $\pi_0$ be an arbitrary policy, such as a uniformly random policy, and $\mathcal{B}_{\pi_0}(\tau|\tau^i) = P(\tau_t|\tau_t^i, \pi_0)$, a belief model predicting the current state, conditioned on the AOH of agent $i$ and the assumption of co-players playing policy $\pi_0$ until the current time $t$, and optimally according to $\pi_1$ from $t$ and beyond. OBL aims to find the policy $\pi_1$ with the optimal, \emph{counter-factual value function},

\begin{equation}
V^{\pi_0 \rightarrow \pi_1}(\tau^i) = \mathbb{E}_{\tau \sim \mathcal{B}_{\pi_0}(\tau^i)}\left[ V^{\pi_1}(\tau) \right].
\end{equation}

\noindent Thus, the agent conditions its policy on the realized AOH $\tau^i$, while optimizing its policy for transition dynamics based on samples from $\mathcal{B}_{\pi_0}$, which are consistent with the assumption that co-players play according to $\pi_0$ until time $t$. Therefore, if $\pi_0$ is a uniformly random policy, $\pi_1$ can no longer benefit from conditioning on the action sequences of its co-players, thereby preventing the formation of cryptic conventions that harm ZSC.

Similarly, in single-agent curriculum learning, we can view the UED teacher as a co-player that performs a series of environment design decisions that defines the environment configuration $\theta$ at the start of the episode, and subsequently performs no-ops for the remainder of the episode. As discussed in Section~\ref{section:curriculum_bias}, curriculum-induced covariate shifts (CICS) can cause the final policy to be suboptimal with respect to the ground-truth distribution $\PGround$ when the teacher produces a curriculum resulting in the training distribution of aleatoric parameters $\PCur(\Theta')$ deviating from the ground-truth distribution $\PGround(\Theta')$. We then see that the fictitious transitions used by SAMPLR are equivalent to those used by OBL, where the  belief model $\mathcal{B}$ assumes the teacher makes its design choices such that the resulting distribution of aleatoric parameters $\Theta'$ matches the ground-truth $\PGround(\Theta')$. Thus, SAMPLR can be viewed as an adaptation of OBL to the single-agent curriculum learning setting, whereby the UED teacher, which designs the environment configuration, is viewed as the co-player. This correspondence is no accident. Indeed, it is but another instantiation of the fundamental fact that single-agent curriculum learning is inherently a multi-agent problem, and thus problems afflicting multi-agent learning also surface in this setting; moreover, methods addressing such issues in one setting can then be adapted to the other.

\section{Related Work}
\label{section:related_works}
The mismatch between training and testing distributions of input features is referred to as \emph{covariate shift}, and has long served as a fundamental problem for the machine learning community. Covariate shifts have been extensively studied in supervised learning \citep{vapnik1971uniform, huang2006correcting, bickel2009discriminative, arjovsky2019irm}. In RL, prior works have largely focused on covariate shifts due to training on off-policy data \citep{sutton2016emphatic, rowland20a, espeholt18a, hallak17a, gelada2019offpolicy, thomasa16} including the important case of learning from demonstrations \citep{pomerleau1988alvinn, ross2010efficient}. Recent work also aimed to learn invariant representations robust to covariate shifts \citep{zhang2019causalreps,zhang2021invariant}. More generally, \csabbrev{} is a form of sample-selection bias \citep{heckman1979sample}. Previous methods like OFFER \citep{ciosek2017offer} considered correcting biased transitions via importance sampling \citep{Sutton1998} when optimizing for expected return on a single environment setting, rather than robust policies over all environments settings.
We believe our work provides the first general formalization and solution strategy addressing curriculum-induced covariate shifts (CICS) for RL.

The importance of addressing CICS is highlighted by recent results showing curricula to be essential for training RL agents across many of the most challenging domains, including combinatorial gridworlds \citep{rtfm}, Go \citep{alphago}, StarCraft II \citep{alphastar}, and achieving comprehensive task mastery in open-ended environments~\citep{xland}. While this work focuses on \plrabbrev{}, the approach described in this chapter can be applied to nearly all autocurricula methods, including minimax adversarial curricula \citep{pinto2017robust, poet, enhanced_poet}, curricula based on changes in learning progress \citep{tscl,portelas2020alpgmm}, and other UED methods.
Curriculum methods have also been studied in goal-conditioned RL \citep{goalgan, amigo, sukhbaatar2018intrinsic, openai2021asymmetric}, though \csabbrev{} does not occur here as goals are observed by the agent. Lastly, domain randomization~\citep[DR,][]{cad2rl, peng2017dr} can be seen as a degenerate form of UED, and curriculum-based extensions of DR have also been studied \citep{evolutionary_dr, tobin_dr}.

Prior work has also investigated methods for learning Bayes optimal policies under uncertainty about the task \citep{zintgraf2020varibad, osband2013more}, based on the framework of Bayes-adaptive MDPs~\citep[BAMDPs,][]{bellman1956problem, duff2002optimal}. In this setting, the agent can adapt to an unknown MDP over several episodes by acting to reduce its uncertainty about the identity of the MDP. In contrast, SAMPLR learns a robustly Bayes-optimal policy for zero-shot transfer. Further unlike these works, our setting assumes the distribution of some aleatoric parameters is biased during training, which would bias the \emph{a posteriori} uncertainty estimates with respect to the ground-truth distribution when optimizing for the BAMDP objective. Instead, SAMPLR proposes a means to correct for this bias assuming knowledge of the true environment parameters, to which we can often safely assume access in curriculum learning.

\section{Conclusion}
\label{section:conclusion}

This work characterized how curriculum-induced covariate shifts (CICS) over aleatoric environment parameters $\Theta'$ can lead to suboptimal policies under the ground-truth distribution over these parameters, $\PGround(\Theta')$. We introduced a general strategy for correcting CICS, by training the agent on fictitious rewards and next states whose distribution is guaranteed to match what would be experienced under $\PGround(\Theta')$. Our method SAMPLR augments \plrabbrev{} with this correction. By training on fictitious transitions, SAMPLR actively samples specific values of $\theta'$ that induce trajectories with greater learning potential, while still grounding the training data to $\PGround(\Theta')$. Crucially, our experiments in challenging environments with aleatoric uncertainty showed that SAMPLR produces robust policies outperforming those trained with competing baselines that do not correct for CICS. 

A core assumption made by SAMPLR and all other UED methods is the ability to reset the environment to arbitrary configurations of some set of free parameters. While such resets can be difficult or impossible to perform in real world environments, in practice, this assumption is nearly always satisfied, as RL training largely occurs under a sim2real paradigm due to the additional costs of training in the wild. Most RL simulators can either be directly reset to specific environment configurations or be straightforwardly made to do so. SAMPLR thus provides a means to more fully exploit the affordances of a simulator to produce more robust policies: Policies trained with SAMPLR retain optimality when transferred to the ground-truth distribution of aleatoric parameters in the real environment---a crucial property not satisfied by prior UED methods. Importantly, the approach based on fictitious transitions used by SAMPLR can, in principle, be generally applied to prior UED methods to provide them with this desirable property.

\chapter{Afterword}
\label{chapter:conclusions}

In this thesis, we developed a series of scalable autocurriculum methods for RL, with each contribution addressing a critical weakness of prior methods.
In Chapter~\ref{chapter:plr}, we introduced Prioritized Level Replay~(PLR) and demonstrated that selective replay of previously challenging environments leads to autocurricula that significantly improve sample efficiency and test performance in potentially infinite task spaces. Then, in Chapter~\ref{chapter:dcd}, we matured the formulation of PLR under the lens of game theory and decision theory, resulting in Robust PLR (PLR$^\perp$), which has provable minimax-regret properties at the Nash equilibria of the corresponding curriculum game. This framing reveals that PLR effectively performs UED, whereby environments are designed by curation via the level replay mechanism. Remarkably, this gradient-free design mechanism empirically outperforms previous gradient-based design mechanisms. Moreover, we showed it can be directly combined with these previous UED methods to produce more effective autocurricula for robustifying student agents. However, environment curation amounts to random search over the task space. We addressed this potential inefficiency in Chapter~\ref{chapter:accel} by replacing random search with evolutionary search, resulting in a new method, ACCEL, that produces autocurricula exhibiting degrees of environment complexity comparable to that of population-based evolutionary methods, while requiring only a single student agent---consequently a ``generalist" capable of navigating a wide gamut of environments. Finally, in Chapter~\ref{chapter:samplr}, we asked how autocurricula can go wrong, leading to the first characterization of how the inherent data bias introduced by curriculum learning can lead to learning suboptimal policies in stochastic settings. We then introduced a general strategy to combat this bias, ensuring optimization still targets optimal behaviors on the ground-truth distribution. A concrete application of this approach to PLR$^\perp$ resulted in SAMPLR, which we showed produces robust agents while avoiding this pitfall. In sum, these works provide a versatile toolbox of principled autocurriculum methods that can both scale to complex task spaces and avoid common biases in stochastic settings. The rest of this chapter discusses limitations, recent follow-up work addressing some of these limitations, and promising paths for scaling these techniques to fully open-ended task spaces.

\section{Extensions to Other RL Settings}
While the methods introduced in this thesis may conceptually extend to many learning settings, their study was limited to the standard setting of single-agent, model-free RL. Nevertheless, this basic setting captures many of the central challenges in designing effective autocurricula: evaluating tasks for learning potential, scalably searching for the most informative tasks, avoiding inherent data biases induced by curricula, and defining consistent protocols for evaluating the success of such curricula. Thus, these works can serve as useful templates for autocurriculum methods in more complex settings, such as multi-agent RL~\citep[MARL,][]{littman1994markov,shoham2003multi,foerster2018deep}, model-based RL~\citep[MBRL,][]{sutton1990integrated,sutton1991planning, worldmodels,schrittwieser2020mastering,hessel2021muesli}, and meta-learning for RL~\citep{schmidhuber1987evolutionary,cotter1990fixed,hochreiter2001learning,ellefsen2015neural,duan2016rl,finn2017model,soltoggio2018born,oh2020discovering,beck2023survey}. In the time since the results of this thesis were published, I have contributed to follow-up works extending these concepts to exactly these settings.

\emph{Multi-Agent Environment Design Strategist for Open-Ended Learning}~\citep[MAESTRO,][]{samvelyan2023maestro} extends PLR$^\perp$ to the MARL setting of two-player zero-sum games. Such MARL settings introduce the additional challenge of co-player non-stationarity---that is the learning potential (e.g. regret incurred by the student) of each environment instance depends on both the environment configuration and the specific co-player policies. Merely performing level replay over the environment configuration, as done in the single-agent RL setting, is insufficient to recreate previously informative settings---the same co-player policies that induced high-regret must also be made present again. MAESTRO addresses this issue by maintaining a set of historical co-player policies and for each of these policies, maintaining a separate level-replay buffer based on evaluating each level's learning potential when playing against that co-player policy. These co-player policies are generated naturally during self-play training~\citep{silver2016mastering}, in which the agent plays against itself to gradually improve. MAESTRO thus approximates a regret-maximizing autocurriculum over pairs of co-player and environment instances. Empirically, agents trained via MAESTRO outcompete those trained by prioritizing only the environment configuration (DR) or only the co-player (Prioritized Fictitious Self-Play~\citep{alphastar}).

\emph{Weighted Acquisition of Knowledge Across Environments for Robustness}~\citep[WAKER,][]{rigter2023rewardfree} extends PLR to generate replay-based autocurricula for learning robust world models for reward-free MBRL~\citep{sekar2020planning}. In this problem setting, the RL agent performs self-supervised learning (i.e. without task-specific rewards) within a \emph{world model}~\citep{craik1943nature,bryson1969applied,worldmodels,hafner2020dreamer}---usually consisting of DNNs that predict the state transitions and rewards of some environment of interest---with the goal of learning useful representations that can be transferred to downstream tasks. The world model is typically trained concurrently with the agent via intermittent rollouts in the target environment. Transfer is then accomplished by fine-tuning the agent inside the same world model outfitted with a task-specific reward function. WAKER derives from a key theoretical result that says for a given task space, the maximum regret incurred by an agent trained for a specific downstream task in such a reward-free world model is upper bounded by a constant multiple of the maximum world model state-transition prediction error over that same task space. Thus, assuming a resettable simulator of the task space, we can use a PLR-style curriculum that selects environments that maximize the world model's prediction errors. The resulting world model can be expected to be more robust in terms of achieving lower prediction errors across the task space, thereby also minimizing the maximum regret on downstream tasks for agents trained within it. In other words, agents can be expected to implement approximately minimax-regret policies on downstream tasks when trained within such robust world models. Empirically, agents fine-tuned via task-specific reward functions in WAKER world models indeed show improved robustness across task instances.

\emph{General RL Optimizers Obtained via Environment Design}~\citep[GROOVE,][]{jackson2023discovering} extends PLR to the problem setting of policy meta-optimization, where we seek to learn part of the RL algorithm itself~\citep{duan2016rl,finn2017model,oh2020discovering,beck2023survey}. GROOVE uses a PLR-based curriculum over procedurally-generated environments to produce a curriculum for \emph{Learned Policy Gradient}~\citep[LPG,][]{oh2020discovering}. In the outer-loop, LPG trains a neural module outputting per-step training targets for the inner-loop agent's policy logits and critic. This outer-loop module is updated via policy gradient to maximize the return achieved after a fixed number of updates performed by the inner-loop agent, trained using the targets output by the outer-loop module. GROOVE uses PLR with a scoring function equal to the \emph{algorithmic regret}---defined as the regret of LPG compared to a target algorithm (in this case, A2C). After training GROOVE over a task space consisting of procedurally-generated mazes, we find it then trains agents that significantly outperform those produced by LPG in terms of OOD transfer to the Atari Learning Environment. Related to this work, \citet{team2023human} recently made use of PLR$^\perp$ to similarly improve the robustness of a policy meta-optimization algorithm in a large, open-ended pixel-based task domain. In this case, the meta-learning algorithm is based on RL$^2$~\citep{duan2016rl}, which, in this case, trains a transformer-based~\citep{vaswani2017attention} policy to maximally improve its performance given multiple \emph{trials} or attempts in the same task instance. Here, the learned network weights of the Transformer represent the meta-learned algorithm, which then update its activations (the ``fast weights") across multiple trials to implement task adaptation without gradient updates.
 
PLR's rapid uptake by the greater community has been exciting to observe, but these applications are likely only the tip of the iceberg. Many other domains stand to benefit from UED-based methods like PLR. Particularly exciting future applications include improving test-time ``ad-hoc" team-play with held-out co-players in cooperative multi-agent settings~\citep{stone2010ad,hu2020other,oliehoek2016concise}, improving the robustness of large language models (LLMs) fine-tuned via RL from human or model feedback~\citep{ouyang2022training,bai2022constitutional}, and environment design within a rich ``multi-task" world model trained on a wide range of environments~\citep{micheli2022transformers,lee2022multi,chen2022transdreamer}.

\section{Generalized Exploration}
\label{sec:generalized_exploration}

Given an appropriately expressive task space, autocurricula can in principle guide any learning system---including those that perform supervised learning---across a potentially unbounded number of tasks, resulting in models with increasingly-general capabilities. Yet, the task spaces studied in this thesis are limited, focusing exclusively on standard RL problem settings with limited potential for inducing endlessly novel behaviors. Foreseeably, in any of the environments featured in this thesis, the agent will stop learning once it learns to master the ultimately limited assortment of challenges offered, e.g. navigating different local features of mazes or terrain with a small selection of obstacles---thereby thwarting any hope for kickstarting an open-ended learning process. A crucial missing piece is a universal task representation, which can serve as the basis for an open-ended generator of training tasks.\footnote{Importantly, tasks in this space should include a \emph{context} variable that serves to distinguish any two tasks that have incompatible solution behaviors.} Autocurricula over this universal task space would then amount to a form of \emph{generalized exploration}: Just as standard RL exploration methods guide the agent to select parts of the state space of a particular task in order to maximize some notion of learning potential, autocurricula conduct such exploration over the task space~\citep{jiang2023general}. As in general, tasks can start from arbitrary states or have as a goal of returning to specific states, such task-space exploration strictly generalizes classic ideas of state-based exploration in RL. 
\begin{figure}[t!]
    \centering
    \begin{subfigure}[b]{0.325\linewidth}
        \includegraphics[width=1\linewidth]{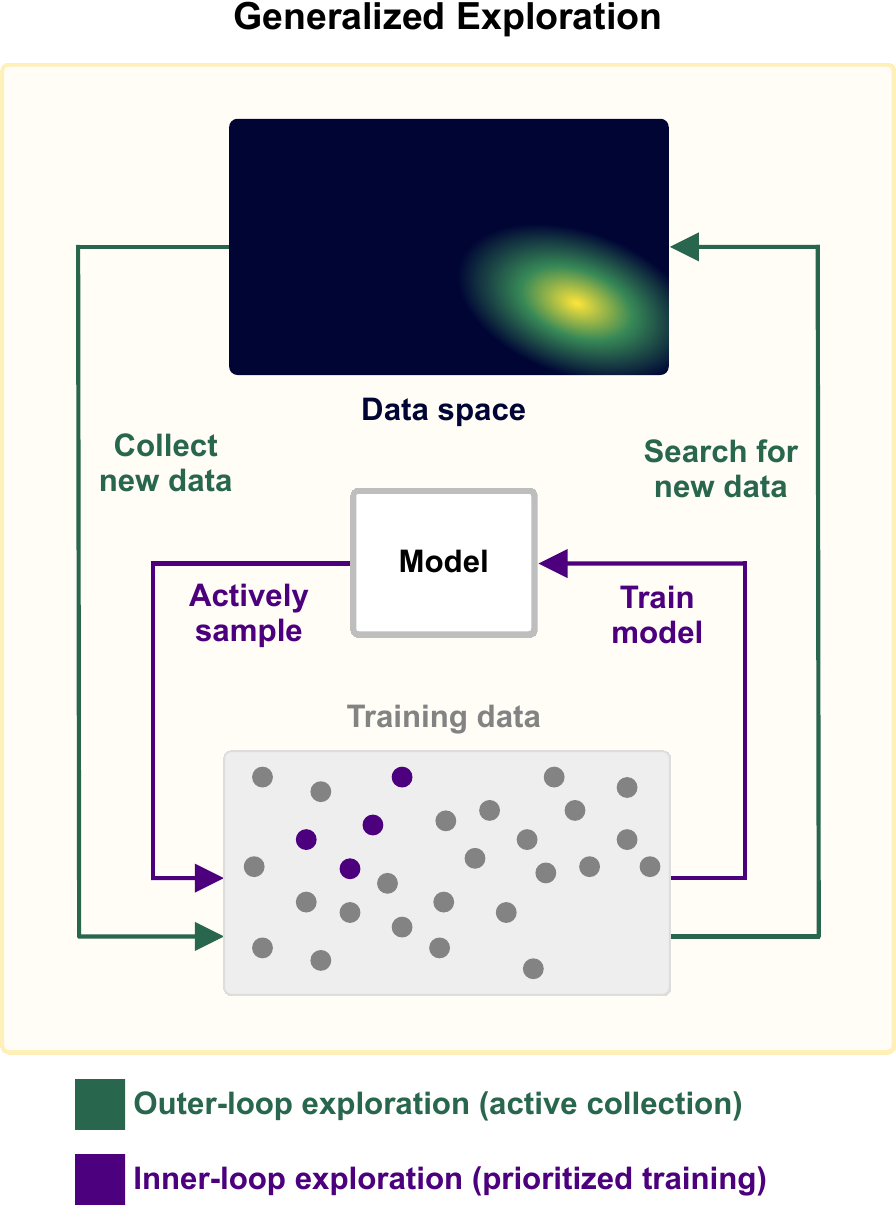}
    \end{subfigure}
    \begin{subfigure}[b]{0.325\linewidth}
        \includegraphics[width=1\linewidth]{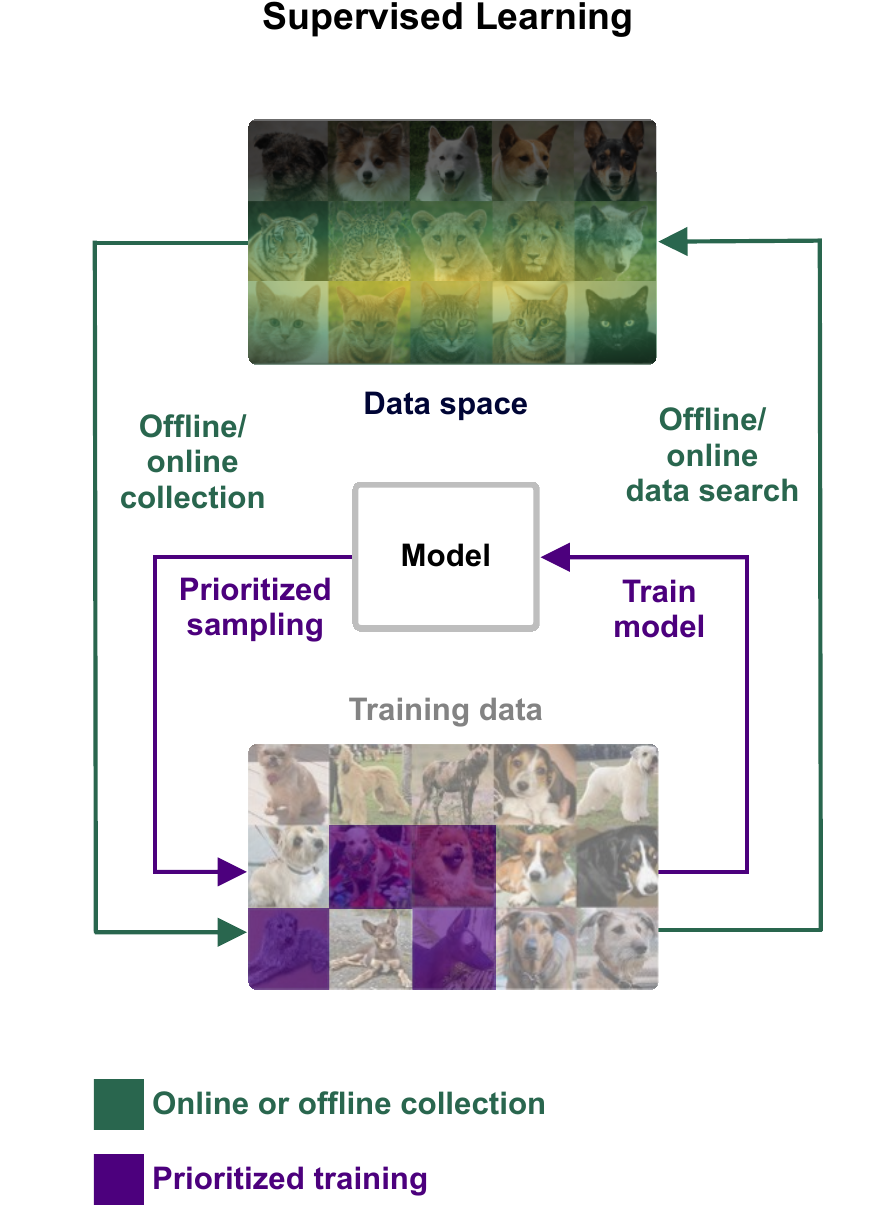}
    \end{subfigure}
    \begin{subfigure}[b]{0.325\linewidth}
        \includegraphics[width=1\linewidth]{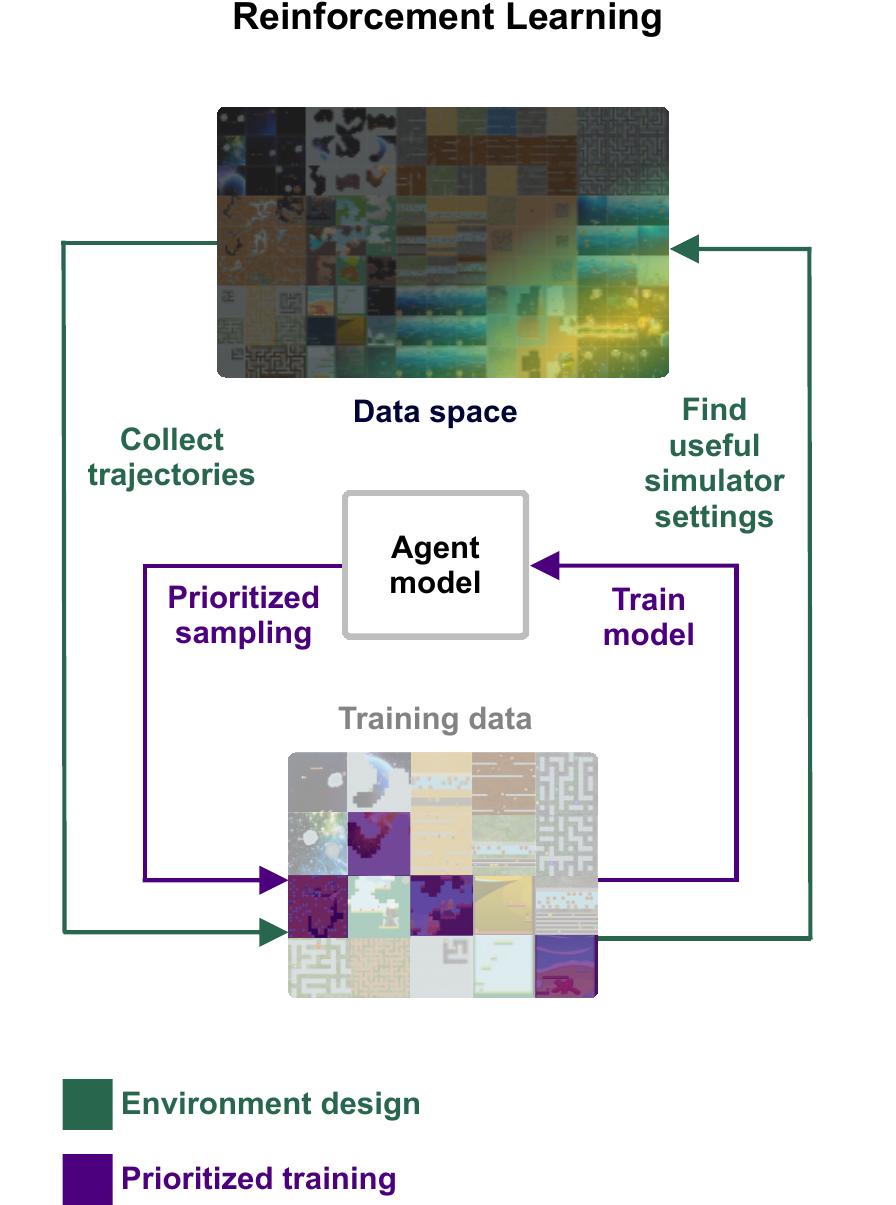}
    \end{subfigure}
    \caption{\small{A general framework for exploration: An outer loop performs active collection of new training data, and an inner loop conducts prioritized training on the current training data. In SL, the outer loop consists of either online or offline data collection. In RL, the outer loop searches for simulator settings that yield useful training data, and the inner loop can perform prioritized sampling, e.g. prioritized experience replay.
    }}
    \label{fig:unified_exploration_overview}
\end{figure}
Moreover, such generalized exploration can occur in a task space that includes SL tasks: After all, any SL problem can be reframed as a single-step MDP, where the observation is the input, the action chooses the correct target output value, and the agent seeks to minimize a distance metric between its chosen action and the target value.\footnote{In this discussion, we treat the term ``supervised learning" as encapsulating the class of methods often called \emph{self-supervised learning}~\citep[SSL,][]{liu2021self,balestriero2023cookbook}. In the case of SSL, the learning process is supervised exactly as in SL, i.e. the prediction targets are directly provided to the learning process per input, with the additional detail that these targets are derived according to some function of the inputs.} This perspective of SL highlights how \emph{active learning}~\citep{settles2009active} plays an analogous role to exploration in SL: By performing prioritized sampling of datapoints that maximize some metric indicative of learning potential, e.g. classifier uncertainty, these methods induce autocurricula over the training data. Generalized exploration then pushes this paradigm further, by exploring not just the datapoints in a single dataset, but across the space of SL tasks. It is important to note that the student model need not use RL to optimize for such SL tasks: The task generator can produce SL tasks by directly generating the labeled dataset, enabling the use of standard SL (or self-supervised) methods. 

By inventing tasks to generate data, generalized exploration blurs the boundary between task and data. We can then think of generalized exploration as exploration over the task data space---the space of all task-relevant input-output pairs. In practice, generalized exploration then proceeds at two levels, as depicted in Figure~\ref{fig:unified_exploration_overview}: Firstly, an outer-loop process continually searches for the most informative training data (i.e. across the task space), and secondly, an inner-loop process performs prioritized sampling of the data already discovered by the outer-loop process. Traditional active learning methods for SL and exploration methods in RL correspond to this inner-loop, focusing on selectively sampling data from a single static dataset or static simulator of a limited range of tasks, while autocurricula methods correspond to the outer loop, searching for informative data across the entire task space. Importantly, the outer-loop process is not necessarily limited to autocurricula. The active collection of new training data can involve humans and other programs in the loop, working in concert to target collection of the most informative data~\citep{ratner2016data,kiela2021dynabench,srivastava2022beyond}. In this way, generalized exploration can make use of both what \citet{schmidhuber1999artificial} calls \emph{artificial curiosity} alongside human insight and domain-specific knowledge. When run for enough time on an open-ended task domain, we can expect such a process to embody an AI generating algorithm~\citep[AI-GA,][]{clune2019ai}---an algorithm that automatically generates an artificial general intelligence, or at least one possessing robust behaviors in a wide diversity of tasks and that continues to improve indefinitely.

Despite the promise of open-ended, generalized exploration, a major question remains: How can we parameterize a universal task space? This problem is especially challenging, considering how real-world data, e.g. natural images and language, are of significantly higher Kolmogorov complexity\footnote{The size of the smallest program that can generate the data.}~\citep{kolmogorov1963tables} than that of the typical RL tasks studied here and in the rest of the literature, which are directly specified by relatively small programs, and therefore trivially bounded in their programmatic complexity. Even sticking to artificial environments still requires a task space parameterizing an open-ended space of programs. Only recently has a viable candidate emerged: Large generative models (themselves giant DNNs) trained on web-scale datasets of text, images, and video~\citep{radford2019language,brown2020language,chen2021evaluating,radford2021learning,ramesh2021zero,ramesh2022hierarchical,rombach2022high,ho2022video,villegas2022phenaki} have been shown capable of simulating the rich dynamics of the real-world phenomena captured in their training data, including complex linguistic and cognitive phenomena such as reasoning~\citep{kojima2022large, wei2022emergent}. Current state-of-the-art video generation models can even output short videos matching an input text prompt~\citep{ho2022video,villegas2022phenaki}. These models serve as promising substrates for scaling the ideas developed in this thesis to the more complex, open-ended domains on which they are trained. Already, researchers are making use of these models as simulators and world models~\citep{todd2023level,sudhakaran2023mariogpt}, yet much remains to be explored. For example, LLMs trained on a dataset of codebases can be prompted to generate many kinds of programs~\citep{chen2021evaluating}. Evolutionary methods that directly use the LLM~\citep{lehman2022evolution,meyerson2023language} for variation and evaluation can then potentially generate both diverse and targeted curricula over programs representing training tasks. Still, it remains unclear whether the current crop of large models supports such use cases due to the high cost of inference\footnote{The forward pass of a vanilla transformer model requires $O(L^3)$ operations, where $L$ is the context length.} and potential biases and limitations in their training data. Ultimately such models are still trained on a finite dataset, and therefore may fail to support fully open-ended learning.

To address this latter issue, a particularly interesting breed of new methods asks LLMs to generate their own training data~\citep{huang2022large,wang2022self,to2023better}. At a high level, these methods query the language model to output example data for a particular kind of task, which may include the task description as well as the corresponding solution. The generated data is then scored according to some heuristic metric for quality, e.g. self-consistency to measure quality or a similarity score with respect to previous generations to maintain diversity. This approach is similar to RL, in which the most successful generations (i.e. sequence of decisions made by the LLM) are reinforced according to the success metric. By modeling tasks and solutions all as text, LLMs enjoy the curious existence of being both the world model and agent acting within it---that is, both the UED teacher and student. Thus, approaches based on adapting ideas from UED to LLMs might be used to simultaneously generate tasks and solutions used to further train the LLM. Moreover, recent works show the LLM may even be queried to produce such self-generated data according to a curriculum, based on the agent's performance on previously generated tasks~\citep{wang2023voyager, zhang2023omni}. Whether such LLM-driven learning processes can unlock the door to fully open-ended learning remains to be seen, as such recursive training can lead to fixed points~\citep{perdomo2020performative,shumailov2023curse} that terminate further evolution.

\section{Open Challenges}

This thesis contributed several advances to the design of UED autocurricula, which may already prove useful in more specialized problem settings. However, many open problems remain on the way to achieving the much grander ambition of scaling these methods to produce an increasingly general agent in a truly open-ended task space. This section concludes the thesis with a brief discussion of several of these open questions.

\medskip
\noindent\emph{Q1. Does the robustness induced by UED persist in sim2real transfer?}
\smallskip

\noindent A notable limitation of UED (and more broadly, curriculum learning) as a method class is that it generally assumes training occurs inside a simulator, over which the training process has some degree of control, e.g. reset based on a particular random seed or more fine-grained control over the environment configuration. Thus far, the robustness gains provided by UED have been demonstrated in simulation only. Much exciting work remains in evaluating and improving UED methods specifically for the sim2real setting, where the transfer domain is no longer simply different settings of the training simulator, but a real-world domain with the potential for dynamics that are not fully modeled by the simulator. 

\medskip
\noindent\emph{Q2. How do we design scalable open-ended data generators?}

\noindent As discussed at length in Section~\ref{sec:generalized_exploration}, a major challenge to scaling UED autocurricula to their full potential is a universal task representation. While any task may be defined as a program implementing a decision process, naively storing and searching all such programs discovered in a non-parameteric fashion, as proposed in prior works \citep{schmidhuber2013powerplay}, is computationally infeasible. Open-ended learning requires a generative process capable of continually inventing new tasks, while storing only the most useful task designs in a compressed representation. Ideally, the number of parameters in such a generator grows much more slowly than the number of tasks represented. While we might imagine the generator as a large generative model, such as a code-generation model or world model, it is unknown whether current model architectures and optimization methods are suitable for this kind of continual invention and compression.

\medskip
\noindent\emph{Q3. How do we determine what data to acquire next?}

\noindent Autocurricula continually seek maximally informative tasks data throughout training, and the curse of dimensionality~\citep{bellman1957dynamic} makes this search more difficult as the task or data generator becomes more complex and open-ended in its possibilities. How to best navigate such vast task spaces serves as a formidable open problem, with two important subproblems: The first being the correct choice of objective for driving the curriculum, and the second, how to efficiently search a high-dimensional space for tasks that maximize this objective. 

While the regret-based UED objectives studied in this thesis show strong empirical performance, there likely remain much room for improvement. Firstly, these objectives can only approximate regret, as a general method for computing the true regret requires knowledge of the optimal policy for each task. Currently, the specific choice of regret-based estimator is largely based on empirical performance, and we lack a principled understanding of their trade-offs, which may also vary by domain. Moreover, minimax regret is but one valid decision rule, and all decision rules require trade-offs~\citep{Milnor1951games}. In particular, minimax regret is sensitive to irrelevant alternatives~\citep{arrow1951social}, which can lead to intransitive pairwise rankings of alternatives~\citep{bell1982regret,bikhchandani2011transitive}. How these defects translate to the RL setting is not understood. Crucially, such regret-based autocurricula assume the agent can indeed learn to solve any high-regret task that is proposed~\citep{dennis2020emergent}. This assumption often does not hold, e.g. when the task is a hard exploration problem. A better UED objective may need to directly consider novelty or diversity to make such sparse reward cases more tractable, as well as to prevent collapse to a limited set of tasks. Lastly, as SAMPLR shows, successful application of autocurricula to a target domain often requires grounding the objective to specific attributes of the target domain~\citep{jiang2021grounding}. At a higher level, the choice of the UED objective requires balancing a desire for fully open-ended exploration of the task space and robustness or safety in a specific target domain---what \citet{ecoffet2020open} calls a ``tension between control and creativity." This trade off between stability and growth appears fundamental to all open-ended systems, with the most innovative phenomena occurring at the edge of chaos~\citep{packard1988adaptation}.

On the problem of search, this thesis considered methods based on random and evolutionary search, as well as those based on RL (e.g. REPAIRED). Other obvious candidates include methods for sequential model-based optimization~\citep{jones1998efficient,hutter2011sequential} and existing methods for quality-diversity search~\citep{pugh2016quality,mouret2015illuminating}. However, in their existing forms, these methods are difficult to scale to higher-dimensional search spaces and struggle with non-stationary objectives, as in the case of adaptive autocurricula. Recent works suggest that pretrained LLMs provide a rich prior that can be adapted to directly propose tasks for training RL agents and other model classes in highly complex domains~\citep{wang2023voyager,zhang2023omni}. This approach presents a truly promising path toward general teacher models capable of rapidly generating informative examples within open-ended task spaces.

\medskip
\noindent\emph{Q4. How should agents interface with open-ended task spaces?}
\smallskip

\noindent Open-ended autocurricula over such a universal task space requires that the agent process inputs from an increasingly diverse observation space and make decisions over an increasingly large action space. The agent thus requires a generic interface between agent and environment, capable of adapting the input and output representations of the agent model to the task at hand. This interface may take the form of tools invented by the agent~\citep{schick2023toolformer}, e.g. real or simulated hardware~\citep{baker2019emergent}, or even a program~\citep{ellis2023dreamcoder,lehman2022evolution,wang2023voyager}. Such tools can be passed on to other agents, which may further evolve the tool for new purposes, leading to new evolutionary dynamics independent of the original inventor. Such tool invention may be critical to the emergence of open-endedness~\citep{lehman2022evolution}. 

\medskip
\noindent\emph{Q5. How do we measure the extent of open-ended learning?}

\noindent There are no commonly-accepted measures for tracking the degree of open-ended learning achieved---that is, some measure of increasing capability. Many proposed measures of open-endedness cannot be adapted for this purpose, as they focus on measuring novelty \citep{bedau1998classification, standish2003open, soros2014identifying}, rather than model capability. In general, such novelty and model capability are unrelated. For example, a process that evolves an agent across an endless range of mazes may score highly in some measures of novelty, but the agent will be limited in capability. Measures based on improved performance, such as the ANNECS metric~\citep{enhanced_poet}, suffer a similar shortcoming: The learning process that fixates on the maze domain may see the agent struggle with new maze variations before solving them, thus propping up such measures without increasing general capabilities. One feasible approach may be to simply track the diversity of tasks based on domain-specific criteria, but such a solution is difficult to scale across domains. An ideal metric for open-ended learning would be domain-agnostic. Such a metric might consider both the agent's behavior in discovered tasks and task novelty based on a general task representation.

\medskip
Resolving these questions may reveal a path to principled, open-ended autocurricula that produce machines that continually self-improve toward increasingly greater degrees of intelligence and capability. In a sense, such an achievement would represent the natural extension of the open-ended evolutionary process that birthed humankind, extending the endless arc of self-organizing intelligence into the full generality and creative potential of the computational realm. The result would be nothing short of a reimagining of the limits of intelligence, and perhaps, of life itself.

\clearpage
\phantomsection
\addcontentsline{toc}{chapter}{Bibliography}
\bibliographystyle{plainnat}
\bibliography{refs}

\appendix
\chapter{Environment Details}

\section{Procgen Benchmark}
\label{appendix:env_procgen}

\begin{figure}[t!]
    \centering
    \includegraphics[width=0.8\linewidth]{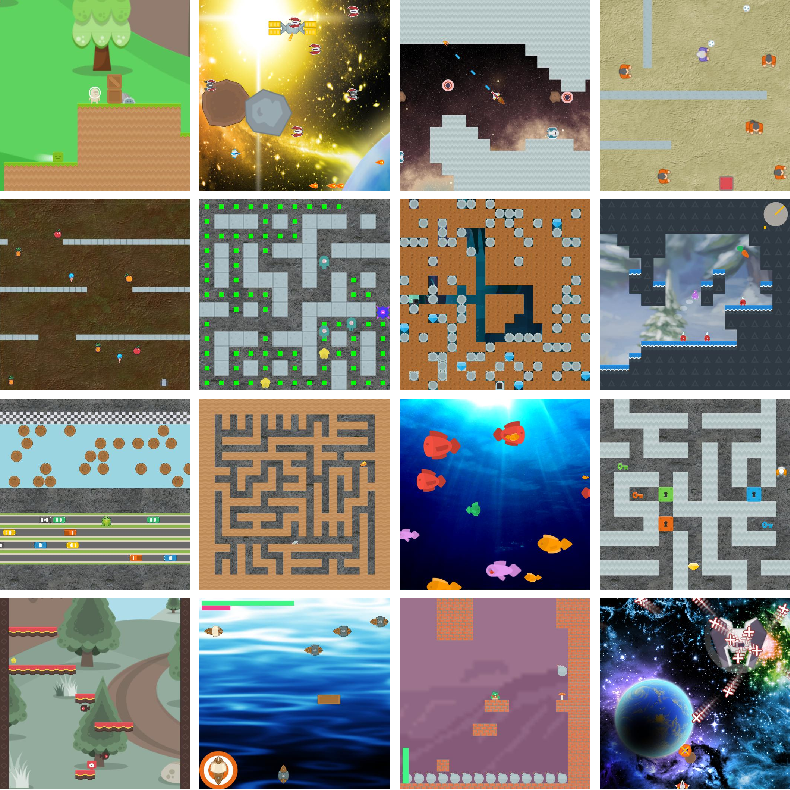}
    \caption{\small{Screenshots of all 16 environments in the Procgen Benchmark.}}
    \label{figure:procgen_games}
\end{figure}

Procgen Benchmark consists of 16 PCG environments of varying styles, exhibiting a diversity of gameplay similar to that of the ALE benchmark. Game levels are determined by a random seed and can vary in navigational layout, visual appearance, and starting positions of entities. All Procgen environments share the same discrete $15$-dimensional action space and produce $64 \times 64 \times 3$ RGB observations. \citet{cobbe2019procgen} provides a comprehensive description of each of the 16 environments, screenshots for each of which are shown in Figure~\ref{figure:procgen_games}. State-of-the-art RL algorithms, like PPO, result in significant generalization gaps between test and train performance in all games, making Procgen a useful benchmark for assessing generalization performance. 

We follow the standard protocol for testing generalization performance on Procgen outlined in \citet{cobbe2019procgen}: We train an agent for each game on a finite number of levels, $N_{\text{train}}$ for a fixed budget $T_{\text{total}}$ of environment steps, and sample test levels from the full distribution of levels. In easy mode, $N_{\text{train}} = 200$ and $T_{\text{total}} = $25M, while in hard mode, $N_{\text{train}} = 500$, and $T_{\text{total}} = $200M. To compare performance across games, normalized test returns are computed as $(R - R_{\text{min}})/(R_{\text{max}} - R_{\text{min}})$, where $R$ is the unnormalized return and each game's minimum return, $R_{\text{min}}$, and maximum return, $R_{\text{max}}$, are provided in \citet{cobbe2019procgen}, which uses this same normalization. 

\newpage

\section{MiniGrid}
\label{appendix:env_minigrid}
The MiniGrid suite \cite{gym_minigrid} features a series of highly structured environments of increasing difficulty. Each environment features a task in a grid world setting, and as in Procgen, environment levels are determined by a seed. Harder levels require the agent to perform longer action sequences over a combinatorially-rich set of game entities, on increasingly larger grids. The clear ordering of difficulty over subsets of MiniGrid environments allows us to track the relative difficulty of levels sampled by PLR over the course of training. The remainder of this section details the specific MiniGrid environments in Chapter~\ref{chapter:plr}. Note we sometimes abbreviate ``ObstructedMazeGamut" as ``OMG."

All MiniGrid environments share a discrete 7-dimensional action space and produce a 3-channel integer state encoding of the $7 \times 7$ grid immediately including and in front of the agent. However, following the training setup in \citet{igl2019generalization}, we modify the environment to produce an $N \times M \times 3$ encoding of the full grid, where $N$ and $M$ vary according to the maximum grid dimensions of each environment. Full observability makes generalization harder, requiring the agent to generalize across different level layouts in their entirety. In all environments, the agent must reach a goal object, upon which the episode terminates and it receives a sparse reward equal to $1.0 - 0.9(T/T_{\text{max}})$, where $T$ is the episode length and $T_{\text{max}}$ is the maximum episode length allowed.

\begin{figure}[htbp]
    \centering
    \includegraphics[width=\linewidth]{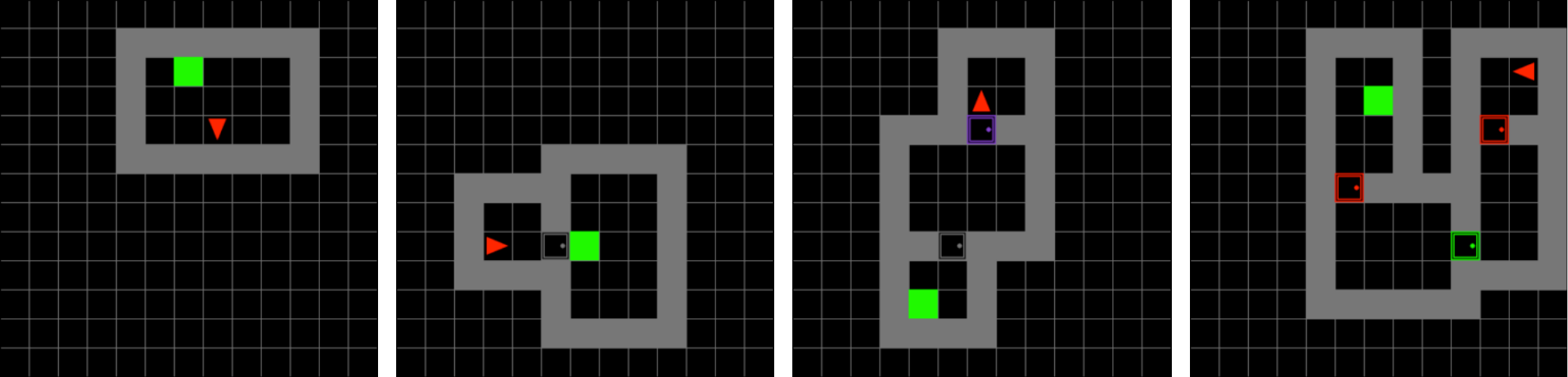}
    \caption{\small{Example levels of each of the four difficulty levels of MultiRoom-N4-Random, in order of increasing difficulty from left to right. The agent (red triangle) must reach the goal (green square).}}
    \label{image:mutliroom-n4-random-examples}
\end{figure}

\medskip
\paragraph{MultiRoom-N4-Random:} This environment requires the agent to navigate through 1, 2, 3, or 4 rooms respectively to reach a goal object, resulting in a natural ordering of levels over four levels of difficulty. The agent always starts at a random position in the furthest room from the goal object, facing a random direction. The goal object is also initialized to a random position within its room. See Figure \ref{image:mutliroom-n4-random-examples} for screenshots of example levels.

\begin{figure}[htbp]
    \centering
    \includegraphics[width=1.0\linewidth]{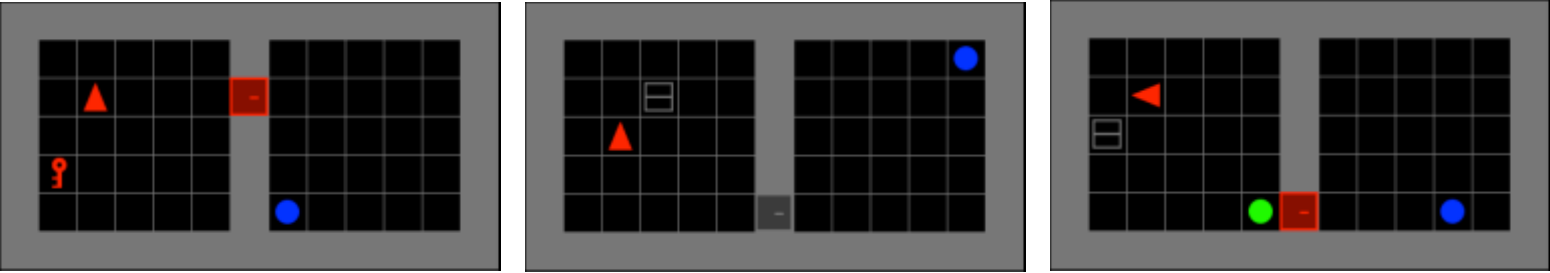}
    \caption{\small{Example levels of each of the three difficulty levels of OMG-Easy, in order of increasing difficulty from left to right. The agent must find the key, which may be hidden under a box, to unlock a door, which may be blocked by an obstacle, to reach the goal object (blue circle).}}
    \label{image:omg-easy-examples}
\end{figure}

\paragraph{ObstructedMazeGamut-Easy:} This environment consists of levels uniformly distributed across the first three difficulty settings of the ObstructedMaze environment, in which the agent must locate and pick up the key in order to unlock the door to pick up a goal object in a second room. The agent and goal object are always initialized in random positions in different rooms separated by the locked door. The second difficulty setting further requires the agent to first uncover the key from under a box before picking up the key. The third difficulty level further requires the agent to first move a ball blocking the door before entering the door. See Figure \ref{image:omg-easy-examples} for screenshots of example levels.

\begin{figure}[htbp]
    \centering
    \includegraphics[width=\linewidth]{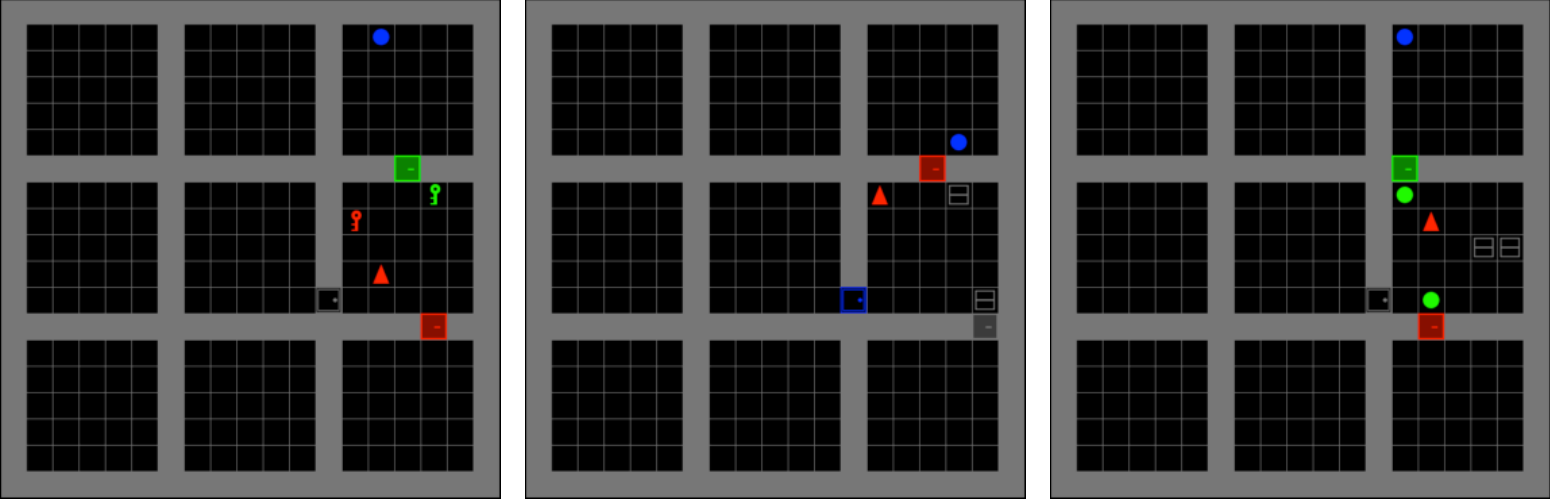}
    \caption{\small{Example levels in increasing difficulty from left to right of each additional difficulty setting introduced by OMG-Hard in addition to those in OMG-Easy.}}
    \label{image:omg-medium-examples}
\end{figure}

\medskip
\paragraph{ObstructedMazeGamut-Hard:} This environment consists of levels uniformly distributed across the first six difficulty levels of the ObstructedMaze environment. Harder levels corresponding to the fourth, fifth, and sixth difficulty settings include two additional rooms with no goal object to distract the agent. Each instance of these harder levels also contain two pairs of keys of different colors, each opening a door of the same color. The agent always starts one room away from the randomly positioned goal object. Each of the two keys is visible in the fourth difficulty setting and doors are unobstructed. The fifth difficulty setting hides the keys under boxes, and the sixth again places obstacles that must be removed before entering two of the doors, one of which is always the door to the goal-containing room. See Figure \ref{image:omg-medium-examples} for example screenshots.

\section{Partially-Observable Navigation}
\label{appendix:env_maze}

The partially-observable mazes are based on MiniGrid~\citep{gym_minigrid}. Each maze consists of a $15 \times 15$ grid, where each cell can contain a wall, the goal, the agent, or navigable space. Like in MiniGrid, the student agent receives a reward of $1-T/T_{\textnormal{max}}$ upon reaching the goal, where $T$ is the episode length and $T_{\textnormal{max}}$ is the maximum episode length (set to 250). Otherwise, the agent receives a reward of 0 if it fails to reach the goal. The observation space consists of the agent's orientation (facing north, south, east, or west) and the $5 \times 5$ grid immediately in front of and including the agent. This grid takes the form of a 3-channel integer encoding. The action space consists of 7 total actions, though mazes only make use of the first three: turn left, turn right, and forward. We do not mask out irrelevant actions.

\begin{figure}[H]
    \centering
    \includegraphics[width=\textwidth]{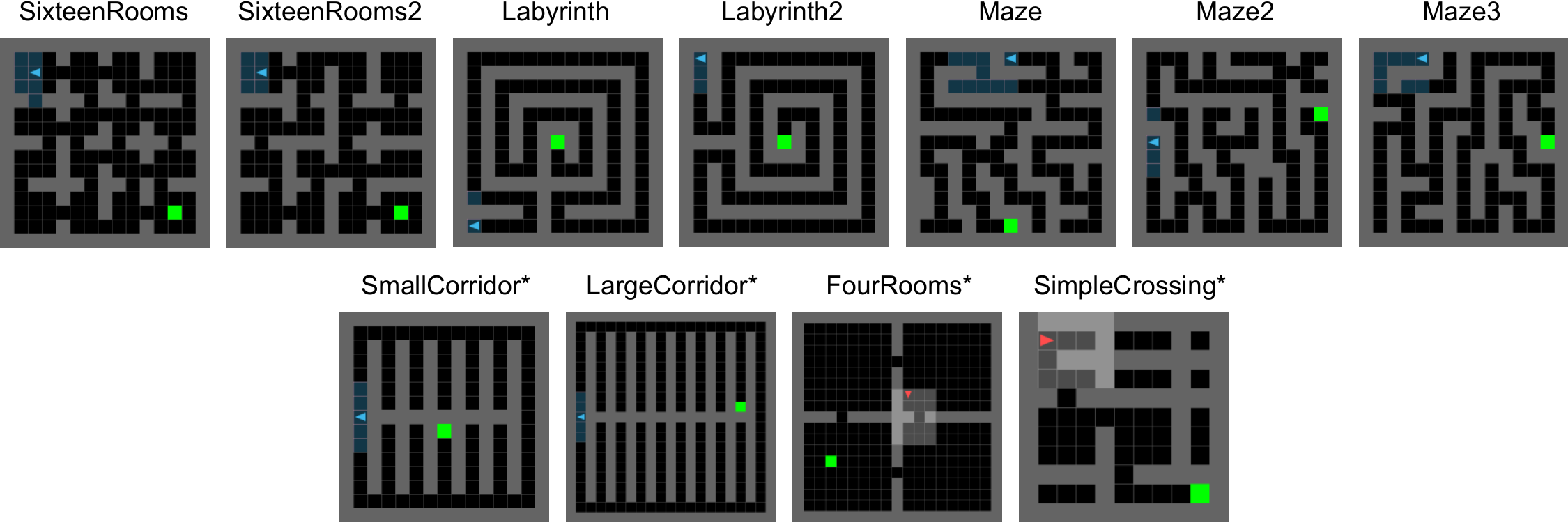}
    \caption{\small{MiniGrid zero-shot Environments. The asterisk * indicates the environment procedurally generates levels: For SmallCorridor and LargeCorridor, the position of the goal can be in any of the corridors. SimpleCrossing randomize vertical and horizontal barriers. FourRooms randomizes the starting location of the agent and the room containing the goal, and the location of room entrances.}}
    \label{fig:minigrid_ood_envs}
\end{figure}

\begin{figure}[h!]
    \centering
    \includegraphics[width=1\linewidth]{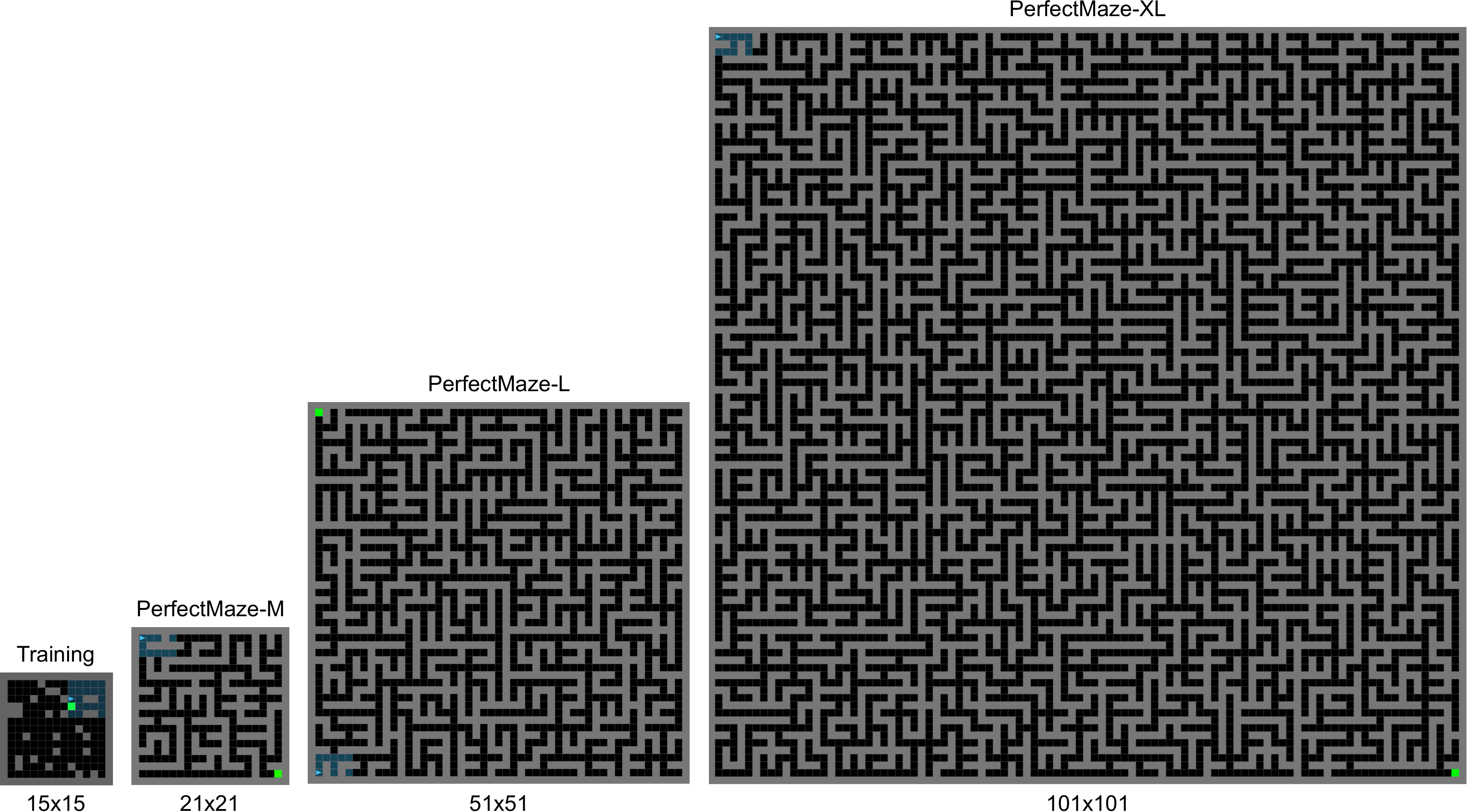}
    \caption{\small{PerfectMaze-(M,L,XL) environments parameterize singly-connected mazes of increasingly larger sizes. The figure depicts the mazes to scale.}}
    \label{figure:perfectmaze_scale}
\end{figure}

For zero-shot evaluation, we use a superset of the challenging test mazes in \citep{paired}: SixteenRooms environments require navigation through up to 16 rooms to find a goal; Labyrinth environments require traversal of a spiral labyrinth; and Maze environments require the agent to find a goal in a binary-tree maze, which requires the agent to successfully backtrack from dead ends. To more comprehensively test the agent's zero-shot transfer performance on OOD mazes, we also use several procedurally-generated mazes: PerfectMaze which parameterizes the set of singly-connected mazes; FourRooms, in which the goal is randomly positioned in one of four rooms, each accessible via a single narrow opening; SimpleCrossing (SimpleX), which requires the agent to weave through a series of horizontal and vertical walls; and finally, SmallCorridor and LargeCorridor, in which the goal position is randomly chosen to lie at the end of one of the corridors, thereby testing the agent's ability to perform backtracking. Figures~\ref{figure:main_minigrid}–\ref{figure:perfectmaze_scale} provide screenshots of the OOD mazes used in  Chapters~\ref{chapter:dcd}–\ref{chapter:accel}.

In order to test agent generalization to much larger mazes, we also define the PerfectMaze-(M,L,XL) environments, shown in Figure~\ref{figure:perfectmaze_scale}, which generate PerfectMaze instances with dimensions $21\times21$, $51\times51$, and $101\times101$ respectively.

\newpage
\section{CarRacing}
\label{appendix:env_carracing}

\begin{figure}[h!]
    \begin{minipage}{1\textwidth}
    \centering
    \begin{subfigure}{.18\textwidth}
        \includegraphics[width=\textwidth]{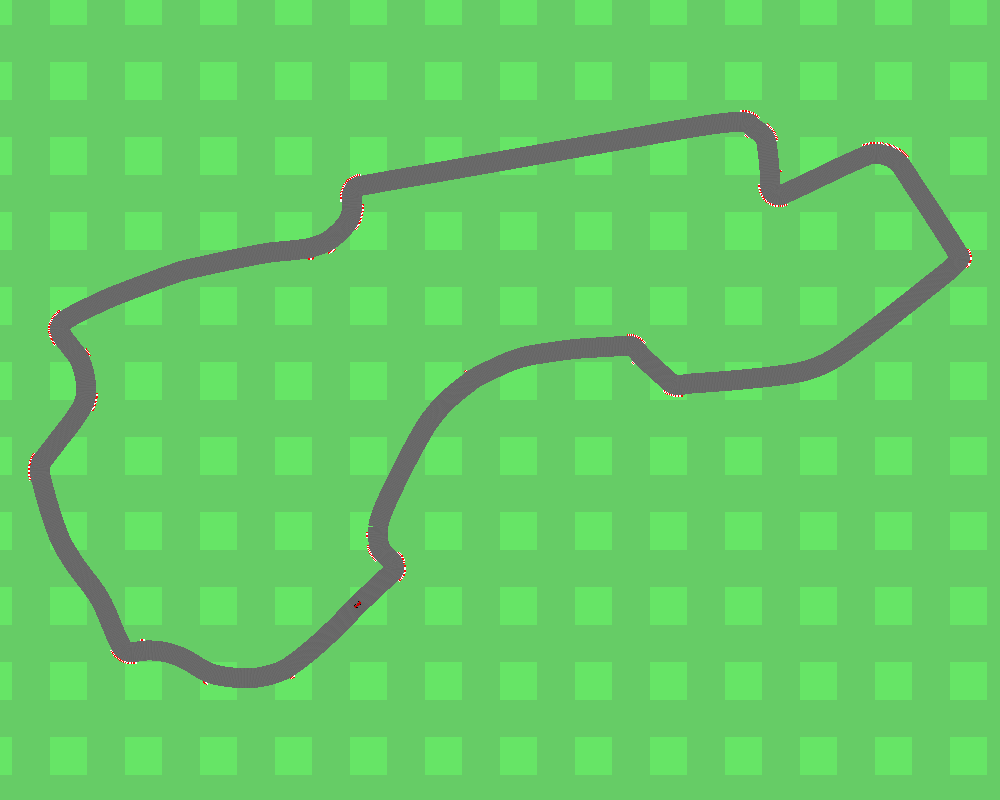}
        \caption{Australia}
    \end{subfigure}
    \begin{subfigure}{.18\textwidth}
        \includegraphics[width=\textwidth]{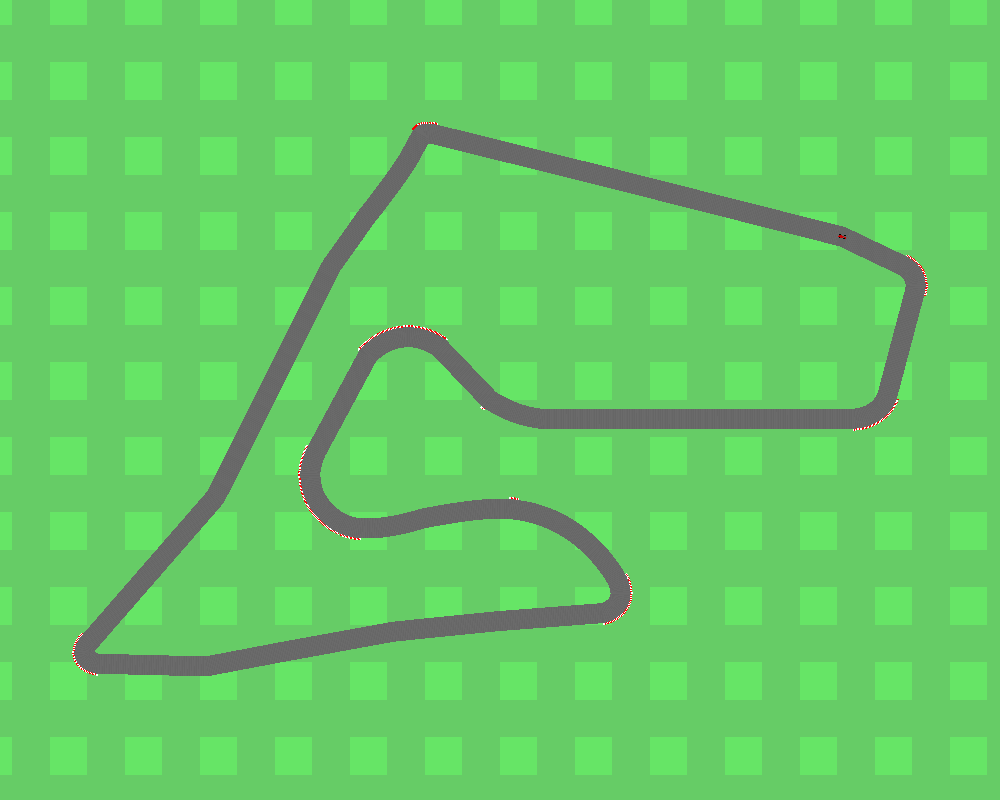}
        \caption{Austria}
    \end{subfigure}
    \begin{subfigure}{.18\textwidth}
        \includegraphics[width=\textwidth]{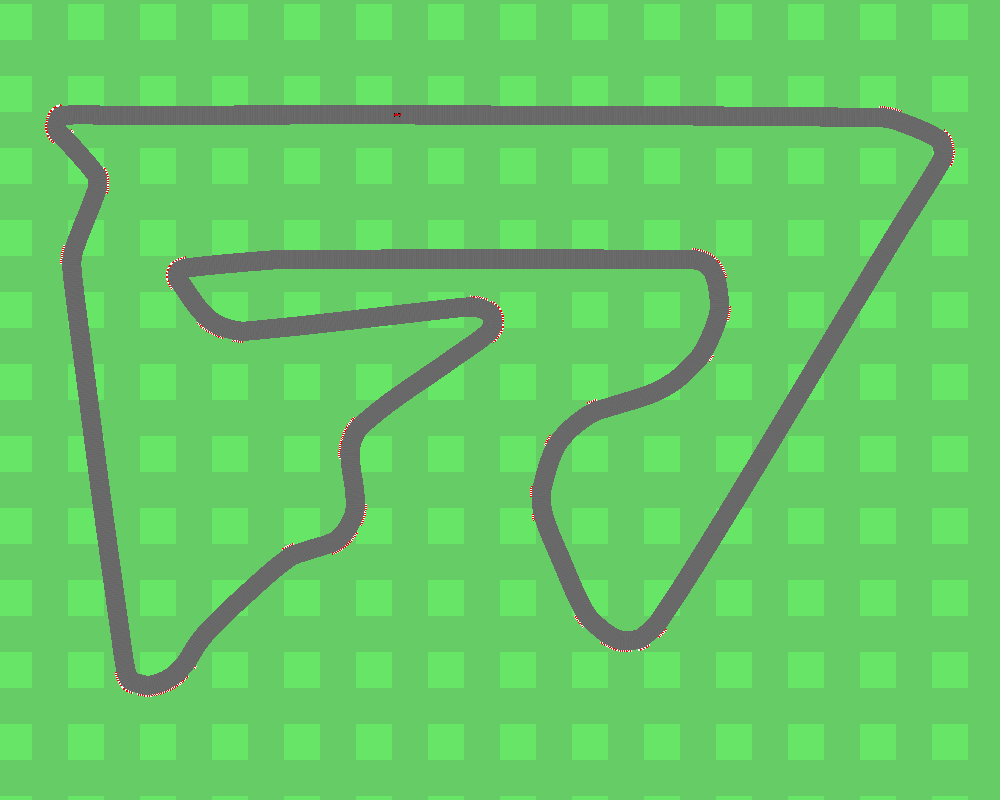}
        \caption{Bahrain}
    \end{subfigure}
    \begin{subfigure}{.18\textwidth}
        \includegraphics[width=\textwidth]{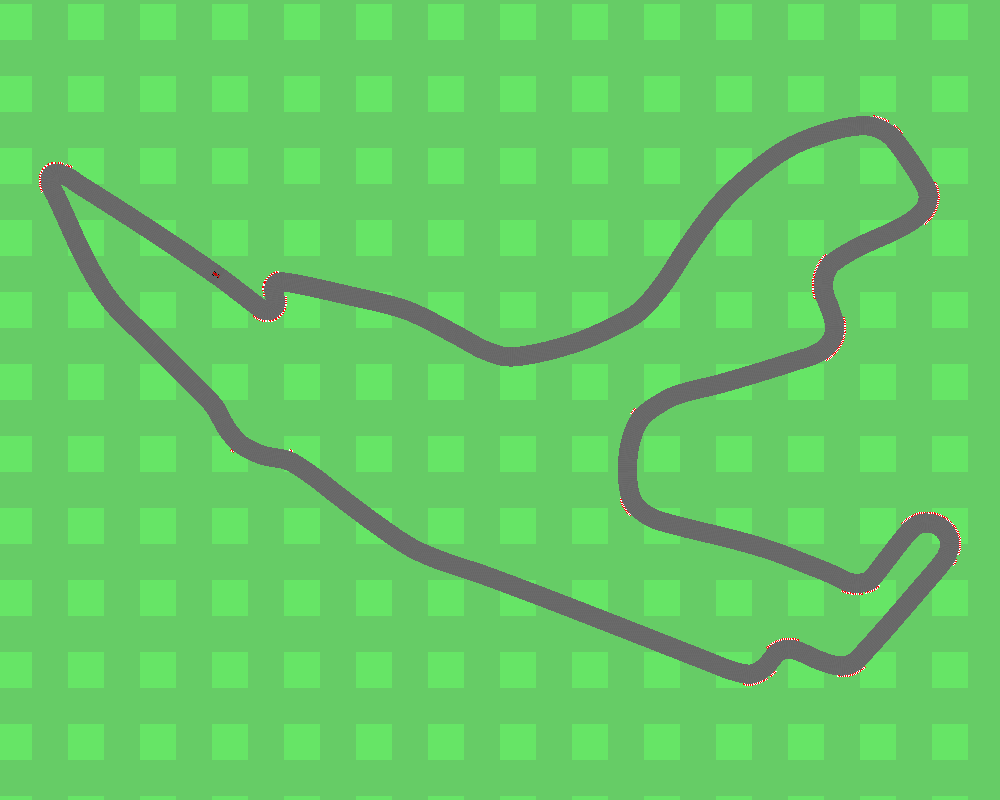}
        \caption{Belgium}
    \end{subfigure}
    \begin{subfigure}{.18\textwidth}
        \includegraphics[width=\textwidth]{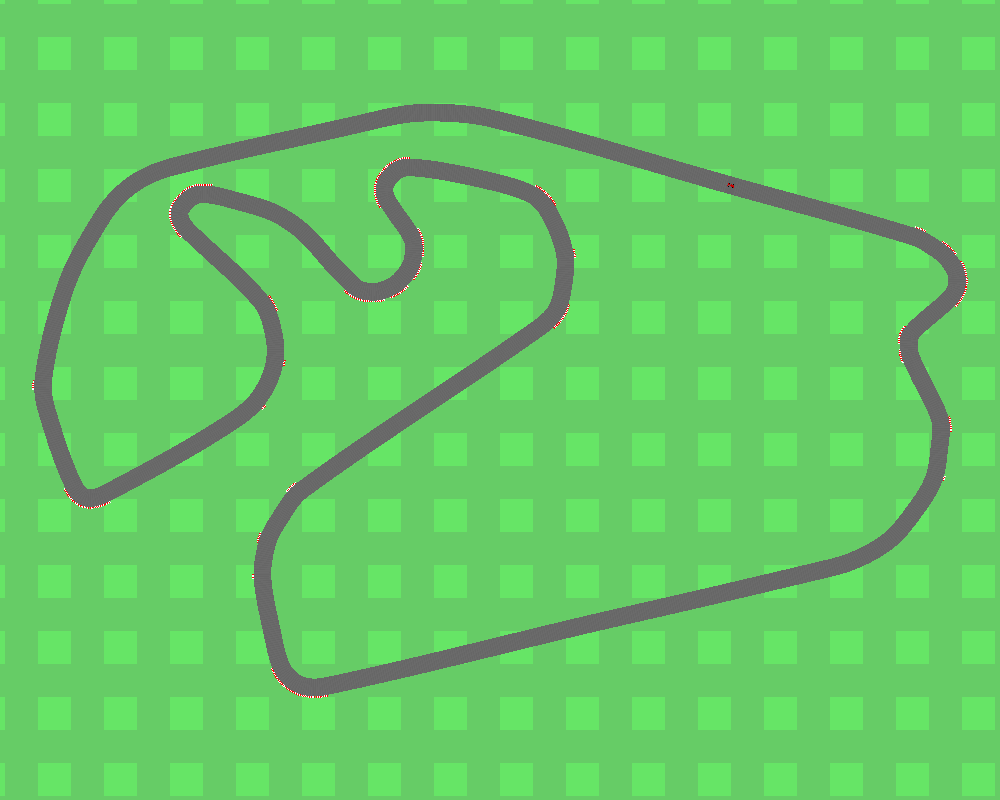}
        \caption{Brazil}
    \end{subfigure}
    \begin{subfigure}{.18\textwidth}
        \includegraphics[width=\textwidth]{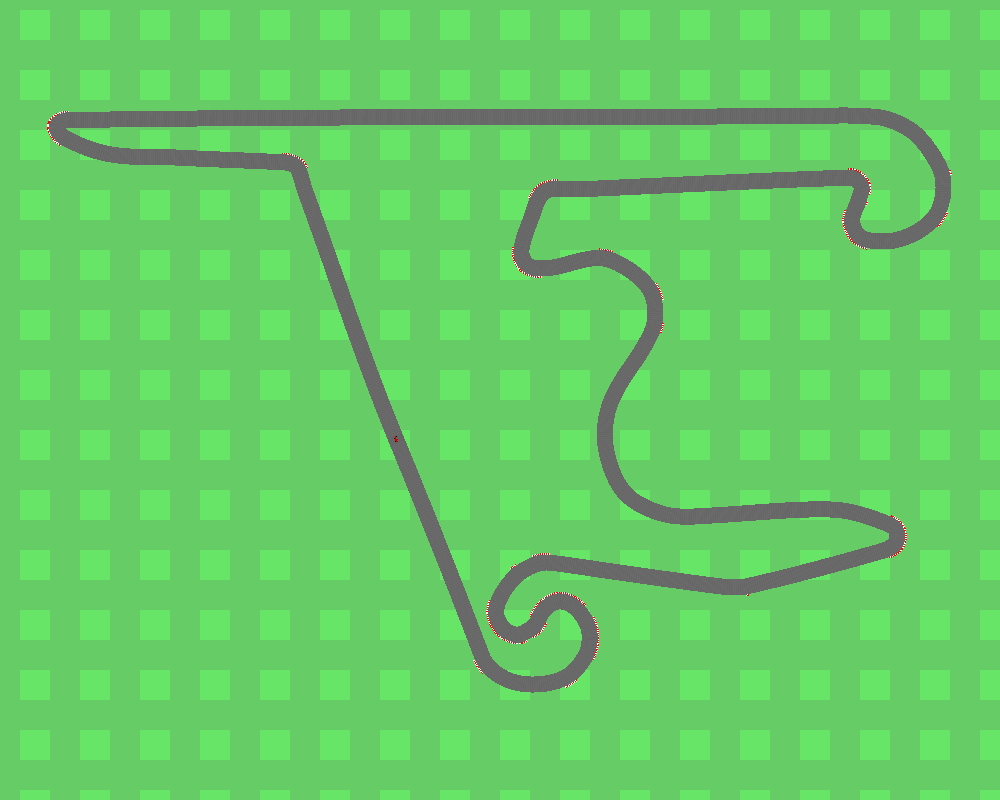}
        \caption{China}
    \end{subfigure}
    \begin{subfigure}{.18\textwidth}
        \includegraphics[width=\textwidth]{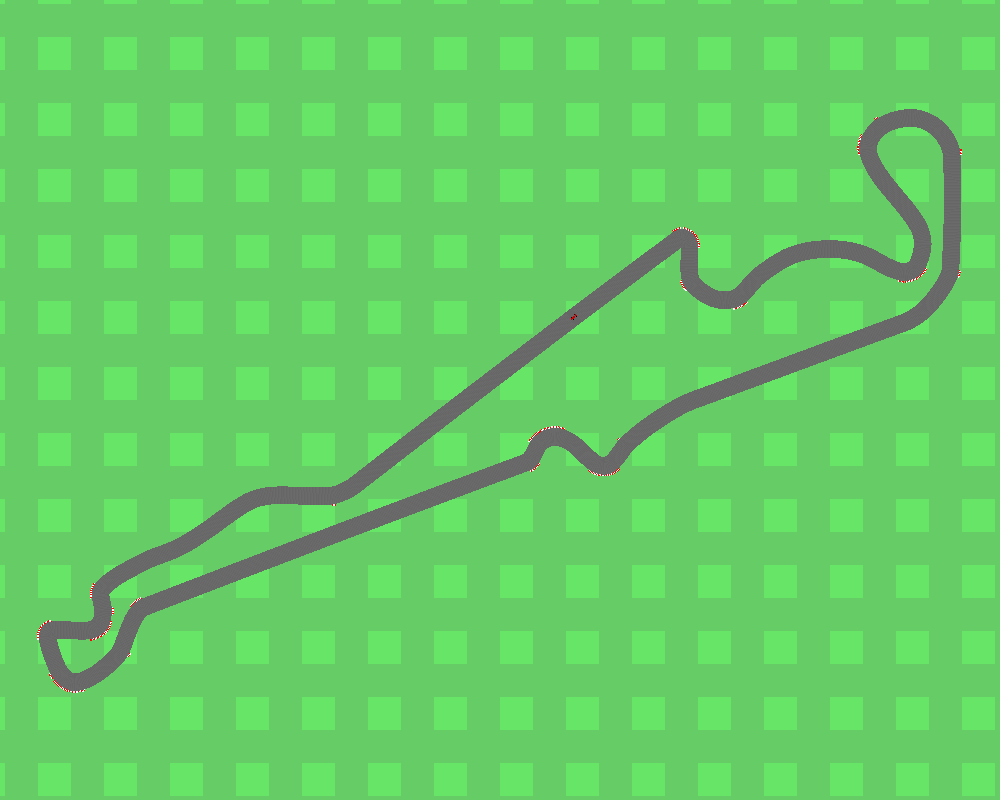}
        \caption{France}
    \end{subfigure}
    \begin{subfigure}{.18\textwidth}
        \includegraphics[width=\textwidth]{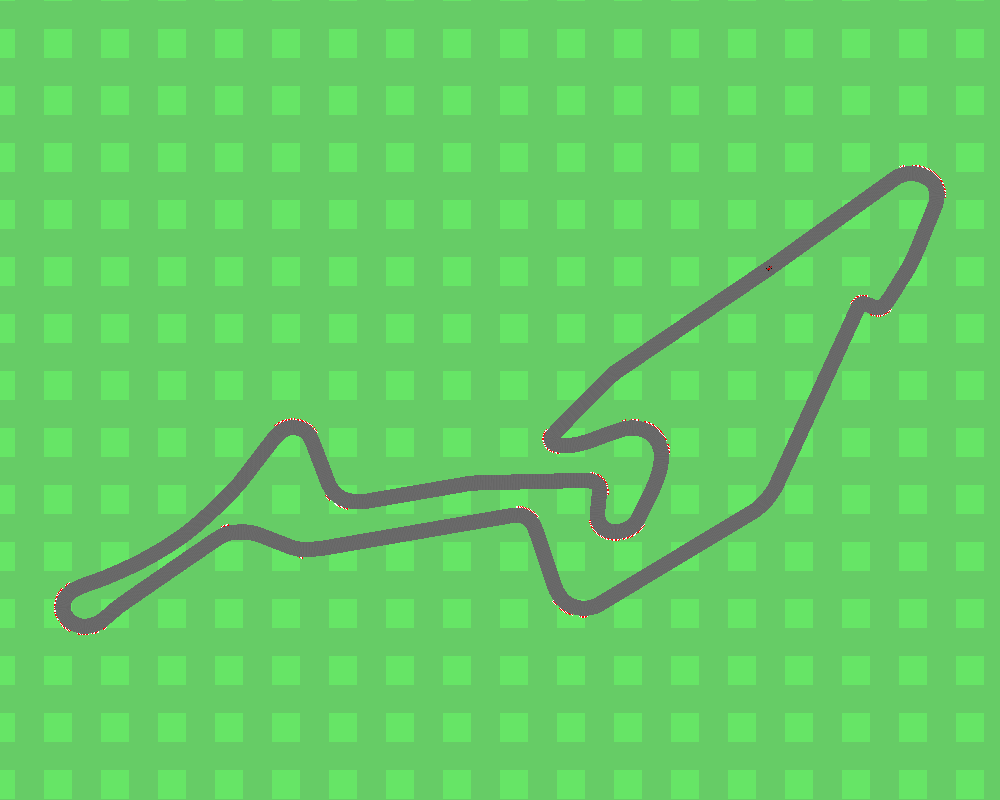}
        \caption{Germany}
    \end{subfigure}
    \begin{subfigure}{.18\textwidth}
        \includegraphics[width=\textwidth]{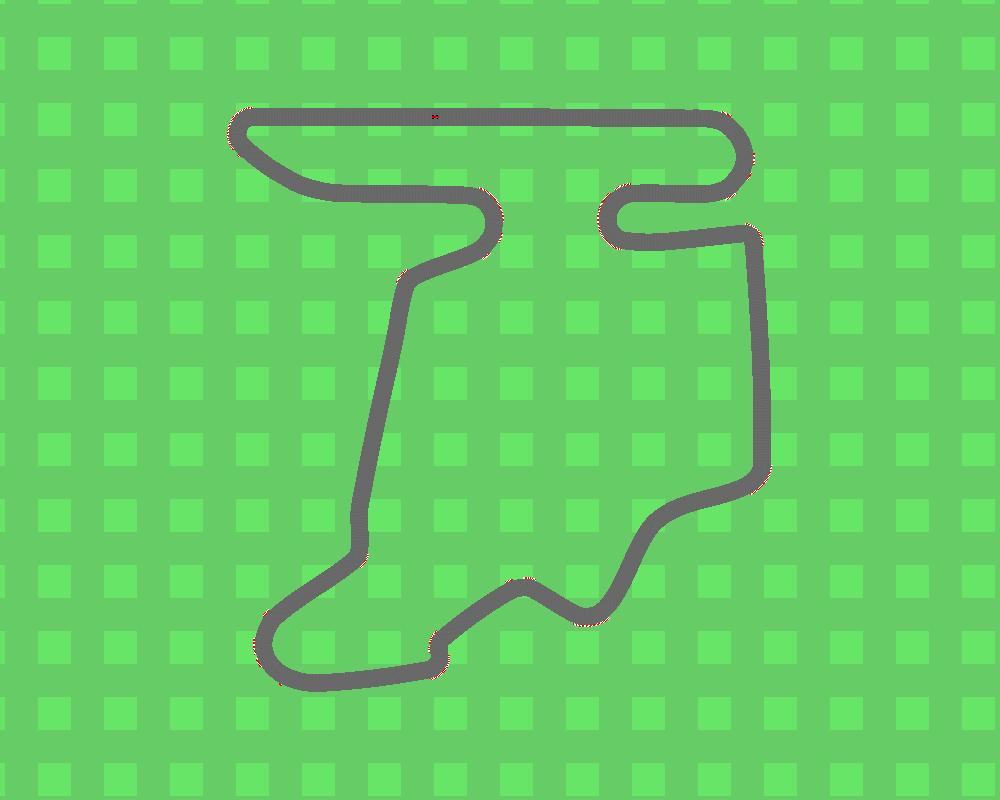}
        \caption{Hungary}
    \end{subfigure}
    \begin{subfigure}{.18\textwidth}
        \includegraphics[width=\textwidth]{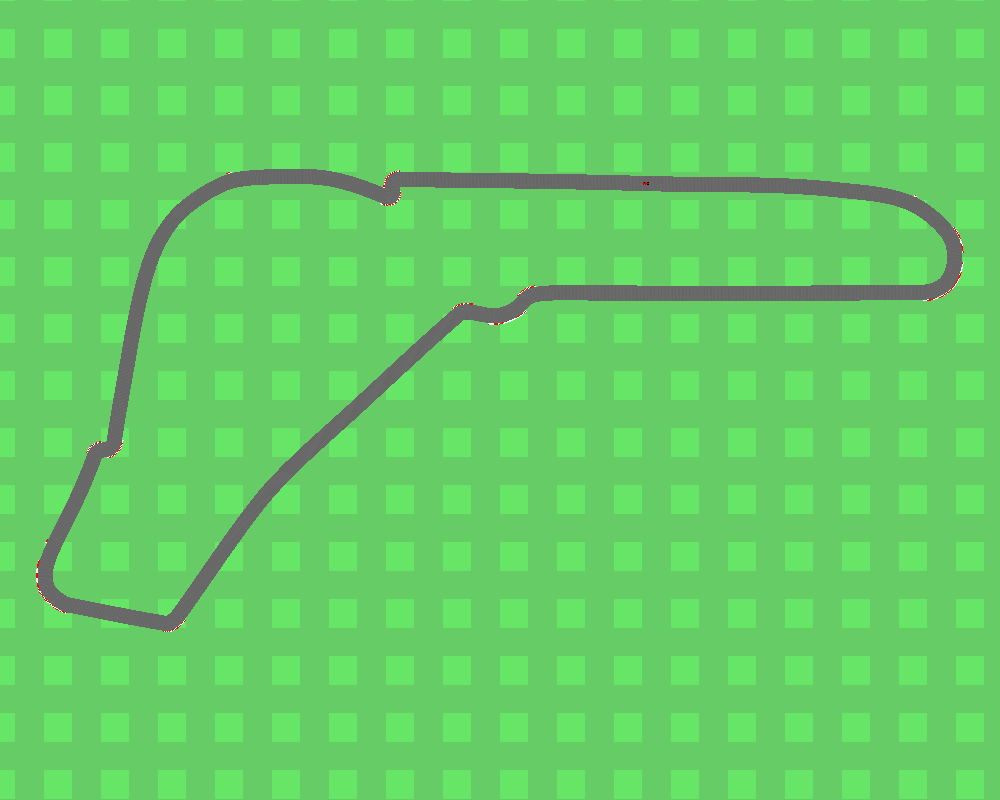}
        \caption{Italy}
    \end{subfigure}
    \begin{subfigure}{.18\textwidth}
        \includegraphics[width=\textwidth]{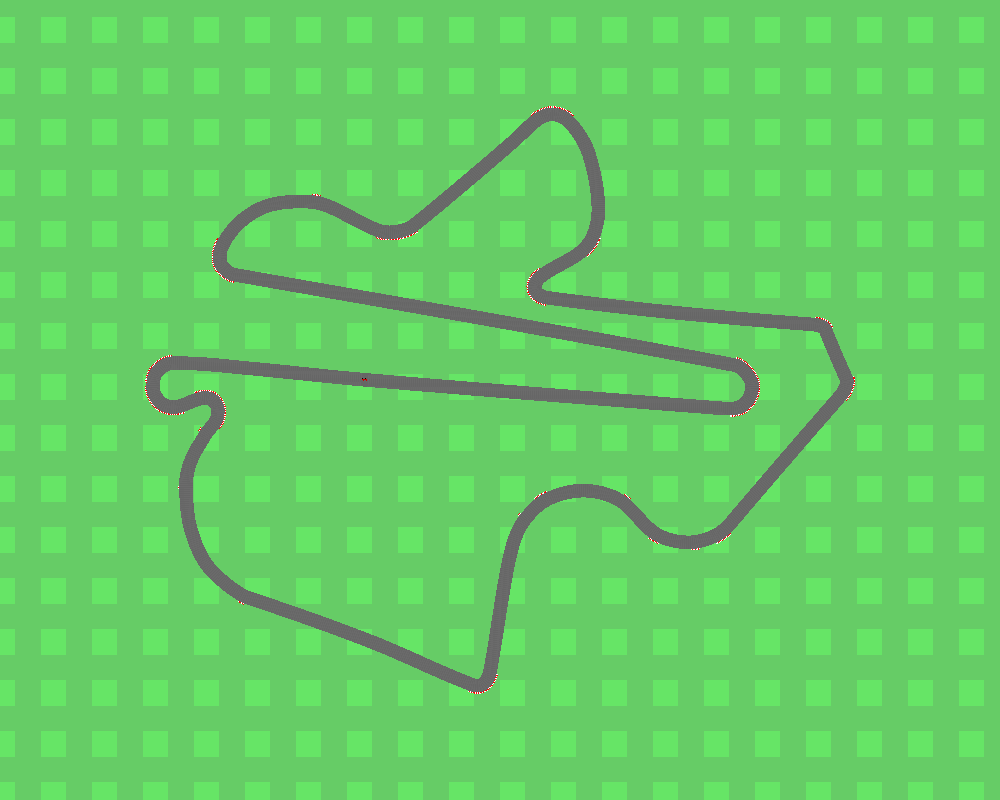}
        \caption{Malaysia}
    \end{subfigure}
    \begin{subfigure}{.18\textwidth}
        \includegraphics[width=\textwidth]{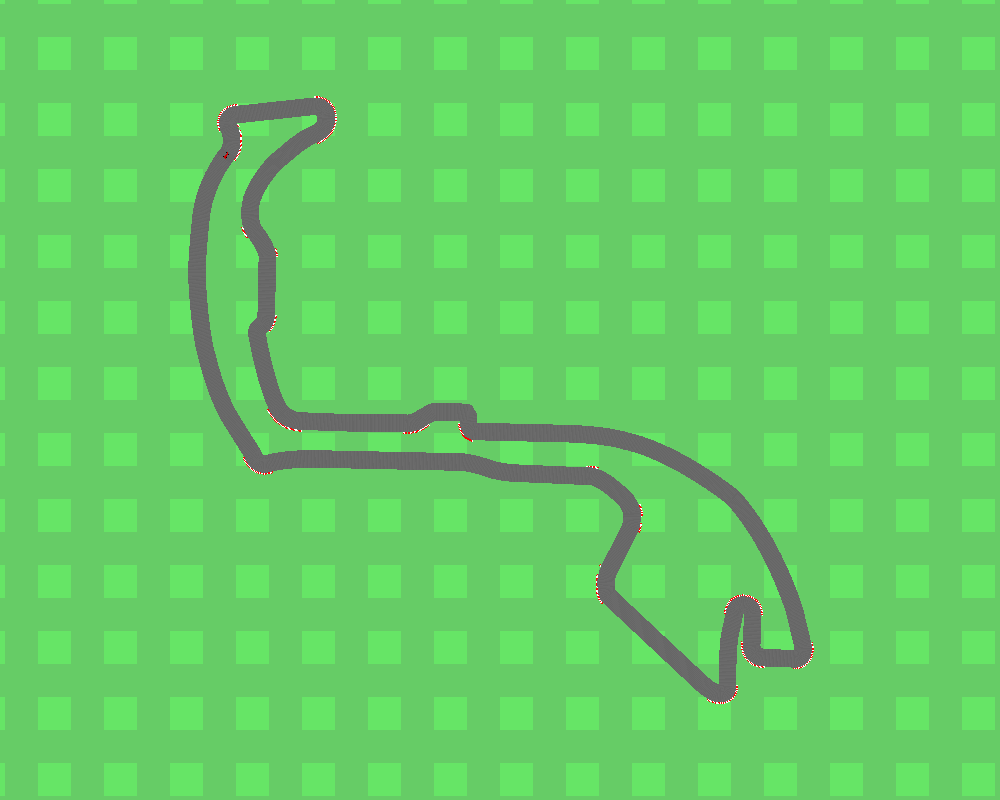}
        \caption{Malaysia}
    \end{subfigure}
    \begin{subfigure}{.18\textwidth}
        \includegraphics[width=\textwidth]{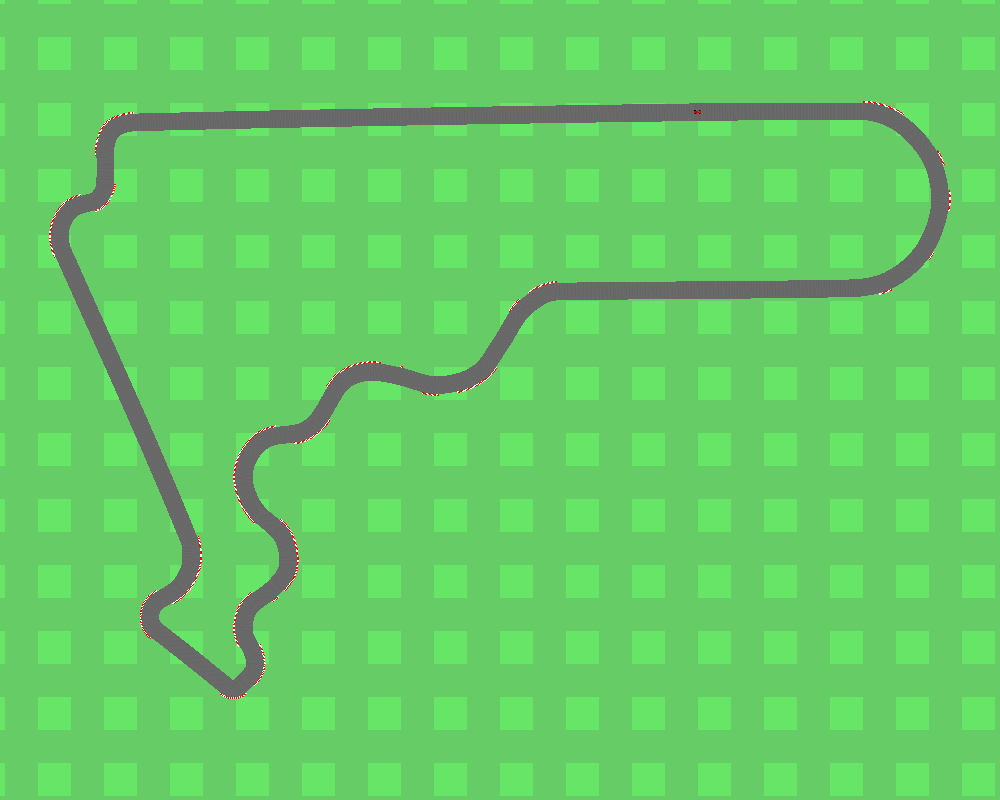}
        \caption{Malaysia}
    \end{subfigure}
    \begin{subfigure}{.18\textwidth}
        \includegraphics[width=\textwidth]{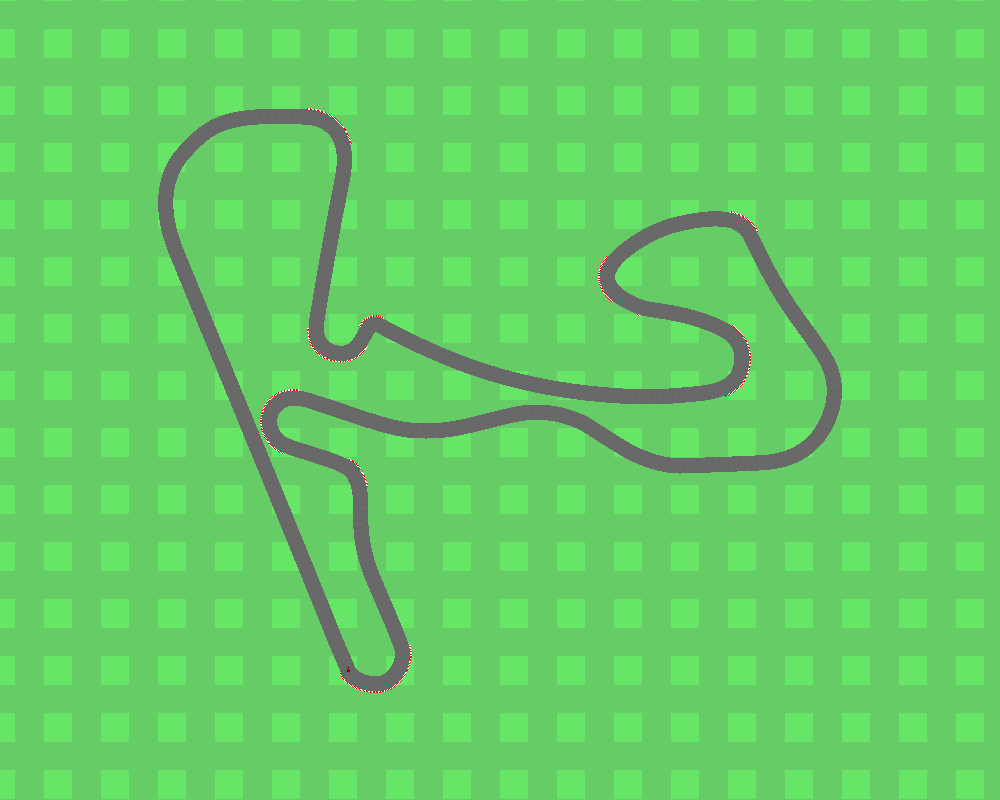}
        \caption{Netherlands}
    \end{subfigure}
    \begin{subfigure}{.18\textwidth}
        \includegraphics[width=\textwidth]{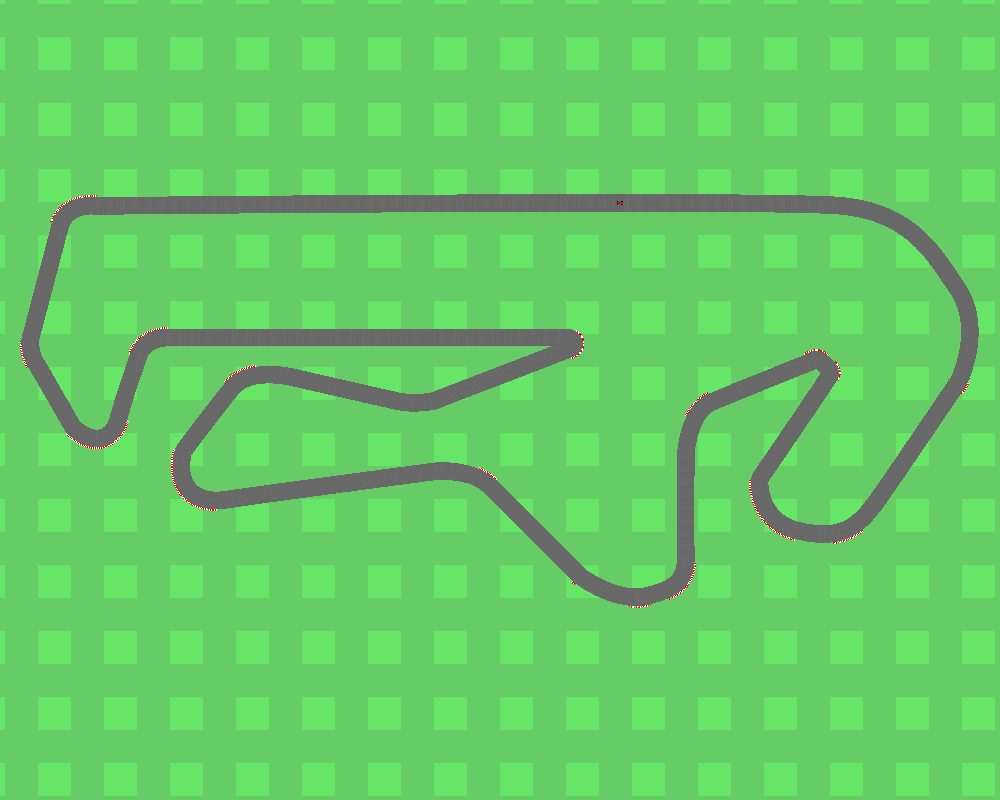}
        \caption{Portugal}
    \end{subfigure}
    \begin{subfigure}{.18\textwidth}
        \includegraphics[width=\textwidth]{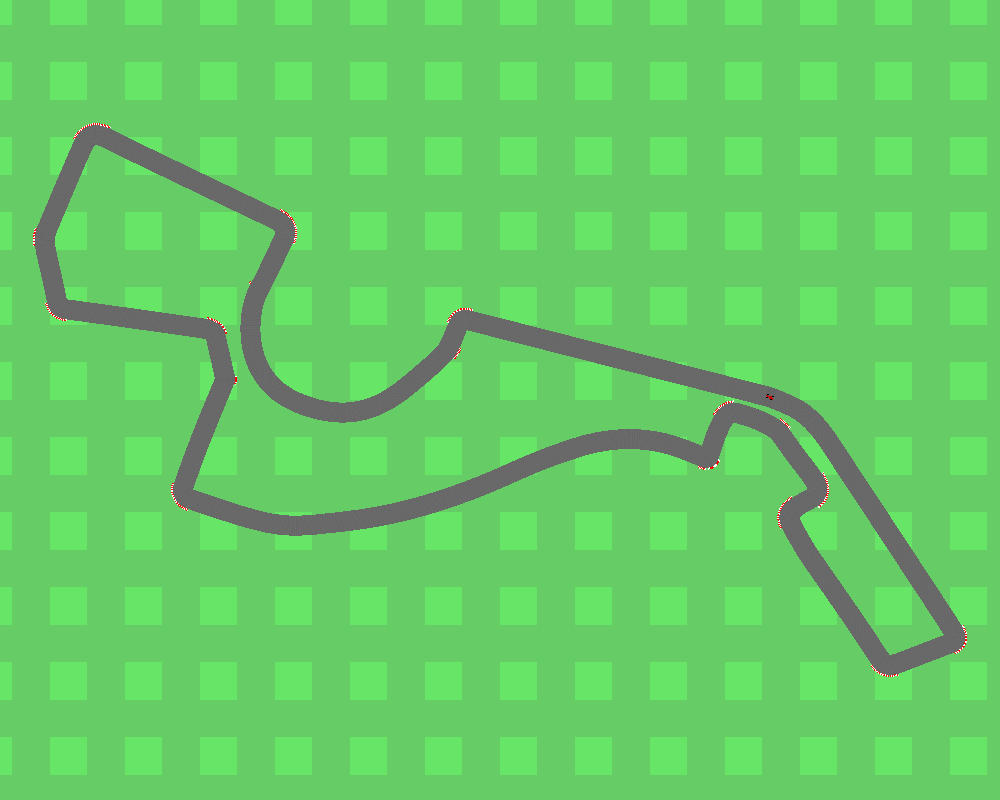}
        \caption{Russia}
    \end{subfigure}
    \begin{subfigure}{.18\textwidth}
        \includegraphics[width=\textwidth]{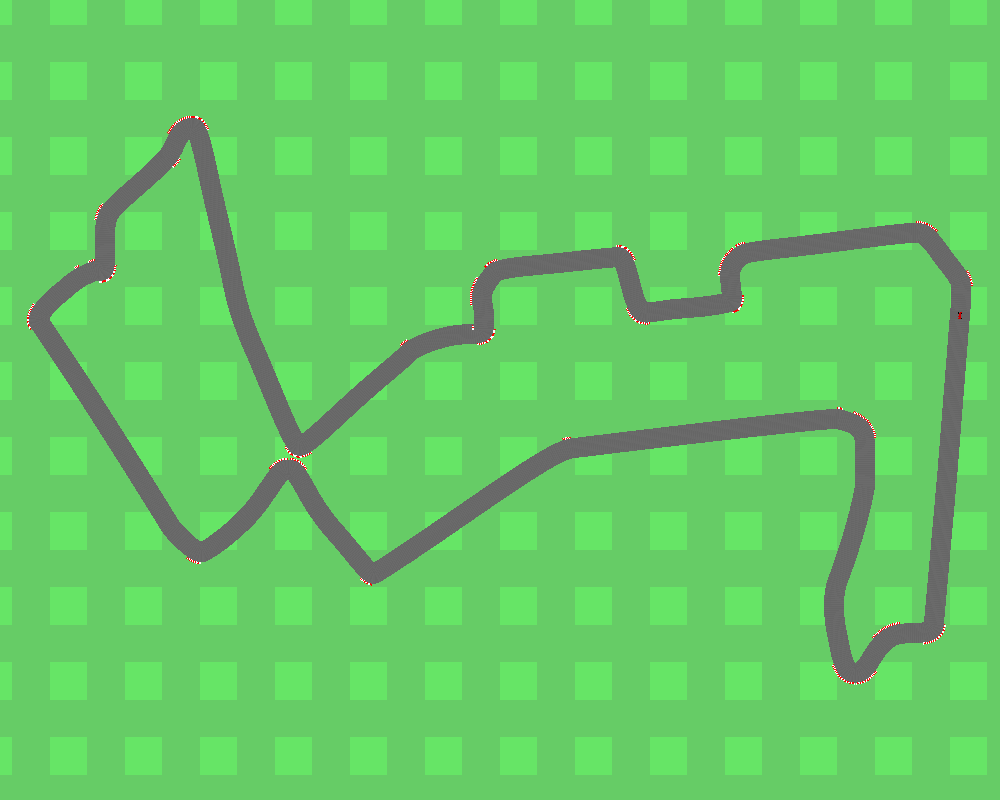}
        \caption{Singapore}
    \end{subfigure}
    \begin{subfigure}{.18\textwidth}
        \includegraphics[width=\textwidth]{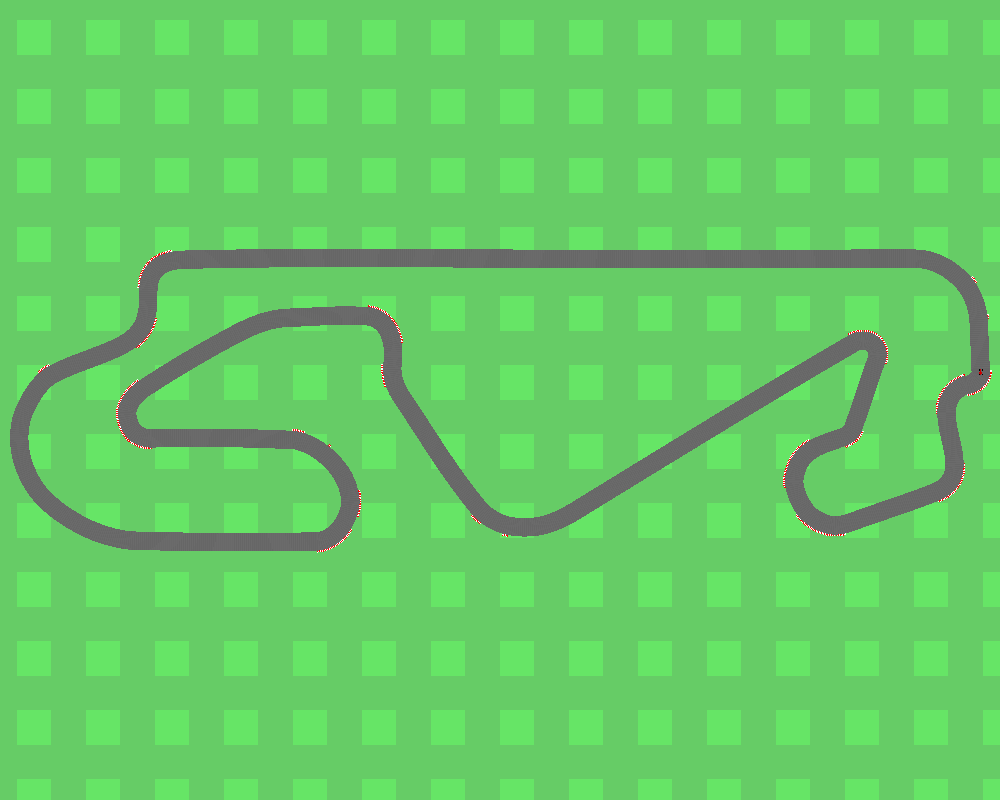}
        \caption{Spain}
    \end{subfigure}
    \begin{subfigure}{.18\textwidth}
        \includegraphics[width=\textwidth]{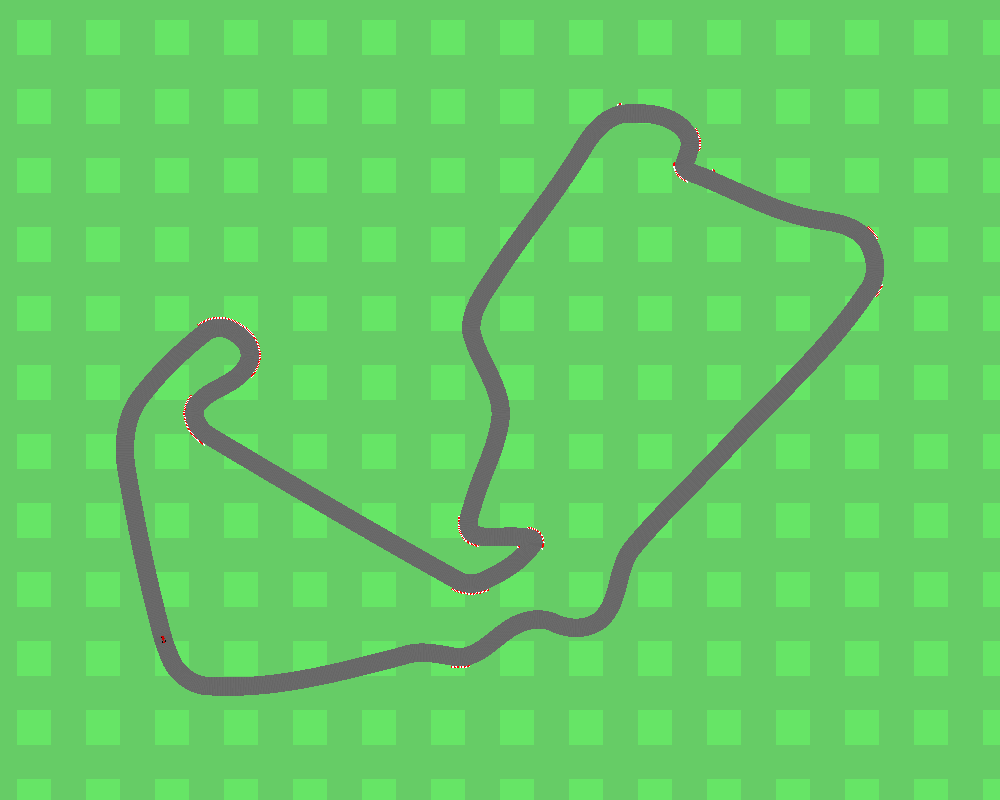}
        \caption{UK}
    \end{subfigure}
    \begin{subfigure}{.18\textwidth}
        \includegraphics[width=\textwidth]{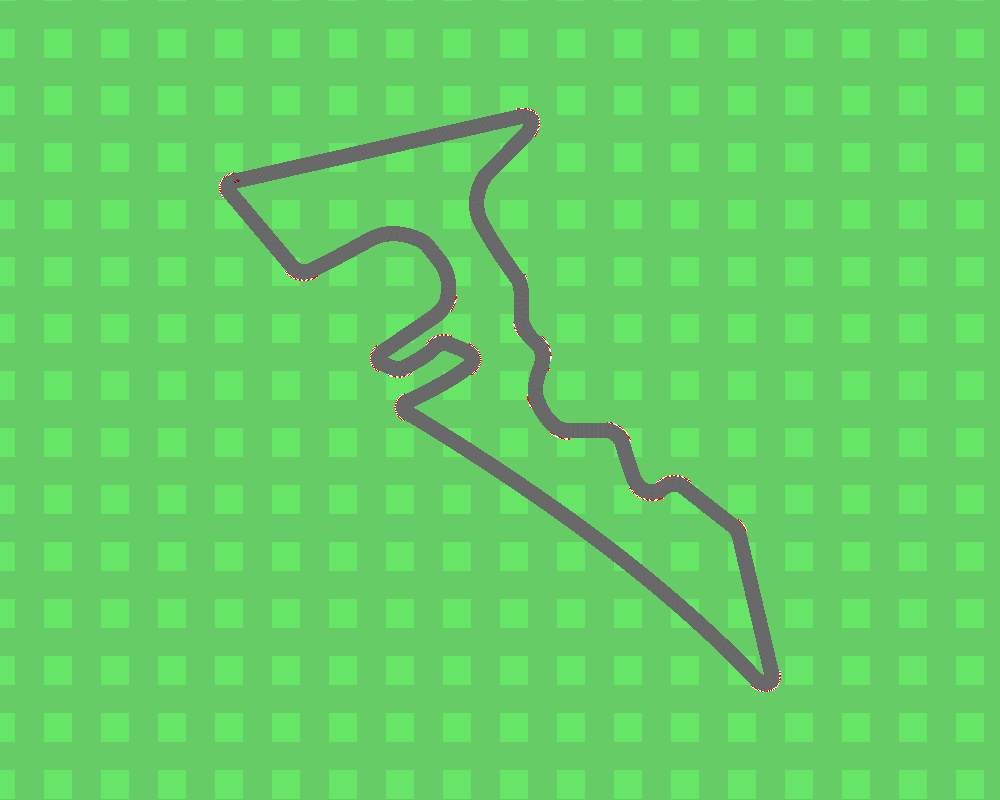}
        \caption{USA}
    \end{subfigure}
    \caption{\small{All tracks in the CarRacing-F1 benchmark used for evaluating zero-shot generalization.}}
    \label{figure:f1_tracks}
    \end{minipage}
\end{figure}

\begin{table}[!htb] 
      \caption{\small{Descriptions for each track in the CarRacing-F1 benchmark.}}
      \centering
      \scalebox{0.85}{
        \begin{tabular}{llr}
        \toprule
         Environment & Real-world track  & Max episode steps   \\
        \midrule
        Australia & Albert Park & 1500 \\
        Austria & Red Bull Ring & 1500 \\
        Bahrain & Bahrain International Circuit & 2500 \\
        Belgium &  Circuit de Spa-Francorchamps & 1500  \\
        Brazil &  Autódromo José Carlos Pace & 2000  \\
        China &  Shanghai International Circuit & 2500 \\
        France & Circuit Paul Ricard & 2000  \\
        Germany & Nürburgring & 2000  \\
        Hungary & Hungaroring & 2000  \\
        Italy & Monza Circuit & 1500  \\
        Malaysia & Sepang International Circuit  & 2500 \\
        Mexico & Autódromo Hermanos Rodríguez  & 2000  \\
        Monaco & Circuit de Monaco  & 1500  \\
        Netherlands & Circuit Zandvoort  & 2000 \\
        Portugal & Algarve International Circuit  & 2500  \\
        Russia &  Sochi Autodrom & 1500  \\
        Singapore &  Marina Bay Street Circuit & 2000  \\
        Spain &  Circuit de Barcelona-Catalunya & 2000  \\
        UK & Silverstone & 2000   \\
        USA & Circuit of the Americas, Austin & 2000  \\

        \bottomrule
        \end{tabular}}
    \label{table:carracing_f1_tracks}
\end{table}

The carracing environments used in Chapters~\ref{chapter:dcd} and \ref{chapter:samplr} are based on CarRacingBezier, which extends CarRacing from OpenAI Gym~\citep{gym} so that tracks are generated as a Bézier curve, so to increase the expressiveness of the environment parameterization to, in principle, support the rendering of any closed curve. Each track consists of a closed loop around which the student agent must drive a full lap. In our experiments, each track consists of a Bézier curve \citep{bezier_ref} based on 12 randomly sampled control points within a fixed radius, $B/2$, of the center of the $B\times B$ playfield. Each track consists of a sequence of $L$ polygons. When driving over each previously unvisited polygon, the agent receives a reward equal to $1000/L$. The student additionally receives a reward of -0.1 at each time step, where the maximum number of episode steps is set to 1000. Aligning with the methodology of \citet{carracing_ppo}, we do not penalize the agent for driving out of the playfield boundaries, terminate episodes if the agent drives too far off track, and repeat every selected action for 8 steps. The student observation space consists of a $96\times96\times3$ pixel observation with RGB channels with a clipped, egocentric, bird's-eye view of the vehicle centered horizontally in the top $84\times96$ portion of the frame. The remaining $12\times96$ portion of the frame consists of the dashboard visualizing the agent's latest action and return. Note that despite the lossiness of the downsampled dashboard, our hyperparameter sweep for the best PPO settings found that including the full frame enabled better performance. Given this observation, the student then decides on a 3-dimensional continuous action, where the components correspond to control values for steer (torque, in $[-1.0, 1.0]$), gas (acceleration, in $[0.0, 1.0]$), and brake (deceleration, in $[0.0, 1.0]$).

In Chapters~\ref{chapter:dcd} and ~\ref{chapter:samplr}, the zero-shot transfer levels are based on 20 real-world Formula One (F1) tracks designed to challenge professional racecar drivers. We predominantly selected tracks based on recent F1 seasons, including some historical favorites such as the Nürburgring Grand Prix.\footnote{We chose not to include the Japanese and Canadian Grand Prix due to the overlapping tracks at Suzuka and the Circuit Gilles Villeneuve.} This collection of tracks, which we call CarRacing-F1, provides a new benchmark for testing robustness and zero-shot generalization in a continuous control setting. Importantly, these tracks are strictly out-of-distribution and of higher complexity with respect to the training levels, as they cannot be represented by Bézier curves limited to 12 control points. Moreover, each F1 track requires more time steps to solve (1500 or 2000) than allotted for the training tracks (1000). Table~\ref{table:carracing_f1_tracks} provides per-track descriptions, and Figure~\ref{figure:f1_tracks} shows bird's-eye views of each track.

Chapter~\ref{chapter:samplr} further extends CarRacingBezier to optionally contain black ice on each track tile with probability $q$, where for a given track, $q$ may first be sampled from an arbitrary prior distribution and after which, ice is sampled I.I.D. per tile. It is impossible to accelerate or brake over ice tiles, which have a friction coefficient of 0, making icy settings much more challenging. Importantly, the identities of the black-ice tiles are not observable by the agent.

\section{BipedalWalker}
\label{appendix:env_bipedal}
We use a modified version of the BipedalWalkerHardcore environment from OpenAI Gym. The agent receives a 24-dimensional proprioceptive state corresponding to inputs from its lidar sensors, joint angles, and contacts. The agent does not have access to its positional coordinates. The action space is four continuous values that control the torques of its four motors. The environment design space is shown in Table~\ref{table:bipedal_params}, where we show the value of the initial environment parameterization for ACCEL, the edit size per parameter, and the maximum possible value of each parameter. In this environment, the UED parameters correspond to the range of values for obstacle attributes. Stair parameters define the number and height of stairs; the pit gap parameters, the width of the pit; the stump parameters, the height of  stumps; and the roughness parameters, the local rate at which the terrain can shift up or down. Under DR, each parameter is i.i.d. sampled uniformly from its corresponding range. Combined with a random seed, the UED parameters thus determine a specific level. For PLR, we combine the environment parameters with the specific random seed that deterministically produces the sampled level, ensuring deterministic generation of the replayed level. ACCEL makes each edit by uniformly sampling one of the eight environment parameters and adding or subtracting the corresponding edit size listed in Table~\ref{table:bipedal_params} from the current parameter value. 

\begin{table}[H]
\begin{center}
\caption{\small Environment design space for the BipedalWalker environment. The UED parameters define the range of values for obstacle attributes. When a specific level is created, each attribute of each obstacle is sampled from the corresponding range. 
}
\label{table:bipedal_params}
\scalebox{0.85}{
\begin{tabular}{ l ccccc } 
\toprule
& \textbf{Stump height} &\textbf{Stair height}  & \textbf{Stair steps} & \textbf{Roughness} & \textbf{Pit gap}  \\ 
\midrule
\textbf{Easy init} & [0,0.4] & [0,0.4] & 1 & Unif(0, 0.6) & [0,0.8] \\
\textbf{Edit size} & 0.2 & 0.2 & 1 & Unif(0, 0.6) & 0.4 \\
\textbf{Max value} & [5,5] & [5,5] & 9 & 10 & [10,10] \\
\bottomrule
\end{tabular}}
\end{center}
\end{table}

To test zero-shot transfer to OOD levels, we test agents on each of the individual challenges encoded in the environment parameterization. Specifically, we evaluate agents in the following four environments: 
\begin{itemize}
    \item Stair: The stair height parameters are set to [2,2] with the number of steps set to 5.
    \item PitGap: The pit gap parameter is set to [5,5].
    \item Stump: The stump parameter is set to [2,2].
    \item Roughness: The ground roughness parameter is set to 5.
\end{itemize}
Each of these environments is visualized in Figure~\ref{fig:bipedal_ood_test_curves}. We also test agents on the simple BipedalWalker-v3 environment and the more challenging BipedalWalkerHardcore-v3 environment. For BipedalWalkerHardcore-v3, we note that none of our agents fully solve the environment, which requires obtaining a mean reward $\ge 300$ over 100 independent trials. To test whether this outcome is possible with our base RL algorithm and agent model, we trained an identical PPO agent from scratch (without any curriculum) directly on the environment for 1B steps. The reward achieved was 239---indistinguishable from that achieved by ACCEL.

\section{Stochastic Fruit Choice}
\label{appendix:env_sfc}
The Stochastic Fruit Choice environment is built using MiniHack~\citep{samvelyan2021minihack}, a library for creating custom environments based on the NetHack Learning Environment~\citep[NLE,][]{nle} runtime. This environment embeds a stochastic binary choice task within a challenging hard-exploration problem. The agent must navigate through up to eight rooms in each level, and in the final room, choose the correct piece of fruit, either the apple or banana to receive a reward. If the agent eats the wrong fruit for the level, it receives a reward of $0$. With probability $q$, the apple is the correct fruit to eat. Eating either fruit terminates the episode. The episode also terminates once the budget of $250$ steps is reached. Notably, passage into adjacent rooms requires first kicking down a locked door. As per NLE game dynamics, locked doors may require a random number of kicks before they give way. To complicate the learning of this kicking skill, kicking the stone walls of the room will lower the agent's health points; multiple misguided kicks can then lead to the agent dying, ending the episode.

The agent's observation consists of two primary elements: The nethack \texttt{glyph} and \texttt{blstats} tensors. The \texttt{glyph} tensor represents a 2D symbolic observation of the dungeon. This glyph tensor contains a $21 \times 79$ window of glyph identifiers, which can each be one of the $5991$ possible glyphs in NetHack, which represent monsters, items, environment features, and other game entities. The \texttt{blstats} vector contains character-centric values, such as the agent's coordinates and the information in the ``bottom-line stats,'' such as the agent's health stats, attribute levels, armor class, and experience points. The action space includes the eight navigational actions, corresponding to moving toward each cell in the agent's Moore neighborhood, in addition to two additional actions for kicking (doors) and eating (apples and bananas).
\chapter{Additional Experiment Details}

\section{Prioritized Level Replay Experiments}
\label{appendix:exp_plr}

\medskip
\subsection*{Procgen Experiments}

To make the most efficient use of our computational resources, we perform hyperparameter sweeps on the easy setting. This also makes our results directly comparable to most prior works benchmarked on Procgen, which have likewise focused on the easy setting. In Procgen easy, our experiments use the recommended settings of $N_{\text{train}} = 200$ and 25M steps of training, as well as the same ResNet policy architecture and PPO hyperparameters shared across all games as in \citet{cobbe2019procgen} and \citet{raileanu2021automatic}. We find 25M steps to be sufficient for uncovering differences in generalization performance among our methods and standard baselines. Moreover, under this setup, we find Procgen training runs require much less wall-clock time than training runs on the two MiniGrid environments of interest over an equivalent number of steps needed to uncover differences in generalization performance. Therefore we survey the empirical differences across various settings of PLR on Procgen easy rather than MiniGrid. 

To find the best hyperparameters for PLR, we evaluate each combination of the scoring function choices in Table~\ref{table:scoring_metrics} with both rank and proportional prioritization, performing a coarse grid search for each pair over different settings of the temperature parameter $\beta$ in $\{0.1, 0.5, 1.0, 1.4, 2.0\}$ and the staleness coefficient $\rho$ in $\{0.1, 0.3, 1.0\}$. For each setting, we run 4 trials across all 16 of games of the Procgen Benchmark, evaluating based on mean unnormalized test return across games. In our TD-error-based scoring functions, we set $\gamma$ and $\lambda$ equal to the same respective values used by the GAE in PPO during training. We found PLR offered the most pronounced gains at $\beta=0.1$ and $\rho=0.1$ on Procgen, but these benefits also held for higher values ($\beta=0.5$ and $\rho=0.3$), though to a lesser degree. 

For UCB-DrAC, we make use of the best-reported hyperparameters on the easy setting of Procgen in \citet{raileanu2021automatic}, listed in Table~\ref{table:plr_hps}.

We found the default setting of mixreg's $\alpha=0.2$ used by \citet{wang2020mixreg} in the hard setting, performs poorly on the easy setting. Instead, we conducted a grid search over $\alpha$ in $\{0.001, 0.005, 0.01, 0.05, 0.1, 0.2, 0.8, 0.2, 0.5, 0.8, 1\}$.
 
Since the TSCL Window algorithm was not previously evaluated on Procgen Benchmark, we perform a grid search over different settings for both Boltzmann and $\epsilon$-greedy variants of the algorithm to determine the best hyperparameter settings for Procgen easy. We searched over window size $K$ in $\{10,100,1000,10000\}$, bandit learning rate $\alpha$ in $\{0.01, 0.1, 0.5, 1.0\}$, random exploration probability $\epsilon$ in $\{0.0, 0.01, 0.1, 0.5\}$ for the $\epsilon$-greedy variant, and temperature $\tau$ in $\{0.1, 0.5, 1.0\}$ for the Boltzmann variant. Additionally, for a fairer comparison to PLR we further evaluated a variant of TSCL Window that, like PLR, incorporates the staleness distribution, by additionally searching over values of the staleness coefficient $\rho$ in $\{0.0, 0.1, 0.3, 0.5\}$, though we ultimately found that TSCL Window performed best without staleness sampling ($\rho = 0$). 

See Table~\ref{table:plr_hps} for a comprehensive overview of the hyperparameters used for PPO, UCB-DrAC, mixreg, and TSCL Window, shared across all Procgen environments in our experiments on Procgen easy.

The evaluation protocol on the hard setting entails training on 500 levels over 200M steps~\citep{cobbe2019procgen}, making it more compute-intensive than the easy setting. To save on computational resources, we make use of the same hyperparameters found in the easy setting for each method on Procgen hard, with one exception: As our PPO implementation does not use multi-GPU training, we were unable to quadruple our GPU actors as done in \citet{cobbe2019procgen} and \citet{wang2020mixreg}. Instead, we resorted to doubling the number of environments in our single actor to 128, resulting in mini-batch sizes half as large as used in these two prior works. As such, our baseline results on hard are not directly comparable to theirs. We found setting mixreg's $\alpha=0.2$ as done in \citet{wang2020mixreg} led to poor performance under this reduced batch size. We conducted an additional grid search, finding $\alpha=0.01$ to perform best, as on Procgen easy.

\subsection*{MiniGrid Experiments}
We evaluate PLR with rank prioritization on two MiniGrid environments whose levels are uniformly distributed across several difficulty settings. Training on levels of varying difficulties helps agents make use of the easier levels as stepping stones to learn useful behaviors that help the agent make progress on harder levels. However, under the uniform-sampling baseline, learning may be inefficient, as the training process does not selectively train the agent on levels of increasing difficulty, leading to wasted training steps when a difficult level is sampled early in training. On the contrary, if PLR scores levels according to the time-averaged L1 value loss of recently experienced level trajectories, the average difficulty of the sampled levels should adapt to the agent's current abilities, following the reasoning outlined in the Value Correction Hypothesis, stated in Section~\ref{section:methods}.

As in \citet{igl2019generalization}, we parameterize the agent policy as a 3-layer CNN with 16, 32, and 32 channels, with a final hidden layer of size 64. All kernels are $2 \times 2$ and use a stride of 1. For the ObstructedMazeGamut environments, we increase the number of channels of the final CNN layer to 64. We follow the same high-level generalization evaluation protocol used for Procgen, training the agent on a fixed set of 4000 levels for MultiRoom-N4-Random, 3000 levels for ObstructedMazeGamut-Easy, and 6000 levels for ObstructedMazeGamut-Medium, and testing on the full level distribution. We chose these values for $|\Lambda_{\text{train}}|$ to ensure roughly 1000 training levels of each difficulty setting of each environment. We model our PPO parameters on those used by \citet{igl2019generalization} in their MiniGrid experiments. We performed a grid search to find that PLR with rank prioritization, $\beta = 0.1$, and $\rho = 0.3$ learned most quickly on the MultiRoom environment, and used this setting for all our MiniGrid experiments. Table~\ref{table:plr_hps} summarizes these hyperparameter choices.

\begin{table}[t!]
\caption{\small{Hyperparameters used for training on Procgen Benchmark and MiniGrid environments.}}
\label{table:plr_hps}
\begin{center}

\scalebox{0.85}{
\begin{tabular}{lrr}
\toprule
\textbf{Parameter} & Procgen & MiniGrid \\
\midrule
\emph{PPO} & & \\
$\gamma$ & 0.999 & 0.999 \\
$\lambda_{\text{GAE}}$ & 0.95 & 0.95 \\
PPO rollout length & 256 & 256 \\
PPO epochs & 3 & 4 \\
PPO minibatches per epoch & 8 & 8 \\
PPO clip range & 0.2 & 0.2 \\
PPO number of workers & 64 & 64 \\
Adam learning rate & 5e-4 & 7e-4 \\
Adam $\epsilon$ & 1e-5 & 1e-5 \\
return normalization & yes & yes \\
entropy bonus coefficient & 0.01 & 0.01 \\
value loss coefficient & 0.5 & 0.5 \\

\addlinespace[10pt]
\emph{PLR} & & \\
Prioritization & rank & rank \\
Temperature, $\beta$, $0.1$ & 0.1 & 0.1 \\
Staleness coefficient, $\rho$ & 0.1 & 0.3 \\

\addlinespace[10pt]
\emph{UCB-DrAC} & & \\
Window size, $K$ & 10 & - \\
Regularization coefficient, $\alpha_r$ & 0.1 & - \\
UCB exploration coefficient, $c$ & 0.1 & - \\

\addlinespace[10pt]
\emph{mixreg} & & \\
Beta shape, $\alpha$ & 0.01 & - \\

\addlinespace[10pt]
\emph{TSCL Window} & & \\
Bandit exploration strategy & $\epsilon$-greedy & - \\
Window size, $K$ & 10 & - \\
Bandit learning rate, $\alpha$ & 1.0 & - \\
Exploration probability, $\epsilon$ & 0.5 & - \\

\bottomrule
\end{tabular}}
\end{center}
\end{table}

\section{Dual Curriculum Design Experiments}
\label{appendix:exp_dcd}

\begin{table}[t!]
\caption{\small{Hyperparameters used for training each method in the maze and car racing environments.}}
\label{table:hp_dcd}
\begin{center}
\scalebox{0.85}{
\begin{tabular}{lrr}
\toprule
\textbf{Parameter} & MiniGrid & CarRacing \\
\midrule
\emph{PPO} & & \\
$\gamma$ & 0.995 & 0.99 \\
$\lambda_{\text{GAE}}$ & 0.95 & 0.9 \\
PPO rollout length & 256 & 125 \\
PPO epochs & 5 & 8 \\
PPO minibatches per epoch & 1 & 4 \\
PPO clip range & 0.2 & 0.2 \\
PPO number of workers & 32 & 16 \\
Adam learning rate & 1e-4 & 3e-4 \\
Adam $\epsilon$ & 1e-5 & 1e-5 \\
PPO max gradient norm & 0.5 & 0.5 \\
PPO value clipping & yes & no \\
Return normalization & no & yes \\
Value loss coefficient & 0.5 & 0.5 \\
Student entropy coefficient & 0.01 & 0.0 \\

\addlinespace[10pt]
\emph{PLR} and \emph{PLR}$^{\bot}$ & & \\
Replay rate, $p$ & 0.5 & 0.5 \\
Buffer size, $K$ & 4000 & 8000 \\
Scoring function & MaxMC & PVL \\
Prioritization & rank & proportional \\
Temperature, $\beta$ & 0.1 & 1.0 \\
Staleness coefficient, $\rho$ & 0.3 & 0.7 \\

\addlinespace[10pt]
\emph{PAIRED} & & \\
Student entropy coefficient & 0.0 & 0.0 \\
Generator entropy coefficient & 0.0 & 0.0 \\

\addlinespace[10pt]
\emph{REPAIRED} & & \\
Generator entropy coefficient & 0.0 & 0.01 \\
Replay rate, $p$ & 0.95 & 0.5 \\
Scoring function & MaxMC & MaxMC \\

\bottomrule 
\end{tabular}
}
\end{center}
\end{table}

This section details the environments, agent architectures, and training procedures used in our experiments discussed in Section~\ref{sec:experiments}. We use PPO to train both student and generator policies in all experiments. Section~\ref{sec:experiments} reports results for each method using the best hyperparameter settings, which we summarize in Figure \ref{table:hyperparams}. Note that unless specified, PPO hyperparameters are shared between student and teacher, and PLR hyperparameters are shared between \plrabbrev{} and REPAIRED. The procedures for determining the hyperparameter choices for each environment are detailed below.

\medskip
\subsection*{Partially-Observable Navigation Experiments}
\medskip
\noindent \textbf{Level generation:} Each maze is fully surrounded by walls, resulting in $13\times13=169$ cells in which the generator can place walls, the goal, and the agent. Starting from an initially empty maze (except the bordering walls), the generator is given a budget of $W=50$ steps in which it can choose a grid cell in which to place a wall. Placing a wall in a cell already containing a wall results in a no-opt. After wall placement, the generator then chooses cells for the goal and the agent's starting position. If either of these cells collides with an existing wall, a random empty cell is chosen. At each time step, the generator teacher receives the full grid observation of the developing maze, the one-hot encoding of the current time step, as well as a 50-dimensional random noise vector, where each component is uniformly sampled from $[0.0, 1.0]$.

\medskip
\noindent \textbf{Generator architecture:} We base the generator architecture on the the original model used for the PAIRED adversary in \citet{paired}. This model encodes the full grid observation using a convolution layer ($3\times 3$ kernel, stride length $1$, 128 filters) followed by a ReLU activation layer over the flattened convolution outputs. The current time step is embedded into a 10-dimensional space, which is concatenated to the grid embedding, along with the random noise vector. This combined representation is then passed through an LSTM with hidden dimension 256, followed by two fully-connected layers, each with a hidden dimension 32 and ReLU activations, to produce the action logits over the 169 possible cell choices. We further ablated the LSTM and found that its absence preserves the performance of the minimax generator in both 25-block and 50-block settings, as well as that of the PAIRED generator in the 50-block setting, as expected given that the full grid and time step form a Markov state. However, the PAIRED generator struggles to learn without an LSTM in the 25-block setting. We believe this improved performance in the 25-block setting is due to the additional network capacity provided by the LSTM. Therefore, in favor of less compute time, our experiments only used an LSTM-based generator for PAIRED in the 25-block setting.

\medskip
\noindent \textbf{Student architecture:} The student policy architecture resembles the LSTM-based generator architecture, except the student model uses a convolution with 16 filters to embed its partial observation; does not use a random noise vector; and instead of embedding the time step, embeds the student's current direction into a 5-dimensional latent space.

\medskip
\noindent \textbf{Choice of hyperparameters:} We base our choice of hyperparameters for student agents and generator (i.e. the teacher) on those used in \citet{paired}. We also performed a coarse grid search over the student entropy coefficient in $\{0.0, 0.01\}$, generator entropy coefficient in $\{0.0, 0.005, 0.01\}$, and number of PPO epochs in $\{5, 20\}$ for both students and generator, as well as the choice of including an LSTM in the student and generator policies. We selected the best performing settings based on average return on the validation levels of SixteenRooms, Labyrinth, and Maze over 3 seeds. Our final choices are summarized in \ref{table:hyperparams}. The main deviations from the settings in \citet{paired} are the choice of removing the generator's LSTM (except for PAIRED with 25 blocks) and using fewer PPO epochs (5 instead of 20). For PLR, we searched over replay rate, $p$, in $\{0.5, 0.95\}$ and level buffer size, $K$, in $\{500, 2000, 4000\}$, temperature $\beta$ in $\{0.1, 0.3\}$, and choice of scoring function in $\{\textnormal{PVL}, \textnormal{MaxMC}\}$. The final PLR hyperparameter selection was also used for \plrabbrev{} and REPAIRED, except for the scoring function, over which we conducted a separate search for each method.

\subsection*{CarRacing Experiments}
\label{appendix:exp_plr}

\medskip
\noindent \textbf{Level generation:} Starting from an empty track, the adversary generates a sequence of 12 control points, one per time step, spaced within a fixed radius, $B/2$ of the center $O$ of the playfield. The agent always begins centered at the track polygon closest to $0^{\circ}$ relative to $O$, facing counterclockwise.

\medskip
\noindent \textbf{Generator architecture:} At each time step, the generator policy receives the set of all control points so far generated, the current time step encoded as a one-hot vector, and a 16-dimensional random noise vector. The control points are spatially encoded in a $10\times10$ grid, called the \emph{sketch}, representing a downsampled and discretized version of the playfield bounds within which the generated track resides. Choosing a control point then corresponds to selecting one of the cells in this grid. After the control points are chosen, each control point's cell coordinates are upscaled to match the  original playfield scale. This ensures no two control points are too close together, preventing areas of excessive track overlapping. The sketch is embedded using two $2\times2$ convolutions using a stride length of 1 with 8 and 16 channels respectively, each followed by a ReLU layer. The flattened outputs of this sequence of convolutions is then concatenated with an 8-dimensional embedding of the time step and the random noise vector. This combined embedding is then fed through two fully connected layers, each with a hidden size of 256, where the first is followed by a ReLU activation, to produce the policy logits over the 100 choices of control points. Note that we mask out any cells in the sketch that have already been chosen to prevent double selection of the same control point. We also experimented with outputing continuous, downsampled control points in $[0.0, 1.0]$ by learning the $\alpha$ and $\beta$ parameters of a $\textnormal{Beta}$ distribution for each of $x$ and $y$ coordinates instead of categorical logits, but found this latter parameterization led to slower learning of generator policies, where the generator policy tended to remain close to or revert to an approximately uniformly random policy.

\medskip
\noindent \textbf{Student architecture:} The student policy architecture is based on the competitive PPO implementation in \citet{carracing_ppo}, which was used as a baseline for AttentionAgent in \citet{attentionagent}. This architecture consists of an image embedding module composed of a stack of 2D convolutions with square kernels of sizes 2, 2, 2, 2, 3, 3, channel outputs of 8, 16, 32, 64, 128, 256, and stride lengths of 2, 2, 2, 2, 1, 1 respectively, resulting in a 256-dimensional image embedding. The image embedding is then passed through a fully connected layer with a hidden size of 100, followed by a ReLU layer. This latter output is then fed through two separate fully-connected layers, each with hidden size of 100 and output dimension equal to the action dimension, followed by softplus activations. We then add 1 to each component of these two output vectors, which serve as the $\alpha$ and $\beta$ parameters respectively for the $\textnormal{Beta}$ distributions used to sample each action dimension. When training the student, we normalize rewards by dividing rewards by the running standard deviation of returns so far encountered. 

\medskip
\noindent \textbf{Choice of hyperparameters:} To determine the best hyperparameters for the student agents, we performed a grid search, in which we trained a  student agent with domain randomization for 300 PPO updates. The grid search covered PPO learning rate in $\{0.001, 0.0003\}$, $\lambda_{\textnormal{GAE}}$ in $\{0.0, 0.5, 0.9\}$, number of PPO epochs in $\{4, 8\}$, PPO number of minibatches per epoch in $\{2, 4, 8\}$, value loss coefficient in $\{0.5, 2.0\}$, whether to grayscale frames, whether to crop frames (i.e remove the dashboard portion), and whether to normalize returns. Further, we found entropy regularization tended to hurt performance of the student policy. Similar to the sharing of PPO hyperparameters between student and generator in \citep{paired}, we then shared the best PPO hyperparameters for the student with the generator, with the exception of searching over separate choices for the entropy coefficient in $\{0.0, 0.01\}$. We selected the best performing settings based on average return on the validation levels of F1-Italy, F1-Singapore, and F1-Germany over 3 seeds. For PLR, we searched over replay rate, $p$, in $\{0.5, 0.95\}$, level buffer size $K$, in $\{500, 2000, 4000, 8000\}$, replay prioritization in $\{\text{rank}, \text{proportional}\}$, staleness coefficient $\rho$ in $\{0.3, 0.7\}$, and replay distribution temperature $\beta$ in \{0.1, 1.0, 2.0\}. The best settings for PLR were then shared with \plrabbrev{} and REPAIRED, except for the scoring function, over which we performed a separate search for each method.

\section{Evolving Curricula Experiments}
\label{appendix:exp_accel}

\medskip
\noindent \textbf{Choice of model and hyperparameters:} The majority of our hyperparameters are inherited from previous works~\citep{paired, plr, jiang2021robustplr}, with a few small changes. For the Lava Grid environment, we use the agent model from the \citet{kuettler2020nethack}, using the \texttt{glyphs} and \texttt{blstats} as observations. The agent observes both a global and a locally cropped view (based on the coordinates in \texttt{blstats}). 

For MiniHack we conduct a grid search across the level replay buffer size $\{4000,10000\}$ for both PLR and ACCEL, and for ACCEL we sweep across the edit method in \{random, PVL\}, where the latter option equates to a learned editor trained with RL to maximize the PVL. For MiniGrid we use the replay buffer size from \citet{jiang2021robustplr} and only conduct the ACCEL grid search over the edit objective, again sweeping across \{random, PVL\}, as well as the replayed levels to edit from \{batch, subbatch (of size 1)\}, and replay rate from \{0.8, 0.9\}. For MiniGrid, we follow the protocol from \citet{jiang2021robustplr} and select the best hyperparameters using the validation levels \{16Rooms, Labyrinth, Maze\}. The final hyperparameters chosen are shown in Table \ref{table:hyperparams}.

For BipedalWalker we used the continuous control policy from the open source implementation of PPO from \citet{pytorchrl}, as well as many of the hyperparameters used in the recommended settings for MuJoCo. This involves a simple feed-forward neural network with two hidden layers of size 64 and tanh activations. We tuned the hyperparameters for our base agent using domain randomization, and conducted a sweep over the learning rate \{3e-4, 3e-5\}, PPO epochs $\{5,20\}$, entropy coefficient \{0, 1e-3\} and number of minibatches $\{4,32\}$, using the validation performance on BipedalWalkerHardcore. We then used these base agent configurations for all UED algorithms. For PLR we further conducted a sweep over the buffer size $\{1000, 5000\}$, replay rate $\{0,9, 0.5\}$ and staleness coefficient $\{0.3, 0.5, 0.7\}$, using the same settings found for both PLR and ACCEL. For ACCEL, we swept over number of edits in \{1, 2, 3, 4\} and whether to edit the full level replay batch or a randomly sampled replay level.

\begin{table}[h!]
\caption{\small{Hyperparameters used for training each method in each environment.}}
\label{table:hyperparams}
\begin{center}
\scalebox{0.85}{
\begin{tabular}{lllr}
\toprule
\textbf{Parameter} & MiniHack (Lava) & MiniGrid & BipedalWalker \\
\midrule
\emph{PPO} & \\
$\gamma$ & 0.995  & 0.995 & 0.99 \\
$\lambda_{\text{GAE}}$ & 0.95  & 0.95 & 0.9 \\
PPO rollout length & 256  & 256 & 2000 \\
PPO epochs & 5  & 5 & 5 \\
PPO minibatches per epoch & 1 & 1 & 32 \\
PPO clip range & 0.2 & 0.2 & 0.2 \\
PPO number of workers & 32 & 32 & 16 \\
Adam learning rate & 1e-4  & 1e-4 & 3e-4 \\
Adam $\epsilon$ & 1e-5 & 1e-5 & 1e-5 \\
PPO max gradient norm & 0.5 & 0.5 & 0.5 \\
PPO value clipping & yes & yes & no \\
return normalization & no  & no & yes \\
value loss coefficient & 0.5  & 0.5 & 0.5 \\
student entropy coefficient & 0.0  & 0.0 & 1e-3 \\
generator entropy coefficient & 0.0 & 0.0 & 0.0 \\

\addlinespace[10pt]
\emph{ACCEL} & & \\
Edit rate, $q$ & 1.0 & 1.0 & 1.0  \\
Replay rate, $p$ & 0.9 & 0.8 & 0.9 \\
Buffer size, $K$ & 10000 & 4000 & 1000 \\
Scoring function & PVL & PVL & PVL \\
Edit method & PVL & random & random \\
Levels edited & full batch & subbatch & subbatch \\
Number of edits & 5 & 5 & 3 \\
Prioritization & rank  & rank & rank \\
Temperature, $\beta$ & 0.3  & 0.3 & 0.1 \\
Staleness coefficient, $\rho$ & 0.3 & 0.3 & 0.5 \\

\addlinespace[10pt]
\emph{PLR} & & \\
Scoring function & PVL & PVL  & PVL \\
Replay rate, $p$ & 0.5 & 0.5 & 0.5 \\
Buffer size, $K$ & 10000 & 4000 & 1000 \\

\bottomrule 
\end{tabular}}
\end{center}
\end{table}

\medskip
\noindent \textbf{Level generation:} For a fair comparison to the PAIRED level generation procedure, DR is implemented by sampling a uniformly random teacher policy to output actions that set the environment parameters, thereby designing each level. Under PAIRED, this policy is no longer uniformly random, but rather optimized to maximize the estimated regret (e.g. PVL) incurred by the student agent on the resulting levels. The environment design procedure for the lava and maze domains is as follows: For each timestep the teacher receives an observation consisting of a map of the entire level and takes chooses a tile in the grid. For the first $N$ steps, where $N$ is teacher's budget of blocks (or lava tiles) the teacher always places a block (or lava tile). In the last two time steps, the teacher chooses a location for the goal and agent. This procedure reflects the approach taken in several recent works \citep{paired, plr, jiang2021robustplr, pcgrl}. For BipedalWalker, the teacher generates each level by choosing a random value between the minimum value of the ``Easy Init'' range in Table~\ref{table:bipedal_params} and the maximum value for each environment parameter. A random integer is then generated to seed the procedural content generation algorithm, which takes the sampled parameters to produce the level. 

\medskip
\noindent \textbf{Level editing:} In Lava Grid, edits only add or remove obstacle tiles (i.e. lava or wall block tiles), while in MiniGrid mazes, edits can also alter the goal location. If an edit places a lava or block tile in the current goal or agent position, then the new tile replaces the goal or agent, which is randomly relocated  after applying all remaining edits. In the BipedalWalker environments, each edit operation first uniformly samples an environment parameter, followed by incrementing or subtracting its value by the edit sizes defined in Table~\ref{table:bipedal_params}. 

\begin{table}[H]
\begin{center}
\caption{\small{Total number of environment steps for a given number of student PPO updates.}}
\label{table:accel_stepcount}
\scalebox{0.95}{
\begin{tabular}{ ll | cc } 
\toprule
\textbf{Environment} & \textbf{PPO Updates} & PLR & ACCEL \\ 
\midrule
MiniGrid & 20k & 327M & 369M \\
BipedalWalker & 30k & 1.96B & 2.07B \\
\bottomrule
\end{tabular}}
\end{center}
\end{table}

\section{Aligning Curricula Experiments}
\label{appendix:exp_samplr}

\medskip
\subsection*{Stochastic Fruit Choice Experiments}
\medskip
\noindent \textbf{Student architecture:} We make use of the same agent architecture from \citet{nle}. The policy applies a ConvNet to all visible glyph embeddings and a separate ConvNet to a $9 \times 9$ egocentric crop around the agent---which was found to improve generalization---producing two latent vectors. These are then concatenated with an MLP encoding of the \texttt{blstats} vector, the resulting vector is further processed by an MLP layer, and finally, input through an LSTM to produce the action distribution. We used the policy architecture provided in {\color{blue} \texttt{\url{https://github.com/facebookresearch/nle}}}.

\medskip
\noindent \textbf{Choice of hyperparameters:} Our choice of PPO hyperparameters, shared across all methods, was based on a grid search, in which we train agents with domain randomization on a $15 \times 15$ maze, in which the goal location and initial agent location, along with up to $50$ walls, are randomly placed. For each setting, we average results over 3 training runs. We chose this environment to perform the grid search, as it allows for significantly faster training than the multi-room environment featured in our main experiments. Specifically, we swept over the following hyperparameter values: number of PPO epochs in $\{5,20\}$, number of PPO minibatches in $\{1,4\}$, PPO clip parameter in $\{0.1,0.2\}$, learning rate in $\{1\text{e-}3, 1\text{e-}4\}$, and discount factor $\gamma$ in $\{0.99, 0.995\}$. Fixing these hyperparameters to the best setting found, we then performed a separate grid search over PLR's replay rate $p$ in $\{0.5, 0.95\}$ and replay buffer size in $\{4000, 5000, 8000\}$, evaluating settings based on evaluation levels sampled via domain randomization after 50M steps of training.

\medskip
\subsection*{CarRacing with Black Ice Experiments}
\medskip
\noindent \textbf{SAMPLR implementation:} In the car racing environment, the aleatoric parameters $\theta'$ determine whether each track tile contains black ice. Thus, the training distribution $\PCur(\Theta')$ directly impacts the distribution over $\tau$. In order to correct for the biased trajectories $\tau$ generated under its minimax regret curriculum, we must train the policy to maximize the \GTU{} conditioned on $\tau$, $\VGroundPiAfterT{\pi}{\tau}$. SAMPLR accomplishes this by training the policy on fictitious transitions that replace the real transitions observed. Each fictitious transition corresponds to the reward $r'_t$ and $s'_{t+1}$ that would be observed if the agent's action were performed in a level such that $\theta' \sim \PGround(\theta' | \tau)$. By training the agent on a POMDP whose future evolution conditioned on $\tau$ is consistent with $\PGround$, we ensure any  optimal policy produced under the biased training distribution will also be optimal under $\PGround(\Theta')$.

Recall from Equation \ref{eq:belief_model} that $\mathcal{B}(s'_t|\tau) = \sum_{\theta'}\PGround(s'_t|\tau, \theta')\PGround(\theta'|\tau)$. We can thus sample a fictitious state $s'_t$ according to $\mathcal{B}$ by first sampling $\theta' \sim \PGround(\theta'|\tau)$, and then $s'_t \sim \PGround(s'_t|\tau, \theta')$. We implement SAMPLR for this domain by assuming perfect models for both the posterior $P(\theta'|\tau)$ and $\PGround(s'_t|\tau, \theta')$. 

Simulating a perfect posterior over $\theta'$ is especially straightforward, as we assume each tile has ice sampled I.I.D. with probability $q \sim \text{Beta}(\alpha, \beta)$, where we make use of the conjugate prior. As $\tau$ contains the entire action-observation history up to the current time, it includes information that can be used to infer how much ice was already seen. In order to simulate a perfect posterior over $\theta'$, we thus track whether each visited track tile has ice and use these counts to update an exact posterior over $q$, equal to $\text{Beta}(\alpha + N_{+}, \beta + N_{-})$, where $N_{+}$ and $N_{-}$ correspond to the number of visited tiles with and without ice respectively. We then effectively sample $\theta' \sim \PGround(\theta'|\tau)$ by resampling all unvisited tiles from this posterior.

In order to sample from $\PGround(s'_t|\tau, \theta')$ and similarly from the grounded transition distribution $\PGround(s'_{t+1}|a_t, \tau, \theta')$, we make use of a second simulator we call the \emph{fictitious simulator}, which acts as a perfect model of the environment. We could otherwise learn this model via online or offline supervised learning. Our design choice using a second simulator in place of such a model allows us to cleanly isolate the effects of SAMPLR's correction for CICS from potential errors due to the inherent difficulties of model-based RL. 

Let us denote the primary simulator by $\mathcal{E}$, and the fictitious simulator by $\mathcal{E}'$. We ensure that the parameters of both simulators always match for $\theta \in \Theta \setminus \Theta'$. Before each training step, we first set the physics state of $\mathcal{E}'$ to that of $\mathcal{E}$ exactly, ensuring both simulators correspond to the same $s_t$, and then resample $\theta' \sim \PGround(\theta'|\tau)$ for the fictitious simulator as described above. We then take the resulting state of $\mathcal{E}'$ as $s'_t$. The agent next samples an action from its policy, $a_t \sim \pi(a_t|s'_t)$. Stepping forward $\mathcal{E}'$ in state $s'_t$ with action $a_t$ then produces a sample of $s'_{t+1}$ from a perfect grounded belief model and the associated reward, $r'_t$. During PPO training, the 1-step TD-errors $\delta_t$ for time $t$ are computed using these fictitious transitions. Similarly, the \plrabbrev{} mechanism underlying SAMPLR estimates regret using $\delta_t$ based on fictitious transitions.

\medskip
\noindent \textbf{Student architecture:} We adopt a policy architecture used across several prior works \citep{jiang2021robustplr, carracing_ppo, attentionagent}, consisting of a stack of 2D convolutions feeding into a fully-connected ReLU layer. The convolutions have square kernels of size 2, 2, 2, 2, 3, 3, output channels of dimension 8, 16, 32, 64, 128, 256, and stride lengths of 2, 2, 2, 2, 1, 1. The resulting 256-dimensional embedding is then fed into alpha, beta, and value heads. The alpha and beta heads are each fully-connected softplus layers, to whose outputs we add 1 to produce the $\alpha$ and $\beta$ parameters for the Beta distribution parameterizing each action dimension (i.e. each action is sampled from $\text{Beta}(\alpha, \beta)$ and then translated into the appropriate range). The value head is a fully-connected ReLU layer. All hidden layers are 100-dimensional. 

\medskip
\noindent \textbf{Choice of hyperparameters:} We selected hyperparameters based on evaluation performance over 5 episodes on the Italy, Singapore, and Germany F1 tracks with ice probability per tile fixed to $q=0.2$, when trained under the ice distribution featured in our main results, where $q \sim \text{Beta(1,15)}$. For each setting, we averaged results over 3 runs. PPO hyperparameters, shared across methods, were selected based on the performance of agents trained with domain randomization across settings in a grid search covering learning rate in $\{0.001, 0.0003, 0.0001, 0.00001\}$, number of epochs in $\{3,8\}$, number of minibatches in $\{2,4,16\}$, and value loss coefficient in $\{0.5,1.0,2.0\}$. The remaining PPO hyperparameters, as well as \plrabbrev{}-specific hyperparameters were based on those used in \citep{jiang2021robustplr}, with the exception of a smaller level buffer size, which we found helped improve validation performance. Additionally, for each method, we also swept over the choice of whether to initialize the policy to ensure actions are initially close to zero. Initializing the policy in this way has been shown to reduce variance in performance across seeds \citep{andrychowicz2020matters}.

\begin{table}[hbtp]
\caption{\small{Hyperparameters used for training each method.}}
\label{tab:hp_samplr}
\begin{center}
\scalebox{0.87}{
\begin{tabular}{lll}
\toprule
\textbf{Parameter} & Stochastic Fruit Choice & Black-Ice Car Racing \\
\midrule
\emph{PPO} & & \\
$\gamma$ & 0.995 & 0.99 \\
$\lambda_{\text{GAE}}$ & 0.95 & 0.9 \\
PPO rollout length & 256 & 125 \\
PPO epochs & 5 & 3 \\
PPO minibatches per epoch & 1 & 4 \\
PPO clip range & 0.2 & 0.2 \\
PPO number of workers & 32 & 16 \\
Adam learning rate & 1e-4 & 1e-4 \\
Adam $\epsilon$ & 1e-5 & 1e-5 \\
PPO max gradient norm & 0.5 & 0.5 \\
PPO value clipping & yes & no \\
return normalization & no & yes \\
value loss coefficient & 0.5 & 1.0 \\
entropy coefficient & 0.0 & 0.0 \\

\addlinespace[10pt]
\emph{PLR$^{\bot}$} and \emph{SAMPLR} & & \\
Replay rate, $p$ & 0.95 & 0.5 \\
Buffer size, $K$ & 4000 & 500 \\
Scoring function & PVL & PVL \\
Prioritization & rank & power \\
Temperature, $\beta$ & 0.3 & 1.0 \\
Staleness coefficient, $\rho$ & 0.3 & 0.7 \\

\bottomrule 
\end{tabular}
}
\end{center}
\end{table}

\end{document}